%% file: activefeatureselection-arxiv.tex
\documentclass[letterpaper]{article} % DO NOT CHANGE THIS
\usepackage{aaai21} 
\usepackage{times}  
\usepackage{helvet} 
\usepackage{courier}  
\usepackage[hyphens]{url} 
\usepackage{graphicx} 
\usepackage{hyperref}
\usepackage{natbib} 
\usepackage{caption}
\usepackage{amsmath,amsfonts,stmaryrd,amsthm,amssymb,enumitem}
\usepackage{algorithm,booktabs,todonotes}

\usepackage{algorithmicx}
\usepackage{algpseudocode}
\algnewcommand\algorithmicinput{\textbf{Input:}}
\algnewcommand\INPUT{\item[\algorithmicinput]}

\algnewcommand\algorithmicoutput{\textbf{Output:}}
\algnewcommand\OUTPUT{\item[\algorithmicoutput]}

\usepackage{pifont}

\usepackage{tikz}
\usetikzlibrary{arrows}
\usetikzlibrary{calc}
\usetikzlibrary{shapes}
\usetikzlibrary{fit}
\usetikzlibrary{matrix} 
\usetikzlibrary{positioning}
\usetikzlibrary{decorations.pathreplacing}

\renewcommand{\eqref}[1]{Eq.~(\ref{eq:#1})}
\newcommand{\figref}[1]{Figure~\ref{fig:#1}}
\newcommand{\tabref}[1]{Table~\ref{tab:#1}}        
\newcommand{\secref}[1]{Section~\ref{sec:#1}}

\newcommand{\myalgref}[1]{Alg.~\ref{alg:#1}}

\renewcommand{\P}{\mathbb{P}}

\newcommand{\Var}{\mathrm{Var}}

\newcommand{\half}{{\frac12}}

\DeclareMathOperator*{\argmax}{argmax}

\newcommand{\err}{\mathrm{err}}

\newcommand{\cD}{\mathcal{D}}

\newcommand{\cX}{\mathcal{X}}
\newcommand{\cY}{\mathcal{Y}}

\usepackage[utf8]{inputenc}
\usepackage{amssymb}
\usepackage{graphicx}
\usepackage{booktabs}			
\usepackage{amsmath}
\usepackage{commath}

\usepackage{amsthm}
\usepackage{mathtools}
\usepackage{caption}

\usepackage{graphicx}
\usepackage{subfigure}
\usepackage{lipsum}

\newcommand{\score}{\mathrm{score}}

\newcommand{\UCB}{\mathrm{UCB}}
\newcommand{\LCB}{\mathrm{LCB}}
\newcommand{\qu}{\hat{q}_v^u}
\newcommand{\ql}{\hat{q}_v^l}

\newcommand{\var}{\mathrm{var}}

\title{Active Feature Selection for the Mutual Information Criterion}
\author{
    Shachar Schnapp and 
    Sivan Sabato\\
}
\affiliations {
  Ben-Gurion University of the Negev\\
  Beer-Sheva, Israel \\
    schnapp@post.bgu.ac.il, sabatos@cs.bgu.ac.il
}

\begin{document}
\maketitle

\begin{abstract}
We study active feature selection, a novel feature selection setting in
  which unlabeled data is available, but the budget for labels is limited, and the examples to label can be actively selected by the algorithm. We focus on feature selection using the classical mutual information
  criterion, which selects the $k$ features with the largest mutual
  information with the label. In the active feature selection setting, the goal is to use significantly
  fewer labels than the data set size and still find $k$ features whose mutual
  information with the label based on the \emph{entire} data set is
  large.  We explain and
  experimentally study the choices that we make in the algorithm, and show
  that they lead to a successful algorithm, compared to other more naive
  approaches. Our design draws on insights which relate the problem of active
  feature selection to the study of pure-exploration multi-armed bandits settings. While we focus here on mutual information, our general methodology can be adapted to other feature-quality measures as well. The code is available at the following url:\\ \url{https://github.com/ShacharSchnapp/ActiveFeatureSelection}.
  
 \end{abstract}

\section{Introduction}\label{sec:intro}
Feature selection is the task of selecting a small
number of features which are the most predictive with respect to the label.
Filter methods for feature selection \citep[see, e.g.,][]{guyon2008feature} attempt to select the best features without committing  to a specific model or task.
We focus here on the classical mutual-information filter method, which selects the features with the largest mutual information with the label.

In the standard feature selection setting, the mutual information of each feature with the label is estimated based on
a fully-labeled data set. In this work, we study a novel active setting, in
which only the unlabeled data is readily available, and there is a limited budget for labels, which may
be obtained upon request. In this setting, the feature selection algorithm can
iteratively select an example from the data set and request its label, while
taking into account in its decision all the labels obtained so far. The goal
is to use significantly fewer labels than the data set size and still find $k$
features whose mutual information with the label based on the \emph{entire}
data set is large. This is the analog, for the feature-selection task, of the
active learning setting, first proposed in \citet{CohnAtLa94,McCallumNi98},
which addresses classification tasks.

Feature selection in such an active setting is useful in applications characterized by a preliminary stage of feature selection with expensive labels but readily available features, followed by a deployment stage  in which only a few features can be collected for each example. For instance, consider
developing a medical protocol which can only use a small number of features, so that patient intake is resource efficient. The goal during protocol design is to select the most informative patient attributes with respect to a
target label, such as the existence of a life-threatening condition. In the
preliminary study, it is possible to collect a large number of features about a cohort of patients, but identifying the condition of the patient (the label) with a
high accuracy may require paying trained specialists. By using an active feature selection algorithm which requires only a small number of labels, the labeling cost at the design stage can be controlled. 

As a second example, consider a design/deploy scenario of a computer system which trains classifiers online. The online system handles large volumes, and thus cannot measure all possible features. Therefore, at the design stage, a small number of features must be selected, which will be the only ones collected later by the deployed system. The feature selection process is performed on an offline system, and therefore does not have easy access to labels, but it does allow collecting all the features for the unlabeled feature selection data set, since it handles smaller volumes of data.
As a concrete example, consider a web-server
which needs to quickly decide where to route client requests, by predicting the  resource type that this client session will later
use. In the live system, it is easy to collect the true labels downstream
and feed them to the training algorithm. However, the server is limited in the
number of features it can retrieve for each request, due to bounded
communication with the client. On the other hand, in the design stage, there
is no limit to communicating with the client, due to the smaller processing volumes. However, simulating the live process in
order to identify the label is expensive, since the databases are not
local.

Our goal is then to interactively select examples to label from the feature selection data set, and to then use these labeled examples to select $k$ features which have a high mutual information with the label. If the budget for labels is limited, a naive approach is to label
a random subset of the unlabeled data set, and to then select the
top-ranking features according to this sub-sample. Our main contribution is a
practical algorithm that selects the examples to label using a smart
interactive procedure. This results in selecting features of a higher
quality, that is, features with a larger (true) mutual information with the label.
Our design draws on insights relating the problem of active feature selection to the study of pure-exploration of multi-armed bandits \citep{antos2008active,carpentier2011upper,chen2014combinatorial,garivier2016optimal}, which we elaborate on below.
While we focus here on mutual information, our general methodology can be adapted to other feature-quality measures, an extension that we leave for future work.

\paragraph{Paper structure.} Related work is discussed in \secref{relwork}. \secref{setting} presents the formal setting and notation. In \secref{single}, we discuss estimating the quality of a single feature is discussed; the solution is an important ingredient in the active feature selection algorithm, which is then presented in \secref{topk}. Experiments are reported in \secref{exps}. We conclude in \secref{discuss}. Full experiment results can be found in the appendices.

\section{Related work}\label{sec:relwork} 
Feature selection for classification based on a fully labeled sample is a widely studied task.
Classical filter methods \citep[see, e.g.,][]{guyon2008feature} select features based on a quality measure assigned to each feature. This approach does not take into account possible correlations between features, which is in general a computationally hard problem \citep{amaldi1998approximability}. Nonetheless, such filter methods are practical and popular due to their computational tractability. The most popular quality measures estimate the predictive power of the feature with respect to the label, based on the labeled sample. Two popular such measures are mutual information and the Gini index \citep[see, e.g.,][]{HastieTiFr01,guyon2008feature,SabatoSh08}. 
A different approach is that of the Relief algorithm \citep{kira1992feature}, which selects features that distinguish similar instances with different labels. 
%The feature selection problem is the need to choose the most relevant attributes/features from a data source for a given predictive modeling problem, \citep{chandrashekar2014survey} is an article that survey this topic.

%Many works attempt to address the possible correlations between features using tractable methods. In \citet{fleuret2004fast}, the mutual information criterion is extended to take into account features that have already been selected. In \citet{zheng2014submodular} and \citet{khanna2017scalable}, greedy selection algorithms are proposed, which are analyzed using sub-modularity properties. In \citet{apolloni2016two}, a wrapper method and a filter are combined. In \citet{emary2016binary}, an optimization procedure for selecting feature subsets is proposed. %We comment on the issue of feature correlations in \secref{discuss}.

 We are not aware of works on feature selection in the active setting studied  here. 
Previous works \cite{liu2003active,liu2004selective} propose to use selective sampling to improve the accuracy of the Relief feature selection algorithm within a limited run time. However, they use the labels of the entire data set, and so they do not reduce the labeling cost.
Alternatively, one might consider using active learning algorithms which output sparse models to actively select features. The theory of such algorithms has been studied for learning half-spaces under specific distributional assumptions \cite{zhang2018efficient}. However, we are not aware of practical active learning algorithms for a given sparsity level, which would allow performing joint active learning and feature selection using a limited label budget.

In this work, we show a connection between the active feature selection problem and exploration-only multi-armed bandit settings. A problem of uniformly estimating the means of several random arms with a small number of arm pulls is studied in \citet{antos2008active} and in \citet{carpentier2011upper}. In \citet{garivier2016optimal}, a proportion-tracking approach for optimal best-arm identification is proposed. 
In \citet{chen2014combinatorial}, the setting of combinatorial pure exploration is studied, in which the goal is to find a subset of arms with a large total quality, using a small number of arm pulls. We discuss these works in more detail in the relevant contexts throughout the paper.

\section{Setting and notations}\label{sec:setting}

For an integer $i$, denote $[i] = \{1,\ldots,i\}$. 
We first formally define the problem of selecting the top-$k$ features based on mutual information. Let $\cD$ be a distribution over $\cX \times \cY$, where $\cX$ is the example domain and $\cY$ is the set of labels. Each example $x \in \cX$ is a vector of $d$ features  $(x(1),\ldots,x(d))$. We assume for simplicity that each feature $j$ accepts one of a finite set of values $V_j$, and that the label $\cY$ is binary: $\cY = \{0,1\}$.\footnote{Generalization to multiclass labels is straight-forward; Real-valued features may be quantized into bins. }
Denote a random labeled example by $(\mathbf{X},Y) \sim \cD$, where $\mathbf{X} = (X(1),\ldots,X(d))$. The quality of feature  $j$ is measured by its mutual information with the label, defined as $I_j := I(X(j); Y)$, equal to
\begin{align}\label{eq:ij}
 &\sum_{\substack{y \in \{0,1\}\\v \in V_j}} \!\!\!\P(X(j)\!\! =\!\! v,Y\!\! =\!\! y) \log(\frac{\P(X(j)\!\! =\!\! v,Y\! =\! y)}{\P(Y\!\! =\!\! y)\P(X(j)\!\!=\!\!v)}).
\end{align}
Given an integer $k$, The goal of the feature selection procedure is to select $k$ features with the largest mutual information with the label. 
In standard (non-active) feature selection, the mutual information of each feature is estimated using a labeled sample of examples. 
In contrast, in active feature selection, the input is an i.i.d.~unlabeled sample $S := (x_1,x_2,...,x_m)$ drawn according to the marginal distribution of $\cD$ over $\cX$. At time step $t$, the algorithm selects an example $x_{i_t}$ from $S$ and requests its
label, $y_{i_t}$, which is drawn according to $\cD$ conditioned on $\mathbf{X} = x_{i_t}$. If the label of the same example is requested twice, the same label is provided again. The selection of $i_t$ at time step $t$ may depend on the labels obtained in previous time steps. When the budget of labels is exhausted, the algorithm outputs a set of $k$ features, with the same goal as the standard feature selection algorithm: selecting the features with the largest mutual information with the label.

If the unlabeled sample $S$ is sufficiently large, then the sample distribution of each feature $X(j)$ is sufficiently similar to the true distribution. Moreover, if all the labels for $S$ were known, as in the classical non-active setting, then the plug-in estimator for mutual information, obtained by replacing each probability value in \eqref{ij} with its empirical counterpart on the full labeled sample, would be a highly accurate estimator of the true mutual information \citep[see, e.g.,][]{Paninski03}. Therefore, for simplicity of presentation, we assume below that the empirical distribution of each feature on the unlabeled sample is the same as the true distribution, and that the best possible choice of features would be achieved by getting the labels for the entire sample $S$, and then selecting the top-$k$ features according to their mutual information plug-in estimate on the labeled sample. The challenge in active feature selection is to approach this optimal selection as much as possible, while observing considerably fewer labels.

\section{Active estimation for a single feature}\label{sec:single}

A core problem for the goal of actively selecting $k$ features based on their quality, is to actively estimate the quality of a single feature. We discuss this problem and its solution, as a stepping stone to the full active feature selection algorithm, which will incorporate this solution.
The quality of a single feature $j$ is $I_j = H(Y) - H(Y\mid X(j)),$
where $H$ is the entropy function.
For $v \in V_j$, denote $p_v := \P[X_j = v]$, $q_v := \P[Y = 1 \mid X(j) = v]$.
Since we are interested in estimating $I_j$ only for the purpose of ranking the features, it is sufficient to estimate the conditional entropy:
\[
H(Y \mid X(j)) \equiv \sum_{v \in V_j} p_v \cdot H(Y \mid X(j) =v) = \sum_{v \in V_j} p_v \cdot H_b(q_v),
\]
where $H_b(q):= q\log(1/q) + (1-q)\log(1/(1-q))$ is the binary entropy. 
This leads to the single-feature estimation problem, described as follows. We observe $m$ i.i.d.~draws $(z_1,\ldots,z_m)$ from the marginal distribution of $\cD$ over $X(j)$. At each time step $t$, the algorithm selects an index $i_t$ and requests the label $y_{i_t} \sim Y | X(j) = z_{i_t}$. The goal is to obtain a good plug-in estimate for $H_j := H(Y|X(j))$ using the available label budget. 
Since the examples $z_1,\ldots,z_m$ are values from  $V_j$, selecting an index $i_t$ is equivalent to selecting a value in $V_j$, and we have $y_{i_t} \sim Y| X(j) = v$. As discussed in \secref{setting}, we assume for simplicity of presentation that the data set is sufficiently large, so that $p_v$ for $v \in V_j$ can be obtained by calculating the empirical probabilities of the sample.
Thus, the single-feature estimation problem can be reformulated as follows: $\{p_v\}_{v \in V_j}$ are provided as input; at each time step, the algorithm selects a value $v \in V_j$ and obtains one random draw of $Y \mid X(j) = v$. Note that this is possible if the data set is large enough, as we assume. When the label budget is exhausted, $H_j$ is estimated based on the obtained samples. 
Since $\{p_v\}$ are known, the goal of the active algorithm is to estimate $\{q_v\}$ to a sufficient accuracy. Let $\hat{q}_v$ be the empirical plug-in estimate of $q_v$, based on the labels obtained for the value $v$. Then the entropy estimate is $\hat{H}_j := \sum_{v \in V_j} p_v H_b(\hat{q}_v)$. We have thus reduced the single-feature estimation problem to the problem of estimating $\{ q_v\}_{v \in V_j}$.

Previous works \citep{antos2008active, carpentier2011upper} study uniform active estimation of the expectations of several bounded random variables, by iteratively selecting from which random variable to draw a sample. In our problem, each value $v \in V_j$ defines the random variable \mbox{$Y \mid X(j) = v$}, with expected value $q_v$. In our notation, these works attempt to minimize the $\ell_\infty$ error measure $\max_{v \in V_j} |q_v - \hat{q}_v|$.
Given the true $q_v$, \citep{antos2008active} defines the \emph{optimal normalized static allocation}, a vector
%$\mathbf{b} := \{b_v\}_{v \in V_j}$ of integers which sum to $B$, which minimizes the expected error when each $q_v$ is estimated using $b_v$ random draws.
$\mathbf{w} = (w(v))_{v\in V_j}$ which sums to $1$, such that $w(v)$ is the fraction of draws that should be devoted to value $v \in V_j$ to minimize the expected $\ell_\infty$ error. 
However, $\mathbf{w}$ depends on the unknown $q_v$ values. The \emph{static-allocation strategy} of \citet{antos2008active} tries to approach the frequencies dictated by $\mathbf{w}$ by calculating an estimate $\mathbf{\hat{w}}$ at each time step, using the plug-in estimate of $q_v$ from the labeled data obtained so far.\footnote{\cite{antos2008active} add a small correction term to the plug-in estimate.} Let $n^{(t)}(v)$ be the number of labels requested for $v \in V_j$ until time step $t$. At time step $t$, the algorithm requests a label for the value $v_t$ which most deviates from the desired fraction of draws as estimated by $\hat{\mathbf{w}}$. Formally, it selects $v_t \in \argmax_{v \in V_j}\hat{w}(v)/n^{(t)}(v).$

In \citet{carpentier2011upper},
an improvement to the estimate of $\mathbf{w}$ is proposed, and shown to generate an empirically superior strategy with improved convergence bounds. This strategy replaces each plug-in estimate of $q_v$ in the calculation of $\mathbf{\hat{w}}$ by its \emph{upper confidence bound} (UCB). The UCB of $q_v$ is a value $\hat{q}_v^u$ such that with a high probability over the random samples, $\hat{q}_v^u \geq q_v$. Intuitively, this is more robust, especially when the static allocation strategy which uses the plug-in estimates may stop querying a random variable because of a wrong estimate, and thus never recover.

The approach of \citet{carpentier2011upper} can be generalized to minimizing any error measure that depends on $q_v$ and $\hat{q}_v$. 
We study three approaches for active single-feature estimation of the mutual information, based on three error measures. In \secref{exps}, we report an extensive experimental comparison.

\paragraph{The $\ell_\infty$ approach.}  Here, we use the algorithms of \citet{carpentier2011upper} for the $\ell_\infty$ error measure as is, thus attempting to make all estimates accurate. However,  this measure ignores the importance of each value, thus many labels could be spent on values $v$ with very small $p_v$.

\paragraph{The variance approach.} Here, we take the weights $p_v$ into account, by replacing the $\ell_\infty$ error with the variance of the weighted estimate for a budget of $B$ samples:
\[
\Var(\sum_v p_v \hat{q}_v) = \sum_v p_v^2 \Var(\hat{q}_v) = \sum_v p_v^2q_v(1-q_v)/(B w(v)).
\]
  Minimizing this objective subject to $\sum_v w(v) = 1$ leads to the following static allocation:
  \[w_{\var}(v) \propto p_v \sqrt{q_v(1 - q_v)}.\] 
  Here and below, the notation $w(v) \propto \alpha(v)$ for some function $\alpha$ indicates that $w(v) = \alpha(v)/\sum_{u \in V_j}\alpha(u)$.
As in the $\ell_\infty$ approach, we calculate $\hat{w}_{\var}(v)$, the estimate of $w_{\var}(v)$,
using UCBs. However, the property of interest here is
$q_v(1-q_v)$, and plugging in $\hat{q}_v^u$ instead of $q_v$ does
not lead to a UCB on this quantity.  Analogously to $\qu$,
let $\ql$ be a lower bound on the value of $q_v$, which holds with a high probability. Denote the upper
confidence bound of a function $f$ of $q_v$ based on these bounds by
\footnote{If no samples are available for the value $v$, we set by convention $\ql = 0$, $\qu = 1$.}
 \[
 \UCB(f,\ql,\qu) := \max_{x \in [\ql,\qu]} f(x).
 \]
Then, if $q_v \in [\ql,\qu]$, we have $f(q_v) \leq \UCB(f,\ql,\qu)$. If $f$ is monotonic increasing in $[0,\half]$ and monotonic decreasing in $[\half,1]$, then 
\begin{gather}\label{eq:ucbmon}
  \UCB(f,\ql,\qu) :=
\begin{cases}
f(\half) &  \ql \leq \half \leq \qu, \\
f(\ql) & \ql > \half,\\
f(\qu)& \text{otherwise } (\qu < \half).
\end{cases}
\end{gather}
This holds for $f= f_{\var}$, where $f_{\var}(x) := x(1-x)$.  
Plugging this into the formula for $w_\var$ given above, we get
$\hat{w}_{\var}(v) \propto p_v\sqrt{\UCB(f_{\var},\ql,\qu)}$. This
approach has the advantage of simplicity. However, it is not optimized for our
goal, which is to estimate the conditional
entropy.

\begin{figure}[htb]
  \begin{center}
    \includegraphics[width=0.4\textwidth]{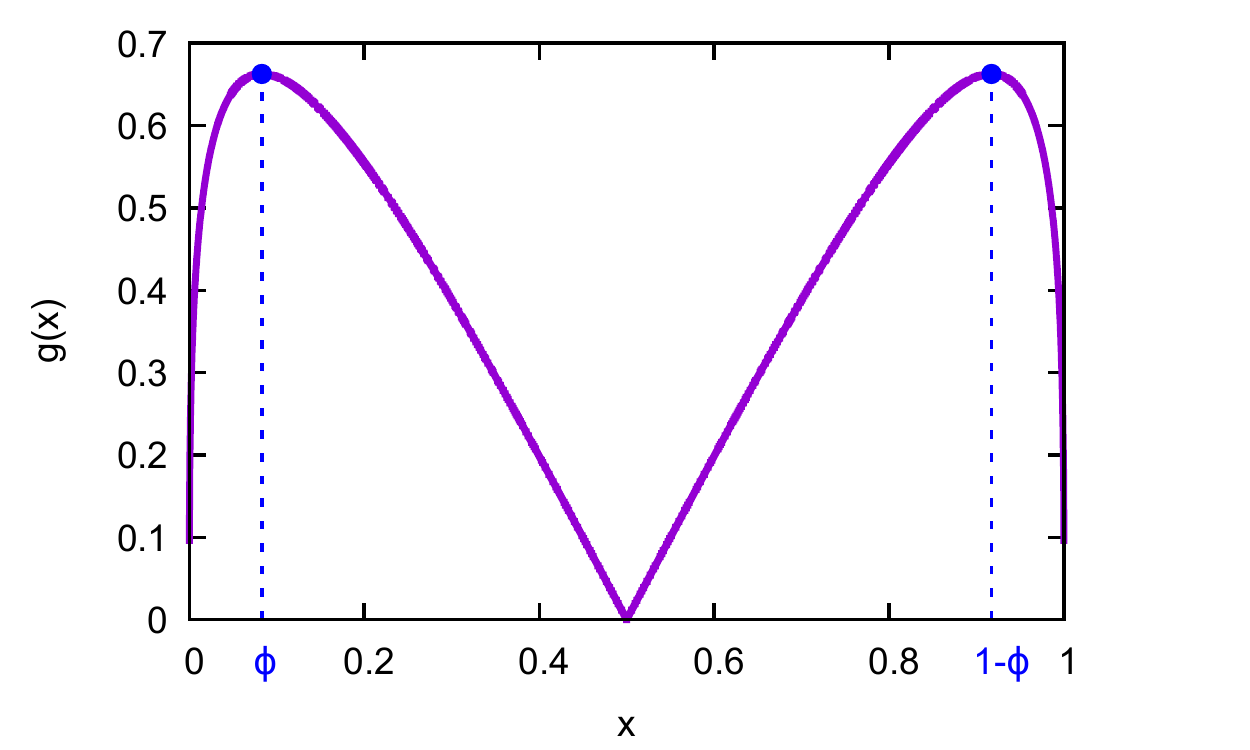}
  \end{center}
  \caption{The function $g(x)$ defined in Active estimation for a single feature section}
  \label{fig:gx}
\end{figure}

\paragraph{The conditional entropy approach.} Here, we set the error to the variance of the
conditional entropy estimate:
$\Var(\hat{H}_j) \equiv \sum_v p_v^2 \cdot \Var(H_b(\hat{q}_v))$.
Since $H_b$ is a smooth function of the random variable $\hat{q}_v$, its variance can be approximated based on its Taylor expansion \citep{benaroya2005probability}. Letting $H_b'$ be the derivative of $H_b$, and $g(x) := \sqrt{x(1-x)}|\log (\frac{x}{1-x})|$, we have \[\Var(H_b(\hat{q}_v)) \approx \Var(\hat{q}_v){H'_b}^2(q_v) = g^2(q_v)/(B w(v)).\]
Therefore, $\Var(\hat{H}_j) \approx \sum_v p_v^2  g^2(q_v)/(B w(v))$. Minimizing the RHS subject to $\sum_v w(v) = 1$ gives the static allocation $w_I(v) \propto p_v  g(q_v)$.
Here too, we estimate $\mathbf{\hat{w}}_I$ using upper confidence bounds. By differentiating $g$, one can see there is a value $\phi\approx 0.083$ such that $g$ is increasing in $[0,\phi]$ and in $[\half, 1-\phi]$ and decreasing in $[\phi,\half]$ and in $[1-\phi,1]$ (see illustration in \figref{gx}). 

This implies the following upper confidence bound for $g$:
\begin{gather*}
\UCB(g, \ql, \qu) =
\begin{cases}
g(\phi),\text{ } \phi \in  [\ql, \qu]\cup [1-\qu,1-\ql] \\
\max\{g(\ql),g(\qu)\}, \text{ } \text{otherwise}.
\end{cases}
\end{gather*}
Replacing $g(q_v)$ in the definition of $w_I(v)$ with its upper confidence bound, we get
\begin{equation}\label{eq:winf}
\hat{w}_I(v) := \frac{\alpha(v)}{\sum_{u \in V_j}\alpha(u)}, \text{ } \alpha(v) := p_v\cdot\UCB(g,\ql,\qu).
\end{equation}

The two new error measures we suggested require calculating $\hat{q}_v^l$ and $\hat{q}_v^u$ for each $v\in V_j$. This can be done  using standard closed-form concentration
inequalities such as Hoeffding's inequality \cite{Hoeffding63} or Bernstein's
inequality \cite{Bernstein24}, as done in \citet{carpentier2011upper}. However, these bounds can be very loose
for small sample sizes. Since $\hat{q}_v$ is Binomially distributed, the tightest concentration bounds are given by the Clopper-Pearson formulas
\citep{clopper1934use}, which can be numerically calculated. Overall, we consider the following options for the single-feature estimation problem, with their abbreviations given in parenthesis:
The two algorithms of \citet{carpentier2011upper} for the $\ell_\infty$ measure, one based on Hoeffding's inequality (MAX-H) and the other on Bernstein's inequality (MAX-B); Static allocation strategy with $\mathbf{\hat{w}}_{\var}$, with each of the concentration inequalities (VAR-H, VAR-B, VAR-CP); Static allocation strategy with $\mathbf{\hat{w}}_{I}$, with each of the concentration inequalities (I-H, I-B, I-CP).
  In addition, we consider a naive baseline which allocates the labels proportionally to $p_v$ (PROP). Our experiments, reported in \secref{exps}, show that I-CP is distinctly the most successful alternative. Indeed, it is the most tailored for the estimation task at hand. In the next section, we describe the full active feature selection algorithm, which incorporate the single-feature allocation strategy as a component.

\section{The active feature selection algorithm}\label{sec:topk}
The active feature selection algorithm attempts to select $k$ features
with the largest mutual information with the label. Naively, one could estimate $H_j$ separately for each of the features using the best single-feature estimation strategy, as described in \secref{single}.
However, this would cause several issues. First, using each label for
estimating only a single feature is extremely wasteful. Indeed, the label
budget might even be smaller than the number of features. On the other hand,
selecting the example to label based on several features may induce a sampling bias. In addition, not all features are as likely to be a part
of the output top-$k$ list. This should be taken into account when selecting examples.
Our approach mitigates these issues, and indeed performs well in practice, as demonstrated in \secref{exps}.

The general framework is as follows. We define a score $\score(x,F)$ for each example $x \in S$ and feature subset $F \subseteq [d]$. The score estimates the utility of labeling the example $x$ in improving the entropy estimates of the features in $F$. In each iteration $t$, we calculate a feature subset $F^{(t)} \subseteq [d]$ based on the information gathered so far, and select an example $x$, out of those not labeled yet, which maximizes $\score(x,F^{(t)})$. We first discuss the calculation of the subset $F^{(t)}$; thereafter, we present
our proposed score. To set $F^{(t)}$, we take inspiration from the problem of
Combinatorial Pure Exploration (CPE) of multi-armed bandits
\citep{chen2014combinatorial}. In this problem, the goal is to select a set of reward
distributions (arms) that maximizes the total expected reward, by
interactively selecting which distribution to sample at each step. A special
case of this problem is the top-$k$ arm-selection problem \citep{kalyanakrishnan2010efficient},
where the selected set of arms must be of size $k$. The CPE problem does not directly
map to the active feature-selection problem, since in CPE, each step provides information about a
single arm, while in active feature-selection, each
label provides information about all the features. Nonetheless,
a shared challenge is to decide which arms/features to consider in each
sampling round; That is, how to set $F^{(t)}$.

\begin{algorithm}[t]
  \begin{algorithmic}[1]
    \INPUT Unlabeled sample $S = (x_1,\ldots,x_m)$, number of features to select $k$, confidence $\delta \in (0,1)$, label budget $B$, number of iterations for testing change $\Lambda$.
    \State For all $j \in [d], v \in V_j$, $p_v(j) \gets \P_{X \sim S}[X(j) = v]$.
    \State $S_1 \gets ()$; $\forall j\in[d],v\in V_j, n^{(1)}(j,v) \gets 0, \hat{q}_v(j) \gets 0$.
    \For{$t \in [B]$}
        
    \State For all $j \in [d]$,
    $\hat{H}_j \gets \sum_{v \in V_j} p_v(j) H_b(\hat{q}_v(j)),$ \\
    $\qquad\hat{H}^l_j \gets \sum_v p_v(j) \cdot \LCB(H_b, \ql(j), \qu(j)),$\\
    $\qquad\hat{H}^u_j \gets \sum_v p_v(j) \cdot \UCB(H_b, \ql(j), \qu(j)).$
    \State $F^{(t)}_k \leftarrow $ Top-$k$ features according to $\hat{H}_1,\ldots,\hat{H}_d$
        \State $\forall j \in F^{(t)}_k$, $\tilde{H}_j \gets \hat{H}_j^l$. $\forall j \notin F^{(t)}_k$, $\tilde{H}_j \gets \hat{H}_j^u$.
        \State $\tilde{F}^{(t)}_k \gets$ Top-$k$ features according to $\tilde{H}_1,\ldots,\tilde{H}_d$.
        \State $F^{(t)} \leftarrow F^{(t)}_k \triangle \tilde{F}^{(t)}_k$ \,\,\,\,\# The symmetric difference
        \State \textbf{if} $F$ is empty, \textbf{then} break from the loop.
         \State Calculate $\mathbf{\hat{w}}_I^{(t)}$ according to \eqref{winf}.

         \State $\displaystyle i_t \gets \textstyle\argmax_{x_i \in S\setminus S_{t}} \score^{(t)}(x_i, F^{(t)}).$ ~\# \eqref{score}
         \State Request the label $y_{i_t}$.
         \State $S_{t+1} \gets S_{t} \circ (x_{i_t}, y_{i_t})$
         \For{$j\in [d]$ set $v:= x_{i_t}(j)$ and}
           \State $n^{(t+1)}(j,v) \gets n^{(t)}(j,v)+1$
           \State \mbox{$\hat{q}_v(j)\! \gets\! |\{ i \mid \! x_i\! \in\! S_t, x_i(j)\! =\! v, y_i\! =\! 1\}|/n^{(t)}(j,v)$}.
           \State Update $\ql(j), \qu(j)$ according to \citet{clopper1934use} with confidence parameter $\delta$.
           
           \EndFor
           \State \textbf{if} $\sum_{i\in F_k^{(t)}} \hat{H}_i$ has not changed in the last $\Lambda$ iterations, break from the loop and select the remaining examples to label at random.
         \EndFor
   \State Calculate $\hat{H}_j$ for all $j$ based on the labeled examples
   \State Return the top-$k$ features according to $\hat{H}_1,\ldots,\hat{H}_d$.
\end{algorithmic}
  \caption{AFS: Active Feature Selection for the Mutual Information Criterion}
  \label{alg:afs}
\end{algorithm}

Our algorithm sets $F^{(t)}$ by adapting the approach of \citet{chen2014combinatorial} to active
feature selection, as follows: Given an estimator $\hat{H}_j$ for $H_j$, and
high-probability lower and upper bounds $\hat{H}^l_j$ and $\hat{H}^u_j$ for $H_j$,
define two sets of $k$ features. The first set, $F^{(t)}_k$, holds the top-$k$
features according to the estimates $\{ \hat{H}_j\}$ (that is, the $k$ features
with the smallest $\hat{H}_j$). The second set, $\tilde{F}^{(t)}_k$, holds an
alternative choice of top-$k$ features: those with the smallest estimates, when the estimates for $j \in F^{(t)}_k$ are set to $\hat{H}_j^l$, and the  estimates  for $j \notin F^{(t)}_k$ are set to $\hat{H}_j^u$. $F^{(t)}$ is set to the
symmetric difference of $F^{(t)}_k$ and $\tilde{F}^{(t)}_k$. Similarly to \citet{chen2014combinatorial},  improving the estimates of these features is expected to have the most impact on the selected top-$k$ set.
To calculate the required $\hat{H}_j,\hat{H}^l_j, \hat{H}^u_j$, denote $p_v(j) := \P[X(j) = v]$. Since $H_j = \sum_v p_v(j) H_b(q_v(j))$, we calculate $\hat{H}_j$ by replacing $H_b(q_v(j))$ with $H_b(\hat{q}_v(j))$. For $\hat{H}^u_j$, replace $H_b(q_v(j))$ with $\UCB(H_b, \ql(j), \qu(j))$, calculated via \eqref{ucbmon}. For $\hat{H}^l_j$, replace $H_b(q_v(j))$ with:
$\LCB(H_b, \ql(j), \qu(j)) := 
 \min_{x \in [\ql,\qu]} H_b(x)= \min\{H_b(\ql(j)), H_b(\qu(j))\}.$

We now discuss the scoring function $\score(x,F)$. First, we define a score for an example $x$ and a single feature $j$. $\score(x,F)$ will aggregate these scores over $j \in F$. Denote the set of examples
labeled before time step $t$ by $S_t$, and let
$n^{(t)}(j,v) :=|\{ x \in S_t \mid x(j)=v\}|$ be the number of examples
requested so far with value $v$ in feature $j$. Based on the single-feature static allocation
strategy, a naive approach would be
to set the score of $x$ for feature $j$ to
$\hat{w}_I^{(t)}(j)/n^{(t)}(j,x(j))$.
However, this may cause significant sampling bias which could skew the entropy estimates, since the aggregate score may encourage labeling specific combinations of feature values, thereby biasing the estimate of $\{q_v\}$ for some features. While complete bias avoidance cannot be guaranteed in the general case, we propose a computationally feasible heuristic for reducing this bias.
We add a correction term to the single-feature score, which rewards feature pairs that have been observed considerably less than their proportion in the data set. We do not consider larger subsets, to keep the heuristic tractable. Denote the sample proportion of examples with values $v_1,v_2$ for features $j_1,j_2$ by
\mbox{$\hat{p}(j_1,j_2,v_1,v_2) := \P_{X \sim S}[X({j_1}) = v_1, X({j_2}) = v_2]$}, and the proportion of these pairs in $S_t$ by
$\hat{p}^{(t)}(j_1,j_2,v_1,v_2) := \P_{X \sim S_t}[X({j_1}) = v_1, X({j_2}) =  v_2].$
If the ratio between these proportions is large, this indicates a strong sampling bias for this feature pair. Denote $\rho^{(t)} := \hat{p}/\!\max(\hat{p}^{(t)},1/|S|)$.\footnote{The maximization circumvents infinity if $\hat{\rho}^{(t)}=0$.}
We aggregate the ratios $\rho^{(t)}$ for the relevant pairs of features using an aggregation function denoted $\psi$, to create a correction term which multiplies the naive single-feature score, thus encouraging labeling examples with under-represented pairs.
$\psi$ maps a real-valued vector to a single value that aggregates its coordinate values. A reasonable function should be symmetric and monotonic; Thus, a natural choice is a vector norm, e.g., $\ell_1$. Our experiments in \secref{exps} show that other reasonable choices work similarly well.
We define the score for a given function $\psi$ as follows: 
At time $t$, the single-feature score  of an example $x$ for feature $j$ is:\footnote{We add $1$ to the denominator to avoid an infinite score; Note that it is
  impractical in this setting to guarantee one sample of each
  feature-value pair, since this could exhaust the labeling budget. }
  \begin{align*}
    &\score^{(t)}(x, j, F^{(t)}) := \\
     & \frac{\hat{w}^{(t)}(j)}{n^{(t)}(j,x(j))+1}\cdot \psi((\rho^{(t)}(j,r,x(j),x(r))_{r \in F^{(t)}\setminus \{j\}}).
  \end{align*}
  At time step $t$, the example with the largest overall score $\score^{(t)}$ is selected, where the score is defined as:
  \begin{equation}\label{eq:score}
    \score^{(t)}(x, F^{(t)})\! =\! \psi((\score^{(t)}(x, j, F^{(t)}))_{j \in F^{(t)}}).
  \end{equation}

  The full active feature selection algorithm, AFS, is given in \myalgref{afs}. AFS gets as input an unlabeled sample of size $m$, the number of features to select $k$, a label budget $B$, and a confidence parameter $\delta$, which is used to set the lower and upper bounds $\ql,\qu$. An additional parameter $\Lambda$ is used for a safeguard described below. AFS outputs $k$ features which are estimated to have the largest mutual information with the label. The computational complexity of the algorithm is $O(B(d + mk+ k\log(d)))$.
  
  AFS includes an additional safeguard, aimed at preventing cases in which 
  the selection strategy of AFS leads to a strongly biased estimate of the mutual information, and the selection strategy itself is too biased to allow collecting labels that correct this estimate.   AFS checks if $\sum_{i\in F_k^{(t)}} \hat{H}_i$ for the current top-$k$ features has changed in the last $\Lambda$ iterations. If it has not changed, AFS selects the remaining examples at random. This guarantees that a wrong estimate will eventually be corrected, while a correct estimate will not be harmed. This safeguard only takes effect in rare cases, but it is important for preventing catastrophic failures.

\section{Experiments}\label{sec:exps}
We first report experiments for the single-feature estimation problem, comparing the approaches suggested in \secref{single}. These clearly show that the I-CP approach is preferable. Then, we report experiments on several benchmark data sets for the AFS algorithm, comparing it to three natural baselines. We further report ablation studies, demonstrating the necessity of each of the mechanisms of the algorithm. Lastly, we compare different choices of $\psi$. Python $3.6$ code for all experiments is available at: \url{https://github.com/ShacharSchnapp/ActiveFeatureSelection}. The full experiment results are reported in the appendices.
All experiments can be run on a standard personal computer.

\textbf{Single-feature estimation.} 
We tested the algorithms listed in \secref{single}
in synthetic and real-data experiments. For the synthetic experiments, we created features with the
same $p_v$ for all feature values. This is a favorable
setting for MAX-H and MAX-B, which do not take $p_v$ into account. We generated two sets of synthetic experiments, with features of cardinality in $\{2,4,6,8,10\}$. In the first set, $q_v$ was drawn uniformly at random from $[0,\half]$; This was repeated $5$ times for each set size, resulting in $25$ experiments. In the second set, we tested cases with extreme $q_v$ values. For each set size, we set $n$ values to have $q_v = 1/2$ and the rest to $q_v = \alpha$, for all
combinations of $\alpha \in \{0.1,0.01,0.001\}$ and
$n \in \{0,1,\ldots,|V_j|\}$. This resulted in $95$ synthetic experiments.
For the real-data experiments, we tested the $14$ features of the
\texttt{Adult} data set \cite{UCI2019} with their true $p_v$ values.  We ran each test scenario 10,000 times and calculated the average estimation error, defined as $\err := |\hat{H}_j - H_j|$.
We further calculated $95\%$ student-t confidence intervals. 
For algorithm $i$, denote its confidence interval $[\err^l_i,\err^u_i]$. We say that $i$ has a ``clear win'' if $\err^u_i \leq \min_{j \neq i} \err^l_j$, and a ``win'' if $\err^l_i \leq \min_{j \neq i} \err^u_j$. 
\tabref{singlewins} reports numbers of clear wins and wins for various budgets, for each experiment and algorithm. The I-CP
approach is clearly the most successful. See \appref{app:single} for detailed results.

\textbf{Active feature selection.}
Our active feature selection experiments were performed on all the large data sets with binary labels and discrete features in the feature-selection repository of \citet{li2018feature}. In addition, we tested on the MUSK data set \citep{UCI2019}, and on the MNIST data set \cite{lecun-mnisthandwrittendigit-2010} for three pairs of digits. Data set properties are listed in \tabref{datasets}.
We tested feature numbers $k \in  \{1,5,10,20\}$.

We compared our algorithm to three natural baselines, which differ in how they select the examples to label. In all baselines, the selected $k$ features were the ones with the largest mutual information estimate, calculated based on the selected labels. The tested baselines were:
\begin{enumerate}
\item

RANDOM: Select the examples to label at random from the data set; This is  the passive baseline, since it is equivalent to running passive feature selection based on the mutual information criterion on a random labeled sample of size $B$.
\item

CORESET: Use a coreset algorithm to select the $B$ most representative examples from the data set. We used the classical Farthest-First Traversal algorithm (see, e.g., \citealp{ros2017dides,ros2020sampling}) with the Hamming distance, which is the most appropriate for categorical values.
\item

DWUS: Label the examples selected by the active learning algorithm Density Weighted Uncertanty Sampling \cite{nguyen2004active},
    implemented in the python package \texttt{libact} \cite{YY2017}.
\end{enumerate}
    
To compare the algorithms, we calculated for each algorithm the mutual information gap between the true top-$k$ features  and the selected features, calculated via $\sum_{j \in F^*} H_j - \sum_{j \in F} H_j$, where $F$ are the top-$k$ features based on the collected labels, and $F^*$ are the top-$k$ features according to the true $H_j$, which was calculated from the full labeled sample. 

\begin{figure}[t]
  \begin{center}
  \includegraphics[width = 0.4\textwidth]{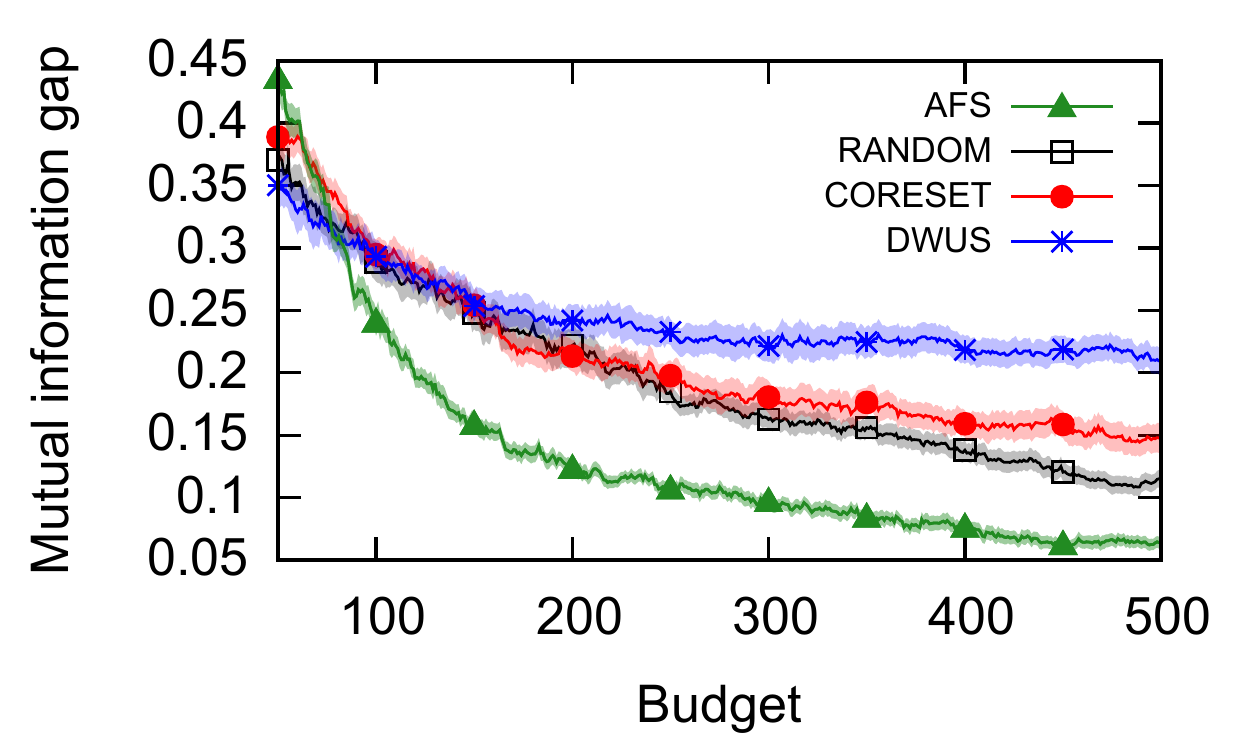}
  \includegraphics[width = 0.4\textwidth]{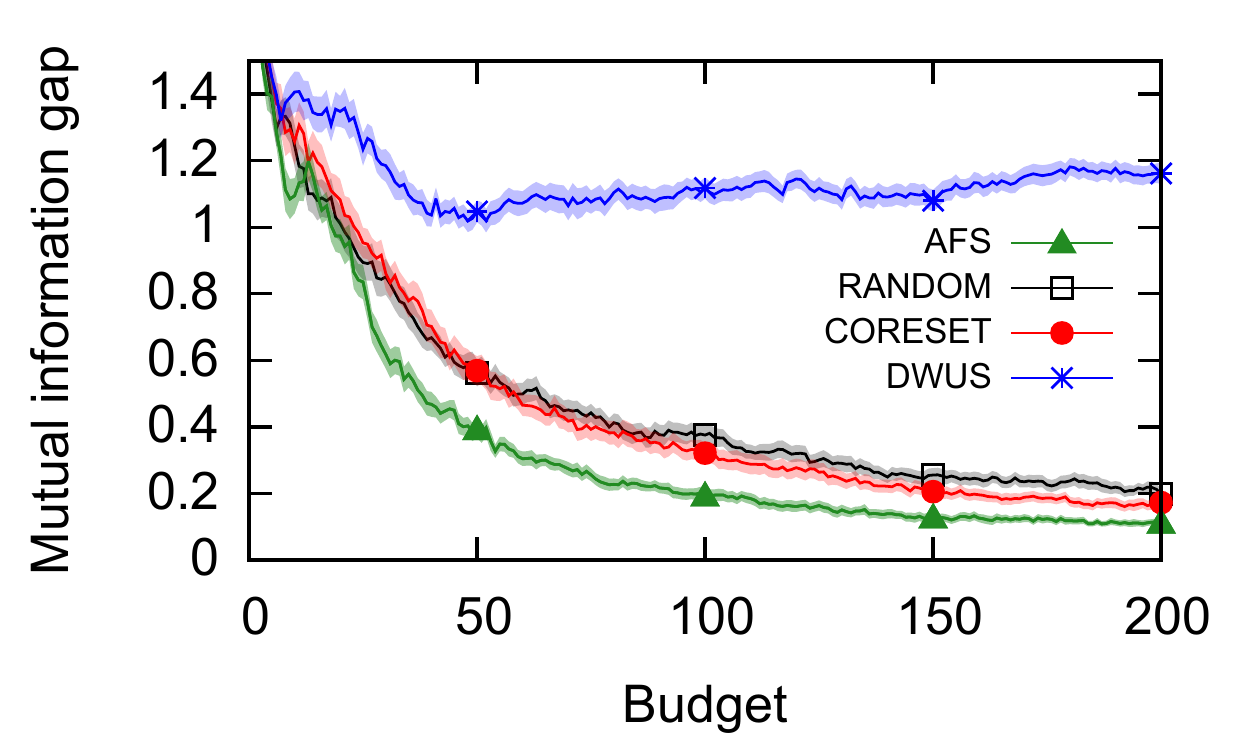}
  \includegraphics[width = 0.4\textwidth]{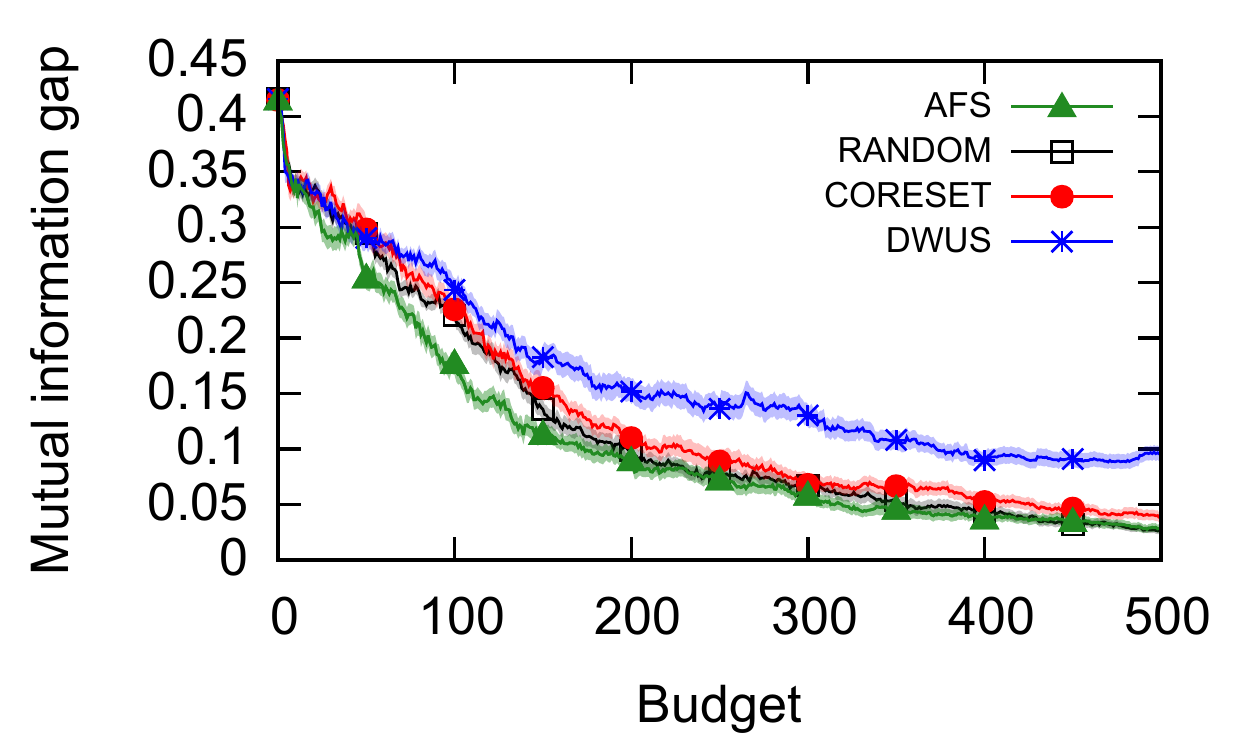}

  \end{center}
  \caption{Some of the experiments comparing to baselines. Top: MUSK, $k=20$. Middle: MNIST, 3 vs 5, $k=20$. Bottom: RELATHE, $k=10$. See full experiment results in Appendix~\ref{app:baseline}.
  }
  \label{fig:baselines}
\end{figure}

We ran each experiment $30$ times. All graphs plot the average score with $95\%$ confidence intervals. We provide in \figref{baselines} the graphs for three of the experiments, demonstrating the advantage of AFS over the baselines. Graphs for all experiments are provided in full in Appendix~\ref{app:baseline}.
In all experiments, AFS performs better or comparably to the best  baseline; its improvement is larger for larger values of $k$.

\begin{table}
  \begin{center}
  \caption{Data set properties}
  \label{tab:datasets}
  {\small
    \begin{tabular}{cccc}
    Data set & Instances & Features & Classes \\
    \toprule
    BASEHOCK & 1993 & 4862 & 2\\
    \midrule
    PCMAC & 1943 & 3289 & 2 \\
    \midrule
      RELATHE & 1427 & 4322 & 2 \\
      \midrule
      MUSK & 6598 & 167 & 2 \\
      \midrule
    MNIST: $0$ vs $1$ & 14,780 & 784 & 2 \\
    \midrule
      MNIST: $3$ vs $5$ & 13,454 & 784 & 2 \\
      \midrule
      MNIST: $4$ vs $6$ & 13,700 & 784 & 2 \\
    \end{tabular}
    }
  \end{center}
\end{table}

\begin{table}[t]
  \begin{center}
    \caption{Single-feature synthetic experiments, reporting numbers of (clear wins, wins) for each algorithm.}
    \label{tab:singlewins}
    {\small Synthetic experimens with uniform $q_v$}\\
	\resizebox{0.5\textwidth}{!}{
	\begin{tabular}{cccccccccc} 
Budget & PROP & MAX-H & MAX-B & VAR-H & VAR-B & VAR-CP & I-H & I-B & I-CP \\ 
\toprule
50 & (0, 12) & (0, 2) & (0, 3) & (0, 12) & (0, 11) & (1, 1) & (0, 12) & (0, 13) & $\mathbf{(10, 23)}$\\ 
\toprule
100 & (0, 4) & (0, 2) & (0, 4) & (0, 4) & (0, 1) & (0, 0) & (0, 4) & (0, 4) & $\mathbf{(21, 25)}$\\ 
\toprule
300 & (0, 4) & (0, 2) & (0, 4) & (0, 4) & (0, 1) & (0, 0) & (0, 4) & (0, 10) & $\mathbf{(15, 25)}$\\ 
\toprule
500 & (0, 3) & (0, 2) & (0, 3) & (0, 3) & (0, 1) & (0, 0) & (0, 3) & (0, 16) & $\mathbf{(9, 25)}$\\ 
\end{tabular} 	
	}
        {\small Synthetic experiments with fixed $q_v$}\\
	\resizebox{0.5\textwidth}{!}{
	\begin{tabular}{cccccccccc} 
Budget & PROP & MAX-H & MAX-B & VAR-H & VAR-B & VAR-CP & I-H & I-B & I-CP \\ 
\toprule
50 & (0, 21) & (0, 6) & (0, 7) & (0, 21) & (0, 18) & (23, 38) & (0, 21) & (0, 19) & $\mathbf{(40, 68)}$\\ 
\toprule
100 & (0, 13) & (0, 8) & (0, 7) & (0, 13) & (0, 10) & (23, 34) & (0, 13) & (0, 13) & $\mathbf{(48, 69)}$\\ 
\toprule
300 & (0, 17) & (0, 5) & (0, 14) & (0, 17) & (0, 3) & (23, 25) & (0, 17) & (0, 32) & $\mathbf{(39, 71)}$\\ 
\toprule
500 & (0, 15) & (0, 6) & (0, 11) & (0, 15) & (0, 2) & (8, 11) & (0, 15) & (4, 42) & $\mathbf{(43, 78)}$\\ 
\end{tabular} 

	}
        {\small Real-data experiments on the \texttt{Adult} data set}\\
	\resizebox{0.5\textwidth}{!}{
\begin{tabular}{cccccccccc} 
Budget & PROP & MAX-H & MAX-B & VAR-H & VAR-B & VAR-CP & I-H & I-B & I-CP \\ 
\toprule
50 & (0, 11) & (0, 1) & (0, 1) & (0, 11) & (0, 9) & (0, 6) & (0, 11) & (0, 11) & $\mathbf{(2, 13)}$\\ 
\toprule
100 & (0, 5) & (0, 0) & (0, 0) & (0, 5) & (0, 6) & (0, 3) & (0, 6) & (0, 7) & $\mathbf{(6, 14)}$\\ 
\toprule
300 & (0, 3) & (0, 1) & (0, 1) & (0, 3) & (0, 5) & (0, 3) & (0, 3) & (0, 8) & $\mathbf{(6, 14)}$\\ 
\toprule
500 & (0, 5) & (0, 1) & (0, 1) & (0, 5) & (0, 4) & (0, 2) & (0, 5) & (0, 7) & $\mathbf{(5, 13)}$\\ 
\end{tabular} 
	
	}
  \end{center}
\end{table}

\begin{figure}[H]
  \begin{center}
      \includegraphics[width = 0.4\textwidth]{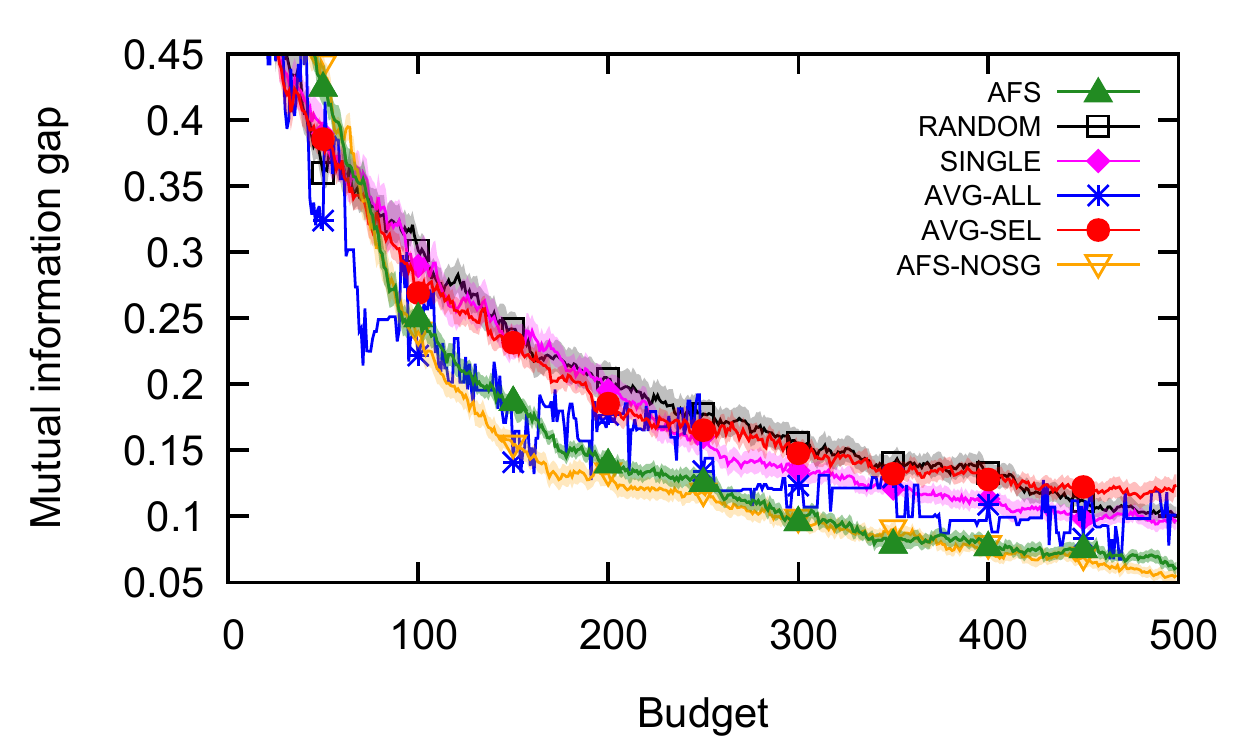} 
        \includegraphics[width = 0.4\textwidth]{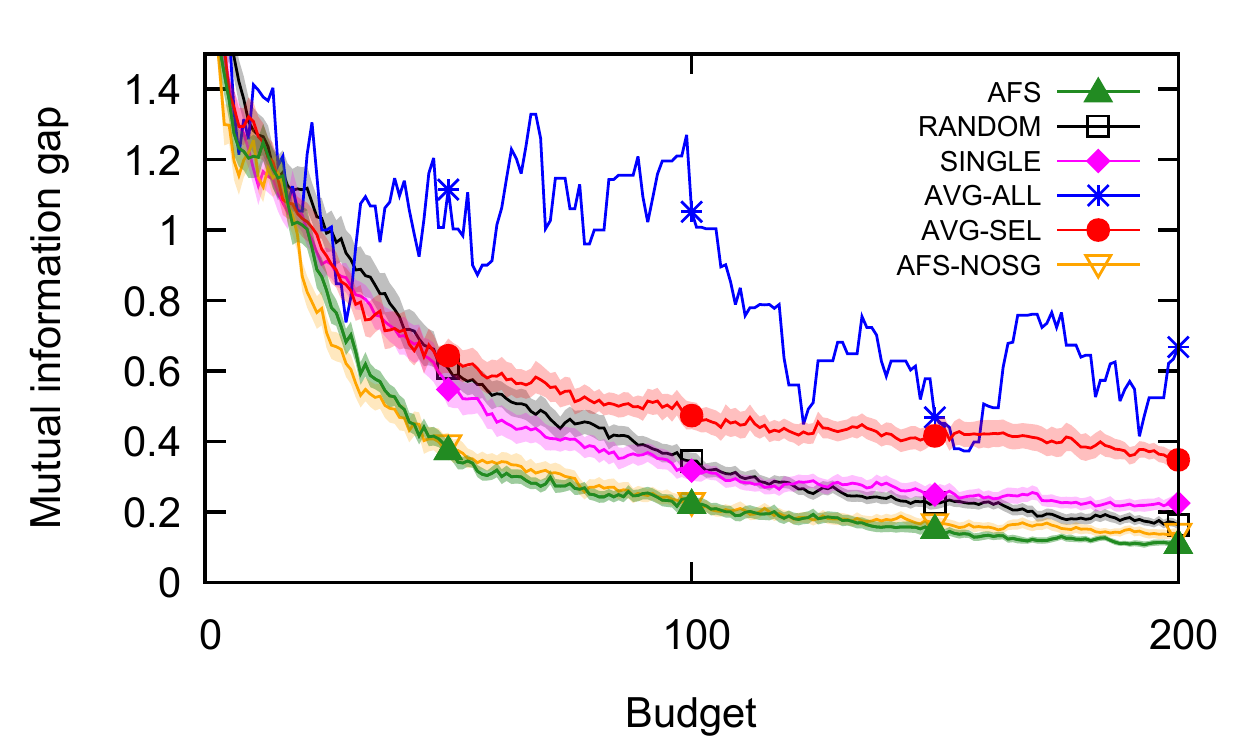} 
        \includegraphics[width = 0.4\textwidth]{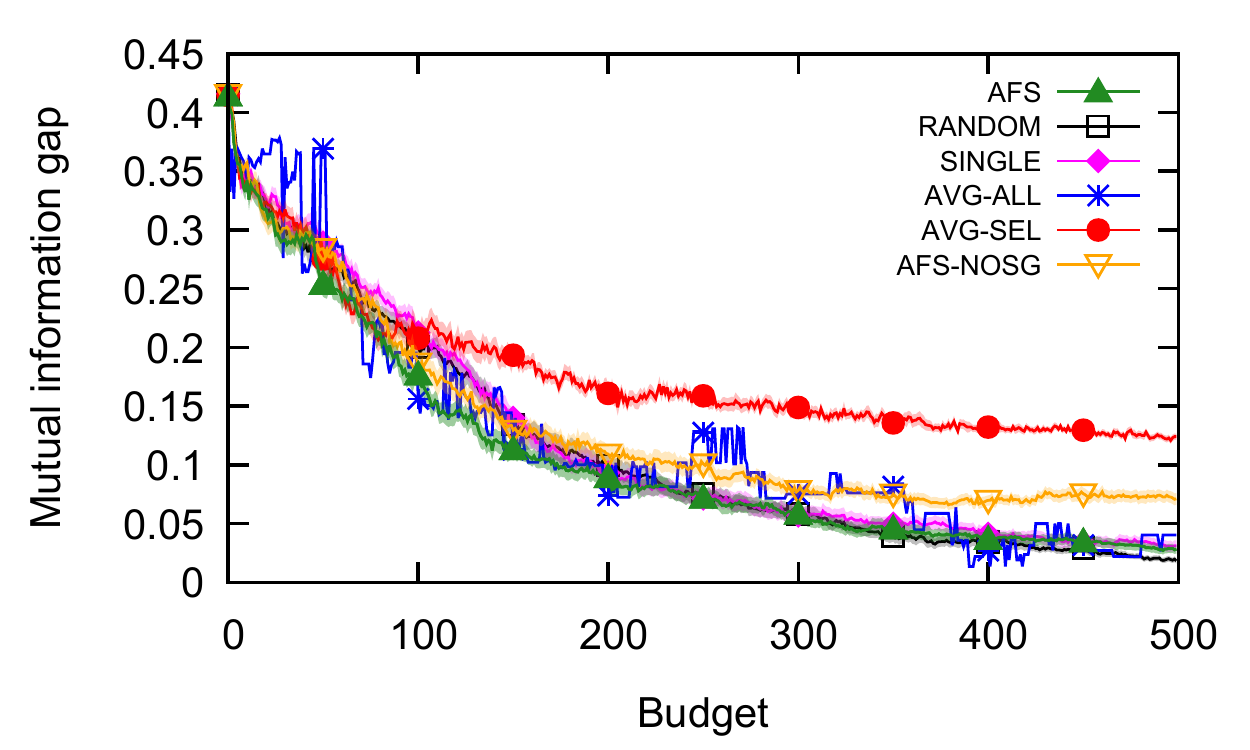}
  \end{center}
  \caption{Some of the ablation tests. Top: MUSK, $k=20$. Middle: MNIST, 3 vs 5. Bottom: RELATHE, $k=10$. See full experiment results in Appendix~\ref{app:ablation}%\appref{ablation}.
  }
  \label{fig:ablations}
\end{figure}

\textbf{Ablation tests.} We study the necessity of the mechanisms used by AFS, by testing the following variations:
\begin{enumerate}
\item Select the example to label by the score of a single randomly-selected feature (SINGLE);
\item Select the example with the largest average score over all features, without a bias correction term (AVG-ALL);
\item Select the example with the largest average score over $F^{(t)}$, without a bias correction term (AVG-SEL);
  \item The full AFS, with $\psi = \ell_1$, $\delta=  0.05$, but without the safeguard (that is, $\Lambda = \infty$) (AFS-NOSG);
  \item The full AFS, with $\psi = \ell_1, \delta = 0.05, \Lambda = 30$ (AFS).
  \end{enumerate}
  
The value of $\Lambda$ for AFS was selected after testing the values $10,20,30,40$.  The five variations above were tested on all the data sets in \tabref{datasets} for $k \in \{5,10,20\}$. Graphs for all experiments are provided in \appref{app:ablation}.
We provide in \figref{ablations} the graphs for three of the ablation tests, which show that AFS is the only algorithm that performs consistently well.
We further compared our choice of $\psi := \ell_1$ to other natural options: $\psi := \ell_\infty$ and 
$\psi := \ell_2$. The graphs are provided in \appref{app:aggregate}. We observed that in most cases, the results are similar for $\ell_1$ and $\ell_2$, while $\ell_\infty$ is sometimes slightly worse. Thus, while we chose $\ell_1$, setting $\psi := \ell_2$ is also a valid choice.

\section{Conclusion}\label{sec:discuss} 
We showed that actively selecting examples to label can improve the performance of feature selection for the mutual information criterion under a limited label budget. We studied the challenges involved in designing such an algorithm, and provided AFS, which improves the quality of the selected features over baselines. In future work, we will study adaptations of the suggested approach for other quality measures, such as the Gini index, and variants of the mutual information criterion that take feature correlations into account.

\bibliography{references}

\appendix

\onecolumn

%\tableofcontents

%\clearpage

\section{Experiments: Single-feature estimation}\label{app:single}
Tables \ref{uniform50}---\ref{adult500} provide the full experiment results for the single-feature experiments. For each experiment, we report the average score and the confidence interval for each of the tested algorithms. Winners (as defined in Section 6) are marked in boldface. If an algorithm is the only winner in the table row, then it is a clear winner.

  \captionof{table}{Single-feature synthetic experiments; Uniform $q_v$; Budget of 50 labels}
  \label{uniform50}
  \begin{center}
	\resizebox{\textwidth}{!}{\input{tables/uniform/uniformtable50.tex}}
	
  \end{center}

\clearpage

  \captionof{table}{Single-feature synthetic experiments; Uniform $q_v$; Budget of 100 labels}
  \begin{center}
	\resizebox{\textwidth}{!}{\input{tables/uniform/uniformtable100.tex}}
	
      \end{center}
      
      \label{uniform100}

  \captionof{table}{Single-feature synthetic experiments; Uniform $q_v$; Budget of 300 labels}
  \begin{center}
	\resizebox{\textwidth}{!}{\input{tables/uniform/uniformtable300.tex}}

\end{center}

\label{uniform300}

  \captionof{table}{Single-feature synthetic experiments; Uniform $q_v$; Budget of 500 labels}
  \begin{center}
	\resizebox{\textwidth}{!}{\input{tables/uniform/uniformtable500.tex}}
	
  \end{center}
  
  	\label{uniform500}

        \clearpage
  \captionof{table}{Single-feature synthetic experiments; Fixed $q_v$; Budget of 50 labels; Part 1}
  \begin{center}
	\resizebox{\textwidth}{!}{\input{tables/min_max_arm/min_max_armstable50.tex}}
	
  \end{center}
  
  \label{mma50a}

  \clearpage
  \captionof{table}{Single-feature synthetic experiments; Fixed $q_v$; Budget of 50 labels; Part 2}
  \begin{center}
    \resizebox{\textwidth}{!}{\input{tables/min_max_arm/min_max_armstable50_1.tex}}
    
  \end{center}
  
  \label{mma50b}

\clearpage

  \captionof{table}{Single-feature synthetic experiments; Fixed $q_v$; Budget of 100 labels; Part 1}
  \begin{center}
	\resizebox{\textwidth}{!}{\input{tables/min_max_arm/min_max_armstable100.tex}}
	
      \end{center}
      
      \label{mma100a}

      \clearpage
  \captionof{table}{Single-feature synthetic experiments; Fixed $q_v$; Budget of 100 labels; Part 2}
  \begin{center}
    \resizebox{\textwidth}{!}{\input{tables/min_max_arm/min_max_armstable100_1.tex}}
    	\label{mma100b}
      \end{center}

      \clearpage
  \captionof{table}{Single-feature synthetic experiments; Fixed $q_v$; Budget of 300 labels; Part 1}
  \begin{center}
    \resizebox{\textwidth}{!}{\input{tables/min_max_arm/min_max_armstable300.tex}}
    \label{mma300a}
  \end{center}

\clearpage
  \captionof{table}{Single-feature synthetic experiments; Fixed $q_v$; Budget of 300 labels; Part 2}
  \begin{center}
    \resizebox{\textwidth}{!}{\input{tables/min_max_arm/min_max_armstable300_1.tex}}
    \label{mma300b}
      \end{center}

\clearpage

  \captionof{table}{Single-feature synthetic experiments; Fixed $q_v$; Budget of 500 labels; Part 1}
  \begin{center}
    \resizebox{\textwidth}{!}{\input{tables/min_max_arm/min_max_armstable500.tex}}
    \label{mma500a}
  \end{center}

\clearpage
  \captionof{table}{Single-feature synthetic experiments; Fixed $q_v$; Budget of 500 labels; Part 2}
  \begin{center}
    \resizebox{\textwidth}{!}{\input{tables/min_max_arm/min_max_armstable500_1.tex}}
    \label{mma500b}
	\label{adult_wins}
      \end{center}
      
  \captionof{table}{Single-feature real-date experiments on the \texttt{Adult} data set; Budget of 50 labels}
  \begin{center}
   \resizebox{\textwidth}{!}{\input{tables/adult/adult_featuretable50.tex}}
	\label{adult50}
      \end{center}

      \clearpage

  \captionof{table}{Single-feature real-date experiments on the \texttt{Adult} data set; Budget of 100 labels}
  \begin{center}
	\resizebox{\textwidth}{!}{\input{tables/adult/adult_featuretable100.tex}}
	\label{adult100}
      \end{center}
      
  \captionof{table}{Single-feature real-date experiments on the \texttt{Adult} data set; Budget of 300 labels}
  \begin{center}
	\resizebox{\textwidth}{!}{\input{tables/adult/adult_featuretable300.tex}}
	\label{adult300}
      \end{center}
      
  \captionof{table}{Single-feature real-date experiments on the \texttt{Adult} data set; Budget of 500 labels}
  \begin{center}
	\resizebox{\textwidth}{!}{\input{tables/adult/adult_featuretable500.tex}}
	\label{adult500}
      \end{center}

  \twocolumn

\newcommand{\myborder}[1]{\fbox{\parbox[c]{0.9\columnwidth}{#1}}}

\section{Experiments: Comparing AFS to the baseline}\label{app:baseline}
In this section, we provide the graphs for all the experiments that compare AFS to the baseline algorithm. The $x$ axis is the label budget, and the
  $y$ axis is the mutual information gap, calculated via $\sum_{j \in F^*} H_j - \sum_{j \in F} H_j$, as
  explained in Section 6. Each graph averages $30$ runs with budgets up to $500$ queries. The shaded areas provide the 95\% confidence intervals. For each experiment, we show the full graph at the top and a zoomed-in version at the bottom, which allows better visibility of the interesting parts of the graph.
  In many cases, the improvement of AFS is more significant in smaller budgets. Note that in the zoomed-in versions, DWUS is often not seen because its error is larger than the maximal value of the $y$ axis.

\subsection{Comparing to baselines: $k=20$}

  The first set of graphs shows the
  experiments for $k=20$. Here, the improvement of AFS compared to the
  baselines is the most pronounced.

\begin{figure}[h]
  \begin{center}
    \myborder{
    \includegraphics[width = 0.4\textwidth]{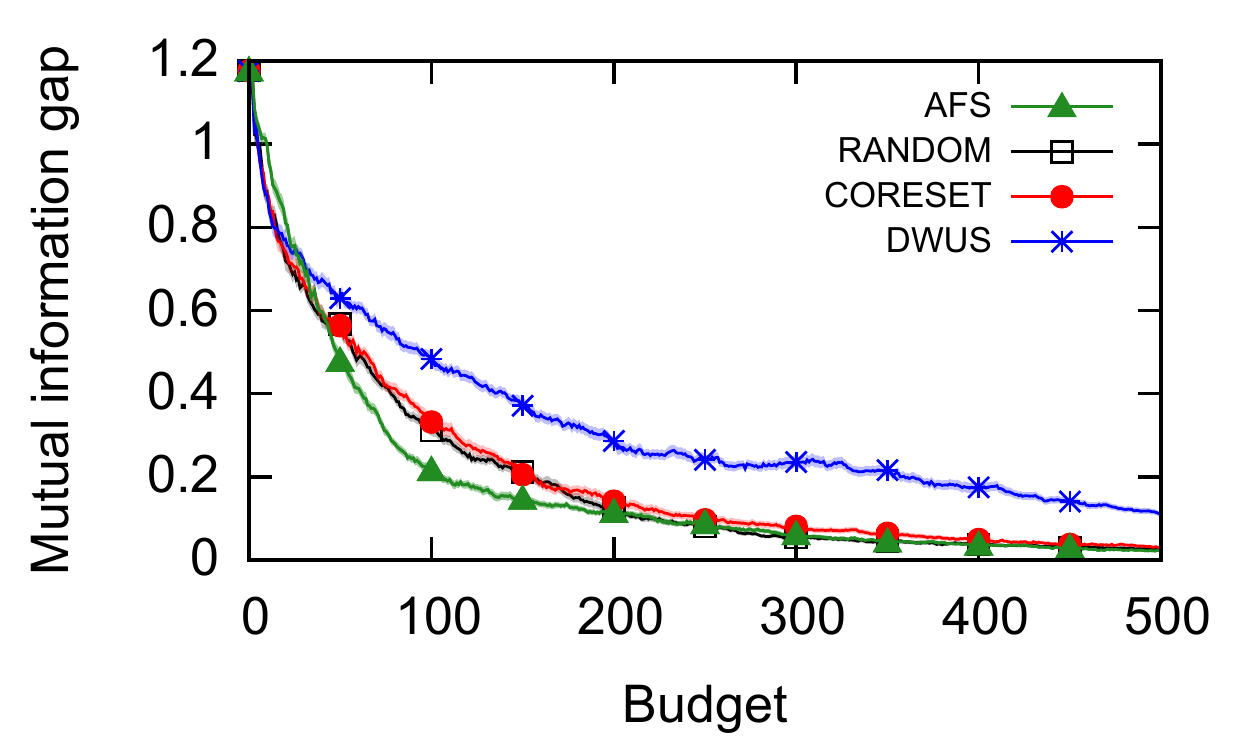} \\
    \includegraphics[width = 0.4\textwidth]{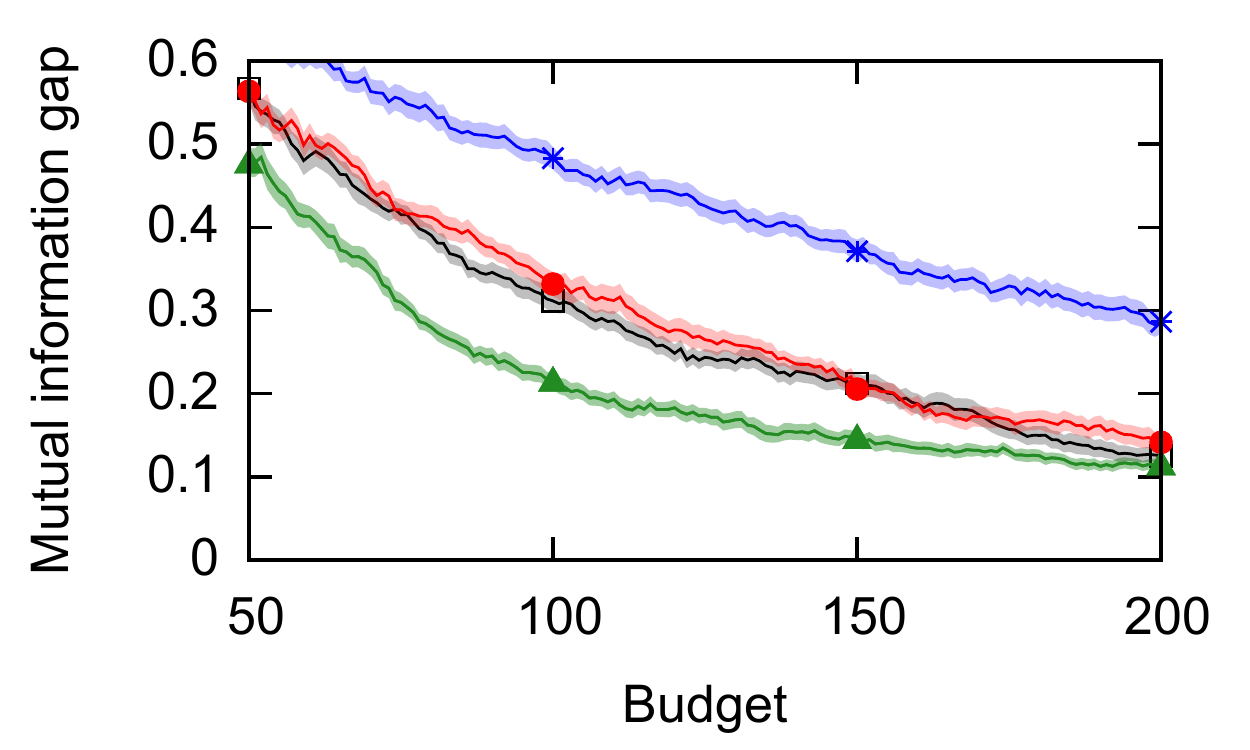}
    }
  \end{center}

\caption{\small AFS vs baseline: BASEHOCK, $k=20$. Top: full experiment. Bottom: Zoom in.}
\end{figure}

\begin{figure}[h]
  \begin{center}
    \myborder{
    \includegraphics[width = 0.4\textwidth]{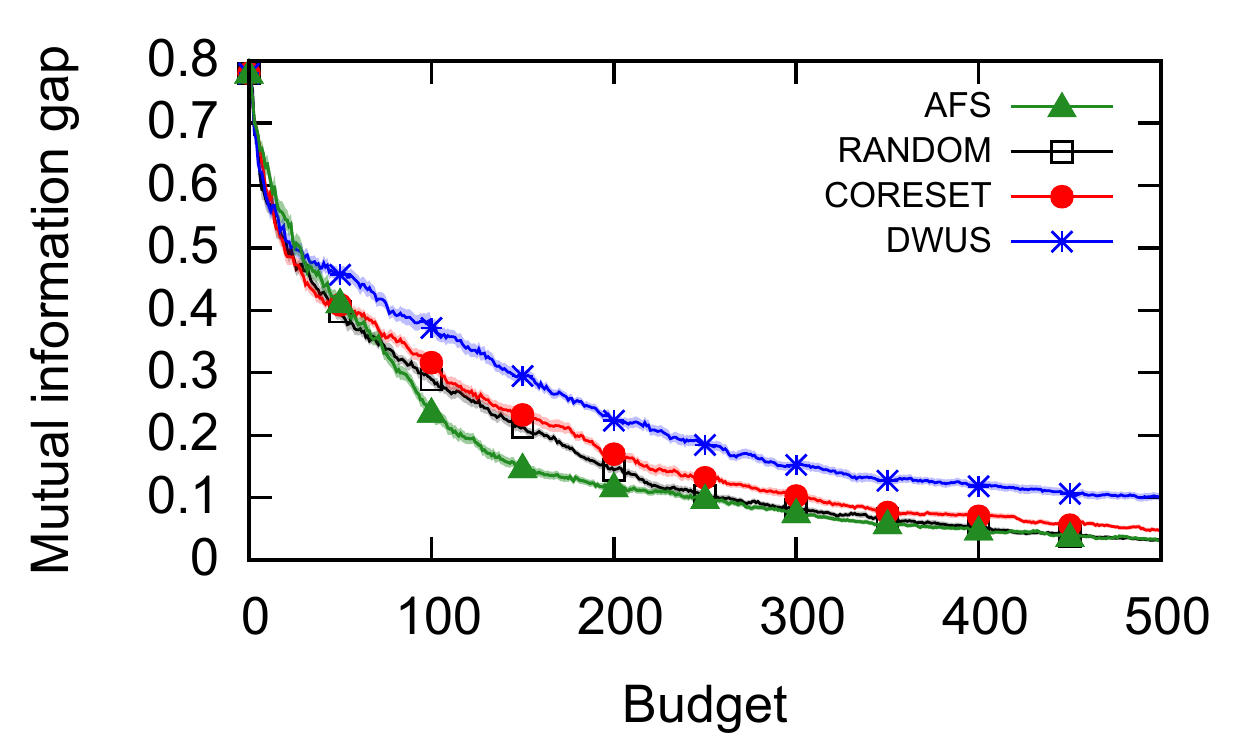} \\
    \includegraphics[width = 0.4\textwidth]{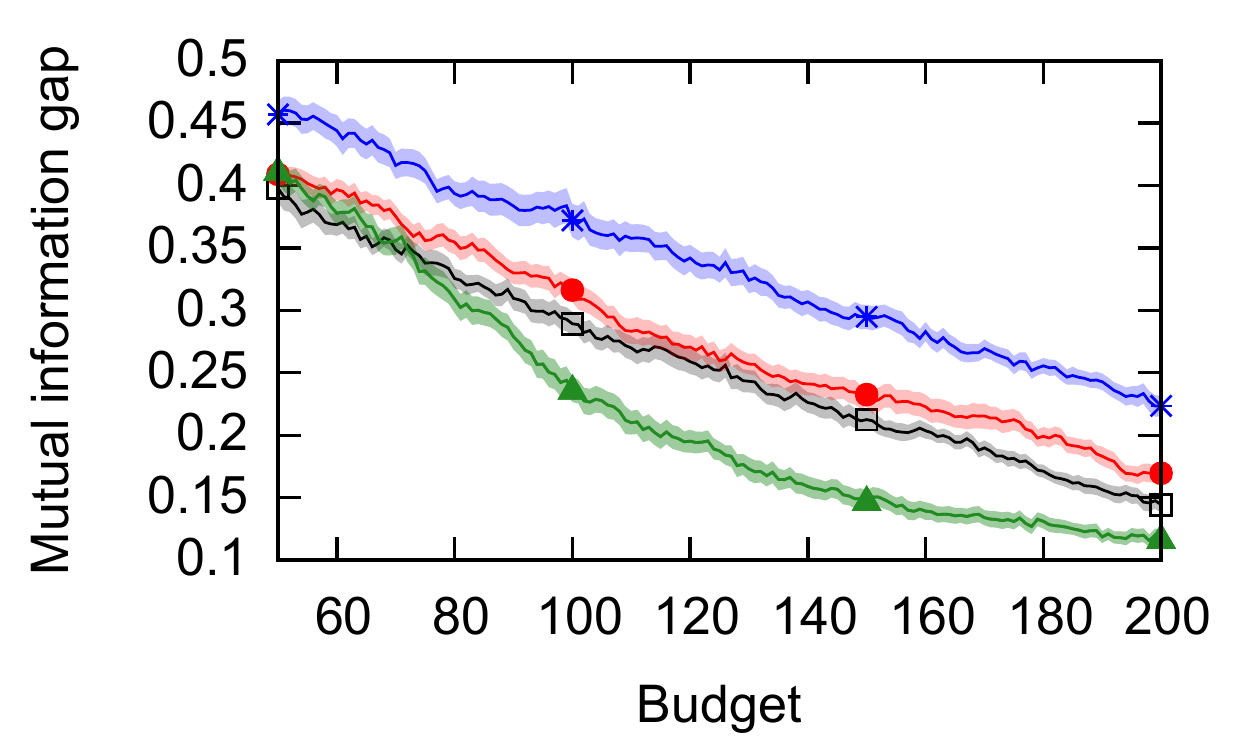}
    }
  \end{center}

\caption{\small AFS vs baseline: PCMAC, $k=20$. Top: full experiment. Bottom: Zoom in.}
\end{figure}

\begin{figure}[h]
  \begin{center}
    \myborder{
    \includegraphics[width = 0.4\textwidth]{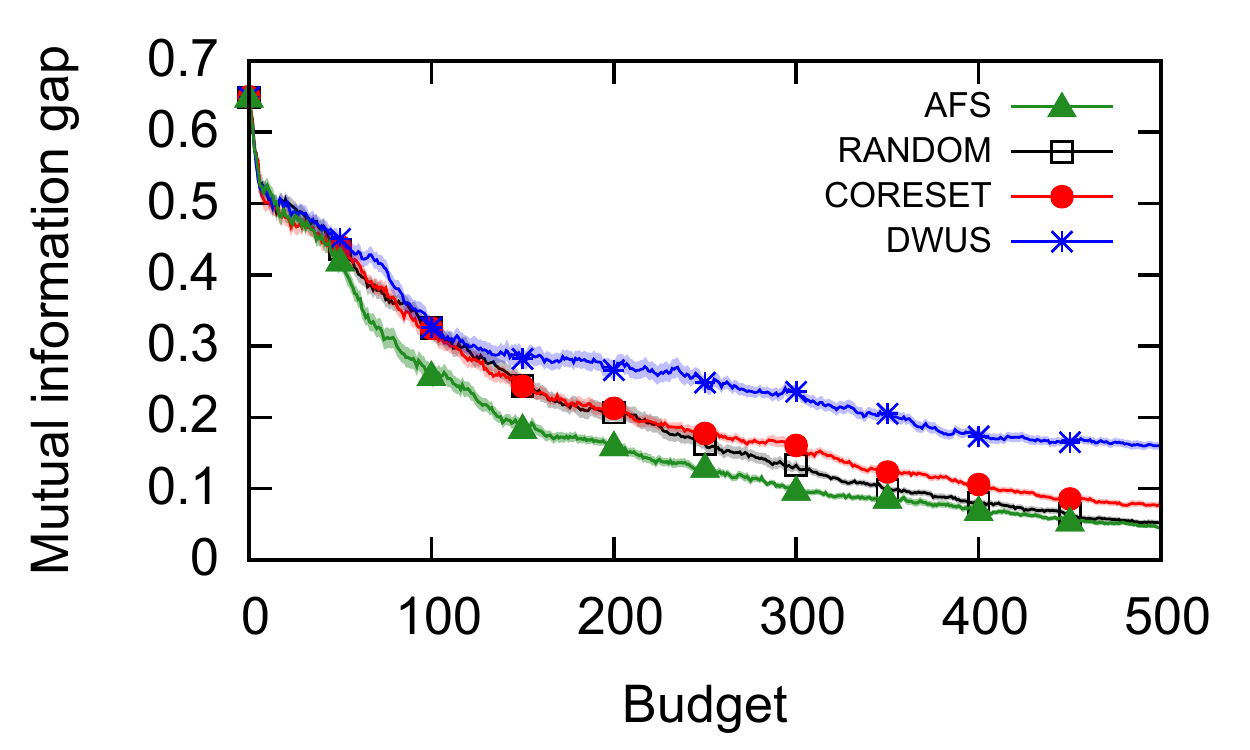} \\
    \includegraphics[width = 0.4\textwidth]{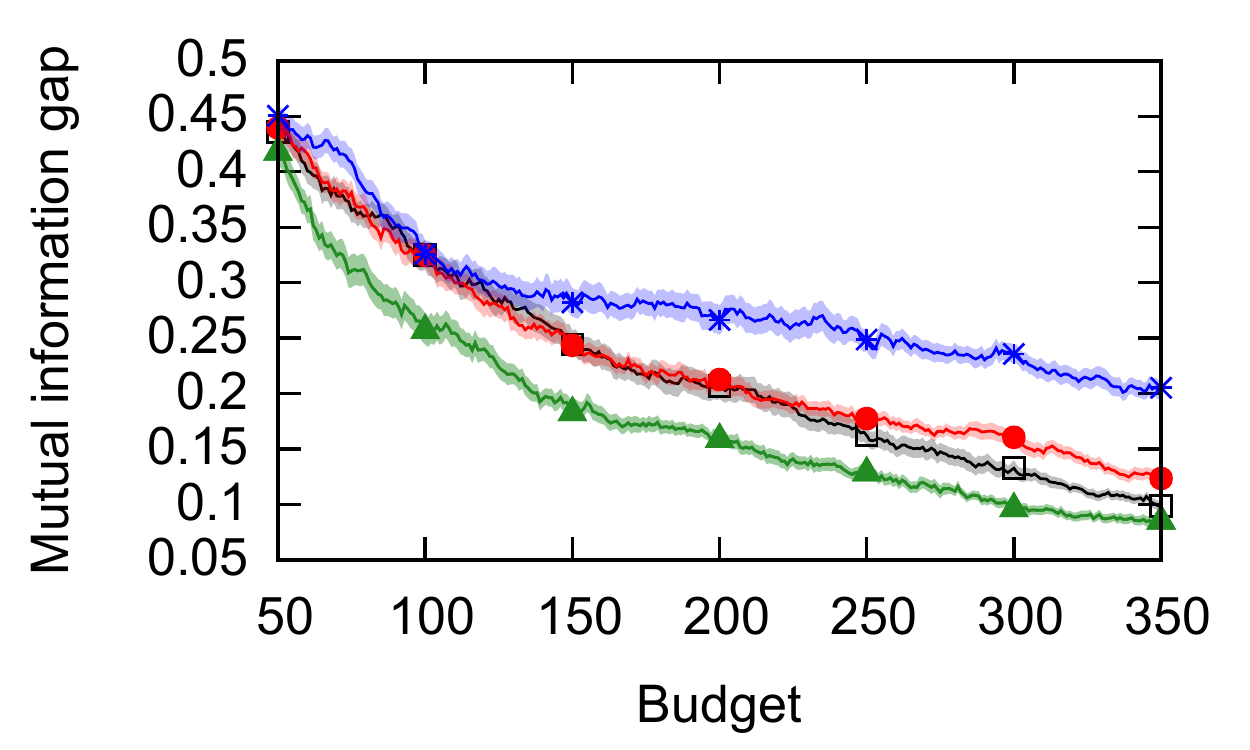}
    }
  \end{center}

\caption{\small AFS vs baseline: RELATHE, $k=20$. Top: full experiment. Bottom: Zoom in.}
\end{figure}

\begin{figure}[h]
  \begin{center}
    \myborder{
    \includegraphics[width = 0.4\textwidth]{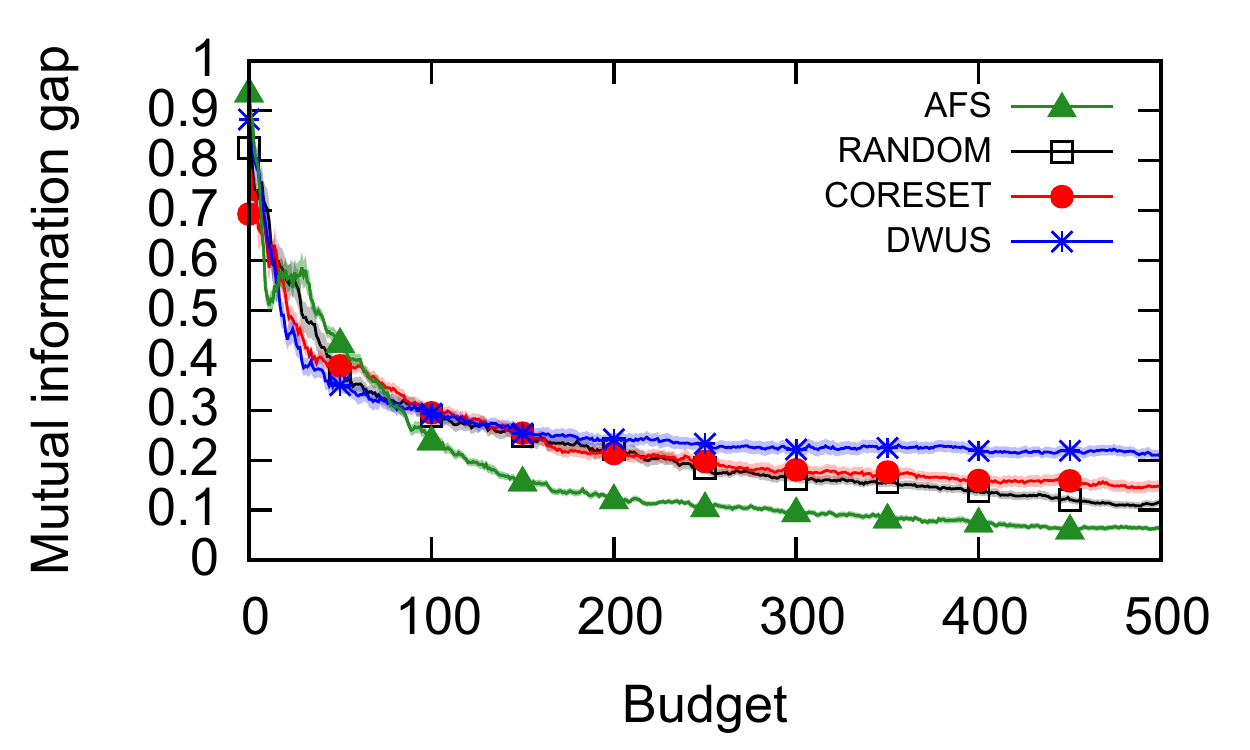} \\
    \includegraphics[width = 0.4\textwidth]{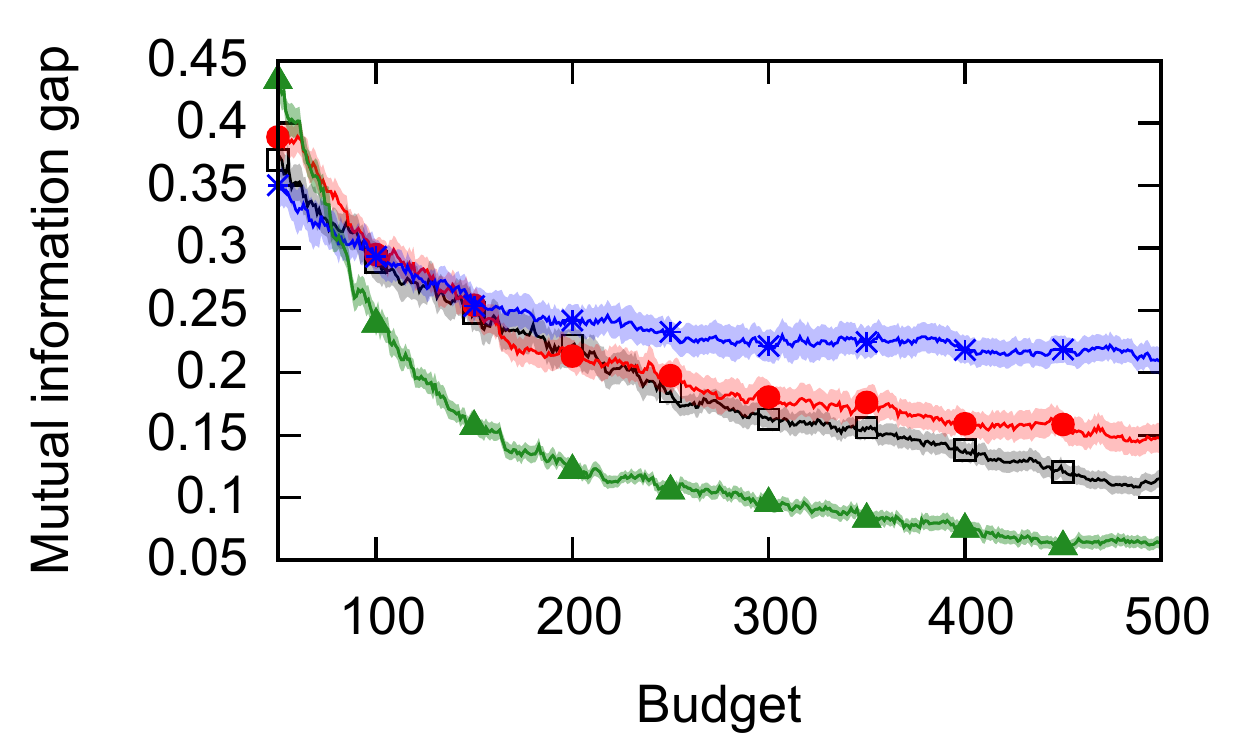}
    }
  \end{center}

\caption{\small AFS vs baseline: MUSK, $k=20$. Top: full experiment. Bottom: Zoom in.}
\end{figure}

\begin{figure}[h]
  \begin{center}
    \myborder{
    \includegraphics[width = 0.4\textwidth]{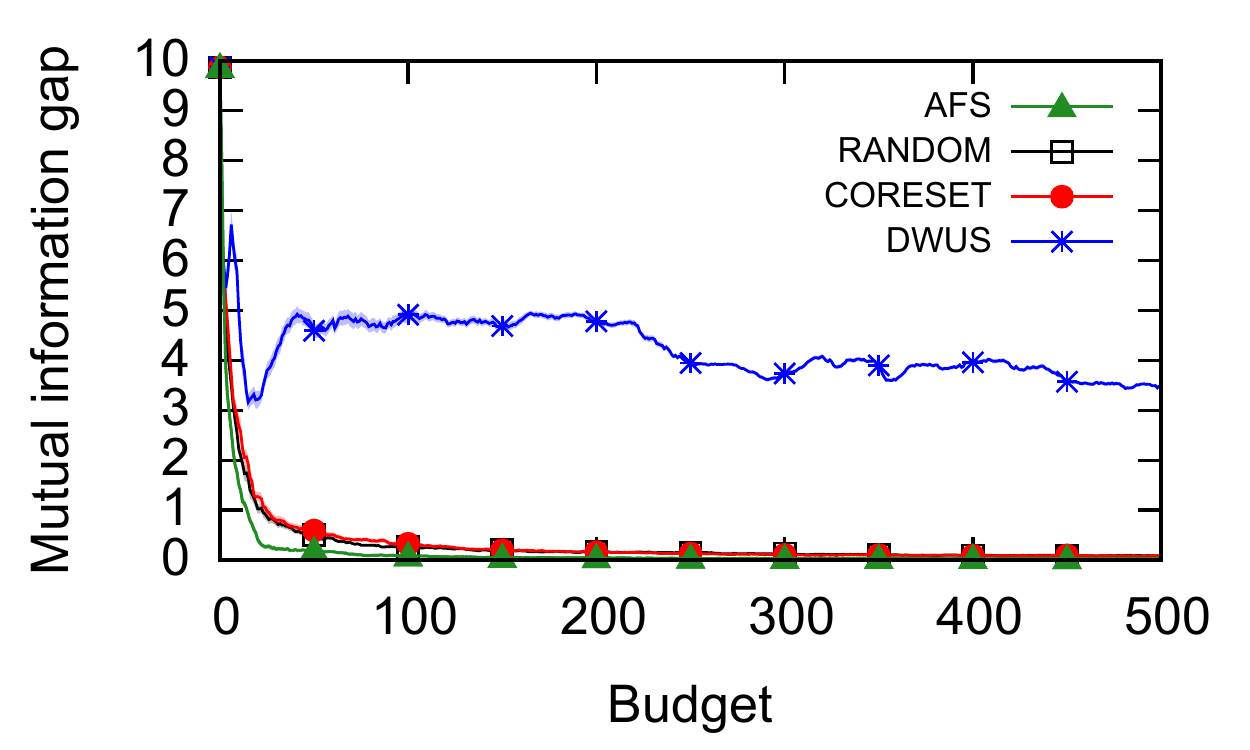} \\
    \includegraphics[width = 0.4\textwidth]{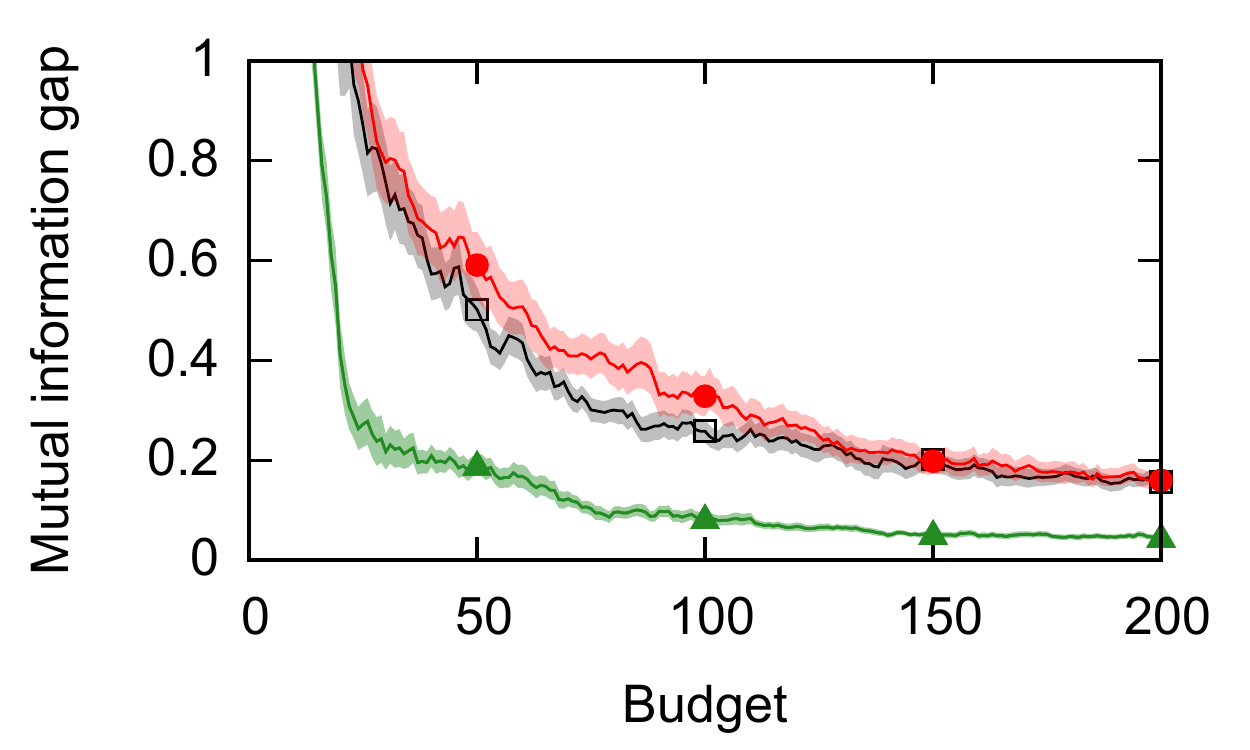}
    }
  \end{center}

\caption{\small AFS vs baseline: MNIST: 0 vs 1, $k=20$. Top: full experiment. Bottom: Zoom in.}
\end{figure}

\begin{figure}[h]
  \begin{center}
    \myborder{
    \includegraphics[width = 0.4\textwidth]{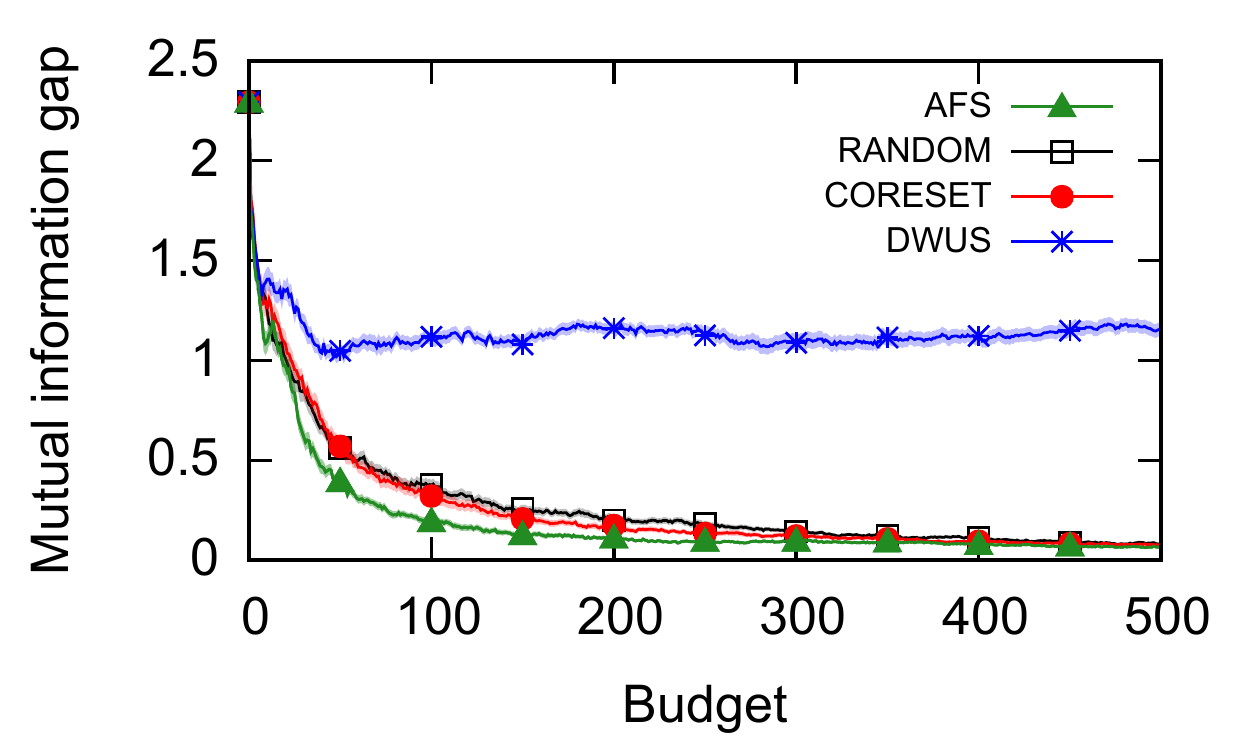} \\
    \includegraphics[width = 0.4\textwidth]{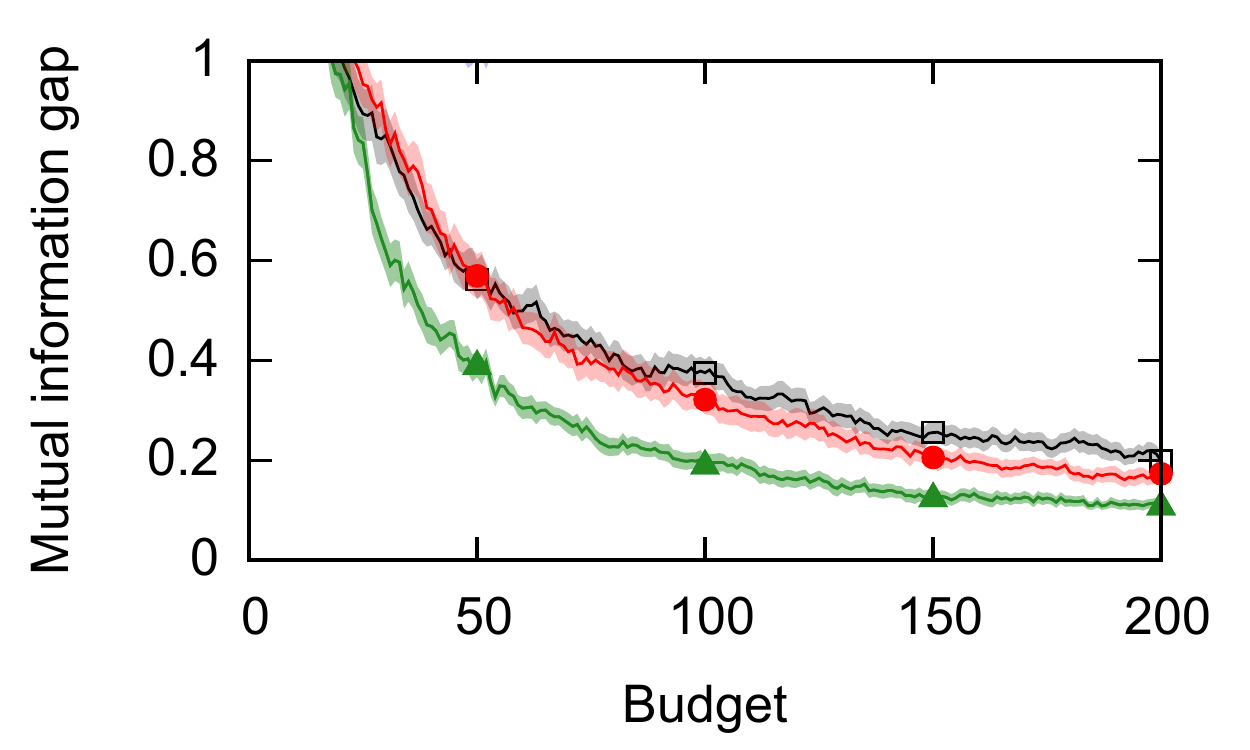}
    }
  \end{center}

\caption{\small AFS vs baseline: MNIST: 3 vs 5, $k=20$. Top: full experiment. Bottom: Zoom in.}
\end{figure}

\begin{figure}[h]
  \begin{center}
    \myborder{
    \includegraphics[width = 0.4\textwidth]{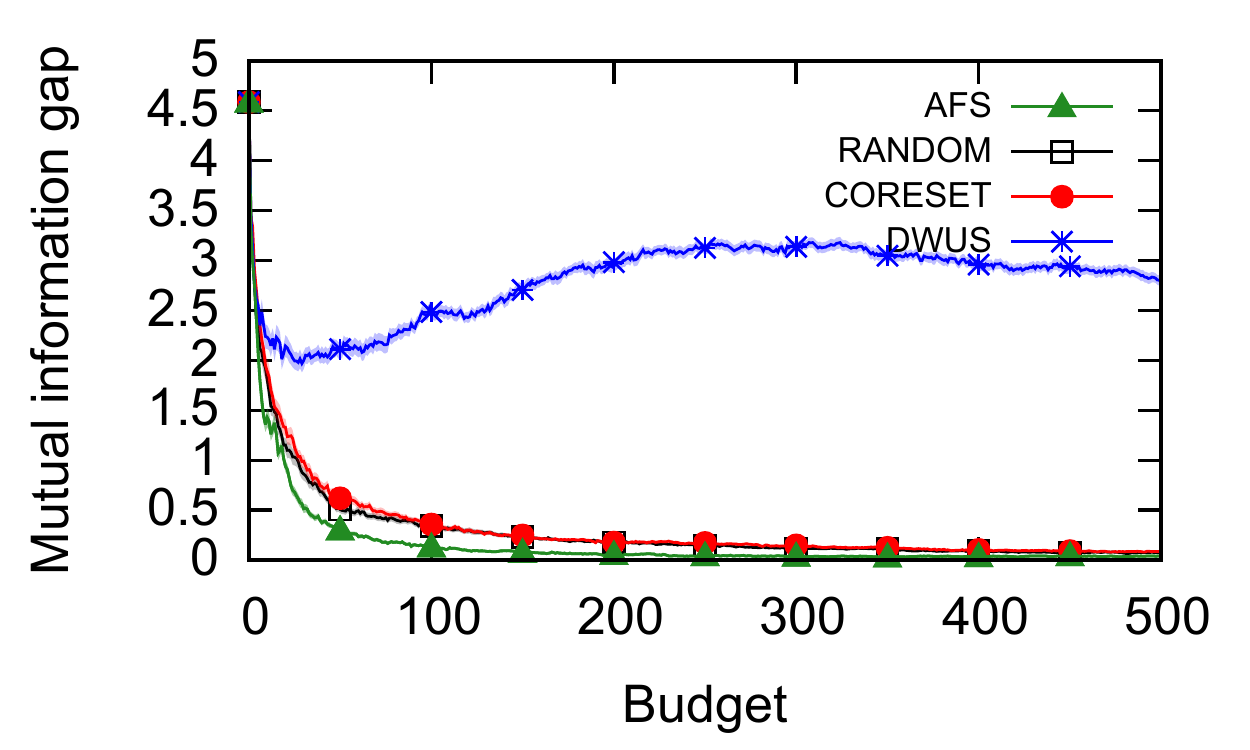} \\
    \includegraphics[width = 0.4\textwidth]{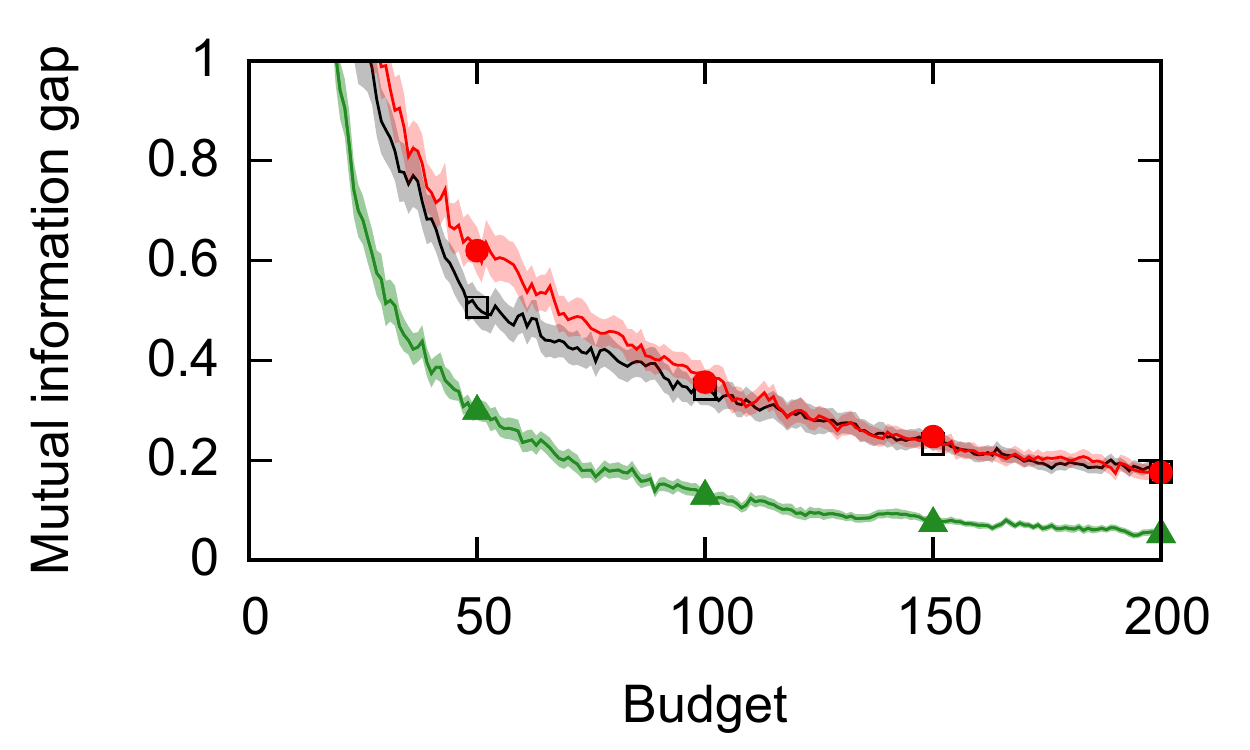}
    }
  \end{center}

\caption{\small AFS vs baseline: MNIST: 4 vs 6, $k=20$. Top: full experiment. Bottom: Zoom in.}
\end{figure}

\clearpage

\subsection{Comparing to baselines: $k=10$}

  The first set of graphs shows the
  experiments for $k=10$. 

\begin{figure}[h]
  \begin{center}
    \myborder{
    \includegraphics[width = 0.4\textwidth]{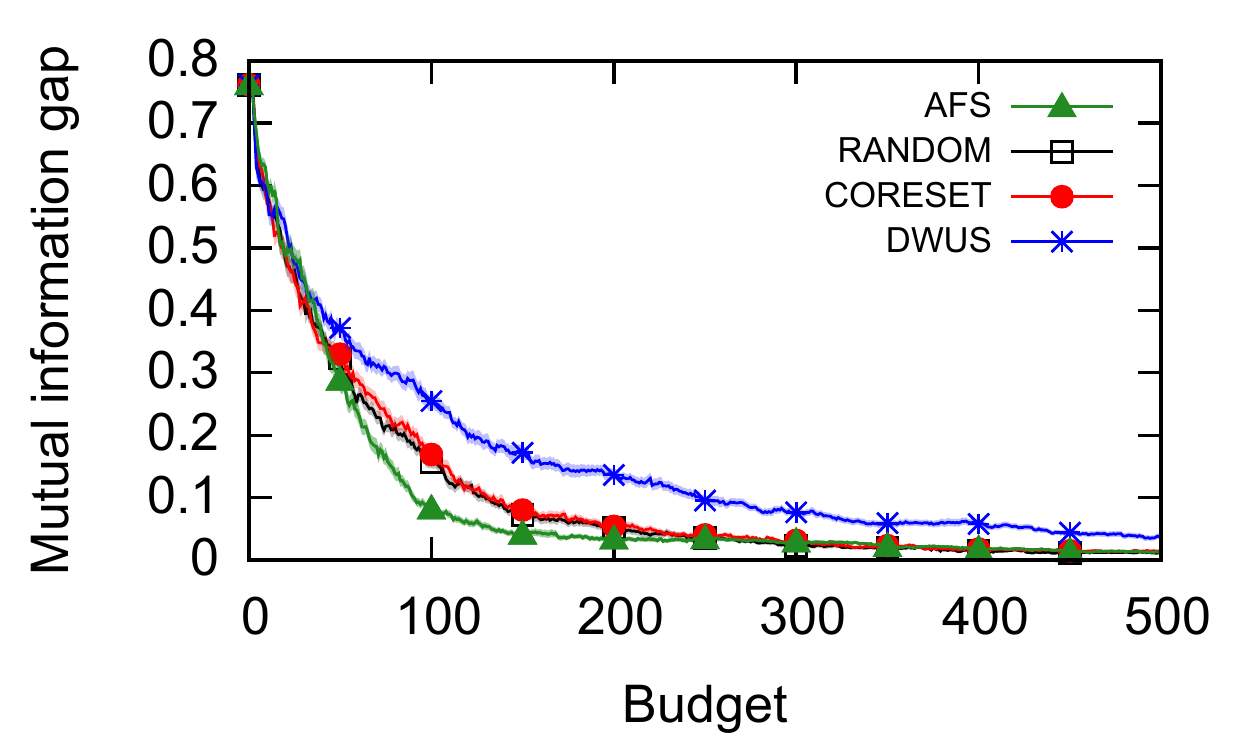} \\
    \includegraphics[width = 0.4\textwidth]{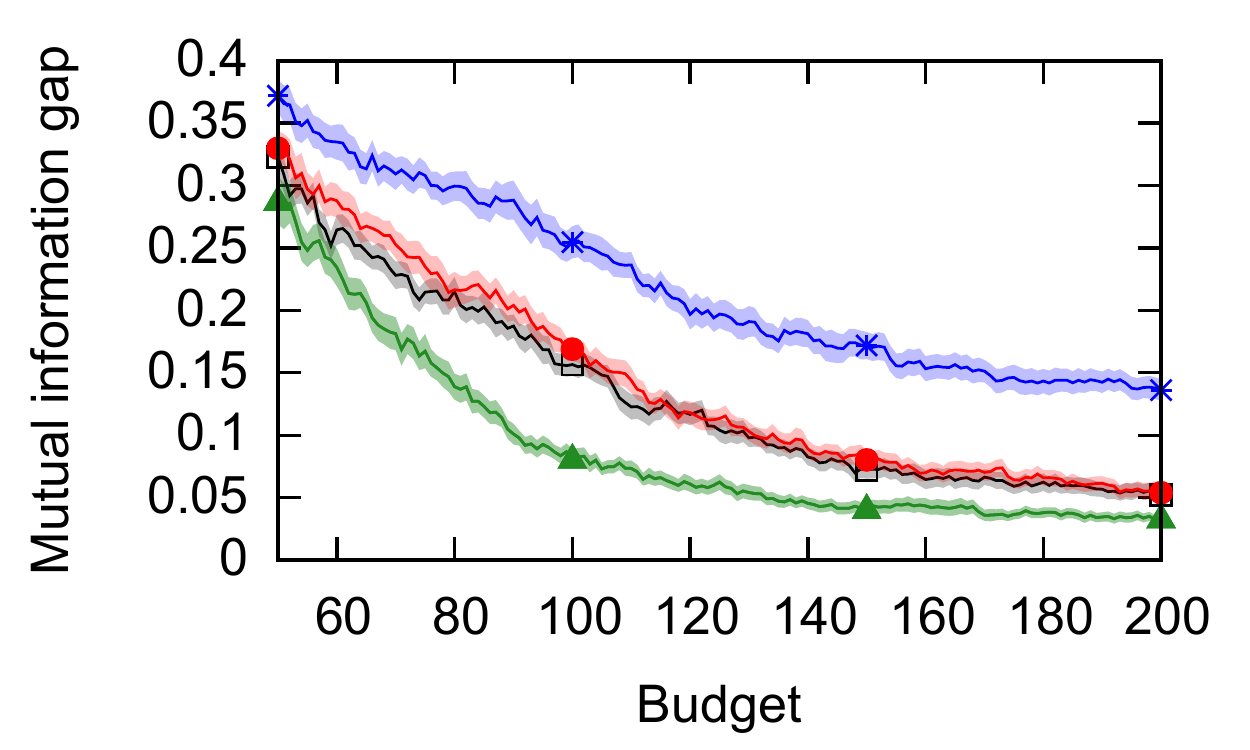}
    }
  \end{center}

\caption{\small AFS vs baseline: BASEHOCK, $k=10$. Top: full experiment. Bottom: Zoom in.}
\end{figure}

\begin{figure}[h]
  \begin{center}
    \myborder{
    \includegraphics[width = 0.4\textwidth]{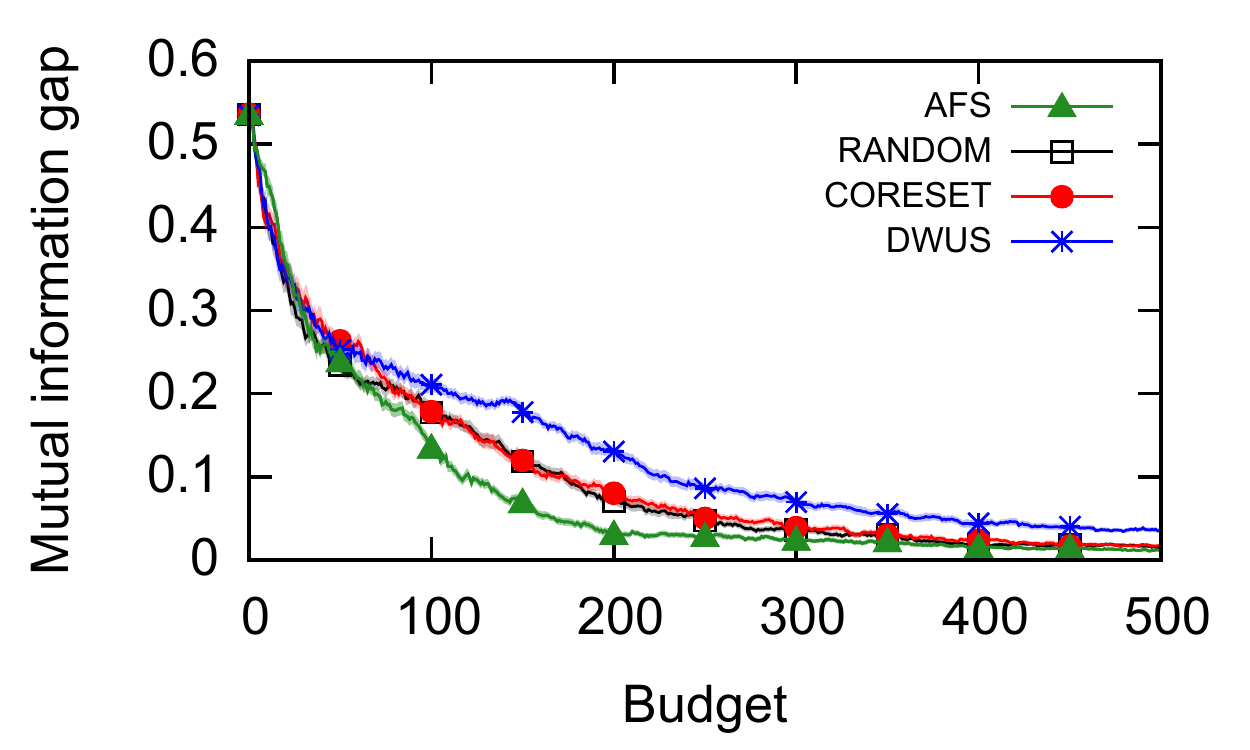} \\
    \includegraphics[width = 0.4\textwidth]{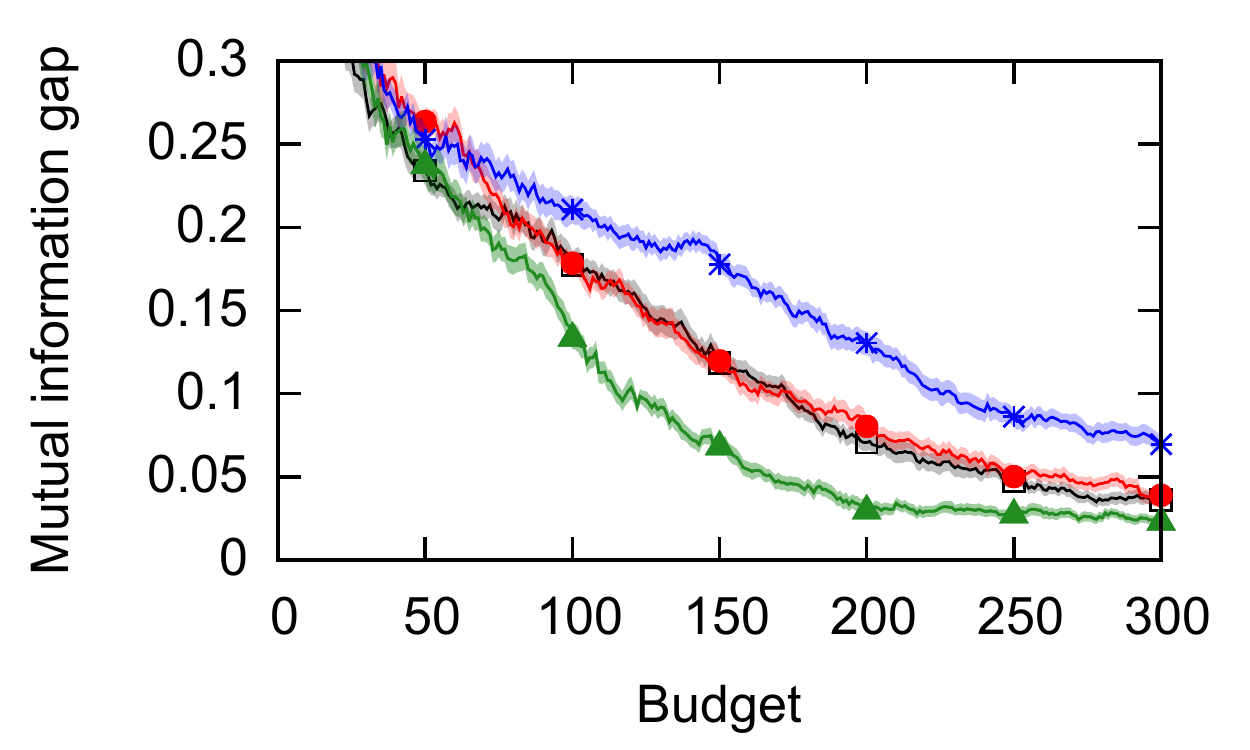}
    }
  \end{center}

\caption{\small AFS vs baseline: PCMAC, $k=10$. Top: full experiment. Bottom: Zoom in.}
\end{figure}

\begin{figure}[h]
  \begin{center}
    \myborder{
    \includegraphics[width = 0.4\textwidth]{graph/baseline/10/RELATHE.pdf} \\
    \includegraphics[width = 0.4\textwidth]{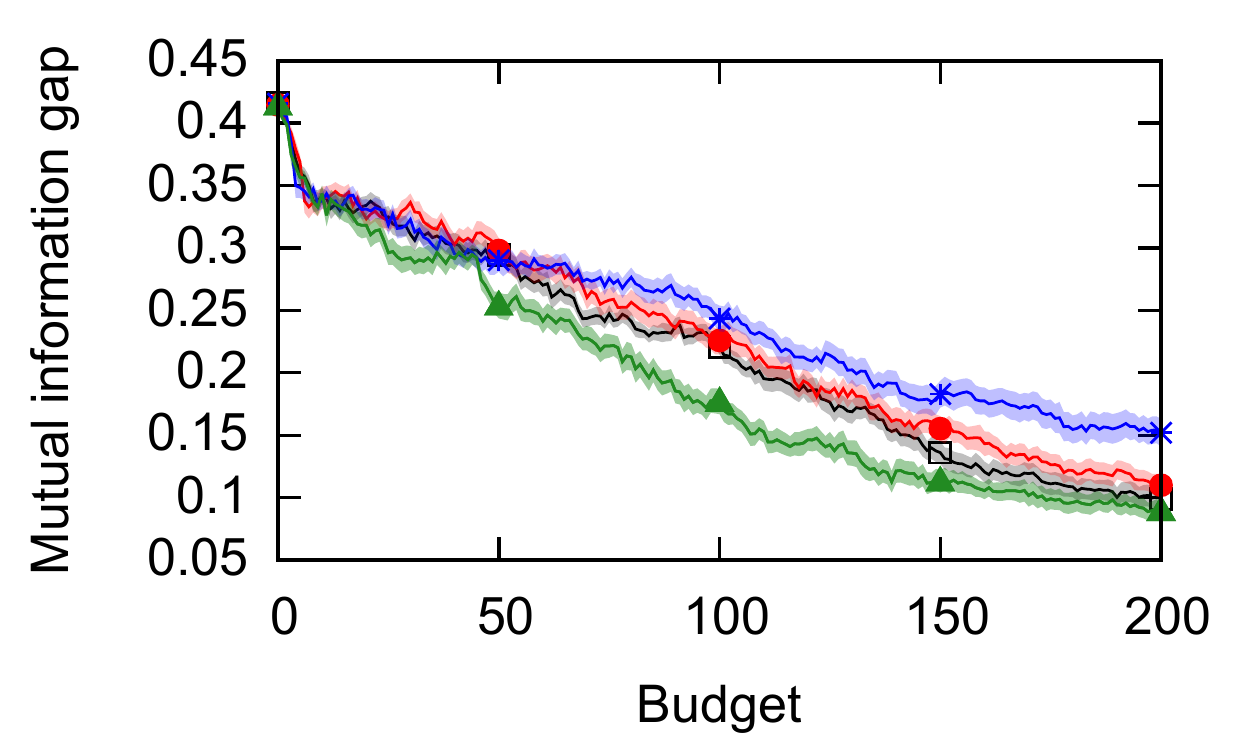}
    }
  \end{center}

\caption{\small AFS vs baseline: RELATHE, $k=10$. Top: full experiment. Bottom: Zoom in.}
\end{figure}

\begin{figure}[h]
  \begin{center}
    \myborder{
    \includegraphics[width = 0.4\textwidth]{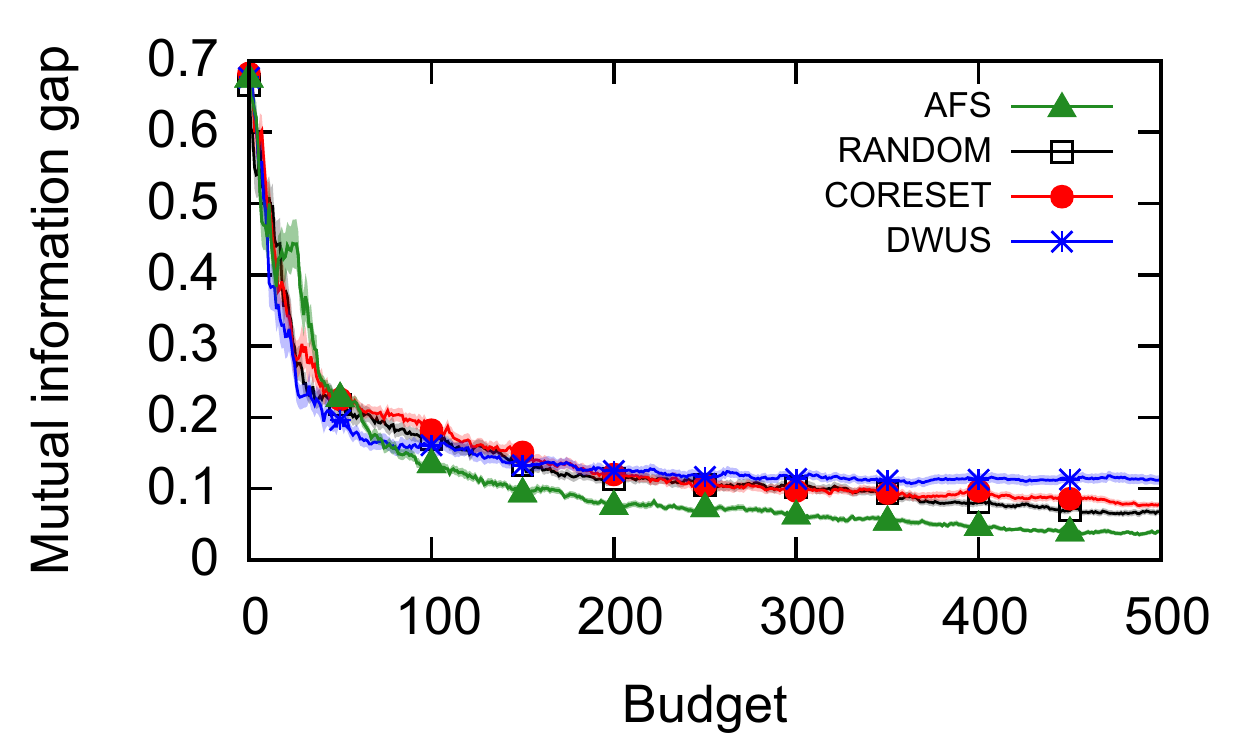} \\
    \includegraphics[width = 0.4\textwidth]{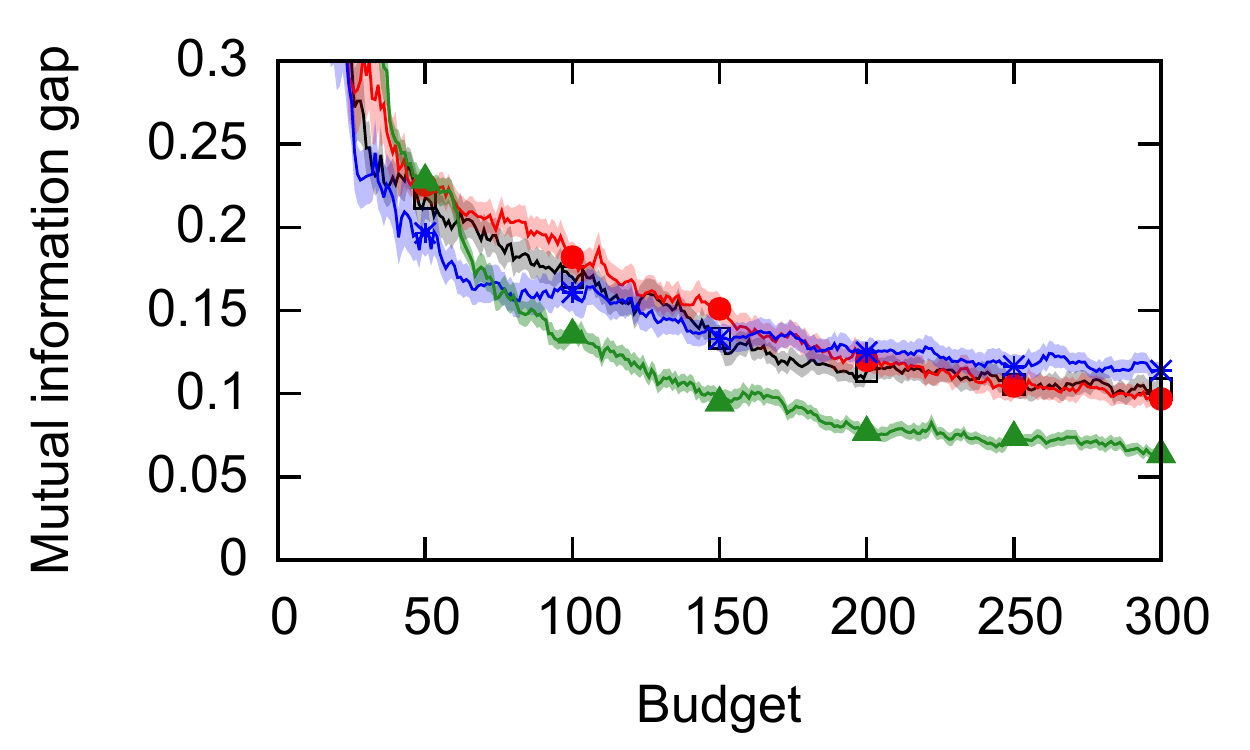}
    }
  \end{center}

\caption{\small AFS vs baseline: MUSK, $k=10$. Top: full experiment. Bottom: Zoom in.}
\end{figure}

\begin{figure}[h]
  \begin{center}
    \myborder{
    \includegraphics[width = 0.4\textwidth]{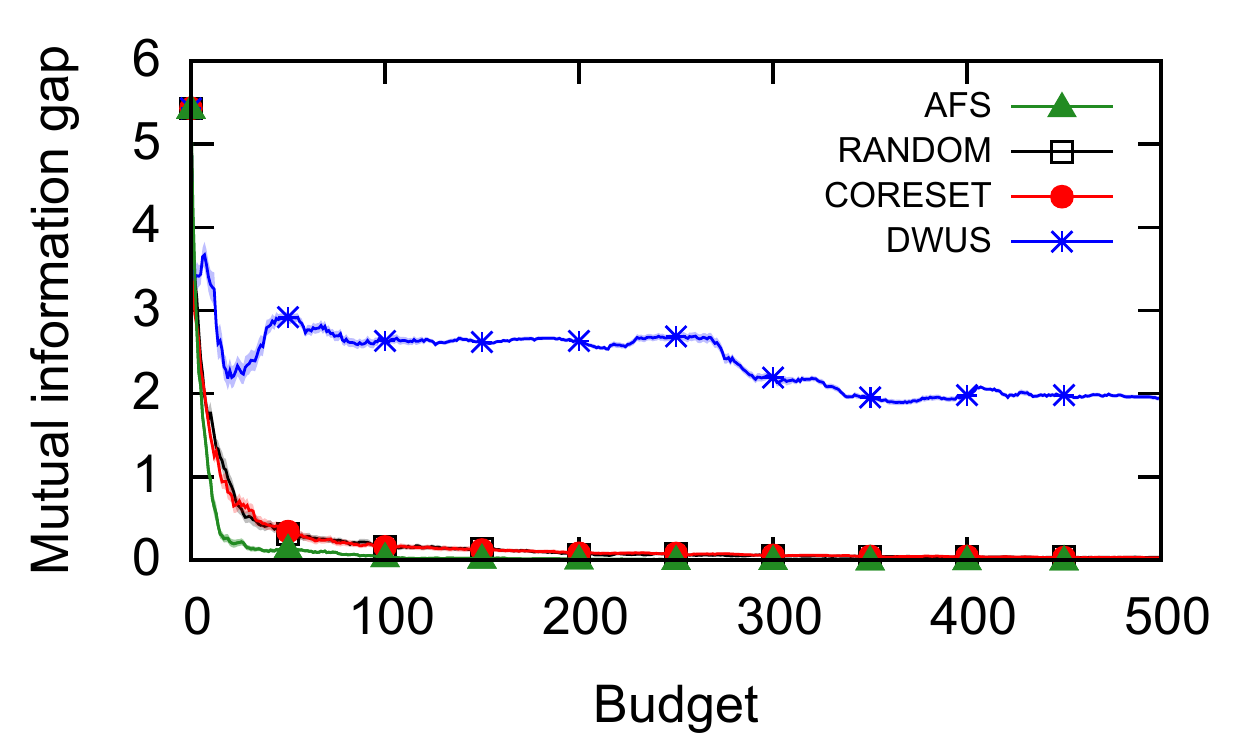} \\
    \includegraphics[width = 0.4\textwidth]{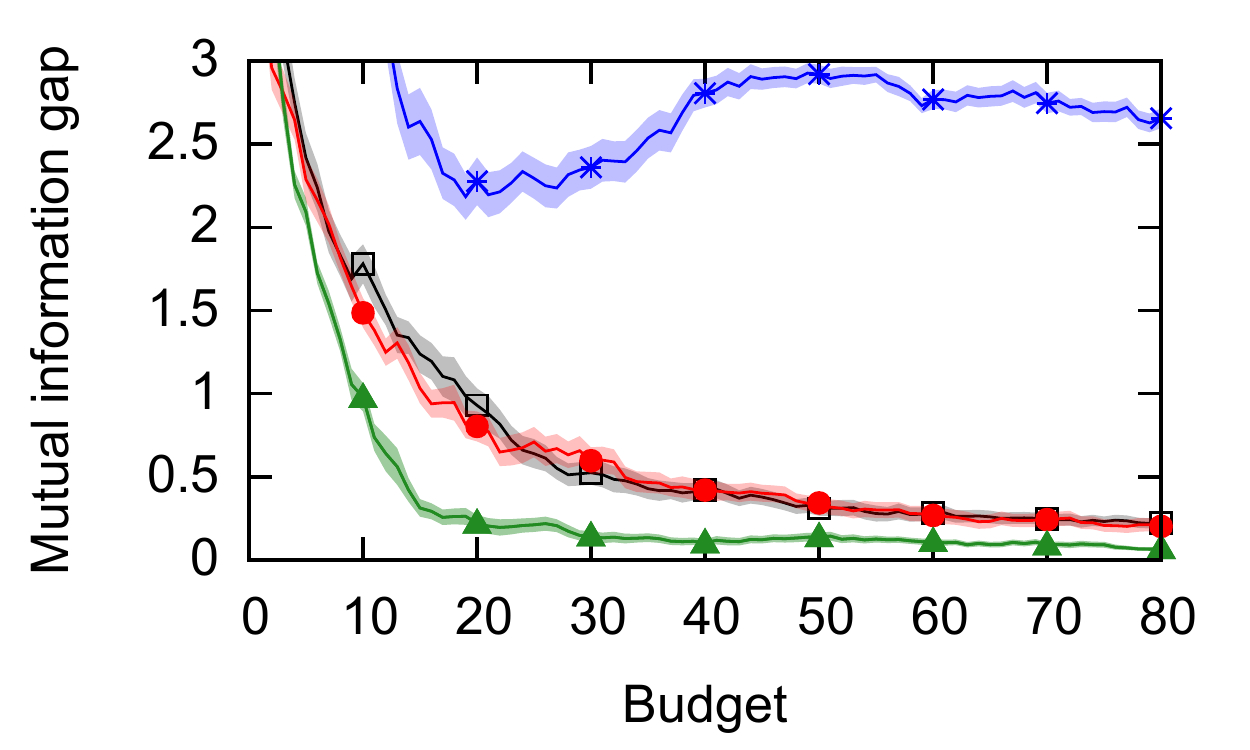}
    }
  \end{center}

\caption{\small AFS vs baseline: MNIST: 0 vs 1, $k=10$. Top: full experiment. Bottom: Zoom in.}
\end{figure}

\begin{figure}[h]
  \begin{center}
    \myborder{
    \includegraphics[width = 0.4\textwidth]{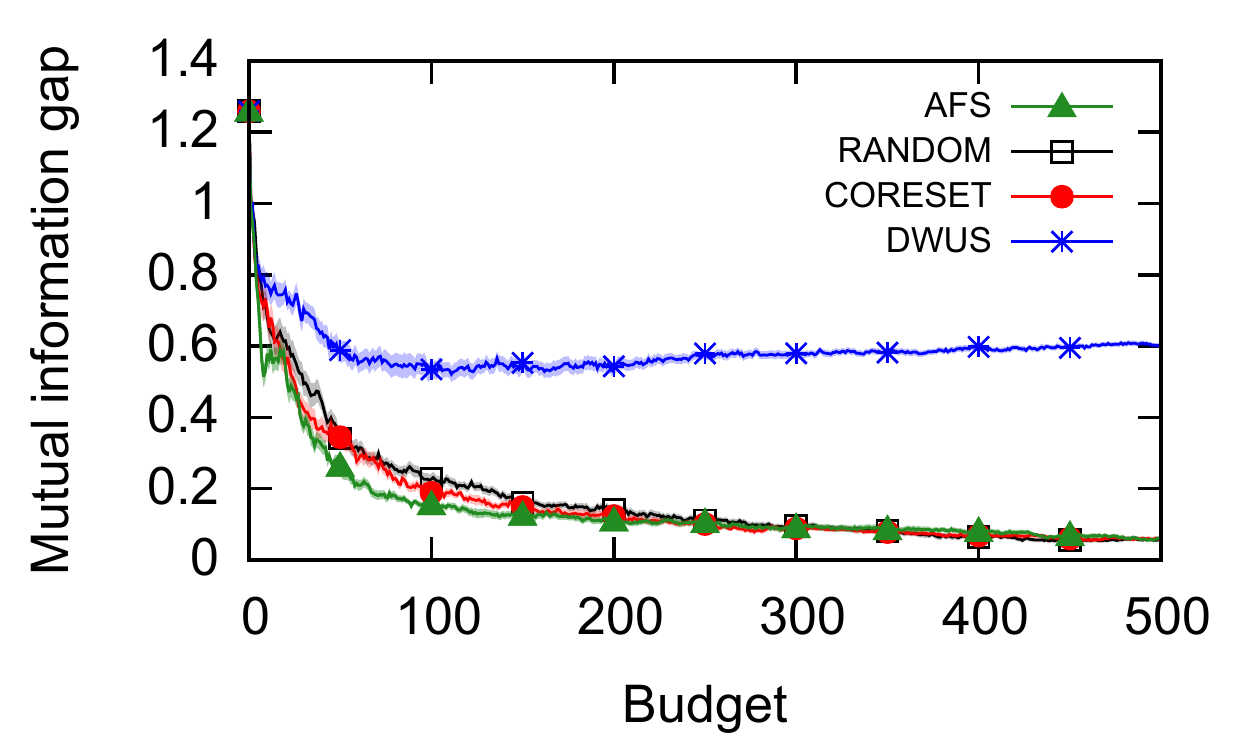} \\
    \includegraphics[width = 0.4\textwidth]{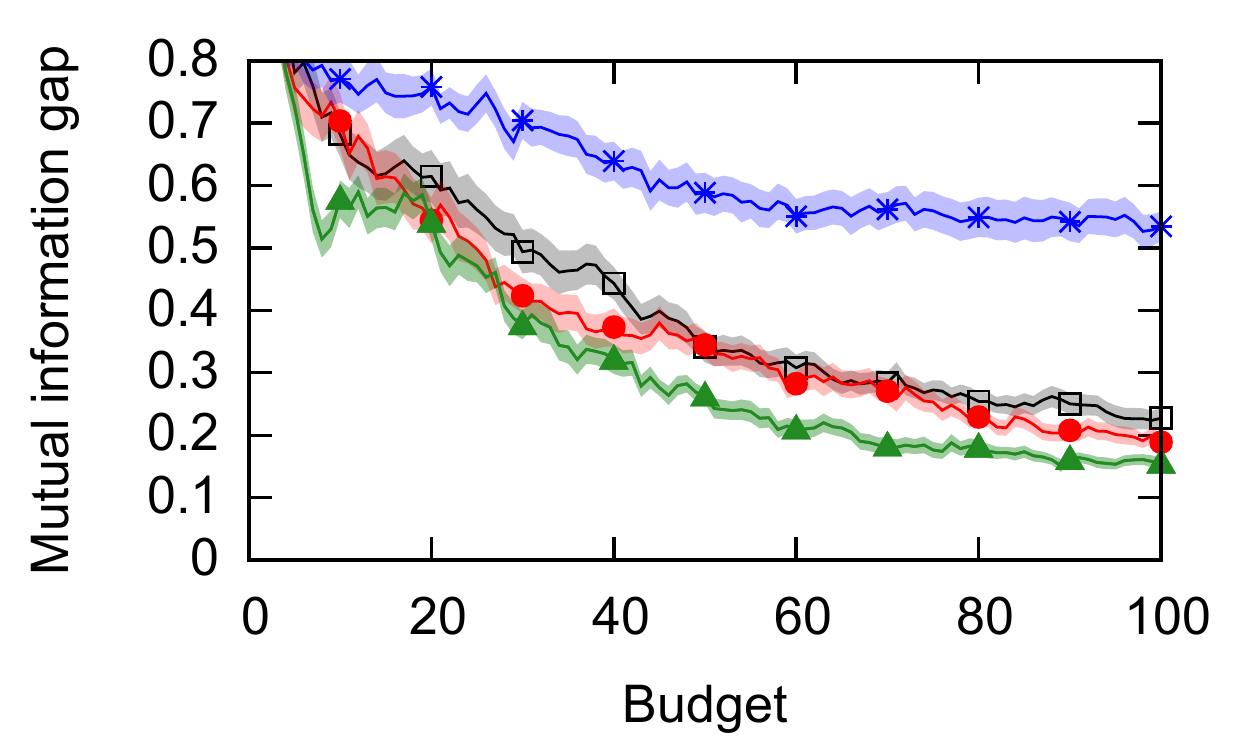}
    }
  \end{center}

\caption{\small AFS vs baseline: MNIST: 3 vs 5, $k=10$. Top: full experiment. Bottom: Zoom in.}
\end{figure}

\begin{figure}[h]
  \begin{center}
    \myborder{
    \includegraphics[width = 0.4\textwidth]{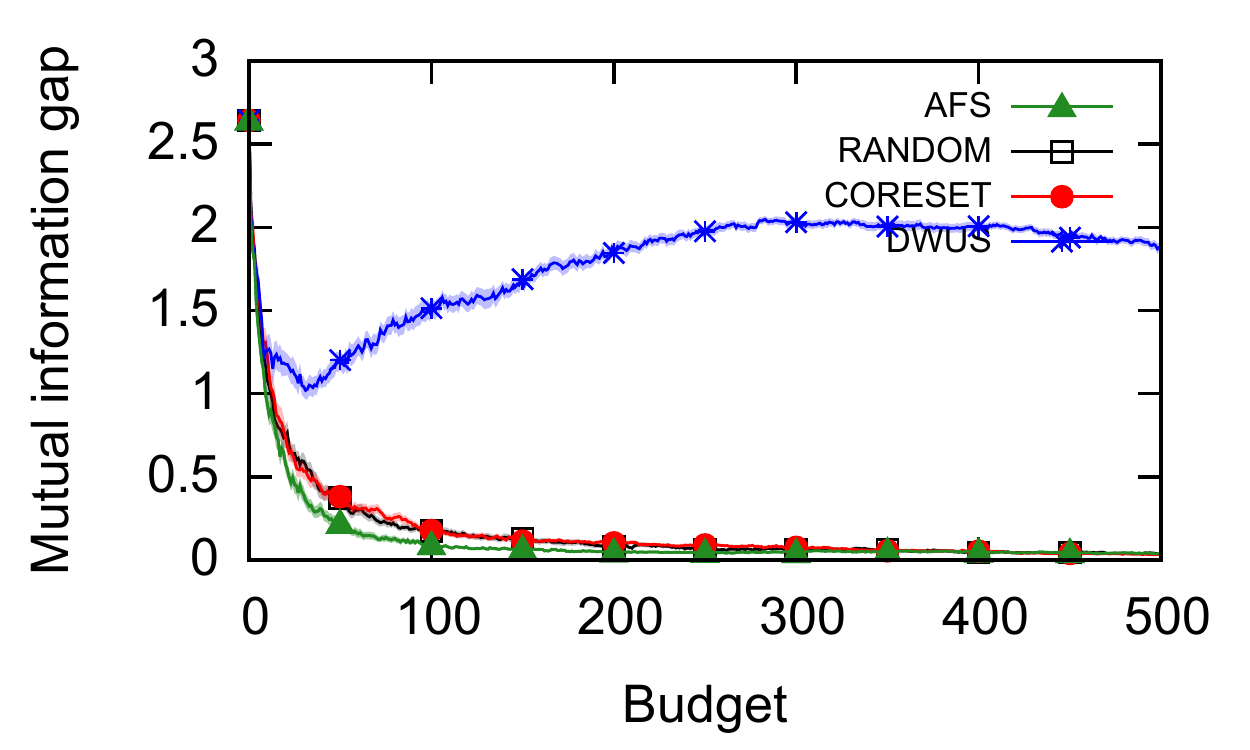} \\
    \includegraphics[width = 0.4\textwidth]{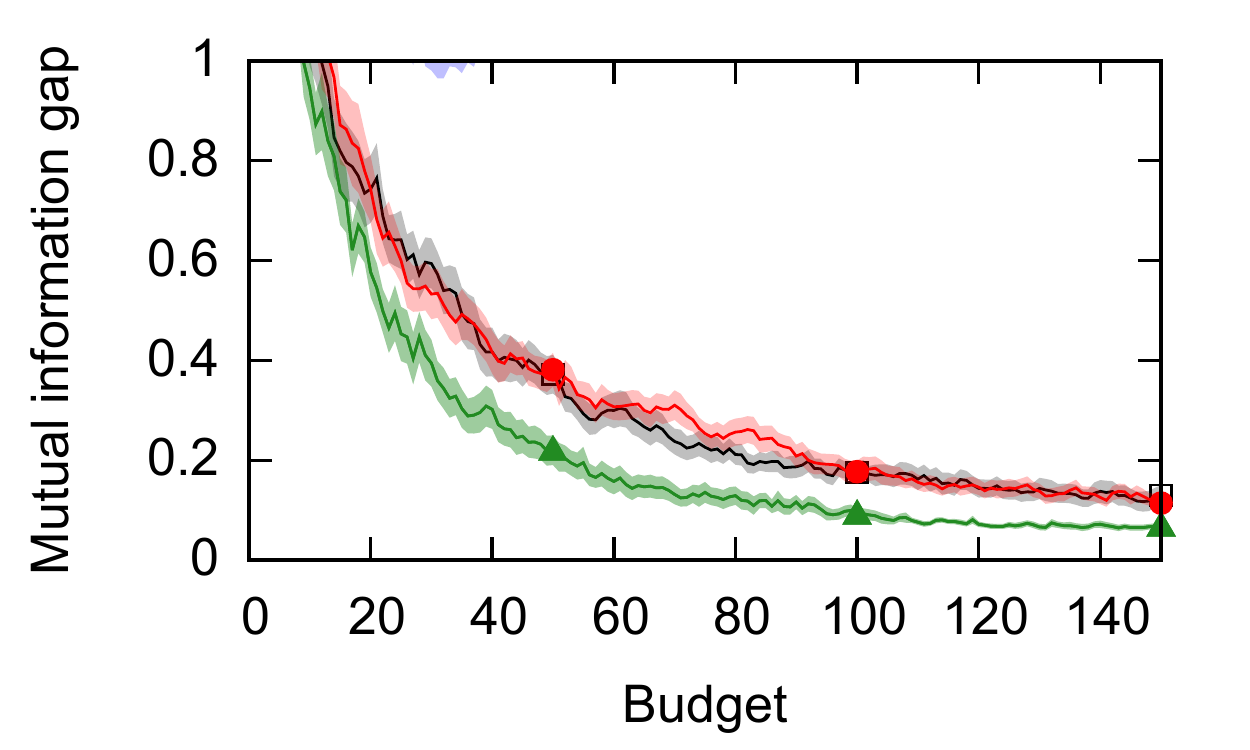}
    }
  \end{center}

\caption{\small AFS vs baseline: MNIST: 4 vs 6, $k=10$. Top: full experiment. Bottom: Zoom in.}
\end{figure}

\clearpage  

\subsection{Comparing to baselines: $k = 5$}

The next set of graphs shows the experiments for $k = 5$. Here too, AFS performs the best compared to the other baselines, although the difference is less pronounced, possibly due to the smaller value of $k$.

\begin{figure}[h]
  \begin{center}
    \myborder{
    \includegraphics[width = 0.4\textwidth]{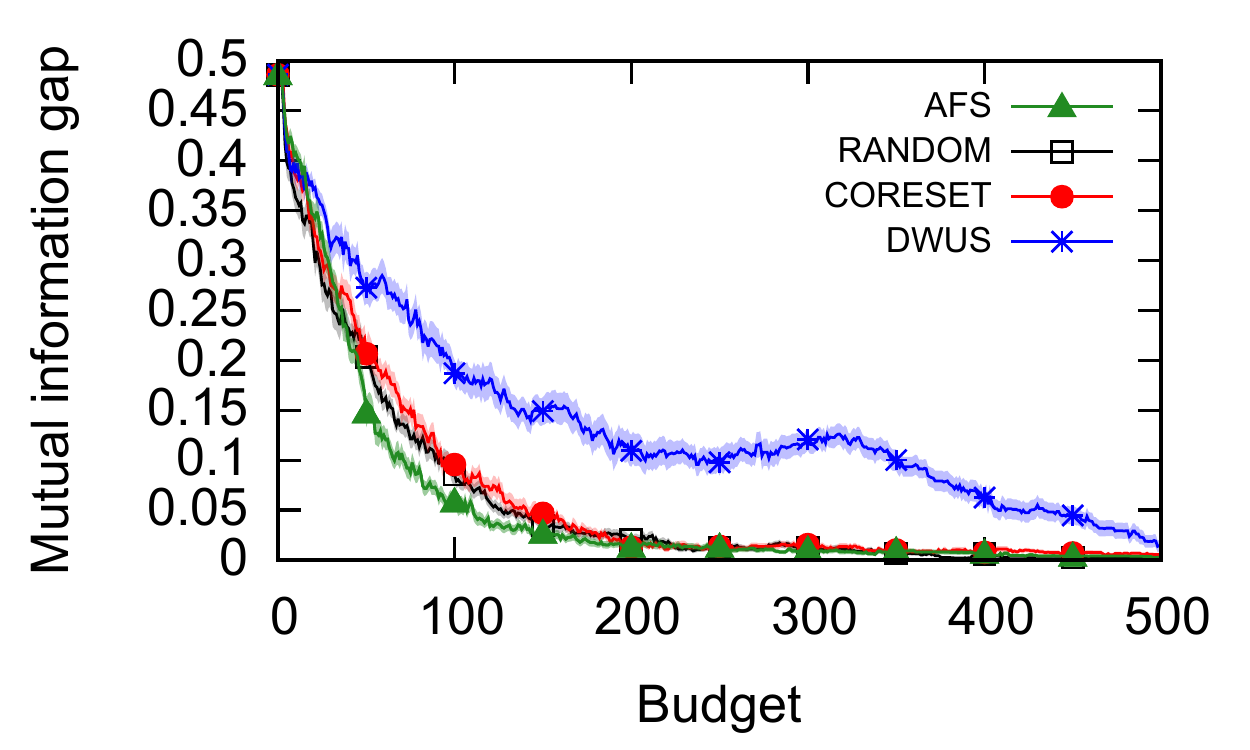} \\
    \includegraphics[width = 0.4\textwidth]{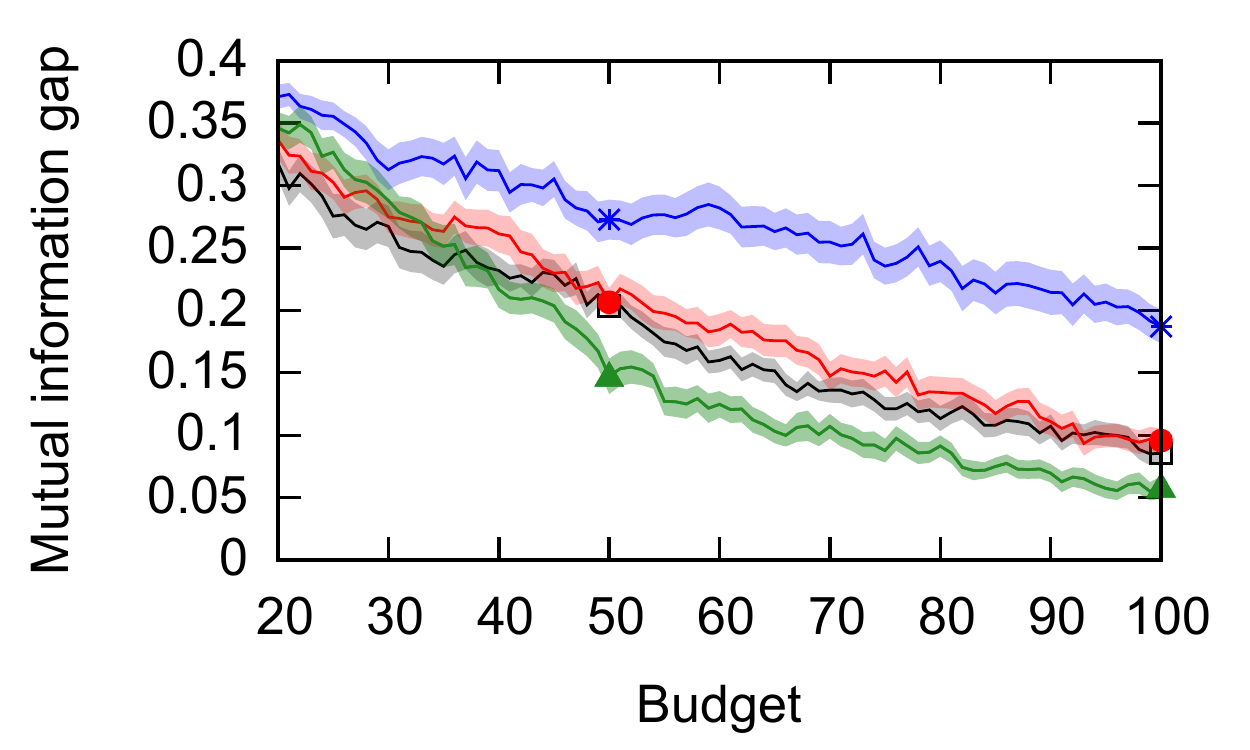}
    }
  \end{center}

\caption{\small AFS vs baseline: BASEHOCK, $k=5$. Top: full experiment. Bottom: Zoom in.}
\end{figure}

\begin{figure}[h]
  \begin{center}
    \myborder{
    \includegraphics[width = 0.4\textwidth]{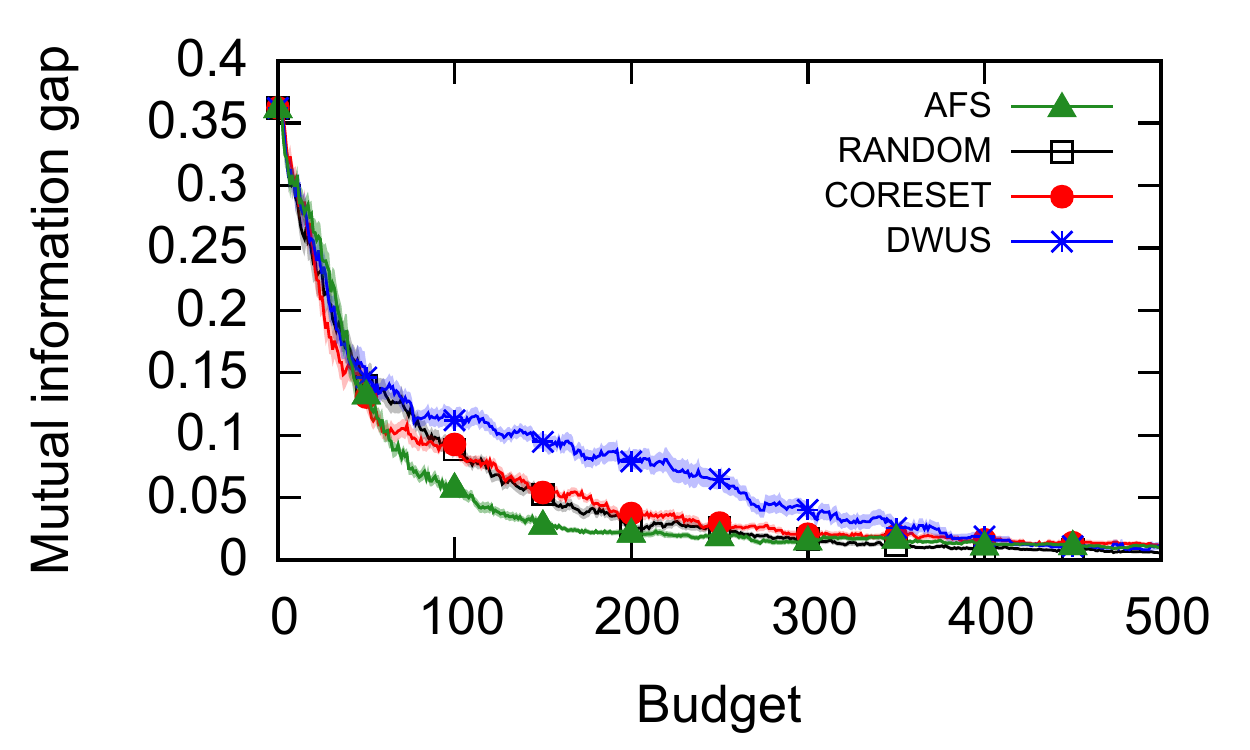} \\
    \includegraphics[width = 0.4\textwidth]{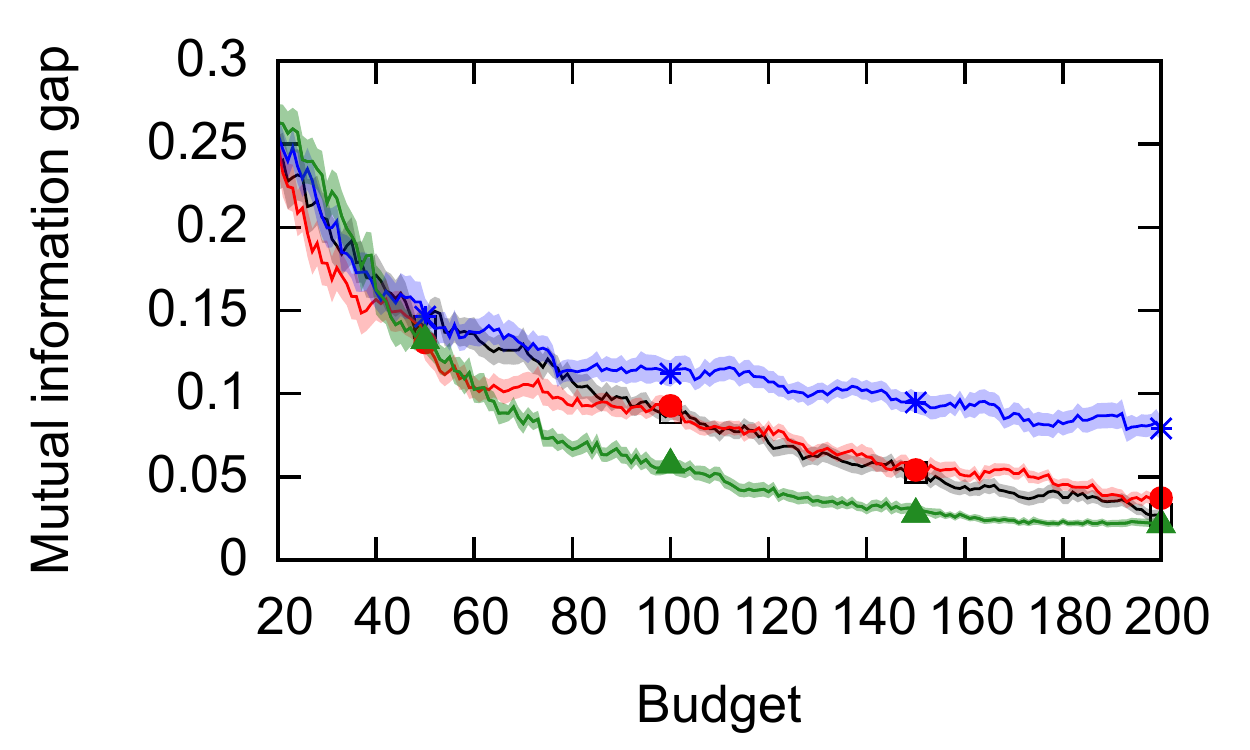}
    }
  \end{center}

\caption{\small AFS vs baseline: PCMAC, $k=5$. Top: full experiment. Bottom: Zoom in.}
\end{figure}

\begin{figure}[h]
  \begin{center}
    \myborder{
    \includegraphics[width = 0.4\textwidth]{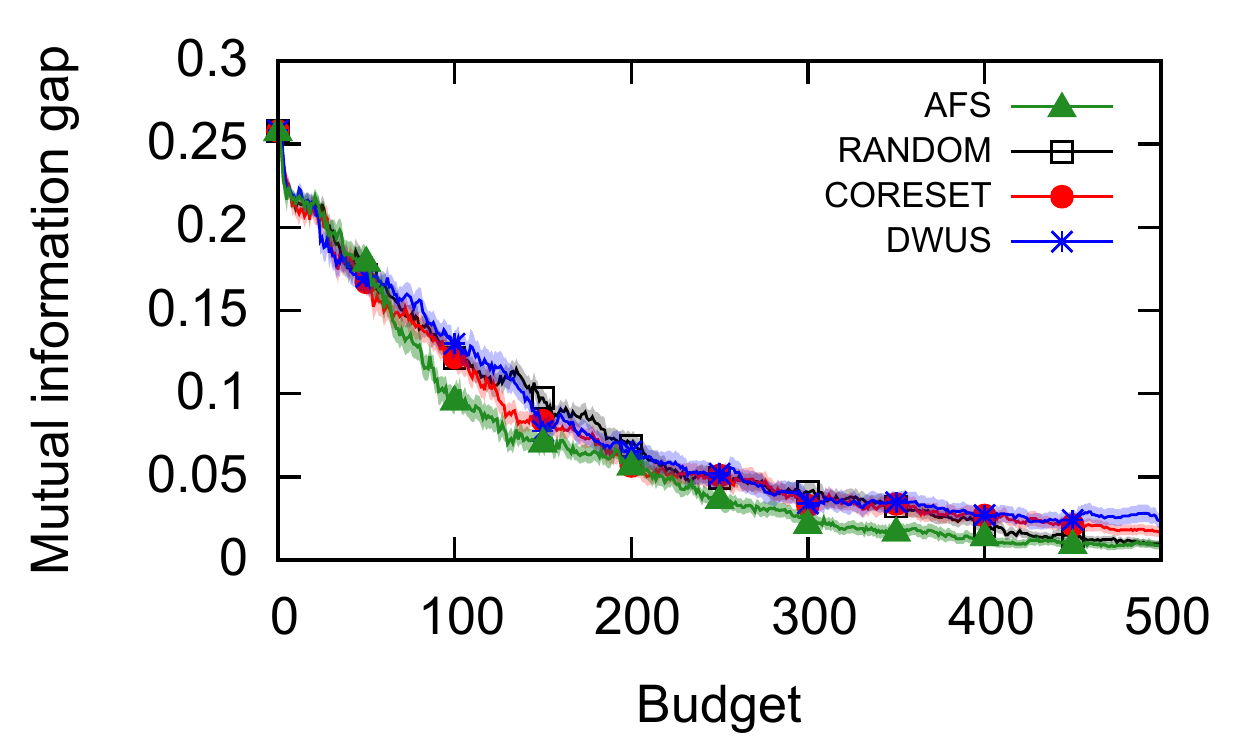} \\
    \includegraphics[width = 0.4\textwidth]{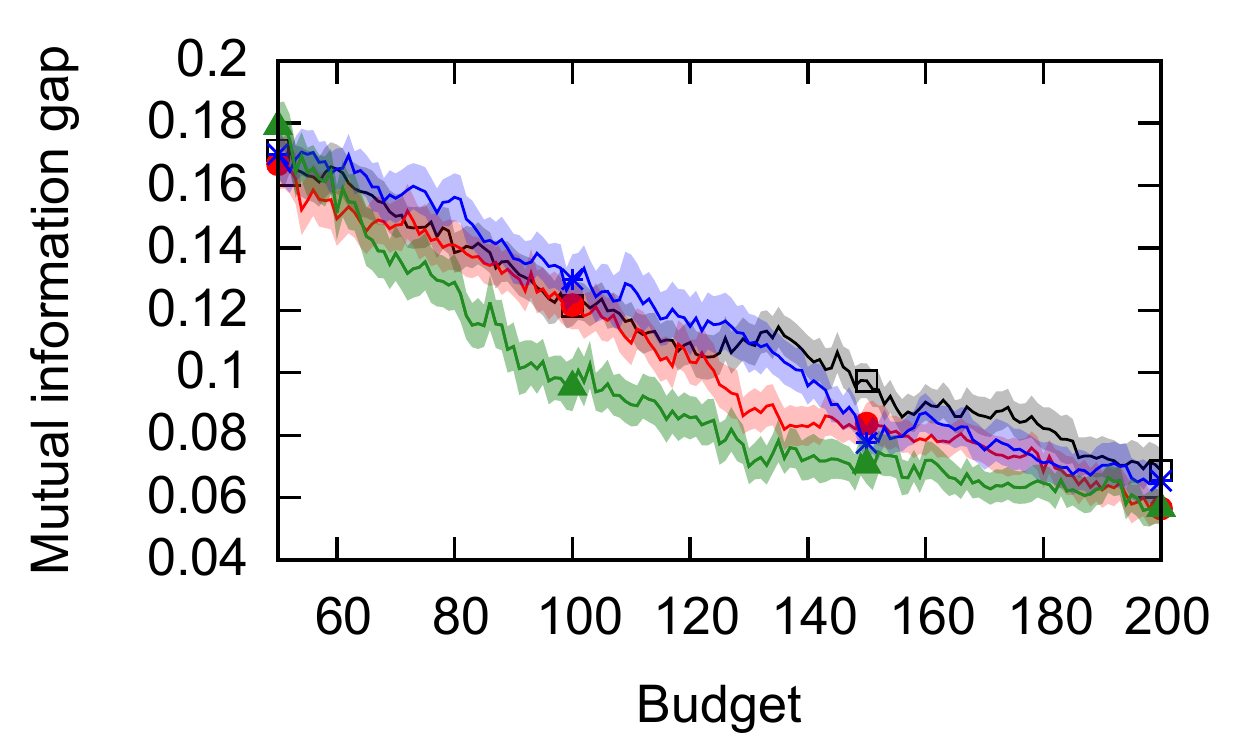}
    }
  \end{center}

\caption{\small AFS vs baseline: RELATHE, $k=5$. Top: full experiment. Bottom: Zoom in.}
\end{figure}

\begin{figure}[h]
  \begin{center}
    \myborder{
    \includegraphics[width = 0.4\textwidth]{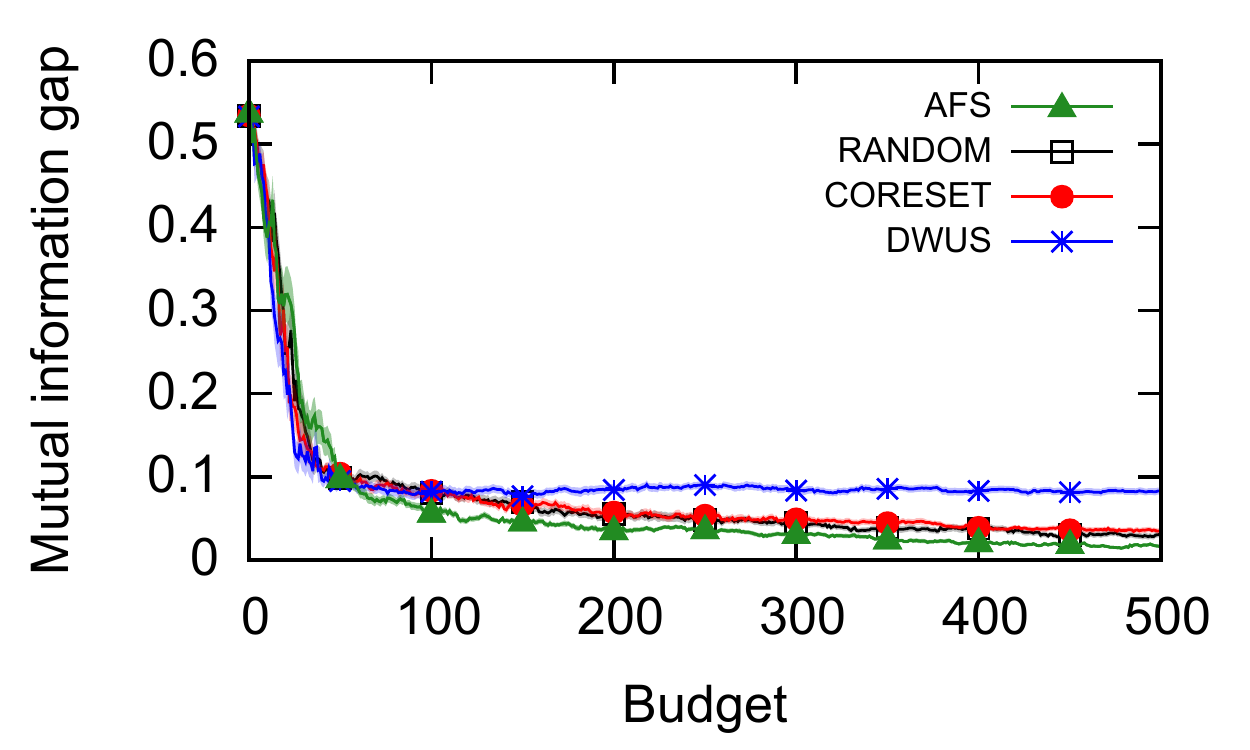} \\
    \includegraphics[width = 0.4\textwidth]{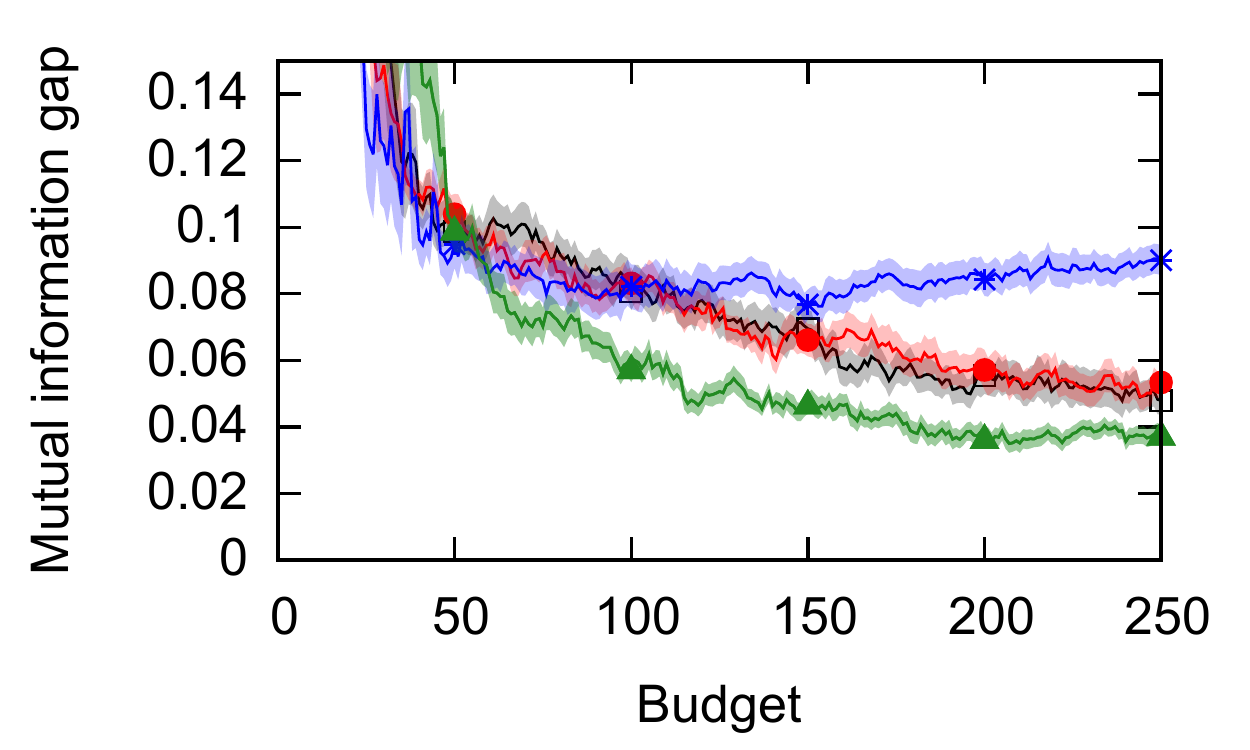}
    }
  \end{center}

\caption{\small AFS vs baseline: MUSK, $k=5$. Top: full experiment. Bottom: Zoom in.}
\end{figure}

\begin{figure}[h]
  \begin{center}
    \myborder{
    \includegraphics[width = 0.4\textwidth]{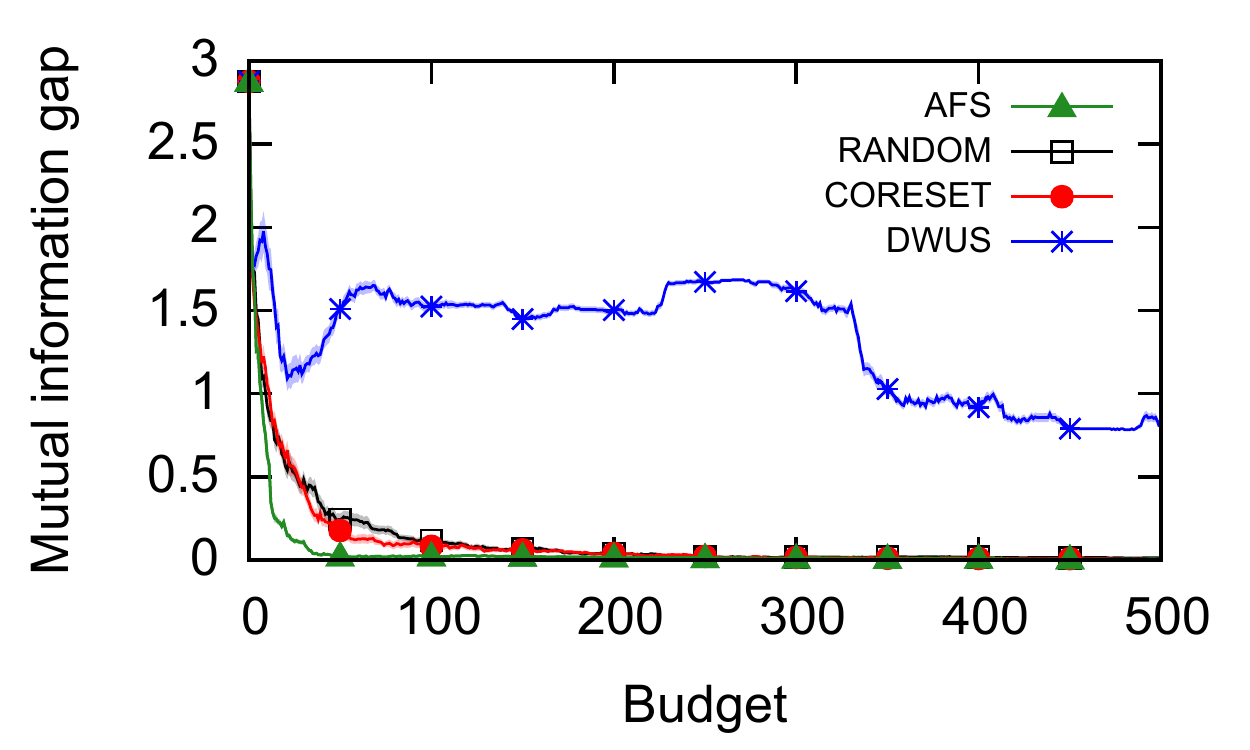} \\
    \includegraphics[width = 0.4\textwidth]{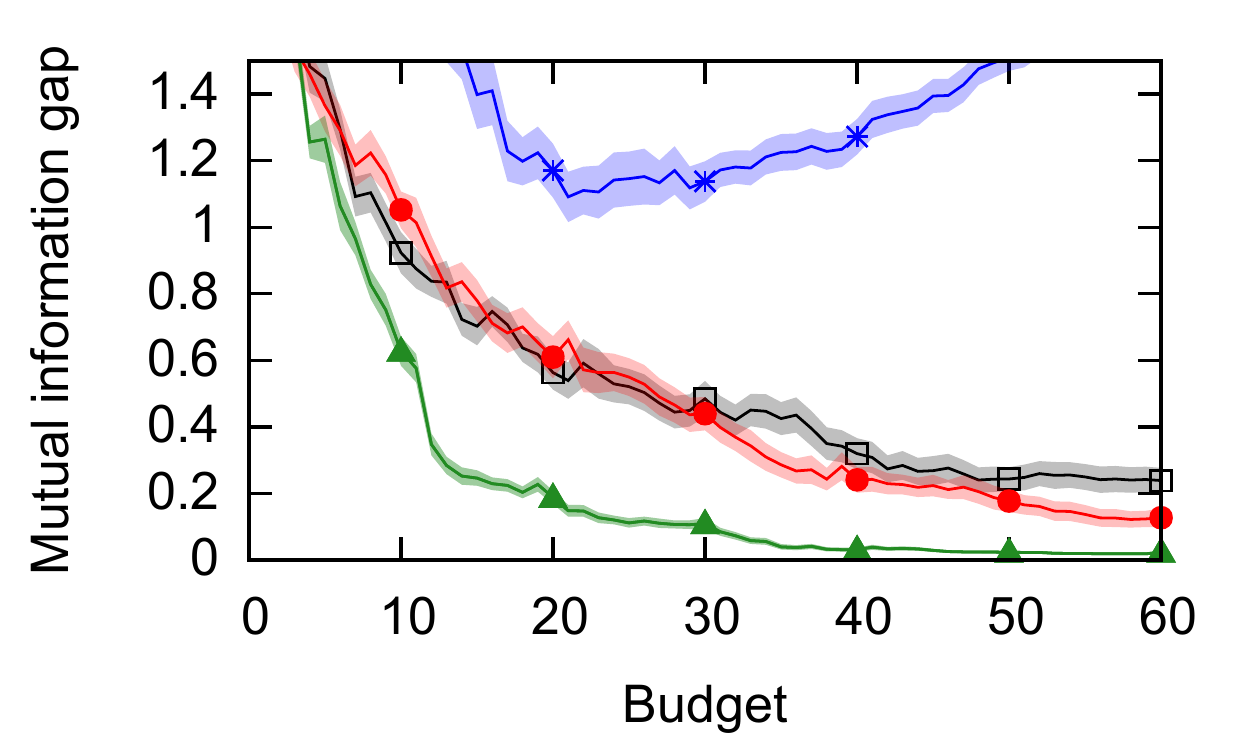}
    }
  \end{center}

\caption{\small AFS vs baseline: MNIST: 0 vs 1, $k=5$. Top: full experiment. Bottom: Zoom in.}
\end{figure}

\begin{figure}[h]
  \begin{center}
    \myborder{
    \includegraphics[width = 0.4\textwidth]{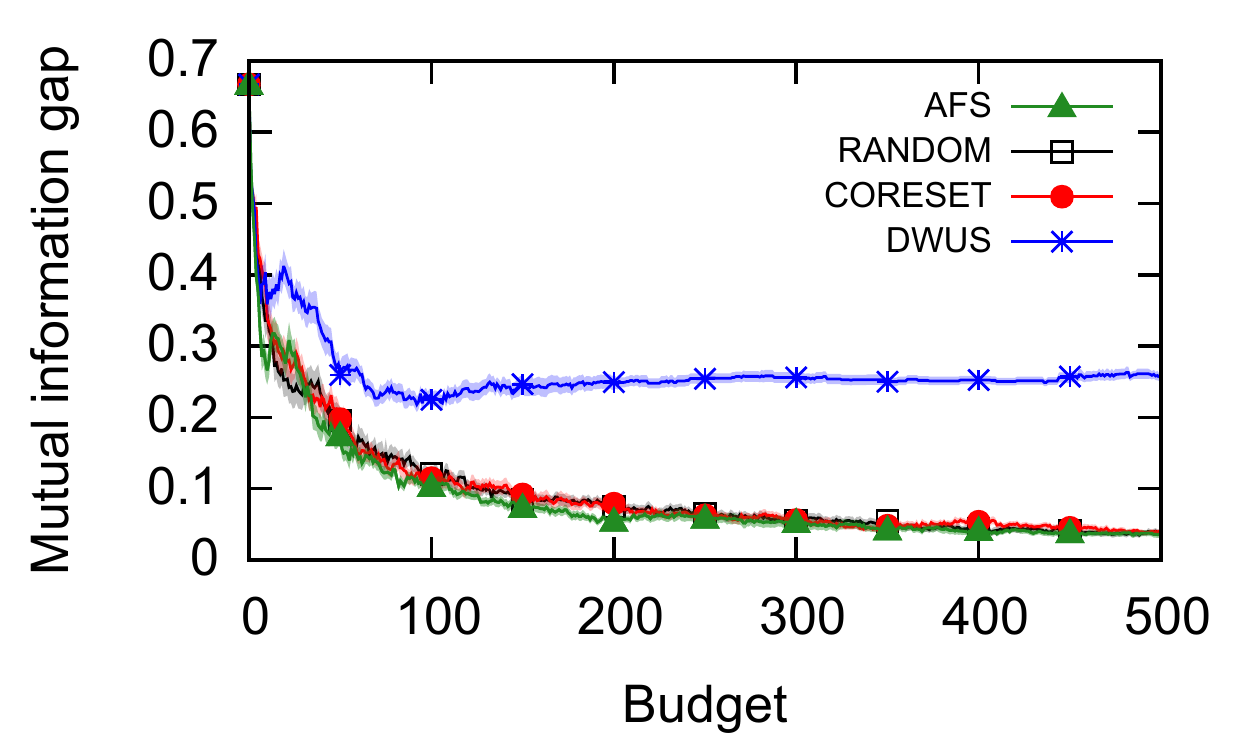} \\
    \includegraphics[width = 0.4\textwidth]{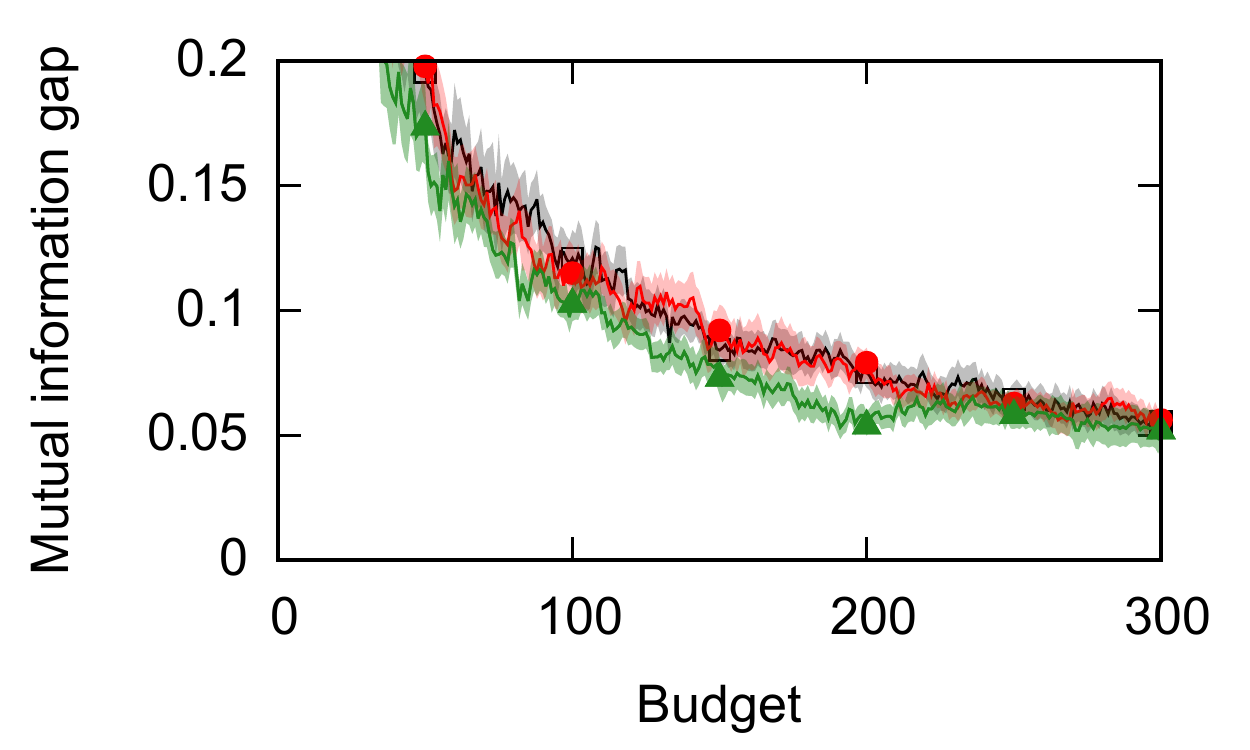}
    }
  \end{center}

\caption{\small AFS vs baseline: MNIST: 3 vs 5, $k=5$. Top: full experiment. Bottom: Zoom in.}
\end{figure}

\begin{figure}[h]
  \begin{center}
    \myborder{
    \includegraphics[width = 0.4\textwidth]{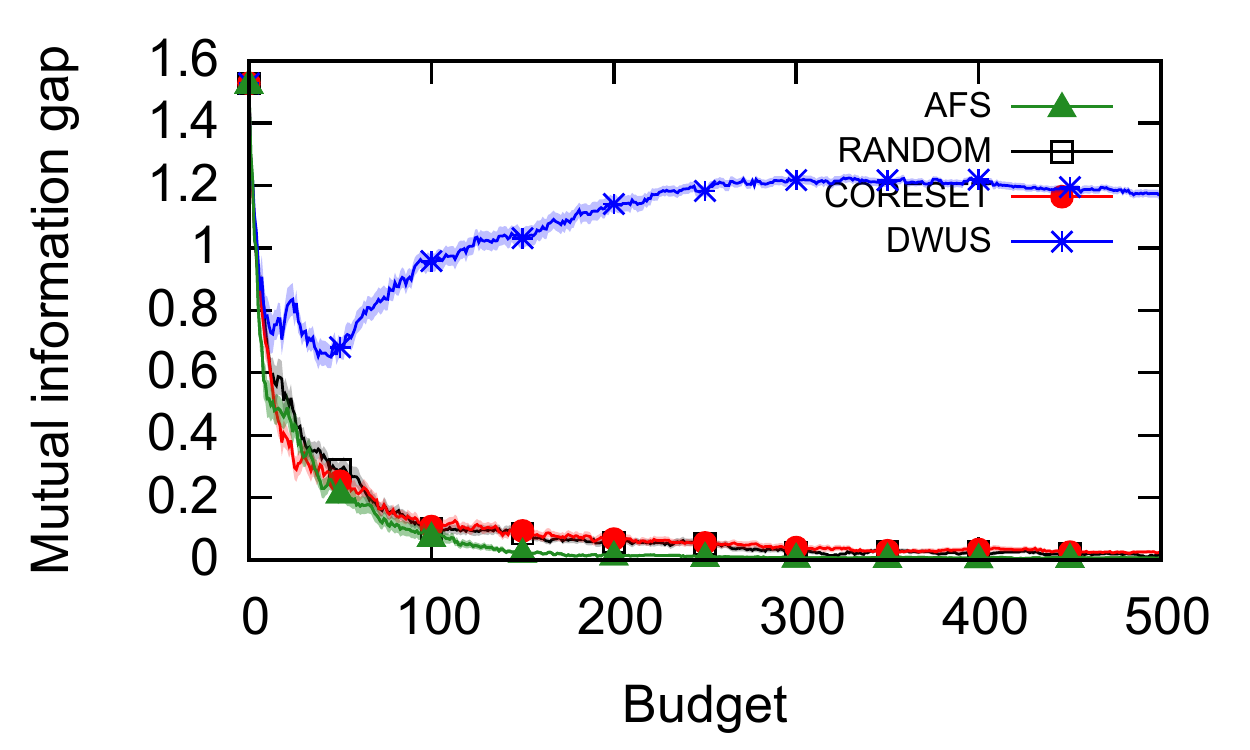} \\
    \includegraphics[width = 0.4\textwidth]{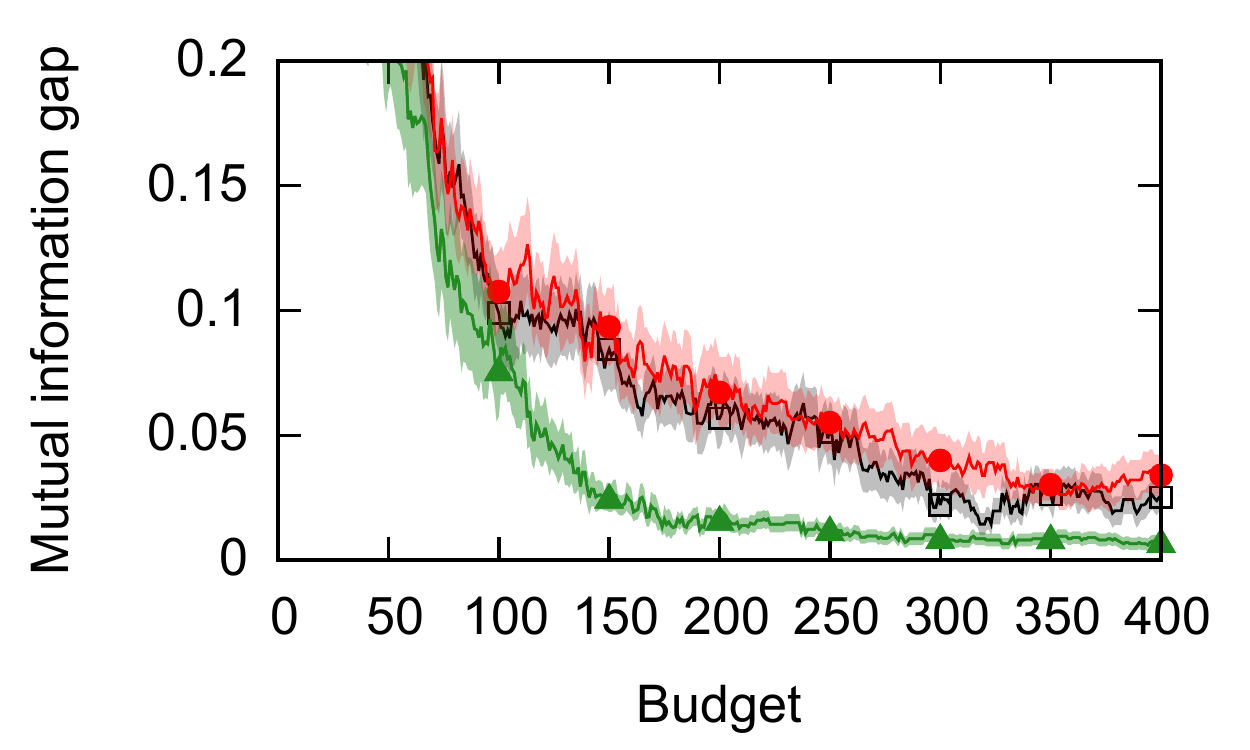}
    }
  \end{center}

\caption{\small AFS vs baseline: MNIST: 4 vs 6, $k=5$. Top: full experiment. Bottom: Zoom in.}
\end{figure}

\clearpage
\subsection{Comparing to baselines: $k = 1 $}
  The last set of graphs shows the experiments comparing AFS to the baseline for $k = 1$. For $k =1$, AFS performs comparably with CORESET and RANDOM. However, the DWUS baseline performs significantly worse in many of the experiments.
\begin{figure}[h]
  \begin{center}
    \myborder{
    \includegraphics[width = 0.4\textwidth]{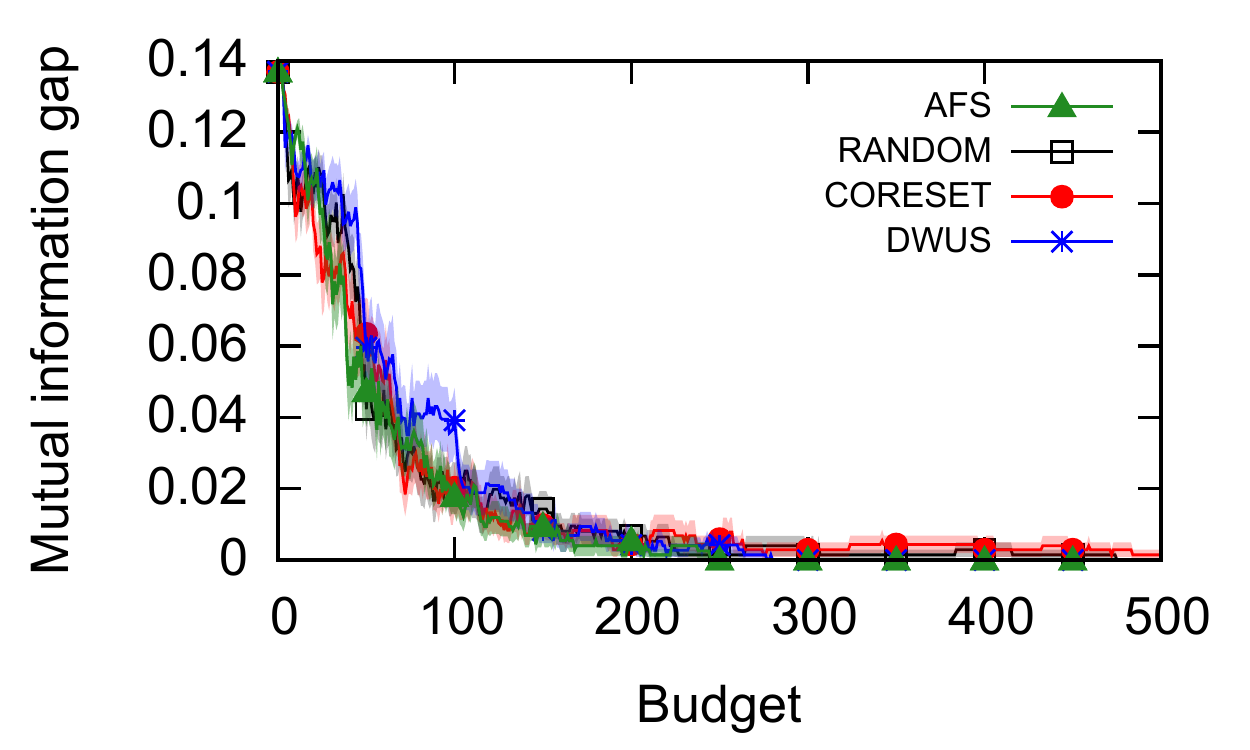}
    }
  \end{center}

\caption{\small AFS vs baseline: BASEHOCK, $k=1$.}
\end{figure}

\begin{figure}[h]
  \begin{center}
    \myborder{
    \includegraphics[width = 0.4\textwidth]{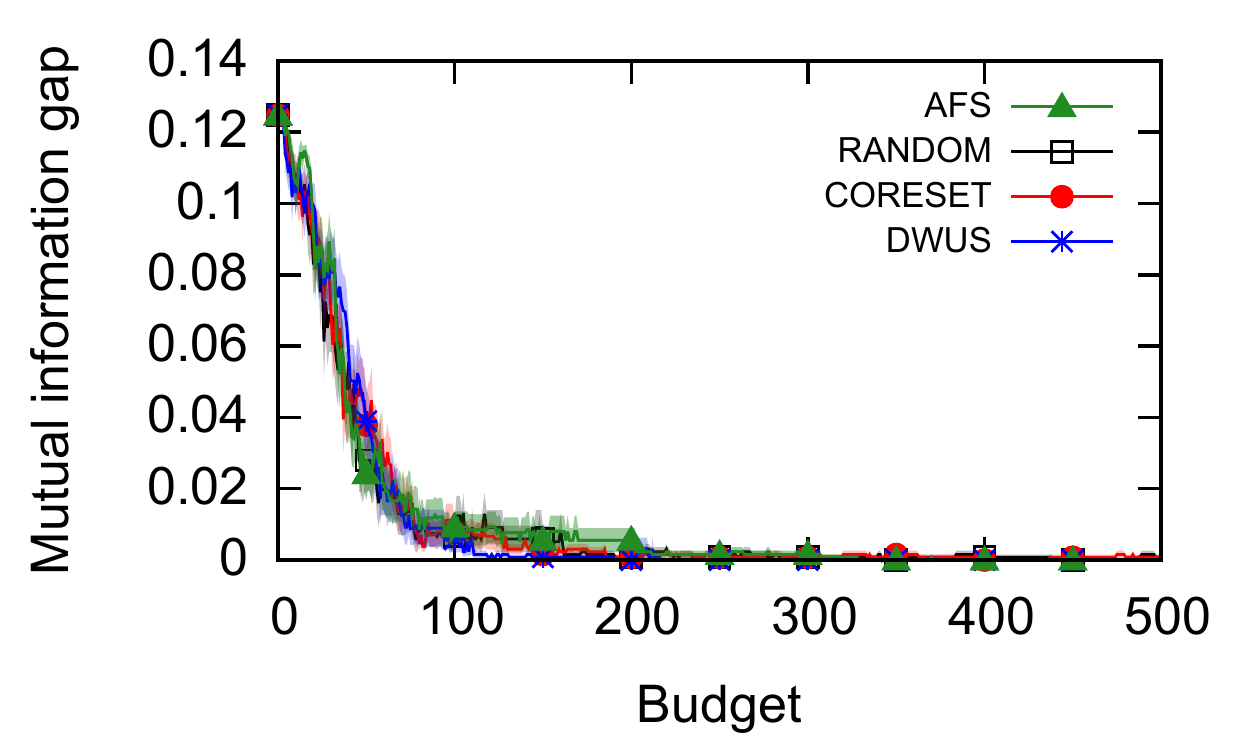}
    }
  \end{center}

\caption{\small AFS vs baseline: PCMAC, $k=1$.}
\end{figure}

\begin{figure}[h]
  \begin{center}
    \myborder{
    \includegraphics[width = 0.4\textwidth]{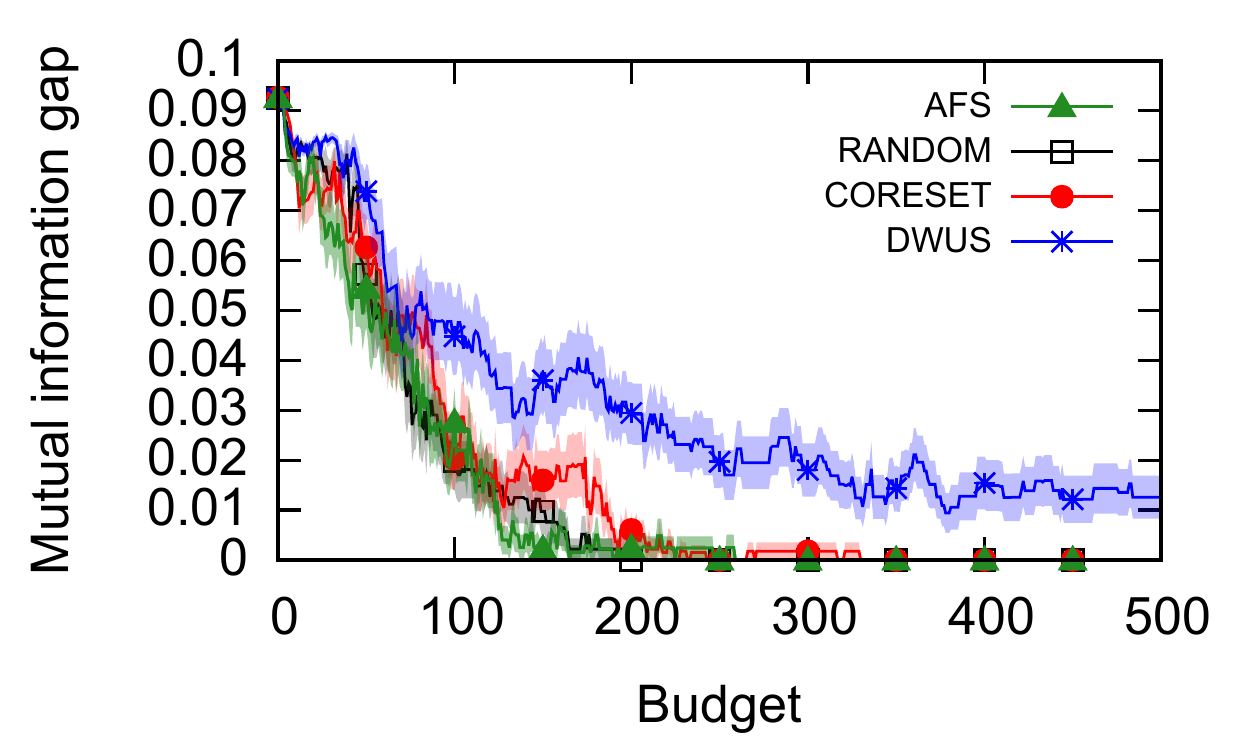}
    }
  \end{center}

\caption{\small AFS vs baseline: RELATHE, $k=1$.}
\end{figure}

\begin{figure}[h]
  \begin{center}
    \myborder{
    \includegraphics[width = 0.4\textwidth]{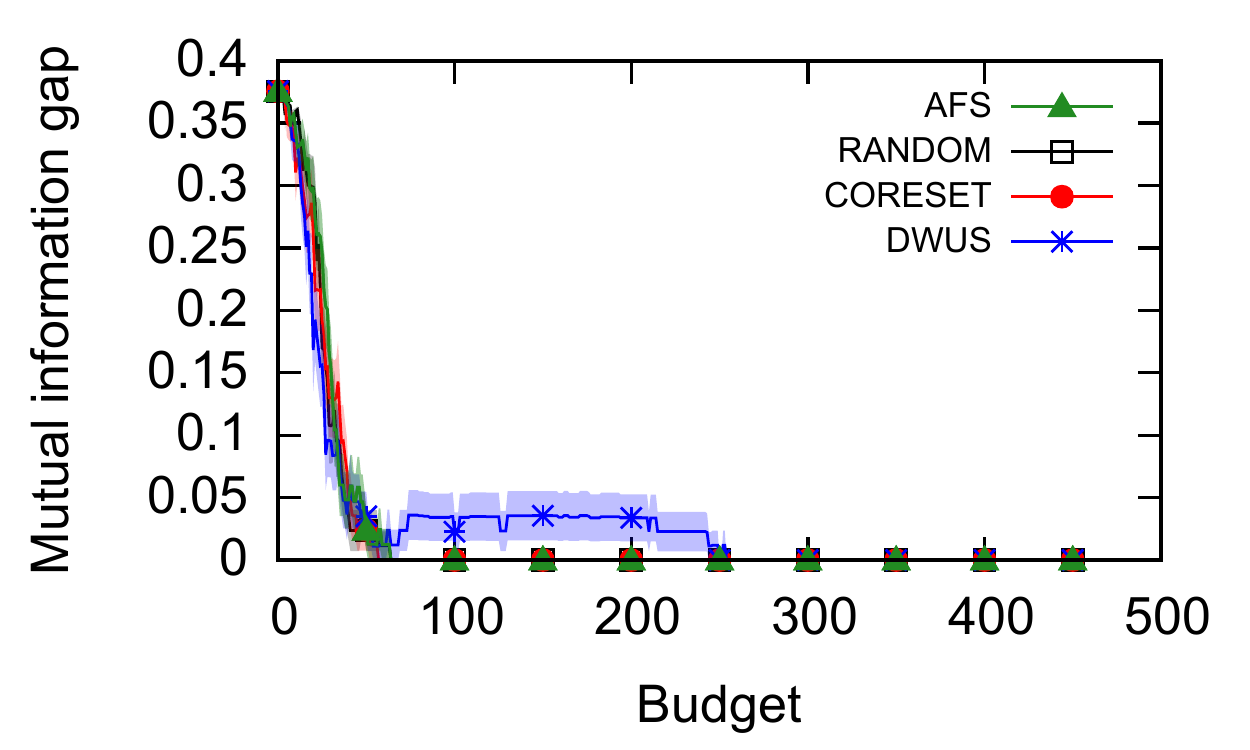}
    }
  \end{center}

\caption{\small AFS vs baseline: MUSK, $k=1$.}
\end{figure}

\begin{figure}[h]
  \begin{center}
    \myborder{
    \includegraphics[width = 0.4\textwidth]{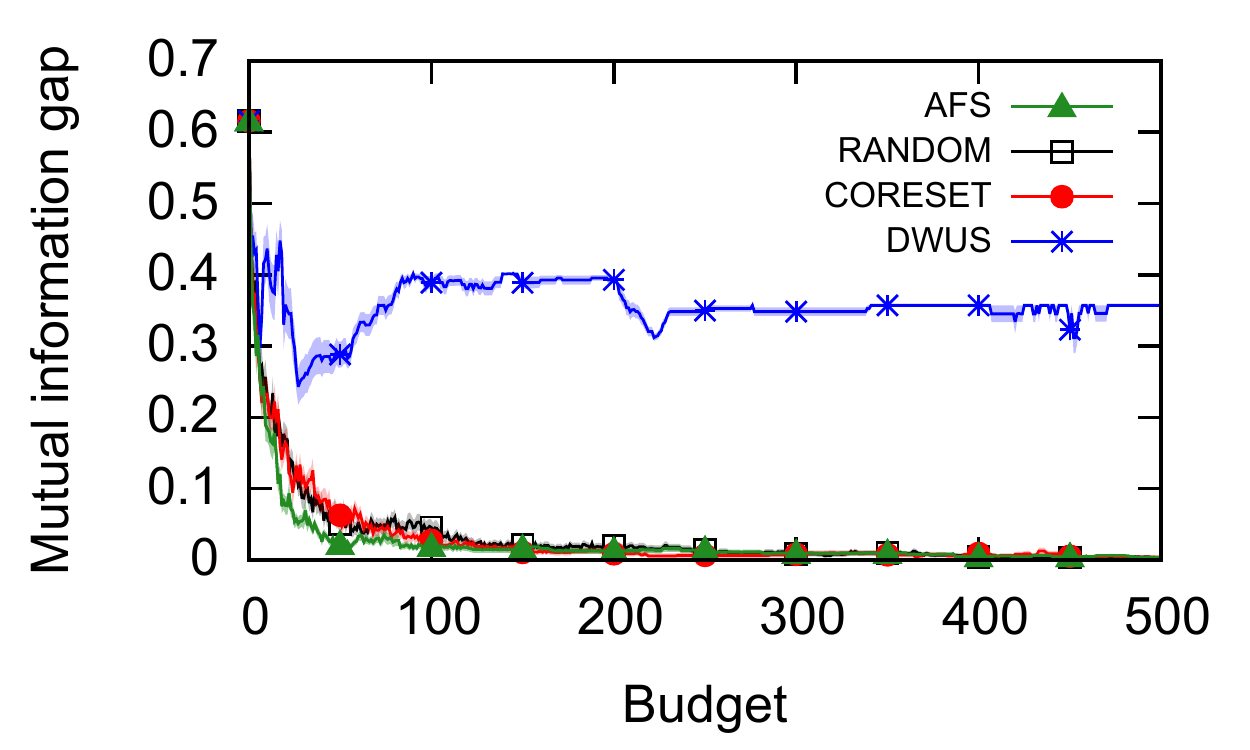}
    }
  \end{center}

\caption{\small AFS vs baseline: MNIST: 0 vs 1, $k=1$.}
\end{figure}

\begin{figure}[h]
  \begin{center}
    \myborder{
    \includegraphics[width = 0.4\textwidth]{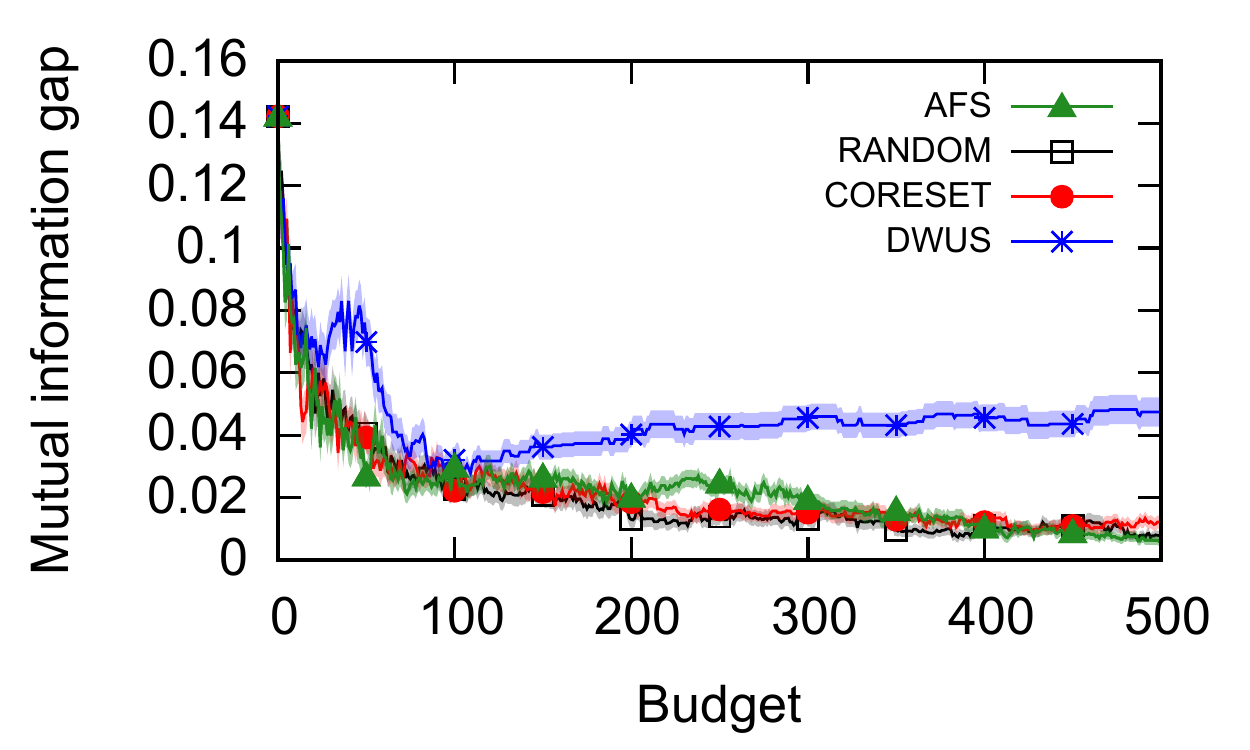}
    }
  \end{center}

\caption{\small AFS vs baseline: MNIST: 3 vs 5, $k=1$.}
\end{figure}

\begin{figure}[H]
  \begin{center}
    \myborder{
    \includegraphics[width = 0.4\textwidth]{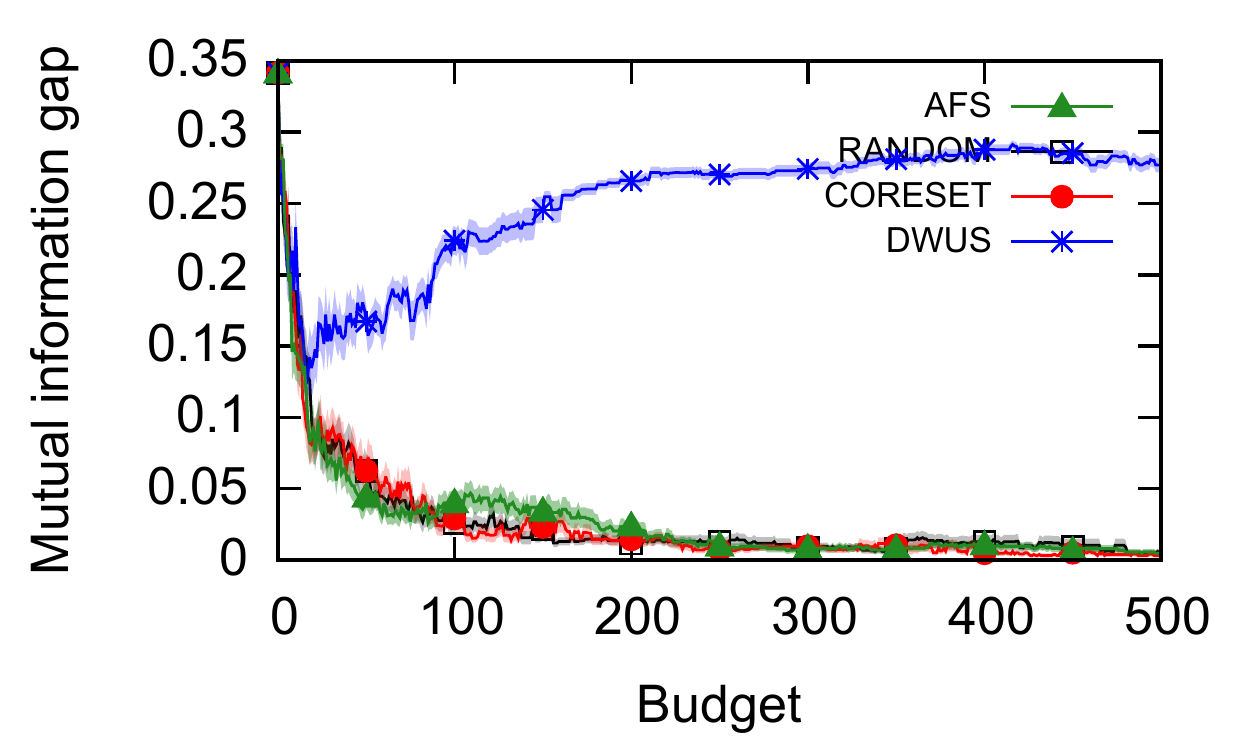}
    }
  \end{center}

\caption{\small AFS vs baseline: MNIST: 4 vs 6, $k=1$.}
\end{figure}

\clearpage
\section{Experiments: Ablation tests}\label{app:ablation}
The following graphs show the results of the ablation experiments.
The $x$ axis is the label budget, and the
  $y$ axis is the mutual information gap, calculated via $\sum_{j \in F^*} H_j - \sum_{j \in F} H_j$, as
  explained in Section 6. Each graph averages $30$ runs with budgets up to $500$ queries. The shaded areas provide the 95\% confidence intervals. For each experiment, we show the full graph at the top and a zoomed-in version at the bottom, which allows better visibility of the interesting parts of the graph.

  AVG-SEL, which avoids the bias-correction term, performs quite poorly in many of the cases. SINGLE, which uses a single feature for scoring, is also usually worse than AFS. AVG-ALL, which averages the score over all the features, is inconsistent and sometimes performs very poorly, although it sometimes performs better than the other options.\footnote{There were no ties when all the features were used to calculate the score, therefore AVG-ALL is deterministic, thus all its runs are exactly the same}

\begin{figure}[h]
  \begin{center}
    \myborder{
    \includegraphics[width = 0.4\textwidth]{graph/ablation/10/RELATHE.pdf} \\
    \includegraphics[width = 0.4\textwidth]{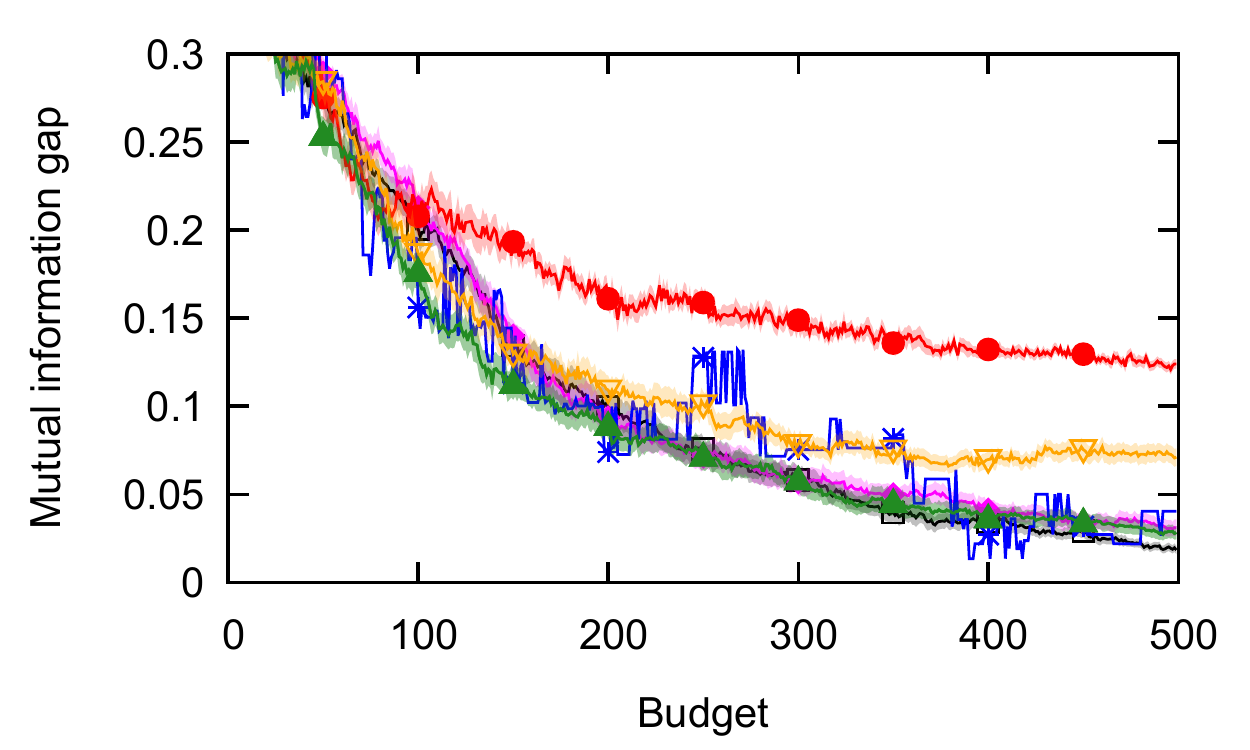}
    }
  \end{center}

  \caption{\small Ablation tests: RELATHE ($k=10$). Top: full experiment. Bottom: Zoom in. Here, AFS-NOSG is significantly worse than AFS and does not converge to a low gap.}
  \label{fig:abrel10}
\end{figure}

\begin{figure}[h]
  \begin{center}
    \myborder{
    \includegraphics[width = 0.4\textwidth]{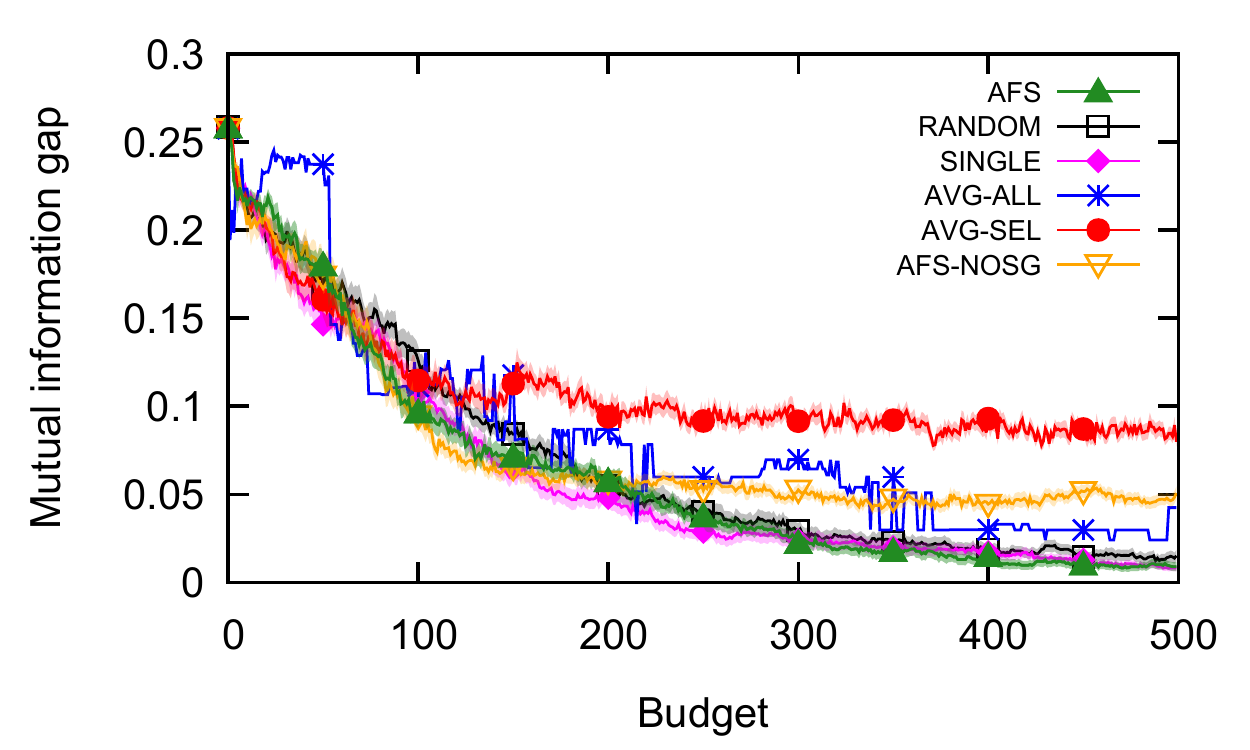} \\
    \includegraphics[width = 0.4\textwidth]{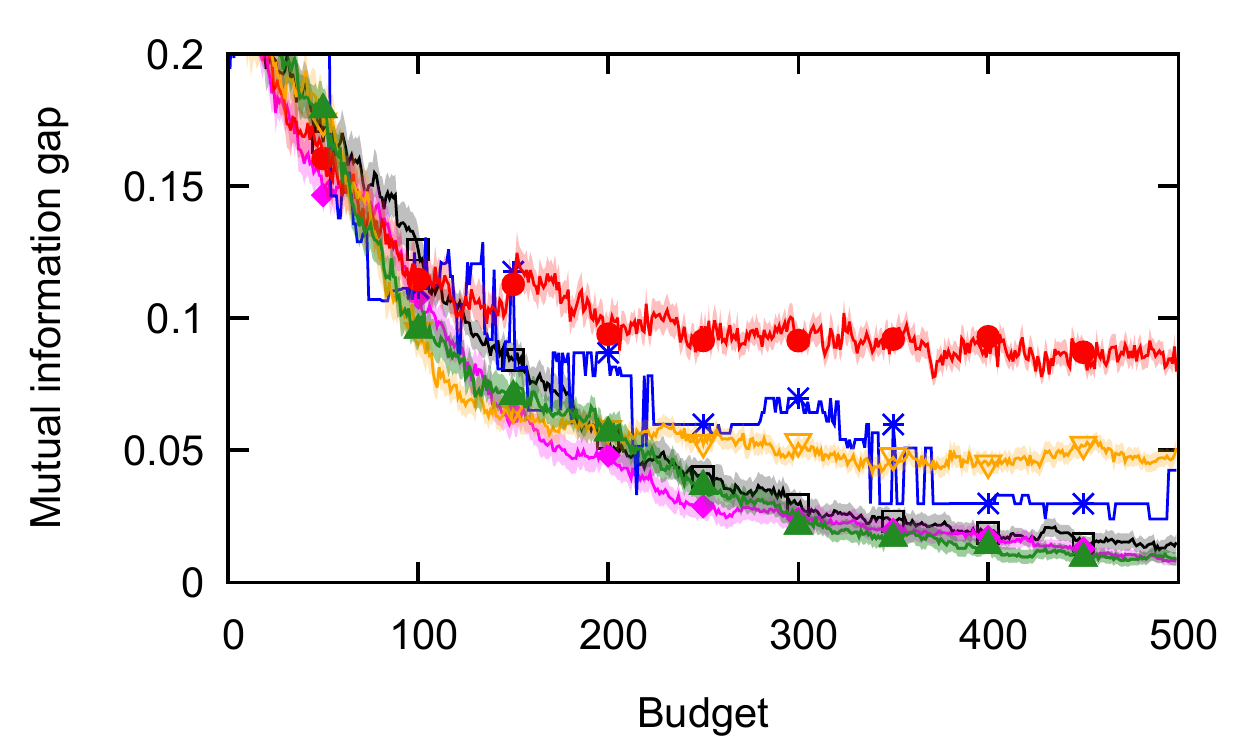}
    }
  \end{center}

  \caption{\small Ablation tests: RELATHE ($k=5$). Top: full experiment. Bottom: Zoom in. Here, AFS-NOSG is significantly worse than AFS and does not converge to a low gap.}
  \label{fig:abrel5}
\end{figure}

Lastly, AFS-NOSG, which has no safeguard in case the estimates converge to a wrong value, usually performs similarly to AFS, but in some cases (most significantly, the RELATHE data set, \figref{abrel10} and \figref{abrel5}) it is significantly worse even compared to the random baseline. It also sometimes performs better than AFS. Nonetheless, AFS-NOSG is overall less robust due to its failure to converge to a low gap in some cases.
These results show that the full AFS algorithm performs well the most consistently.

\subsection{Ablation tests: $k = 20$}
The following graphs gives the results of the ablation tests for $k = 20$.
\begin{figure}[h]
  \begin{center}
    \myborder{
    \includegraphics[width = 0.4\textwidth]{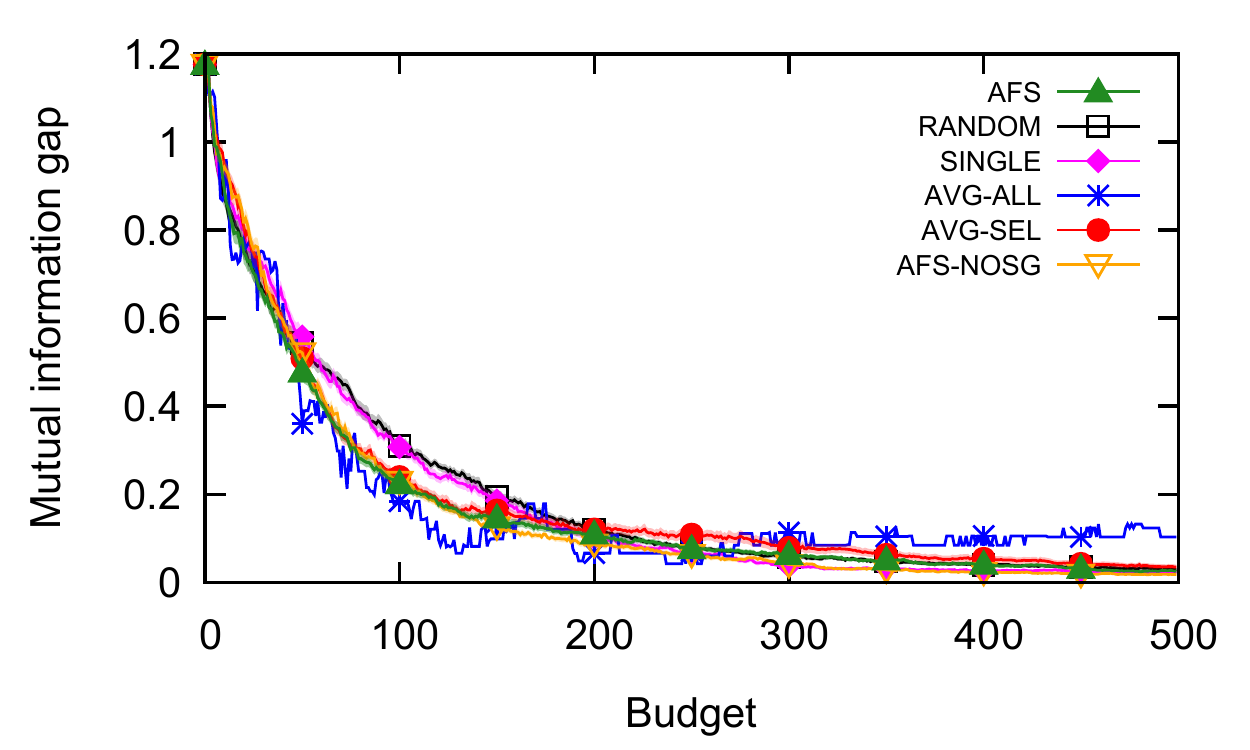} \\
    \includegraphics[width = 0.4\textwidth]{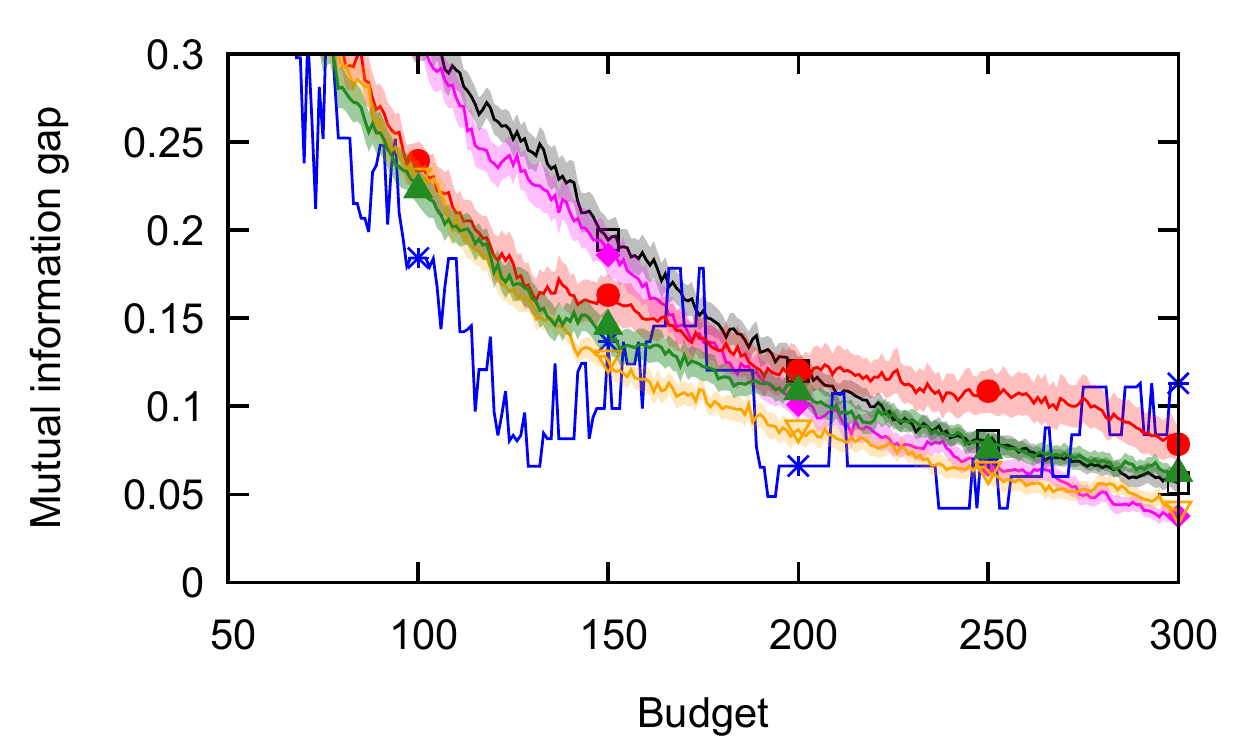}
    }
  \end{center}

\caption{\small Ablation tests: BASEHOCK ($k=20$). Top: full experiment. Bottom: Zoom in.}
\end{figure}
\begin{figure}[h]
  \begin{center}
    \myborder{
    \includegraphics[width = 0.4\textwidth]{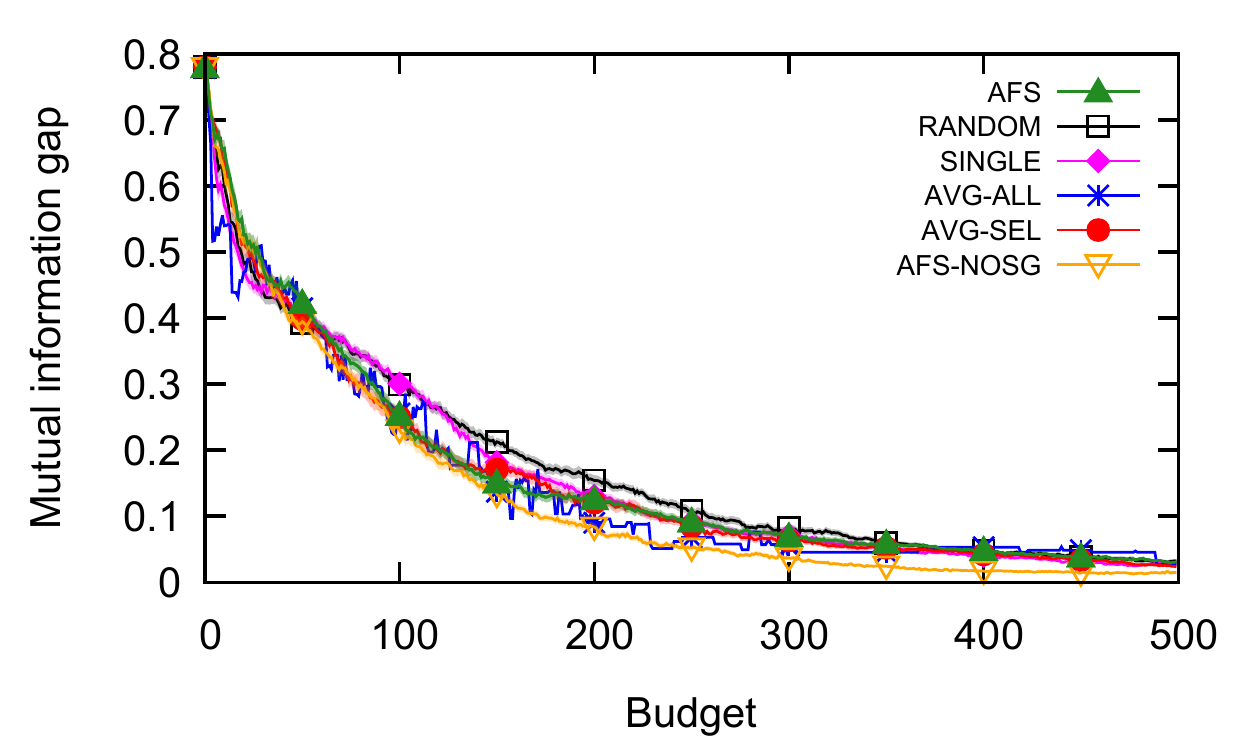} \\
    \includegraphics[width = 0.4\textwidth]{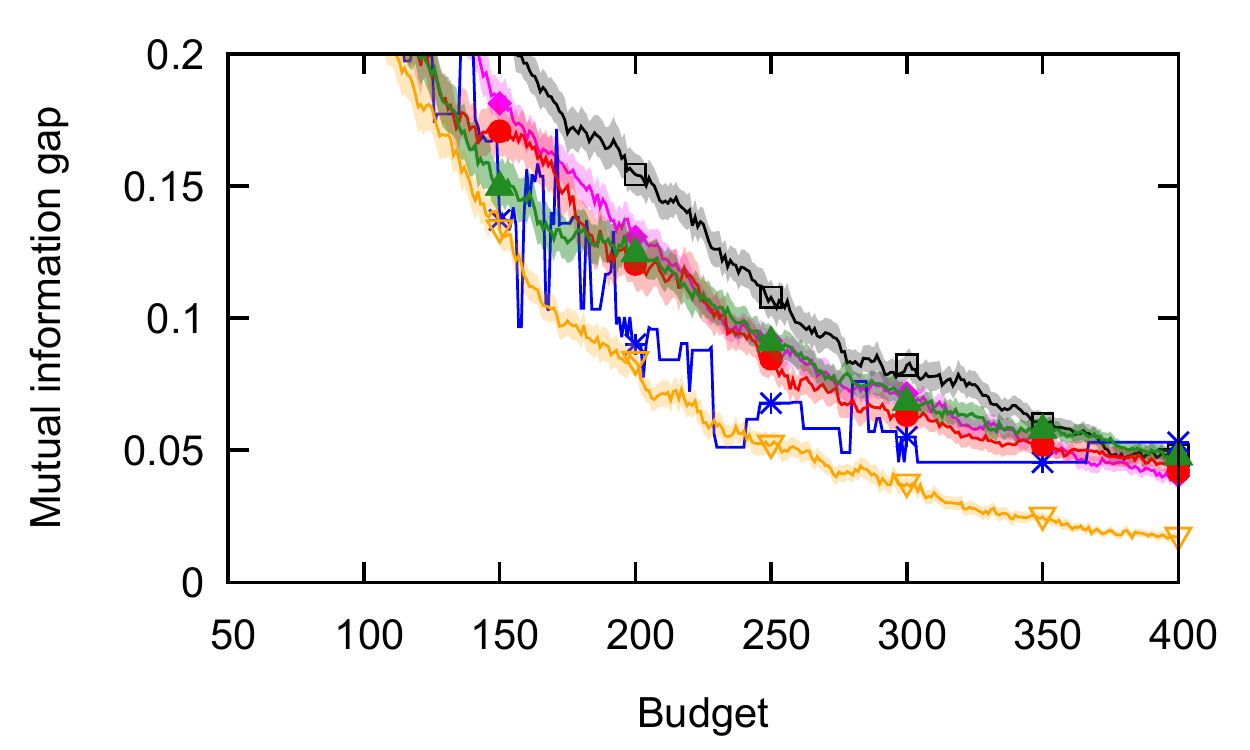}
    }
  \end{center}

\caption{\small Ablation tests: PCMAC ($k=20$). Top: full experiment. Bottom: Zoom in.}
\end{figure}

\begin{figure}[h]
  \begin{center}
    \myborder{
    \includegraphics[width = 0.4\textwidth]{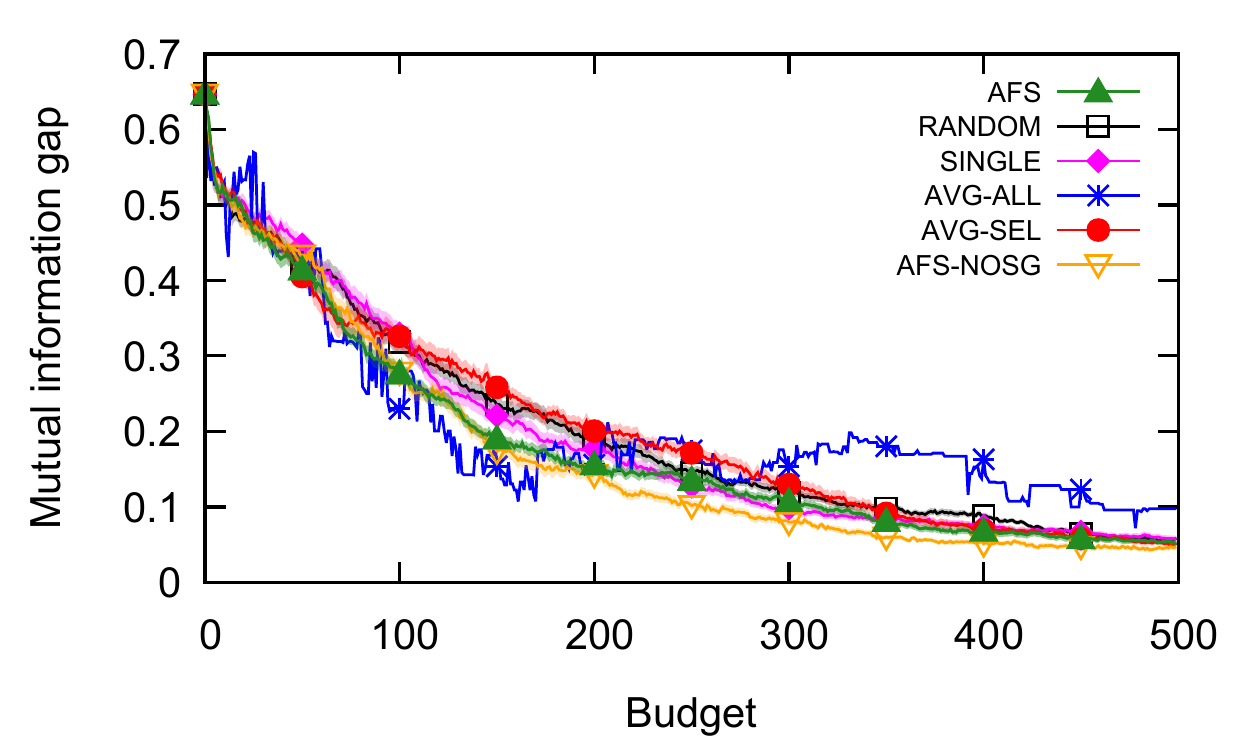} \\
    \includegraphics[width = 0.4\textwidth]{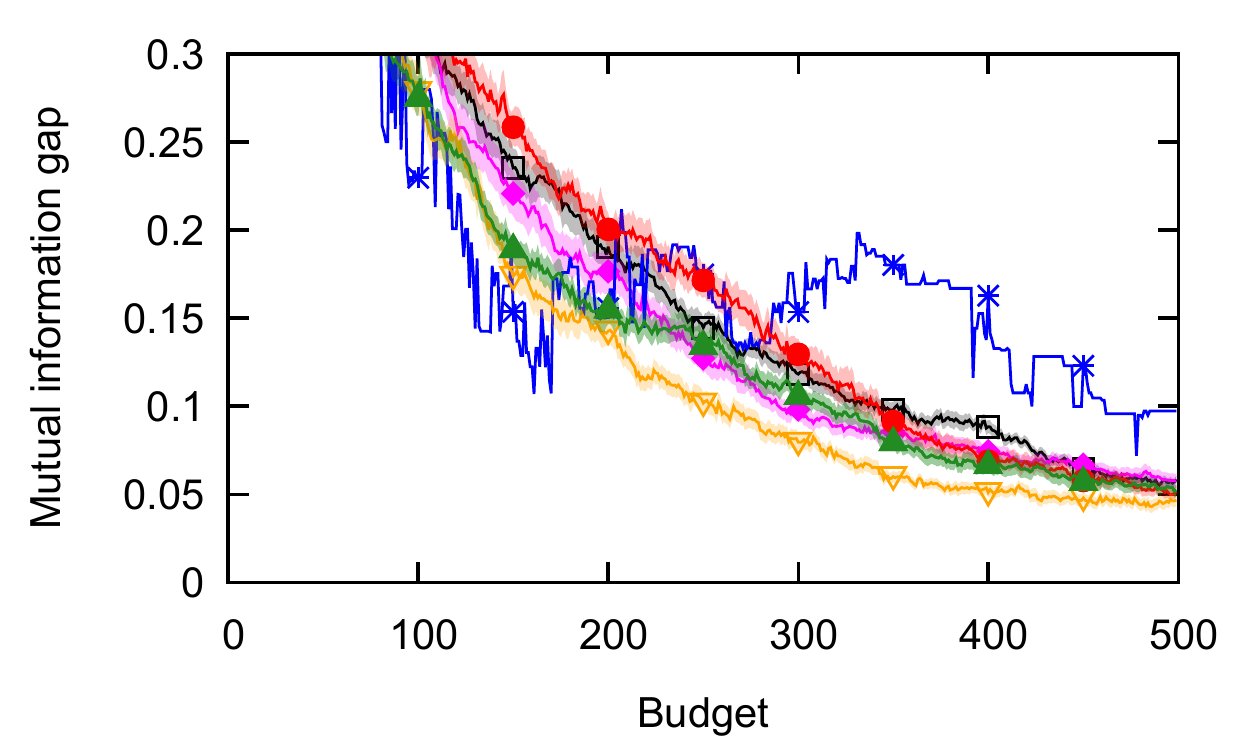}
    }
  \end{center}

  \caption{\small Ablation tests: RELATHE ($k=20$). Top: full experiment. Bottom: Zoom in.}
  \label{fig:abrel20}
\end{figure}

\begin{figure}[h]
  \begin{center}
    \myborder{
    \includegraphics[width = 0.4\textwidth]{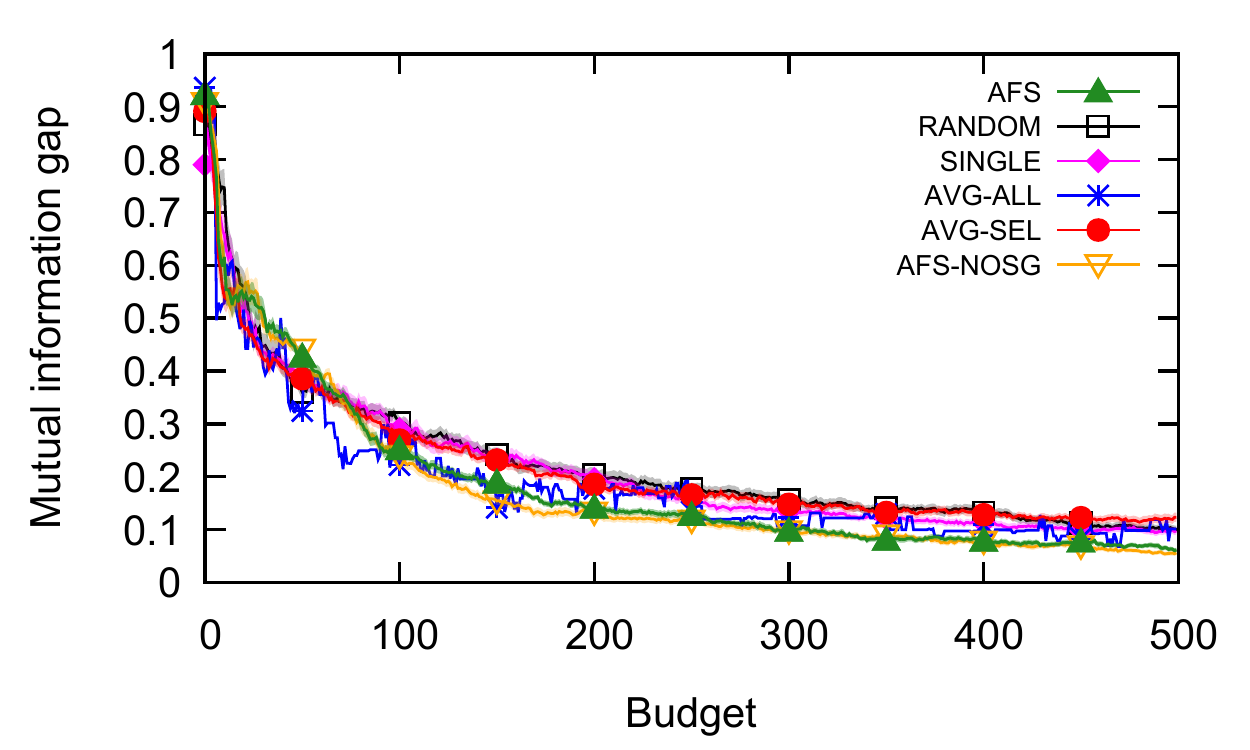} \\
    \includegraphics[width = 0.4\textwidth]{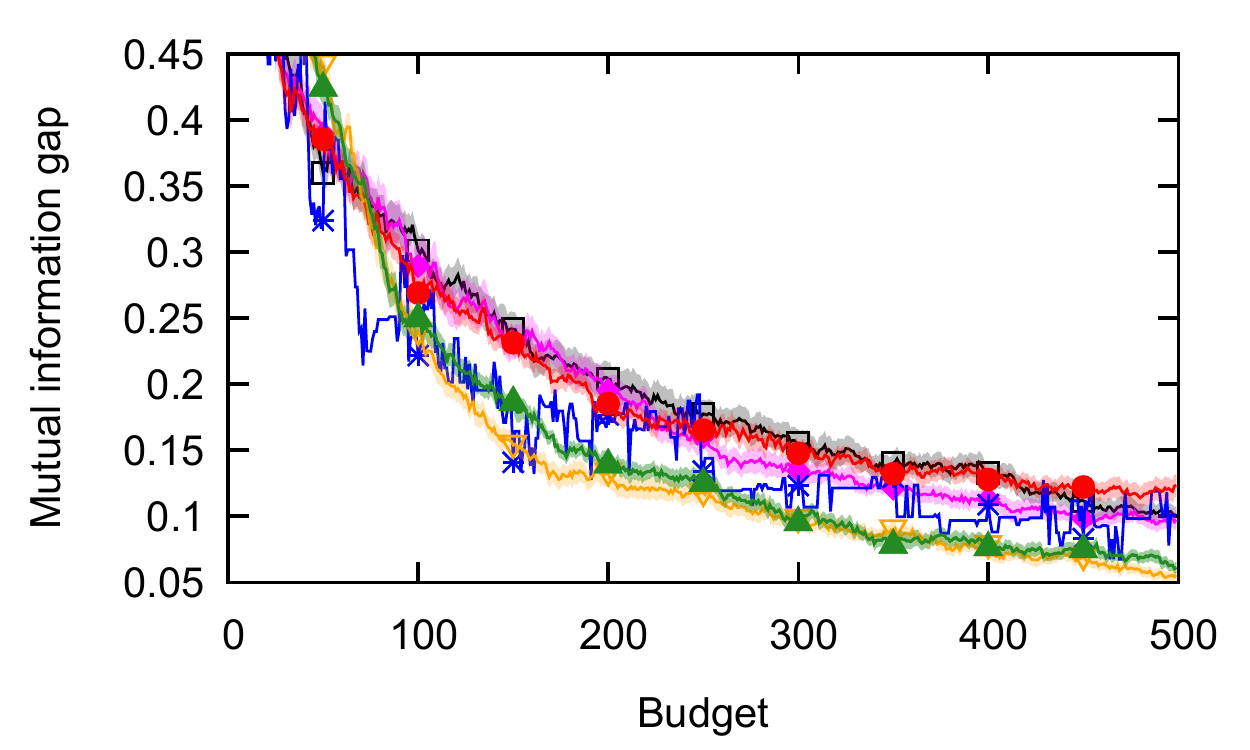}
    }
  \end{center}

\caption{\small Ablation tests: MUSK ($k=20$). Top: full experiment. Bottom: Zoom in.}
\end{figure}

\begin{figure}[h]
  \begin{center}
    \myborder{
    \includegraphics[width = 0.4\textwidth]{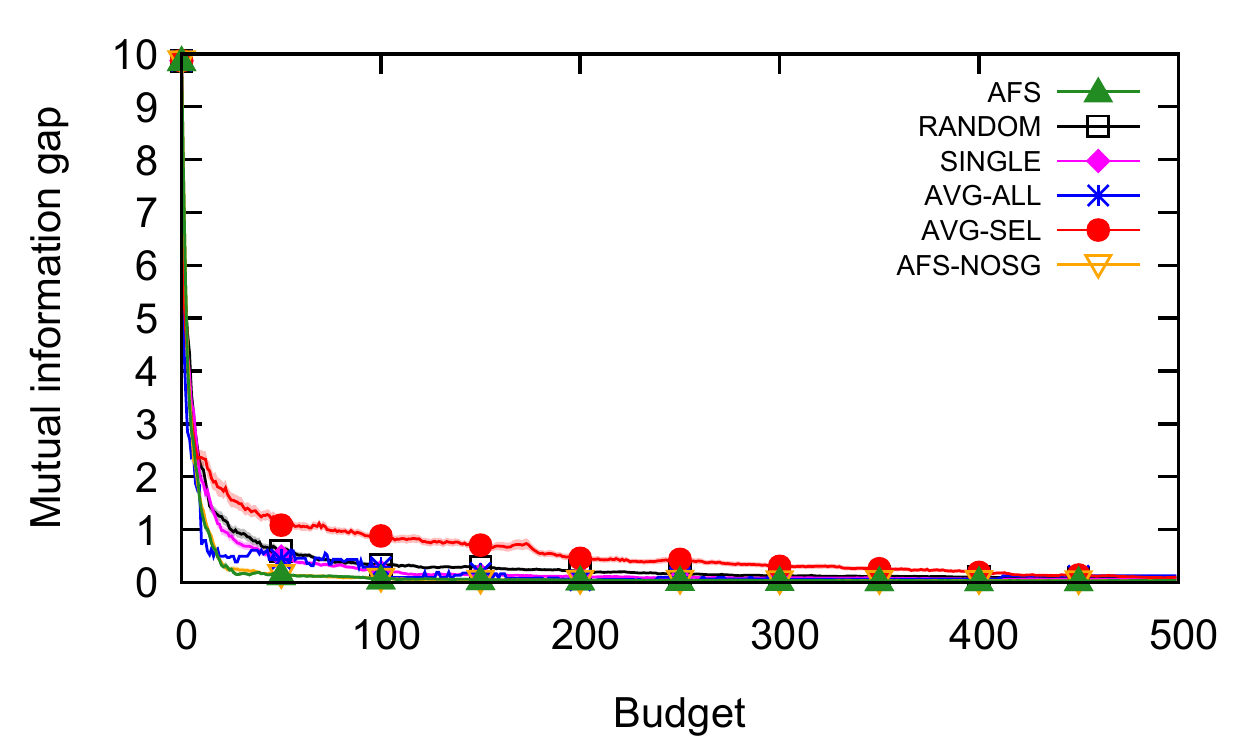} \\
    \includegraphics[width = 0.4\textwidth]{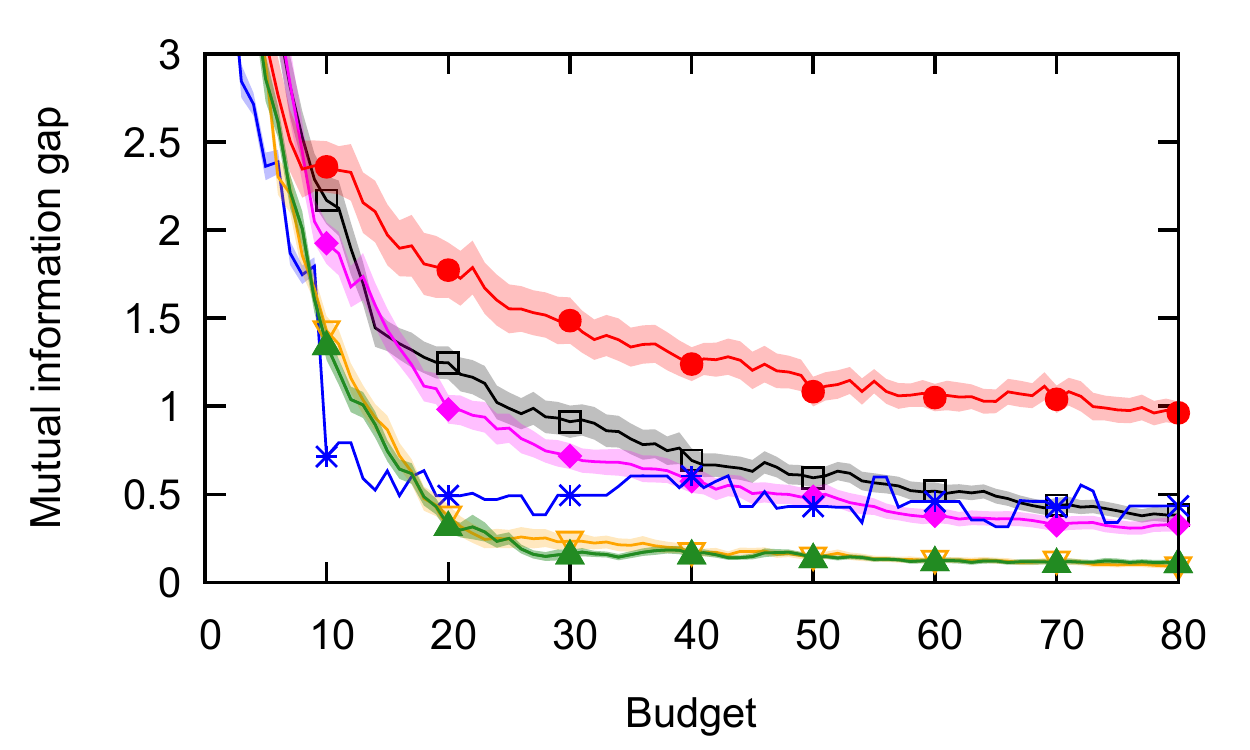}
    }
  \end{center}

\caption{\small Ablation tests: MNIST: 0 vs 1 ($k=20$). Top: full experiment. Bottom: Zoom in.}
\end{figure}

\begin{figure}[h]
  \begin{center}
    \myborder{
    \includegraphics[width = 0.4\textwidth]{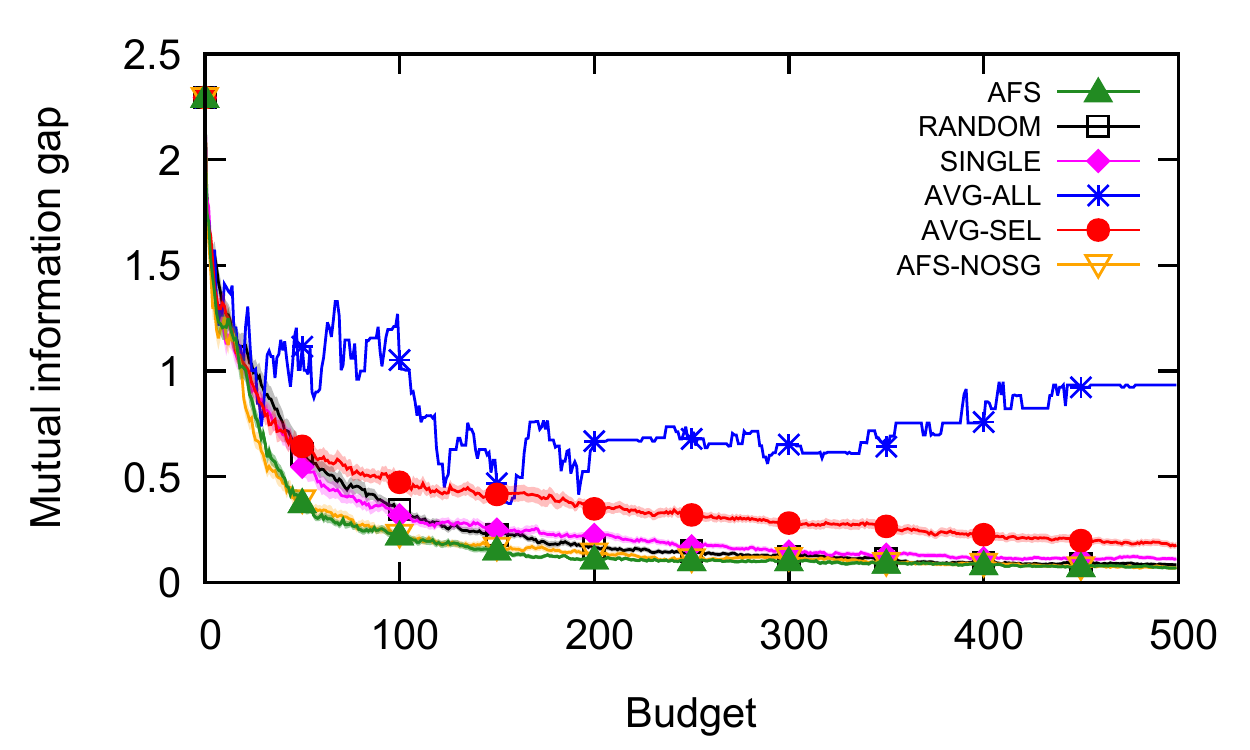} \\
    \includegraphics[width = 0.4\textwidth]{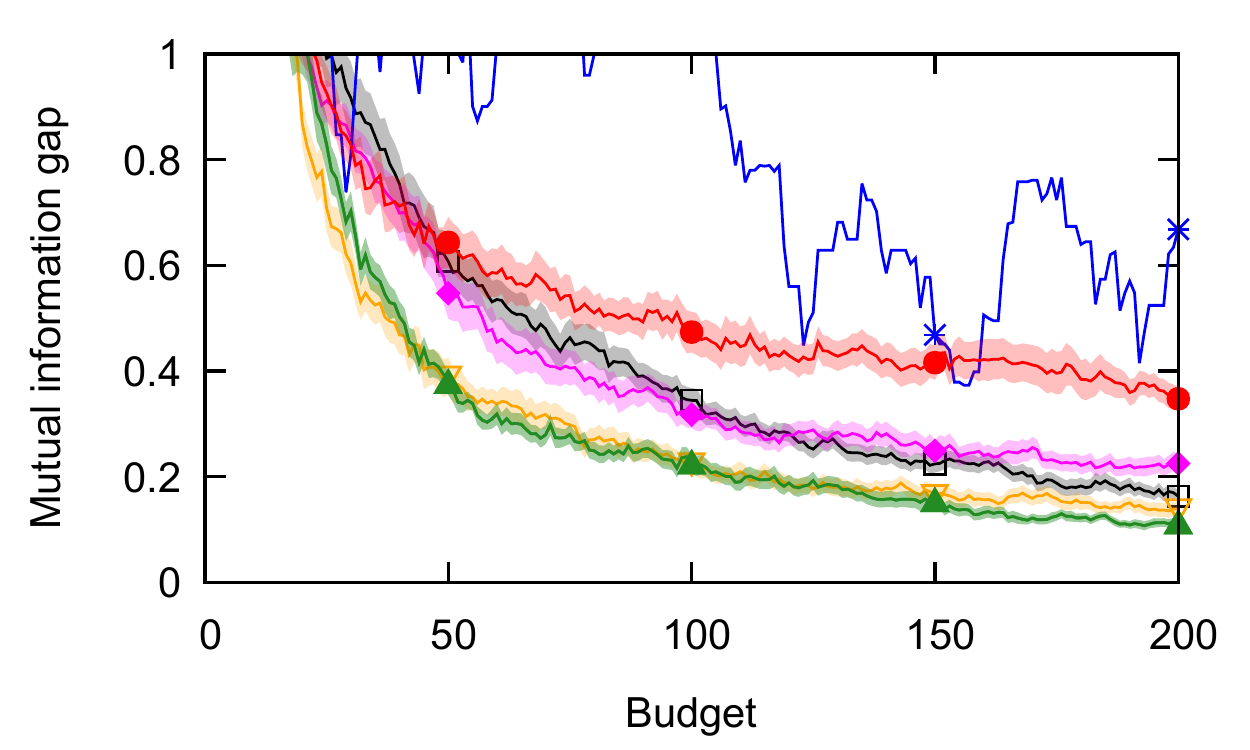}
    }
  \end{center}

\caption{\small Ablation tests: MNIST: 3 vs 5 ($k=20$). Top: full experiment. Bottom: Zoom in.}
\end{figure}

\begin{figure}[h]
  \begin{center}
    \myborder{
    \includegraphics[width = 0.4\textwidth]{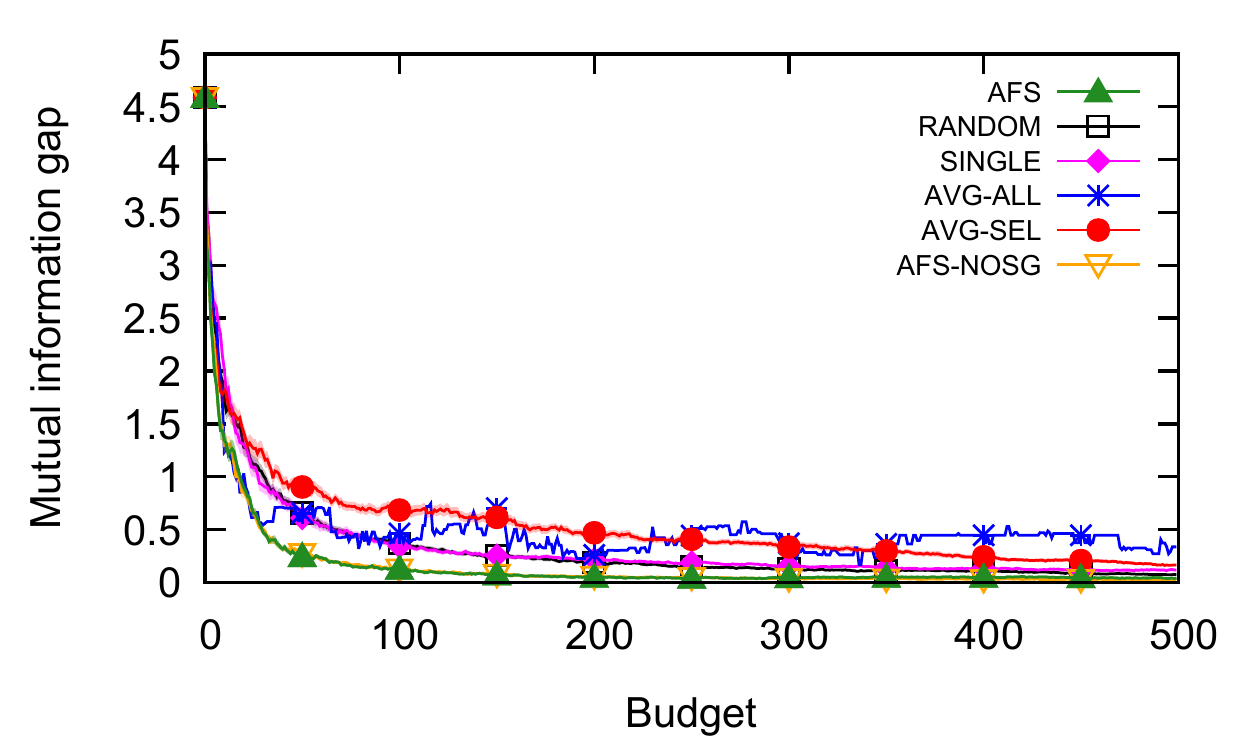} \\
    \includegraphics[width = 0.4\textwidth]{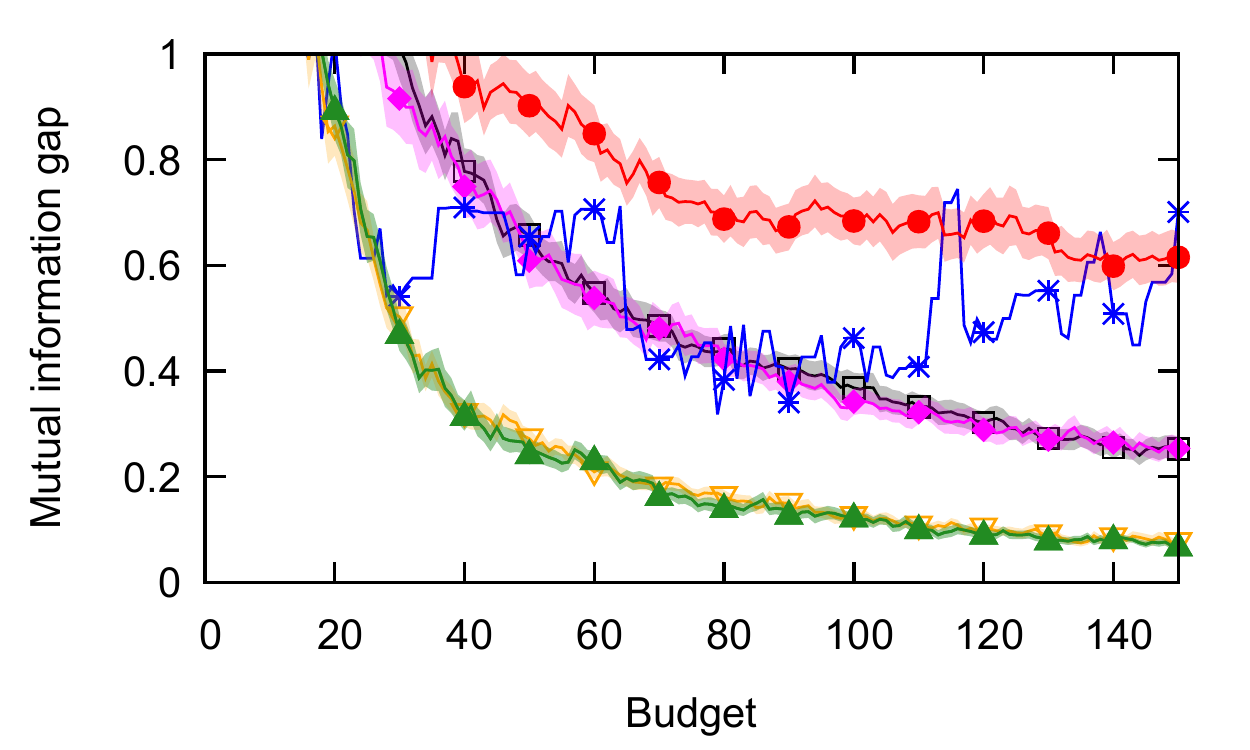}
    }
  \end{center}

\caption{\small Ablation tests: MNIST: 4 vs 6 ($k=20$). Top: full experiment. Bottom: Zoom in.}
\end{figure}

\clearpage
\subsection{Ablation tests: $k = 10$}

\begin{figure}[h]
  \begin{center}
    \myborder{
    \includegraphics[width = 0.4\textwidth]{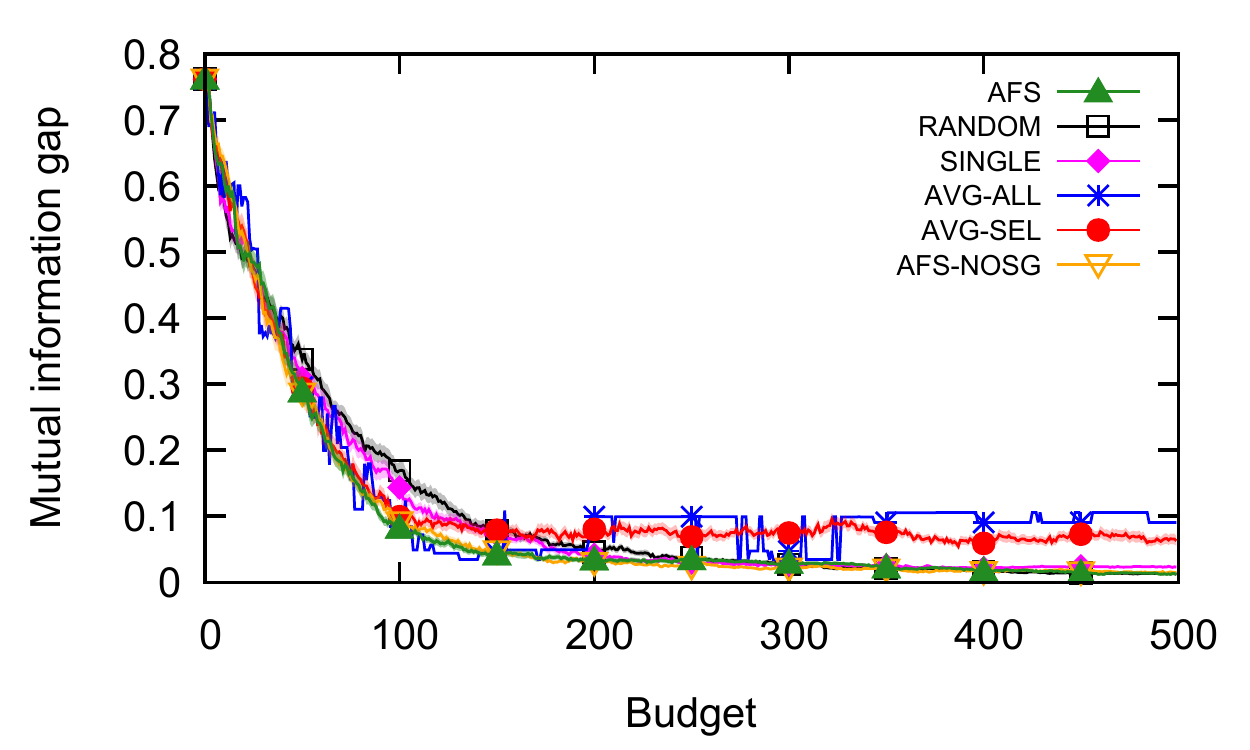} \\
    \includegraphics[width = 0.4\textwidth]{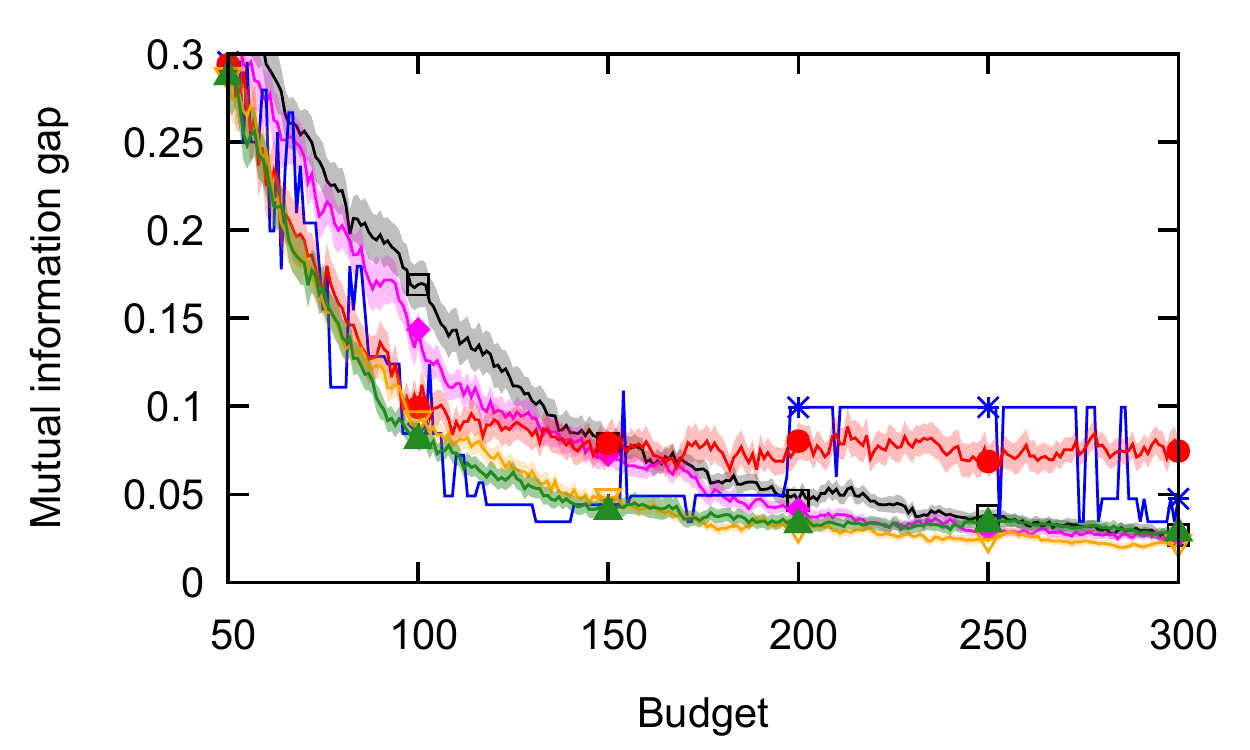}
    }
  \end{center}

\caption{\small Ablation tests: BASEHOCK ($k=10$). Top: full experiment. Bottom: Zoom in.}
\end{figure}

\begin{figure}[h]
  \begin{center}
    \myborder{
    \includegraphics[width = 0.4\textwidth]{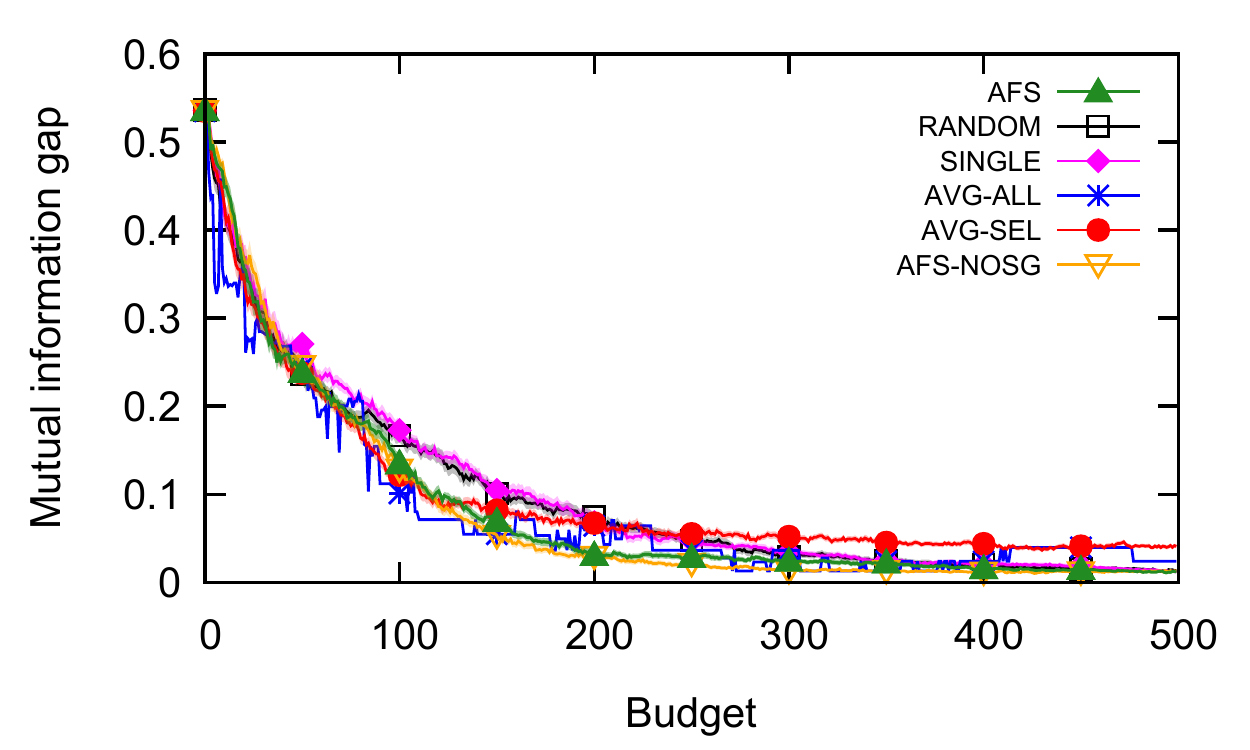} \\
    \includegraphics[width = 0.4\textwidth]{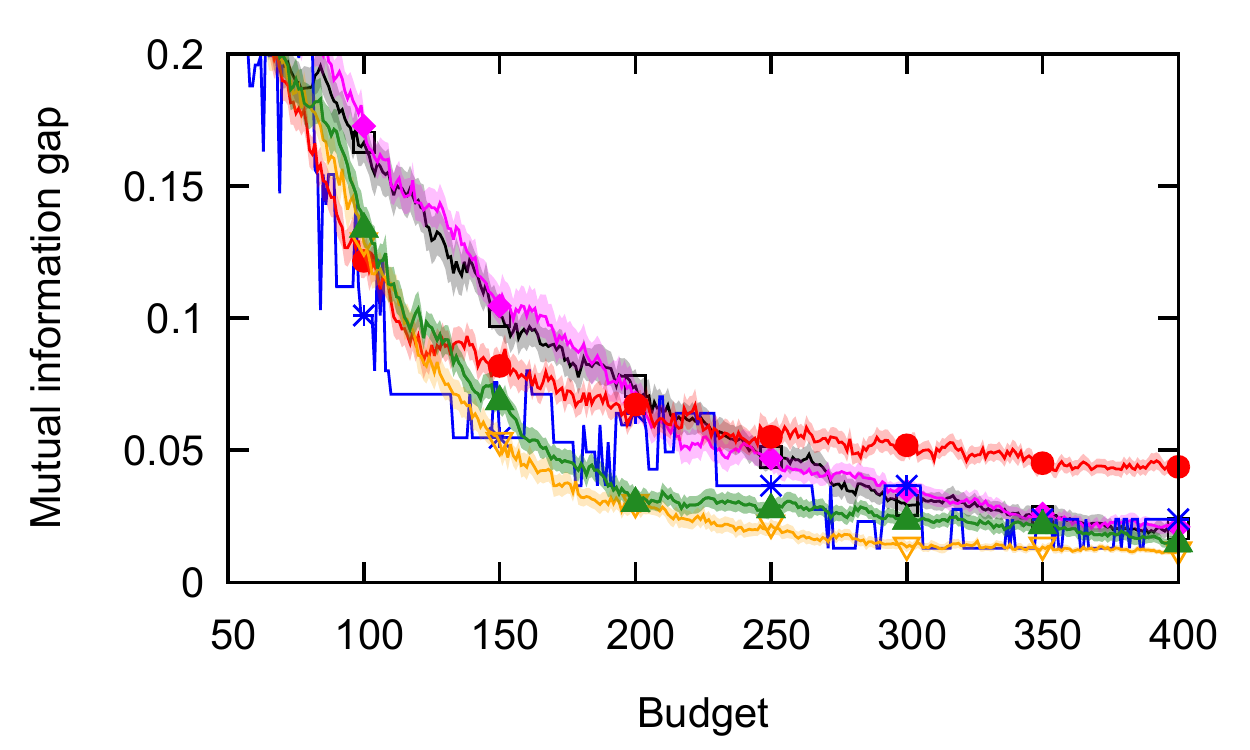}
    }
  \end{center}

\caption{\small Ablation tests: PCMAC ($k=10$). Top: full experiment. Bottom: Zoom in.}
\end{figure}

\begin{figure}[h]
  \begin{center}
    \myborder{
    \includegraphics[width = 0.4\textwidth]{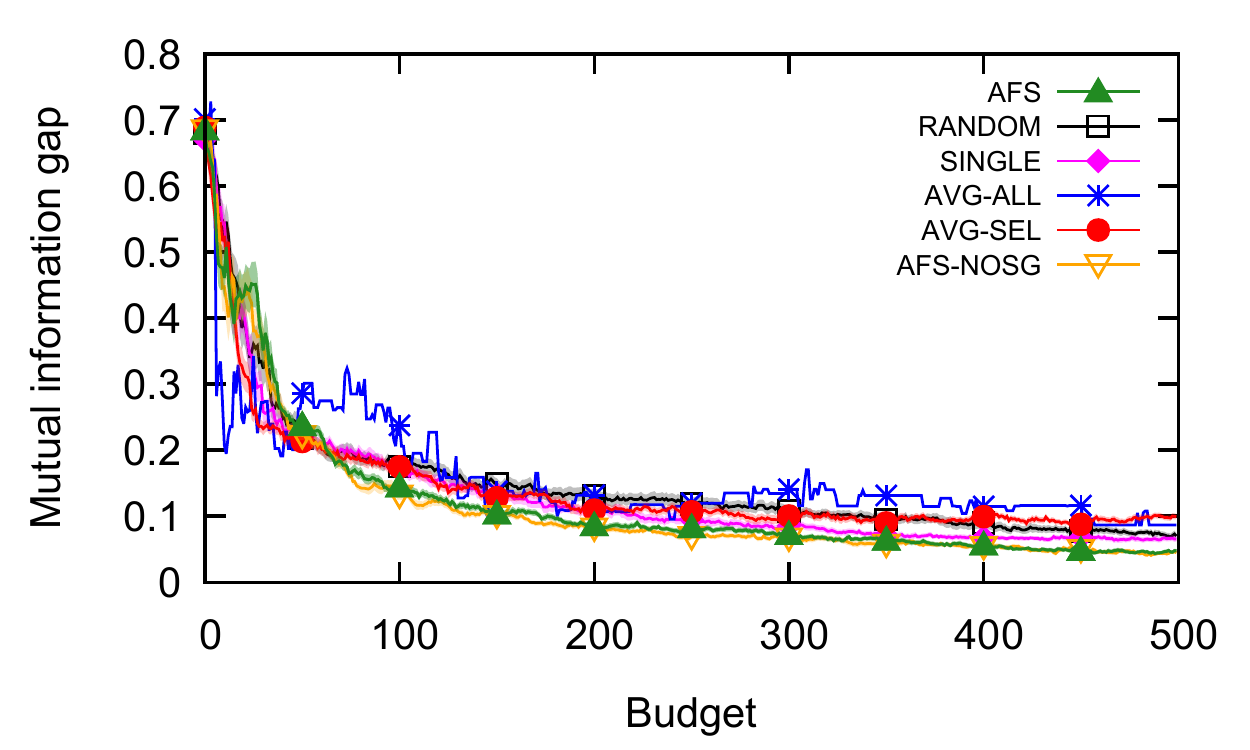} \\
    \includegraphics[width = 0.4\textwidth]{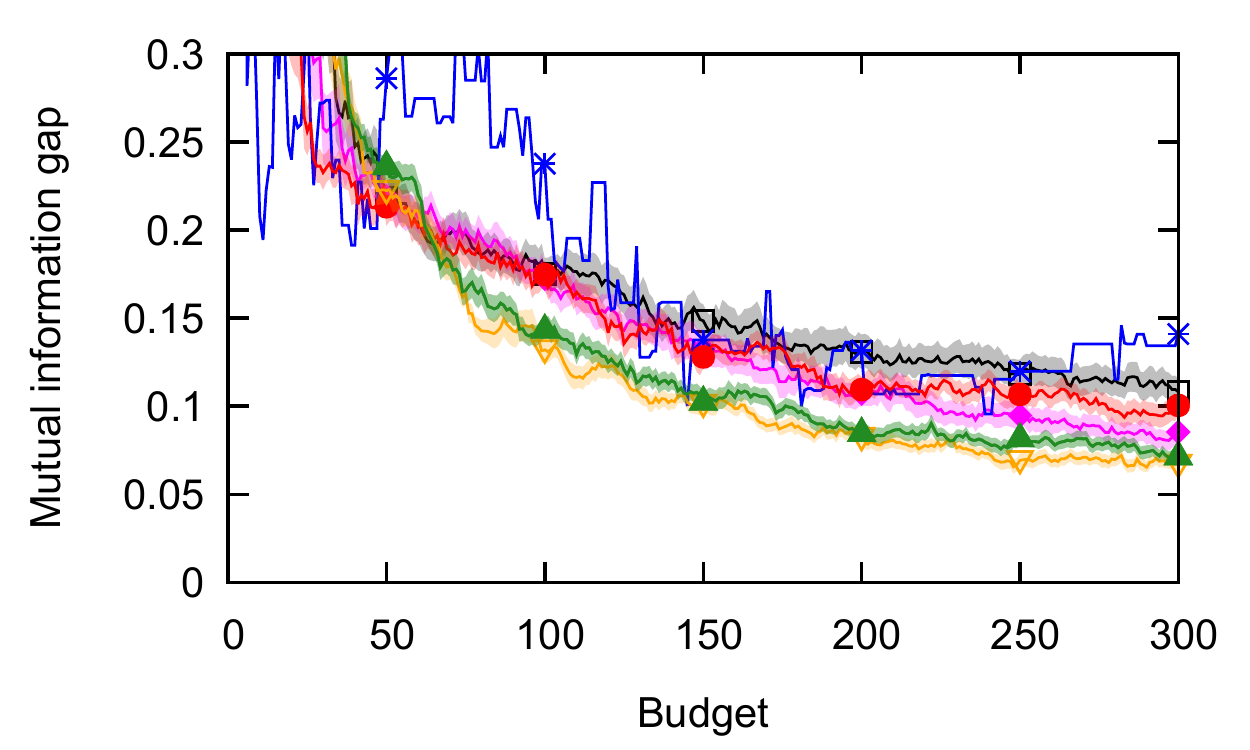}
    }
  \end{center}

\caption{\small Ablation tests: MUSK ($k=10$). Top: full experiment. Bottom: Zoom in.}
\end{figure}

\begin{figure}[h]
  \begin{center}
    \myborder{
    \includegraphics[width = 0.4\textwidth]{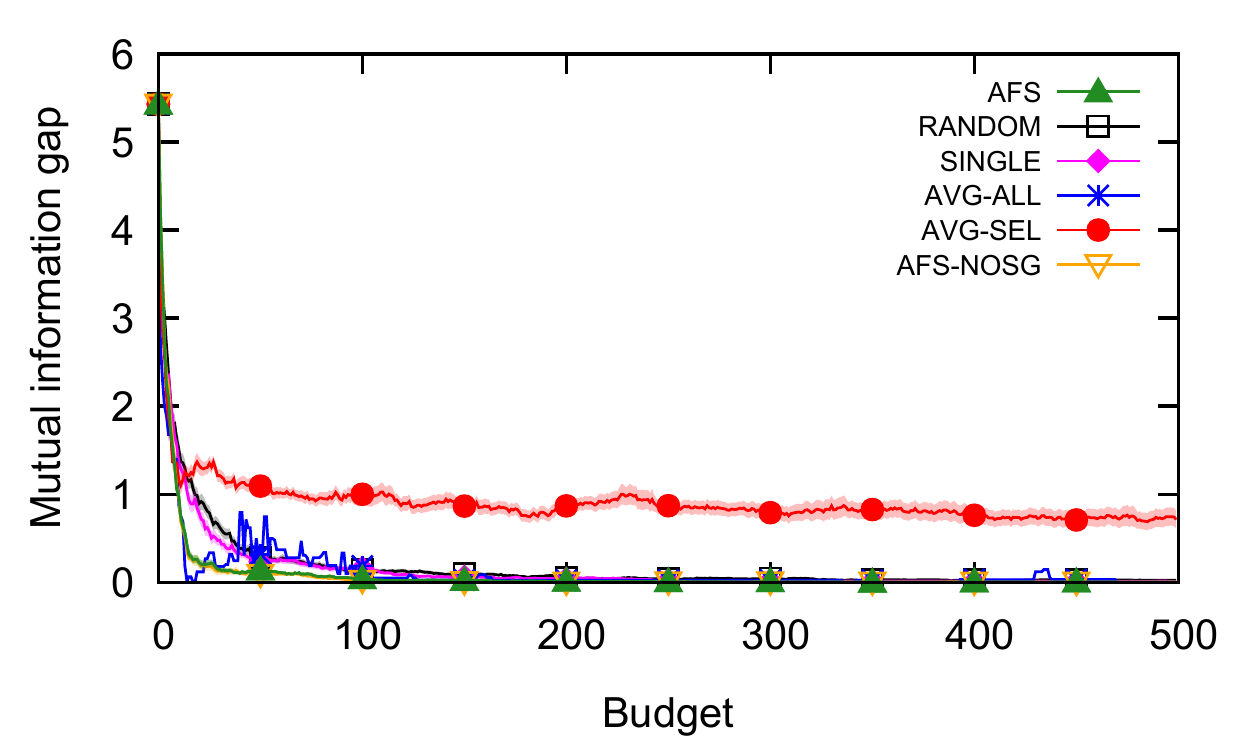} \\
    \includegraphics[width = 0.4\textwidth]{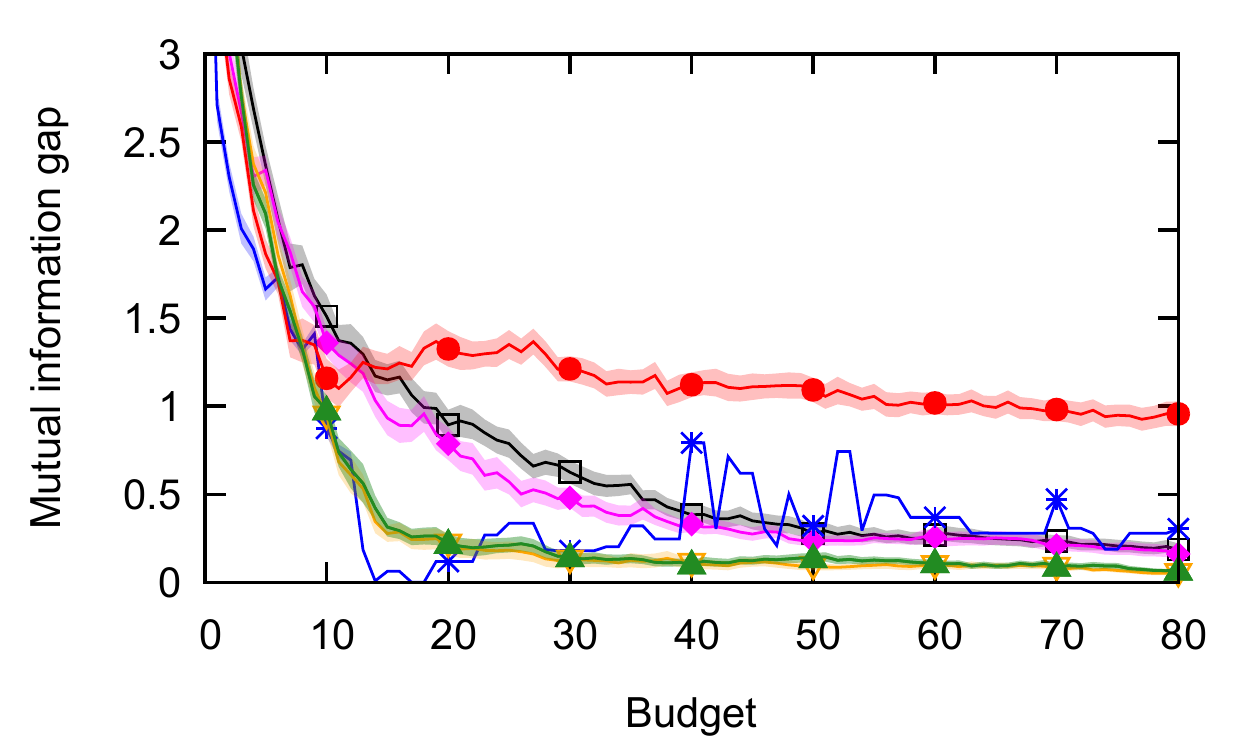}
    }
  \end{center}

\caption{\small Ablation tests: MNIST: 0 vs 1 ($k=10$). Top: full experiment. Bottom: Zoom in.}
\end{figure}

\begin{figure}[h]
  \begin{center}
    \myborder{
    \includegraphics[width = 0.4\textwidth]{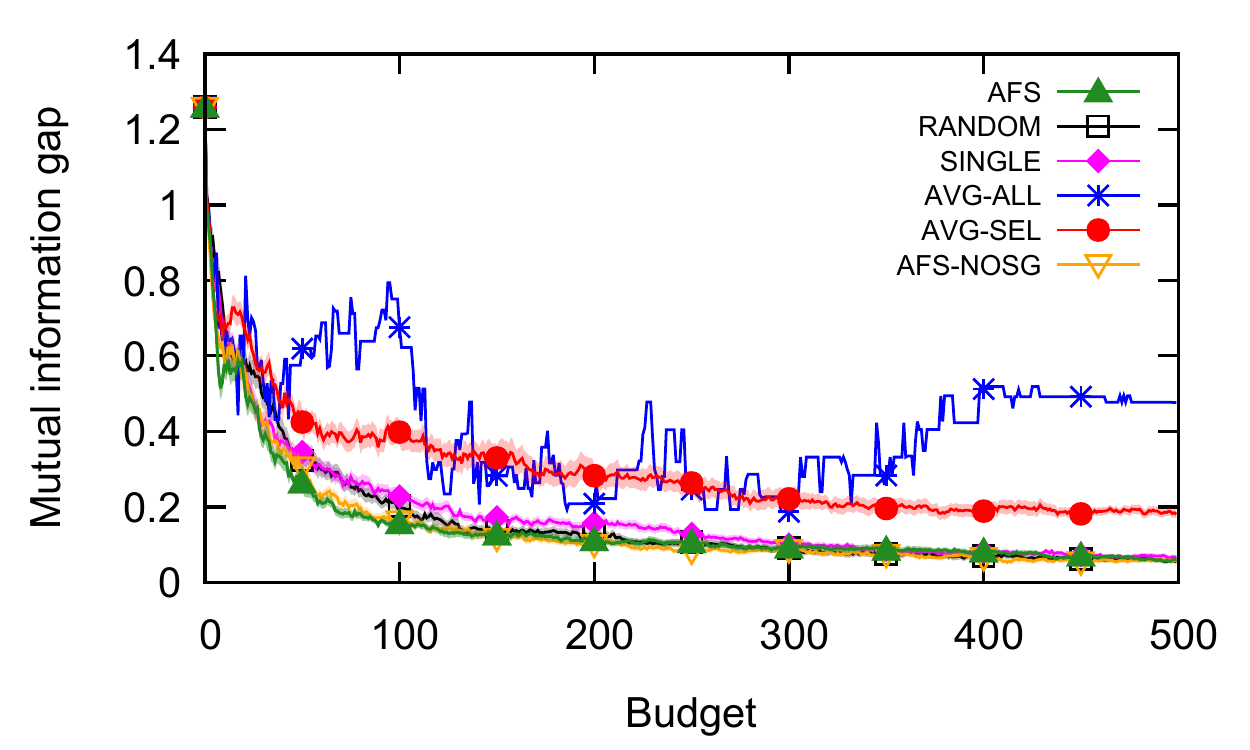} \\
    \includegraphics[width = 0.4\textwidth]{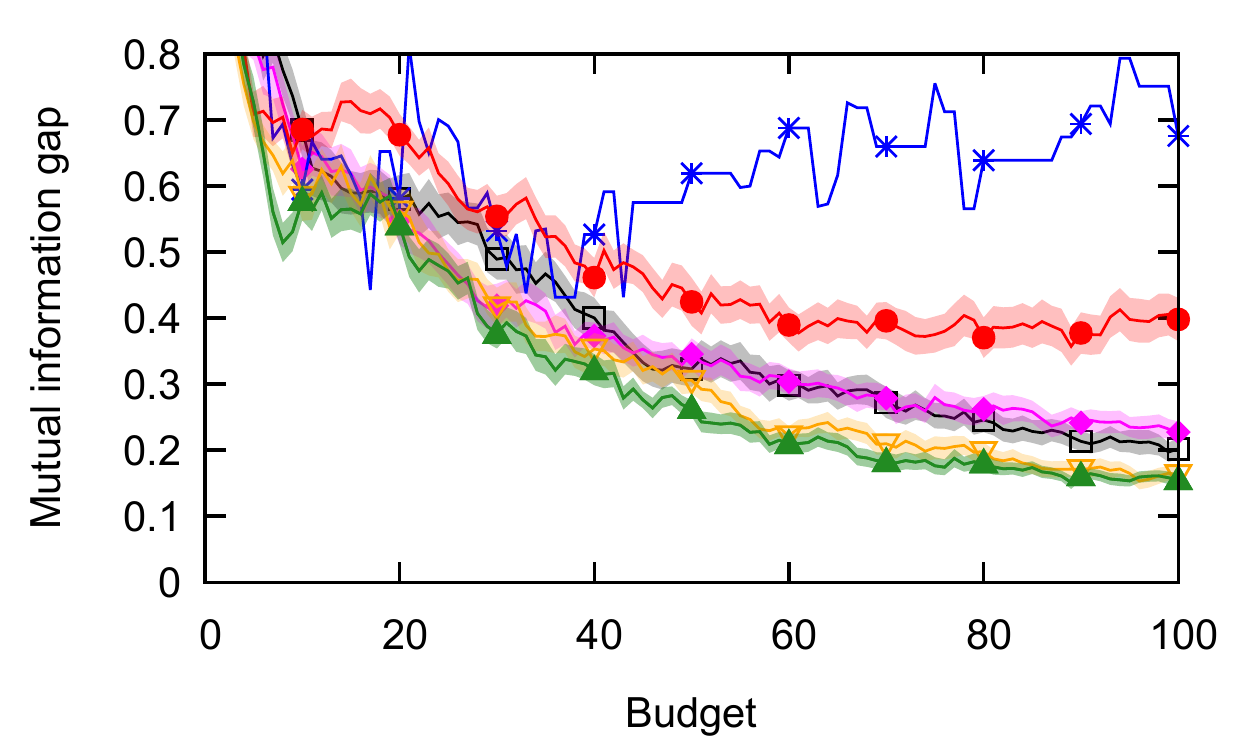}
    }
  \end{center}

\caption{\small Ablation tests: MNIST: 3 vs 5 ($k=10$). Top: full experiment. Bottom: Zoom in.}
\end{figure}

\begin{figure}[h]
  \begin{center}
    \myborder{
    \includegraphics[width = 0.4\textwidth]{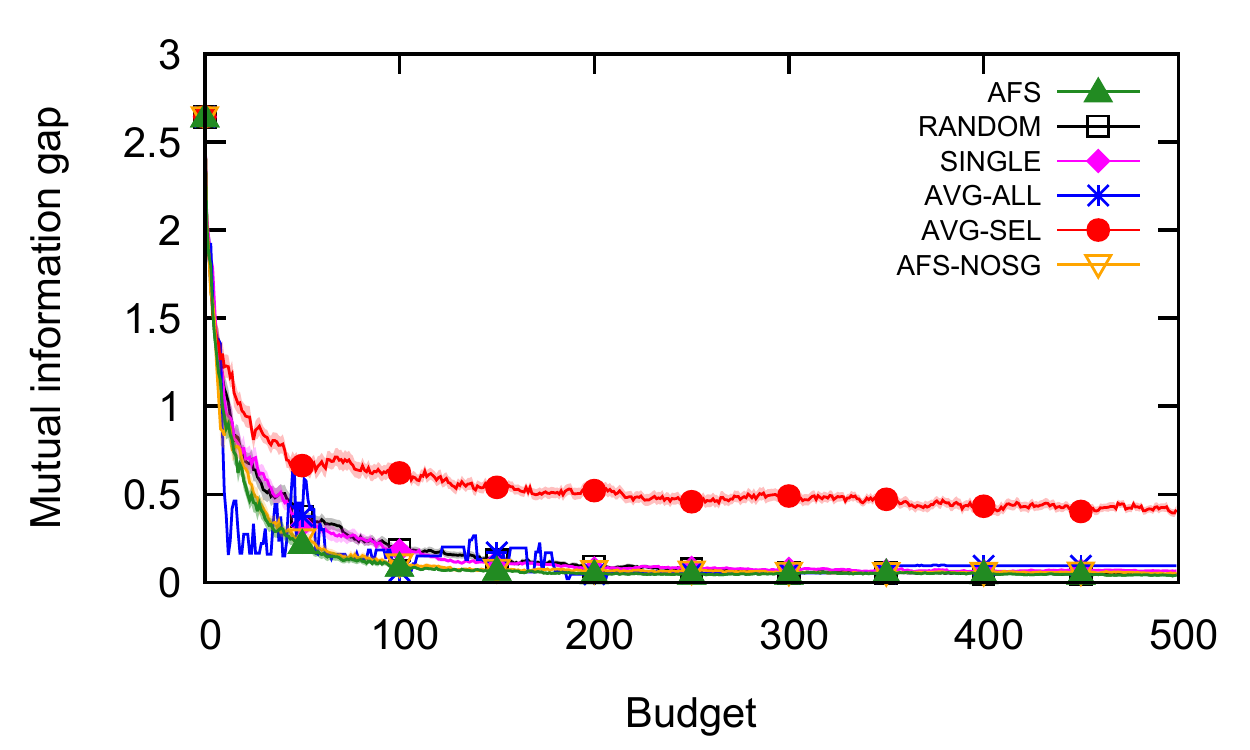} \\
    \includegraphics[width = 0.4\textwidth]{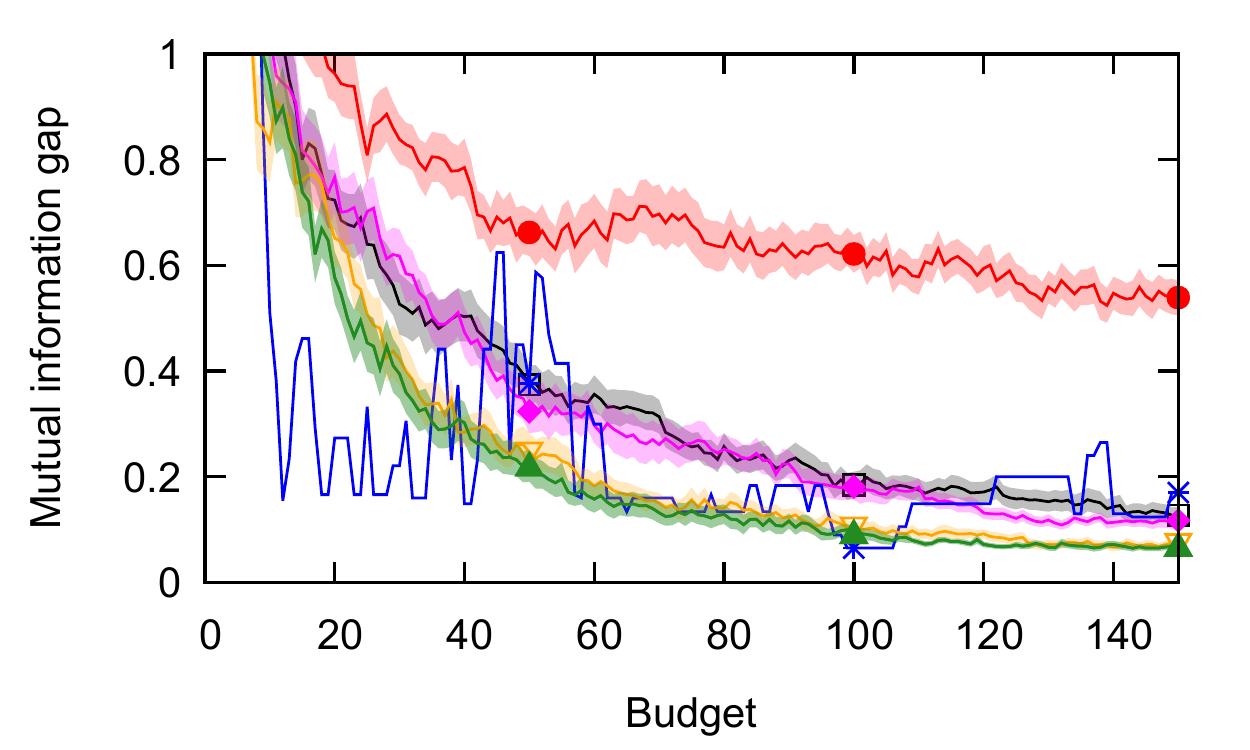}
    }
  \end{center}

\caption{\small Ablation tests: MNIST: 4 vs 6 ($k=10$). Top: full experiment. Bottom: Zoom in.}
\end{figure}

\clearpage
\subsection{Ablation tests: $k = 5$}

\begin{figure}[h]
  \begin{center}
    \myborder{
    \includegraphics[width = 0.4\textwidth]{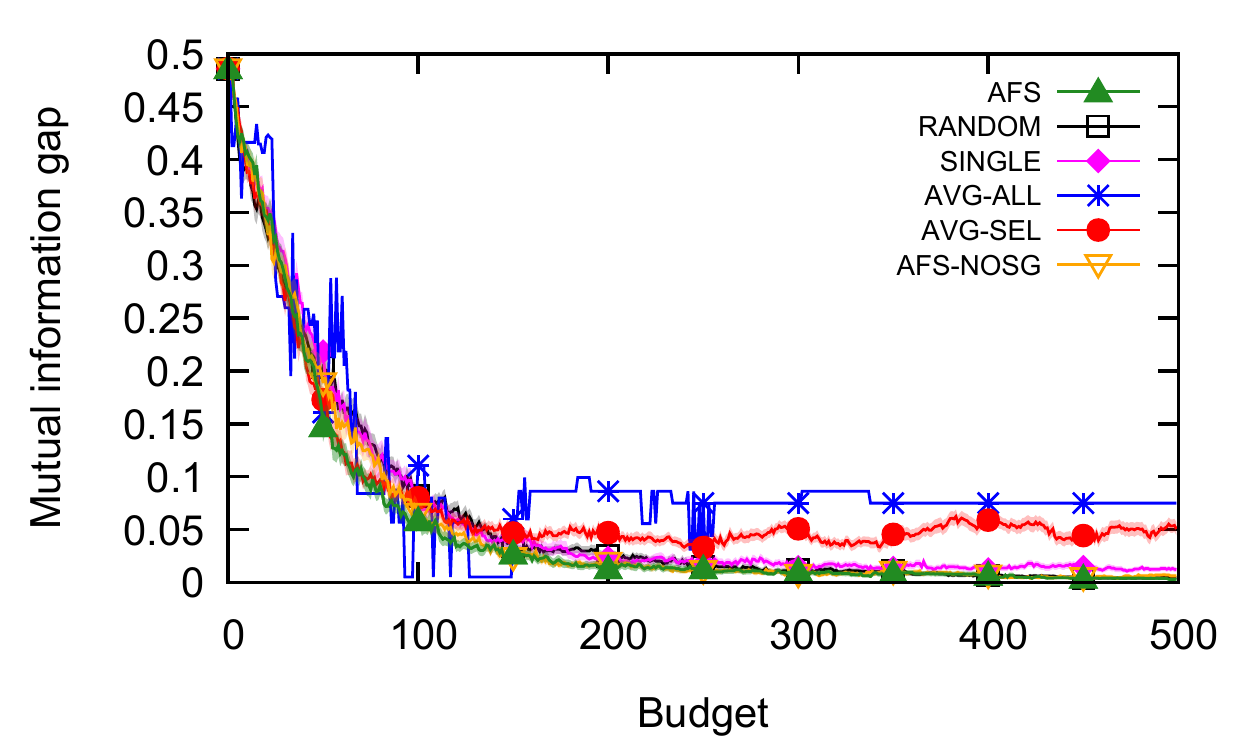} \\
    \includegraphics[width = 0.4\textwidth]{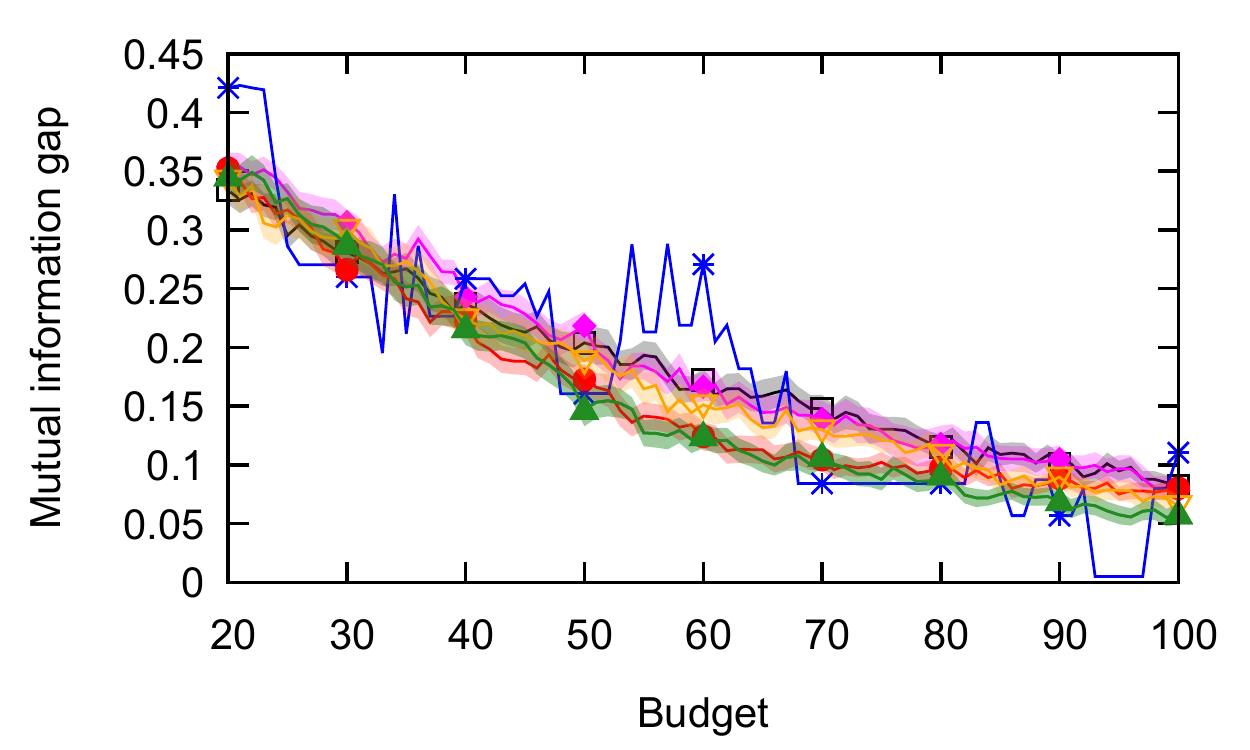}
    }
  \end{center}

\caption{\small Ablation tests: BASEHOCK ($k=5$). Top: full experiment. Bottom: Zoom in.}
\end{figure}

\begin{figure}[h]
  \begin{center}
    \myborder{
    \includegraphics[width = 0.4\textwidth]{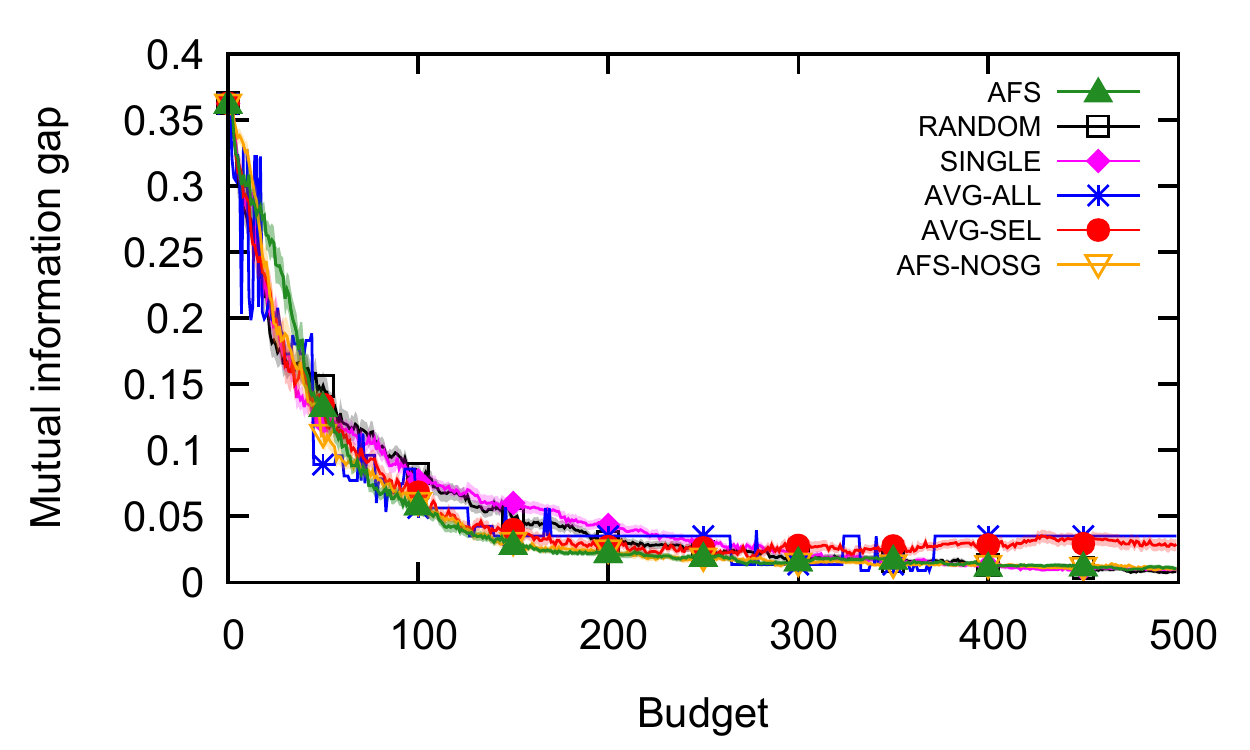} \\
    \includegraphics[width = 0.4\textwidth]{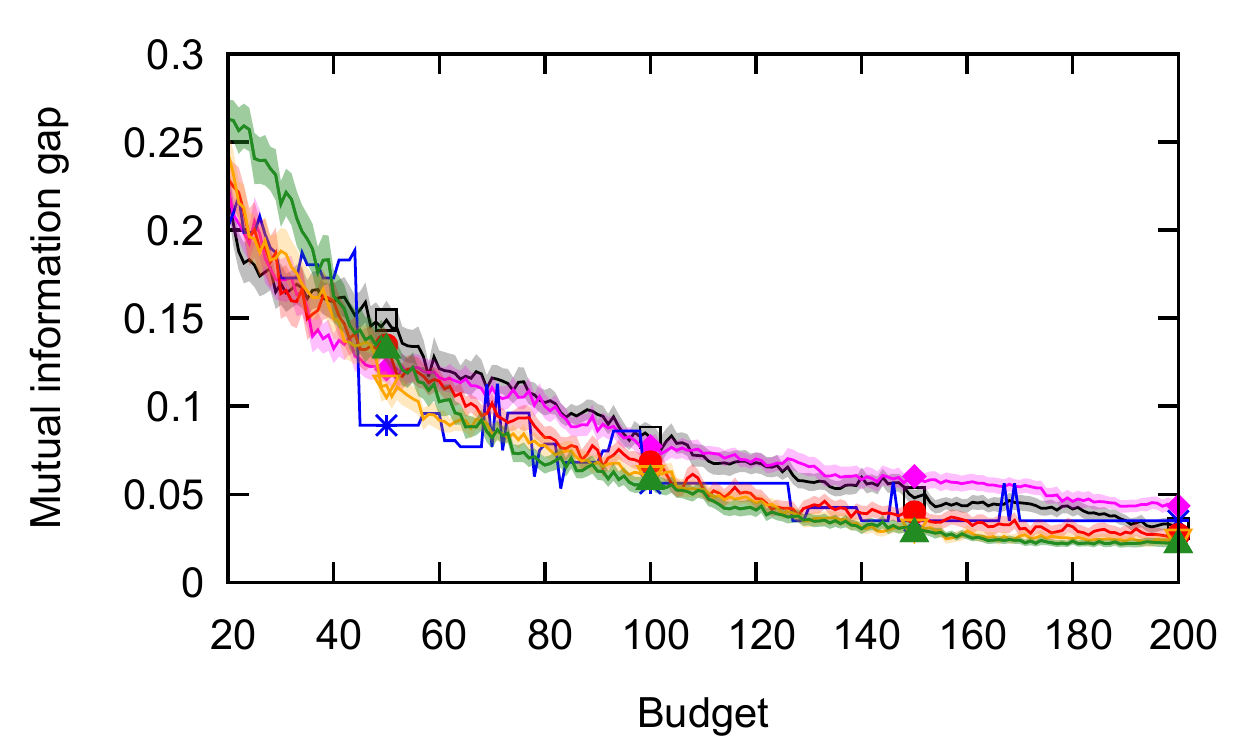}
    }
  \end{center}

\caption{\small Ablation tests: PCMAC ($k=5$). Top: full experiment. Bottom: Zoom in.}
\end{figure}

\begin{figure}[h]
  \begin{center}
    \myborder{
    \includegraphics[width = 0.4\textwidth]{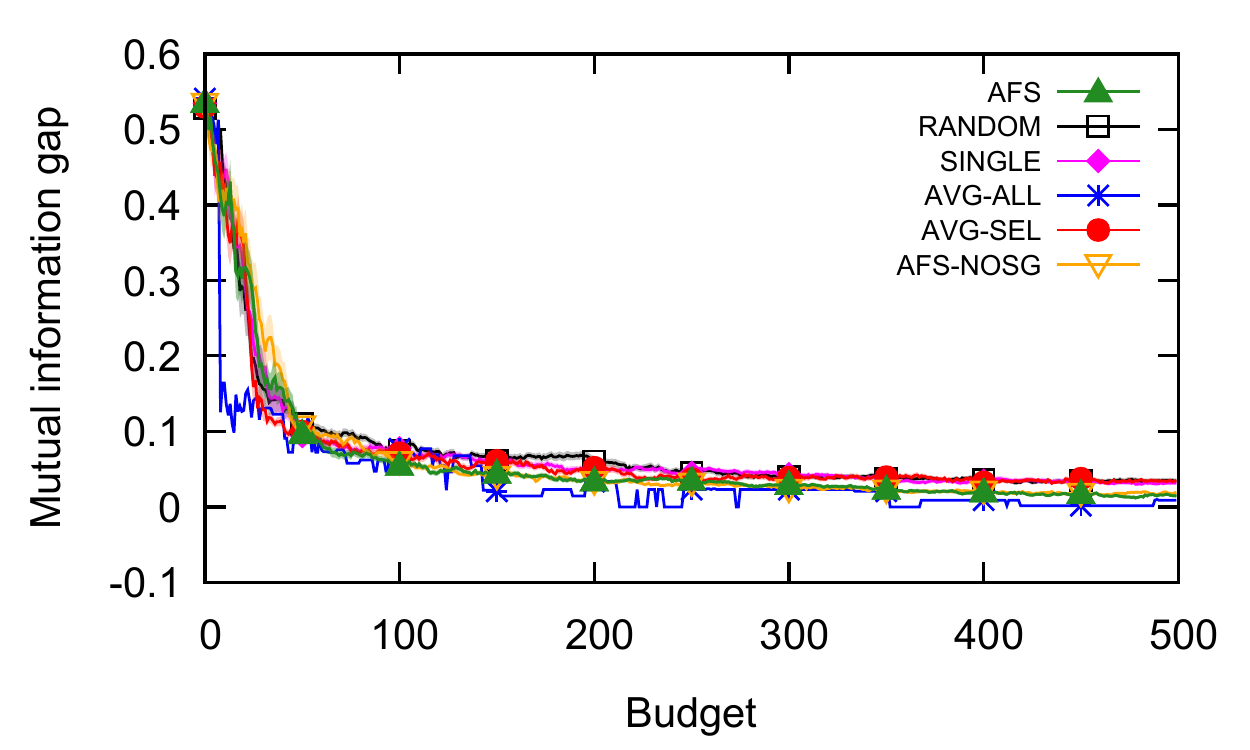} \\
    \includegraphics[width = 0.4\textwidth]{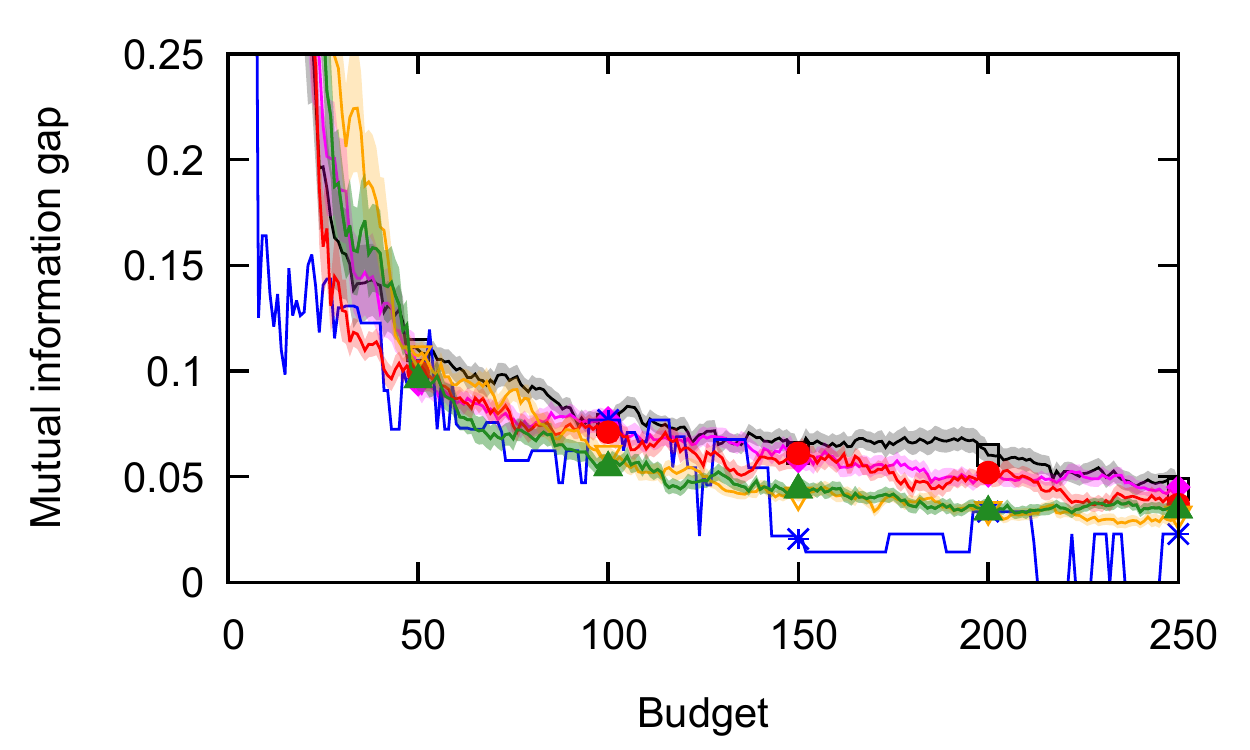}
    }
  \end{center}

\caption{\small Ablation tests: MUSK ($k=5$). Top: full experiment. Bottom: Zoom in.}
\end{figure}

\begin{figure}[h]
  \begin{center}
    \myborder{
    \includegraphics[width = 0.4\textwidth]{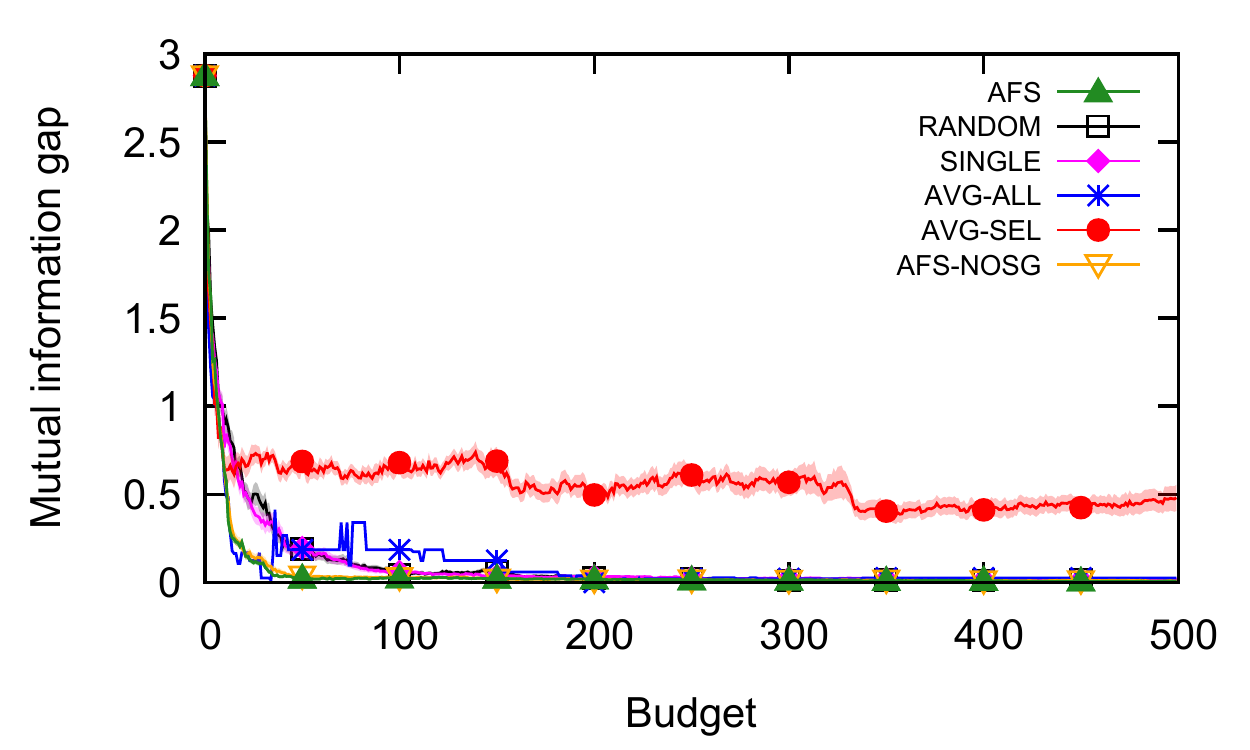} \\
    \includegraphics[width = 0.4\textwidth]{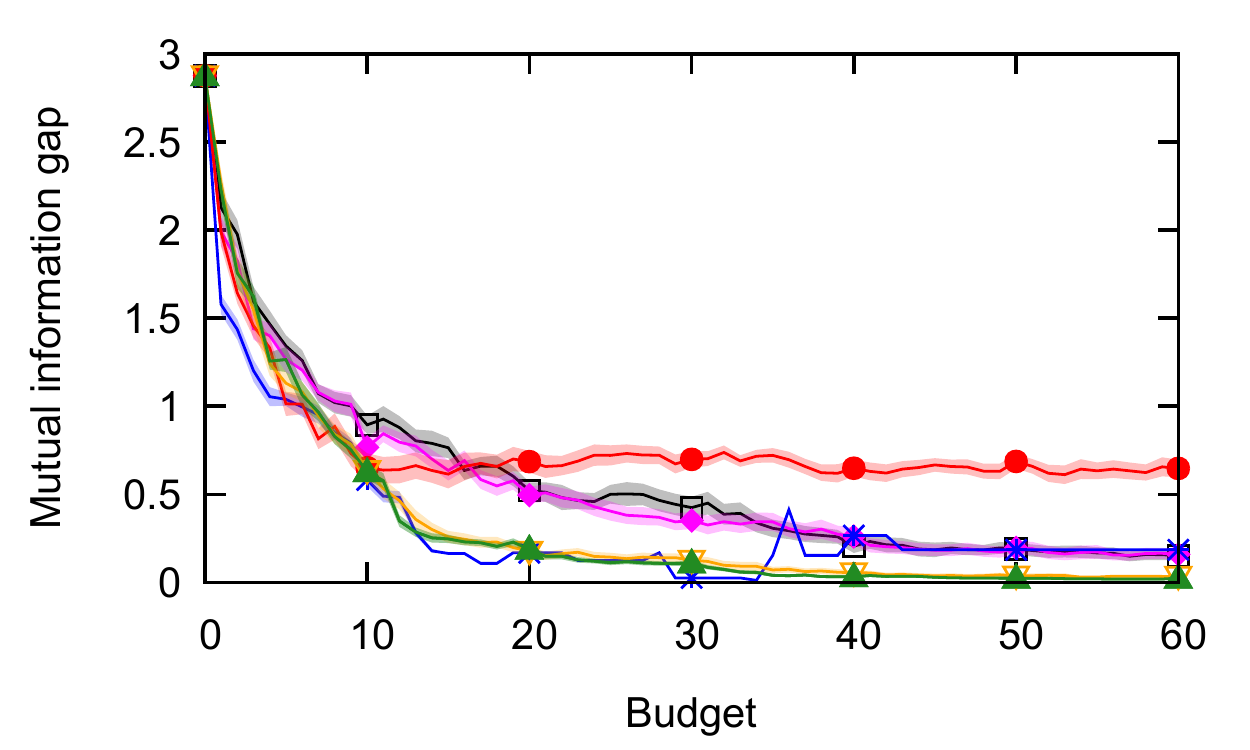}
    }
  \end{center}

\caption{\small Ablation tests: MNIST: 0 vs 1 ($k=5$). Top: full experiment. Bottom: Zoom in.}
\end{figure}

\begin{figure}[h]
  \begin{center}
    \myborder{
    \includegraphics[width = 0.4\textwidth]{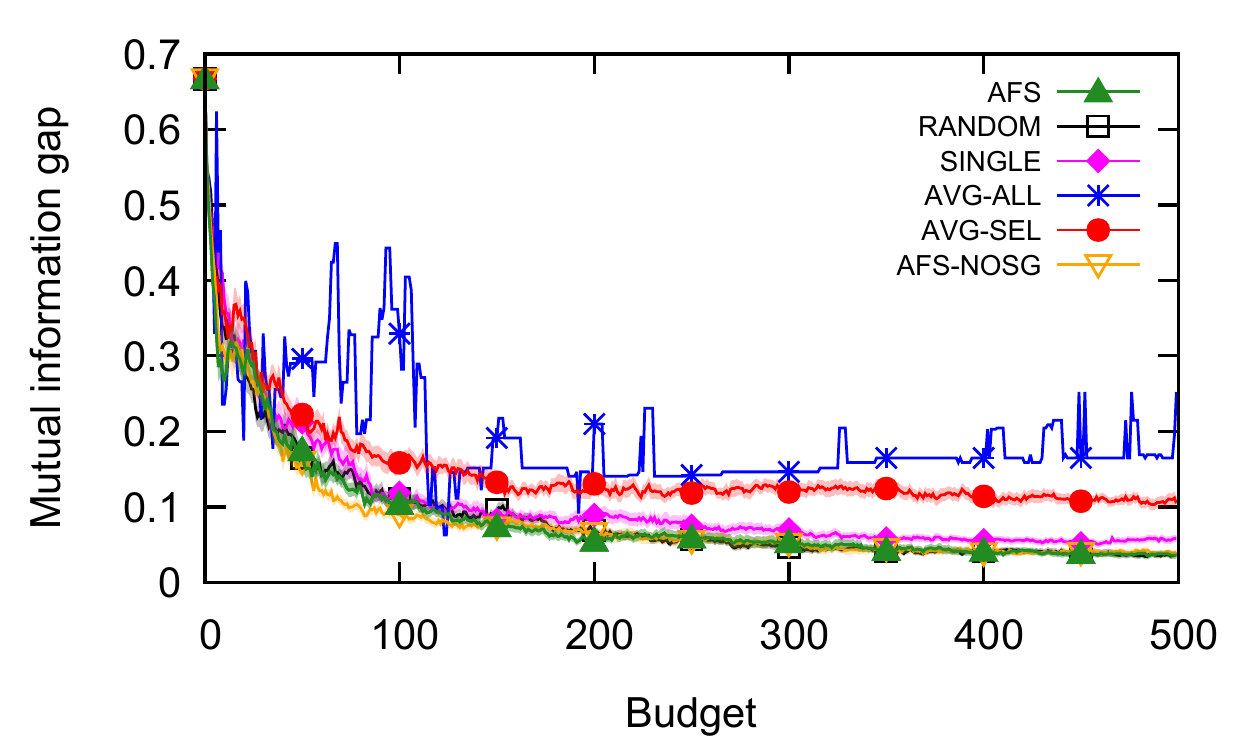} \\
    \includegraphics[width = 0.4\textwidth]{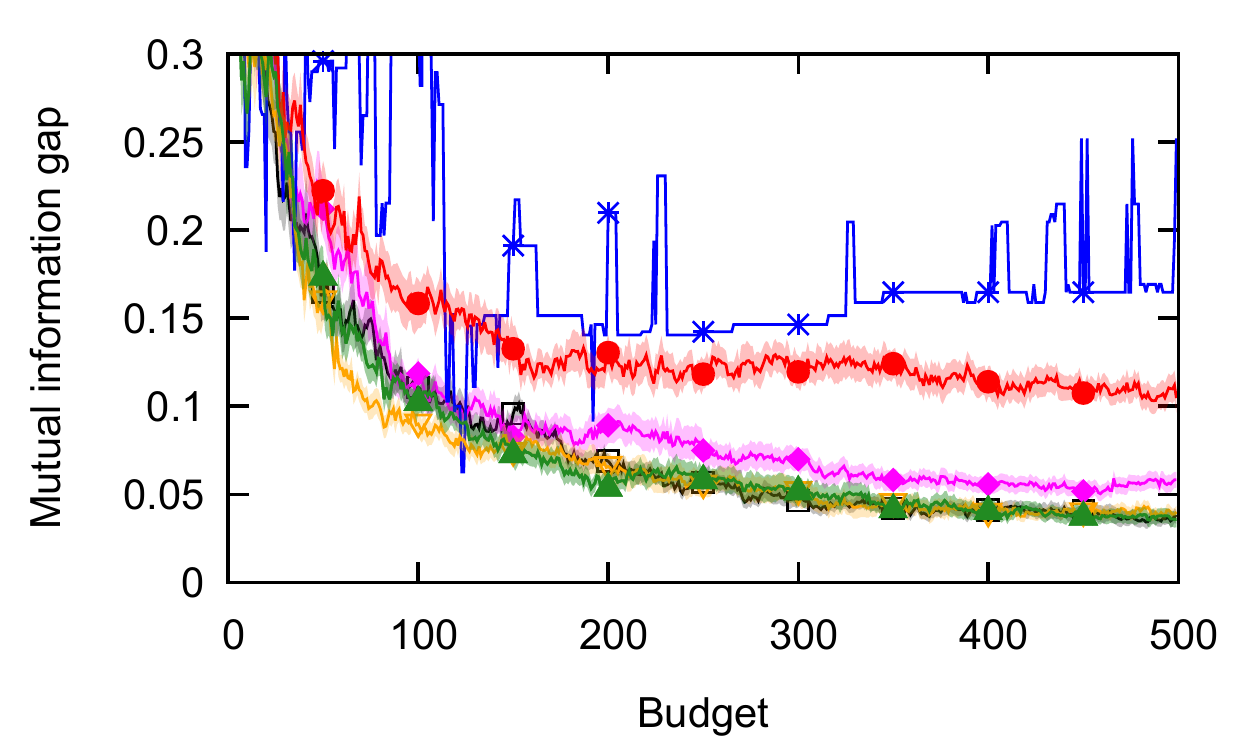}
    }
  \end{center}

\caption{\small Ablation tests: MNIST: 3 vs 5 ($k=5$). Top: full experiment. Bottom: Zoom in.}
\end{figure}

\begin{figure}[h]
  \begin{center}
    \myborder{
    \includegraphics[width = 0.4\textwidth]{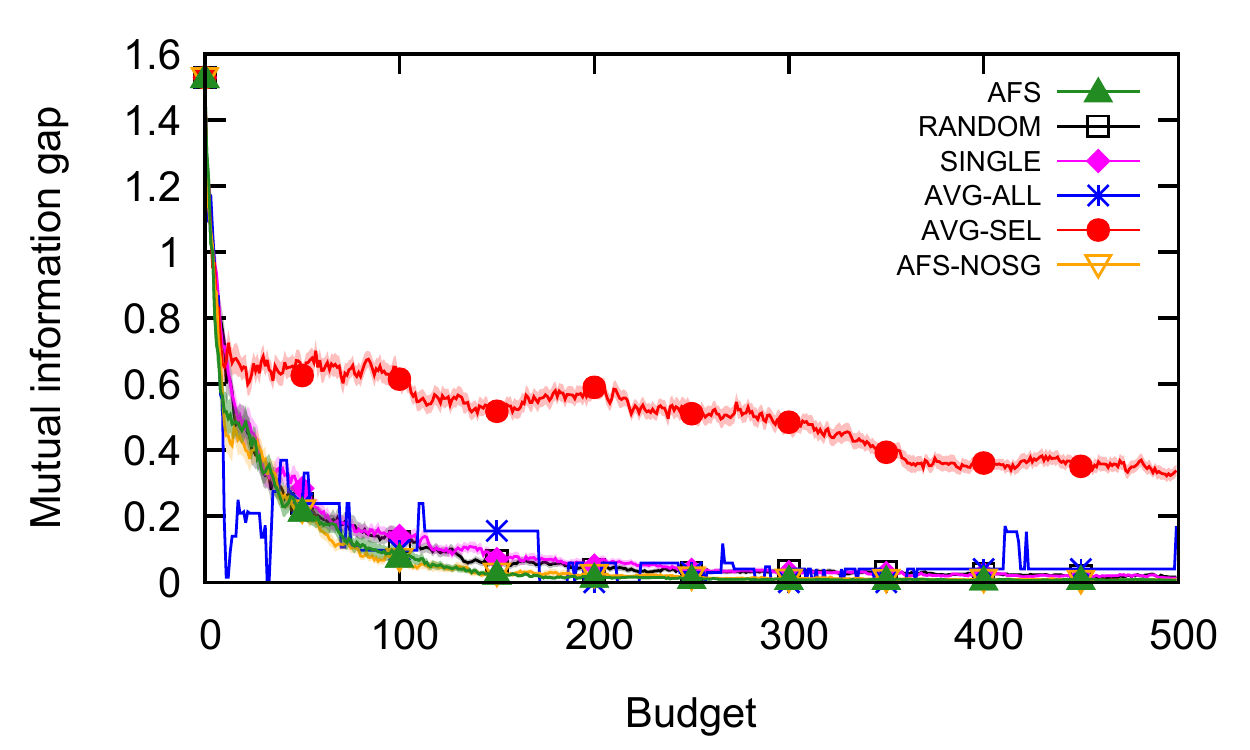} \\
    \includegraphics[width = 0.4\textwidth]{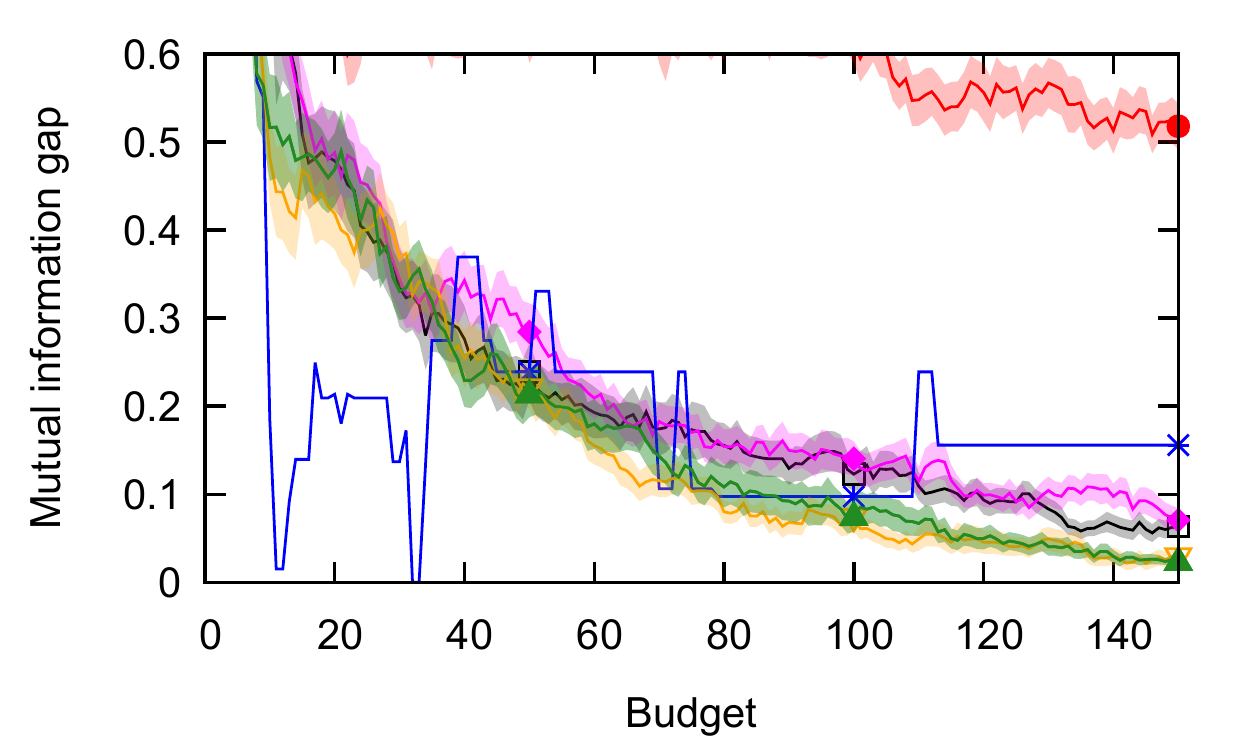}
    }
  \end{center}

\caption{\small Ablation tests: MNIST: 4 vs 6 ($k=5$). Top: full experiment. Bottom: Zoom in.}
\end{figure}

\clearpage
\section{Experiments: Comparing aggregation functions}\label{app:aggregate}
We report experiments comparing possible choices of the aggregation function
$\psi$. We tested the $\ell_1,\ell_{\infty}$ and $\ell_2$ norms. It can be
seen that the results are similar for $\ell_1$ and $\ell_2$, while $\ell_\infty$ is sometimes slightly worse. Thus, while we chose
$\ell_1$, setting $\psi :=
\ell_2$ is also a valid choice.  For easy comparison, the graphs show the
RANDOM baseline as well.

\subsection{Comparing aggregation functions: $k = 20$}

\begin{figure}[h]
  \begin{center}
    \myborder{
    \includegraphics[width = 0.4\textwidth]{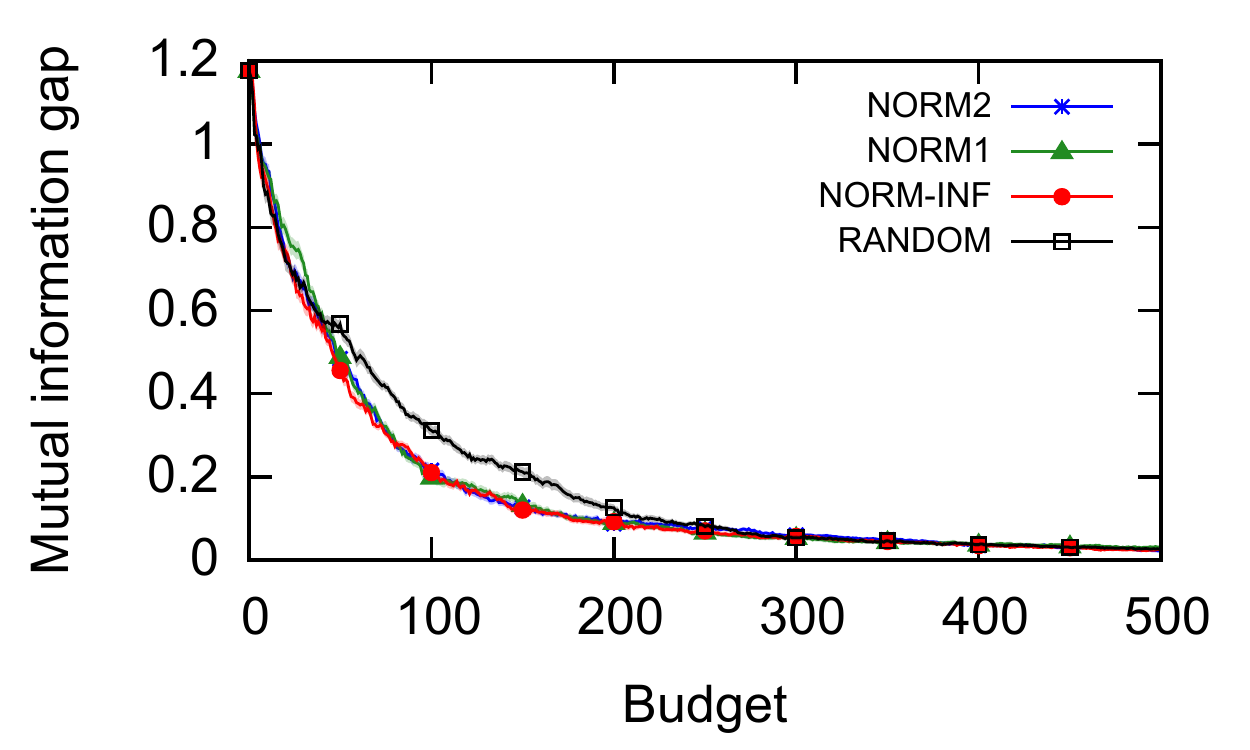} \\
    \includegraphics[width = 0.4\textwidth]{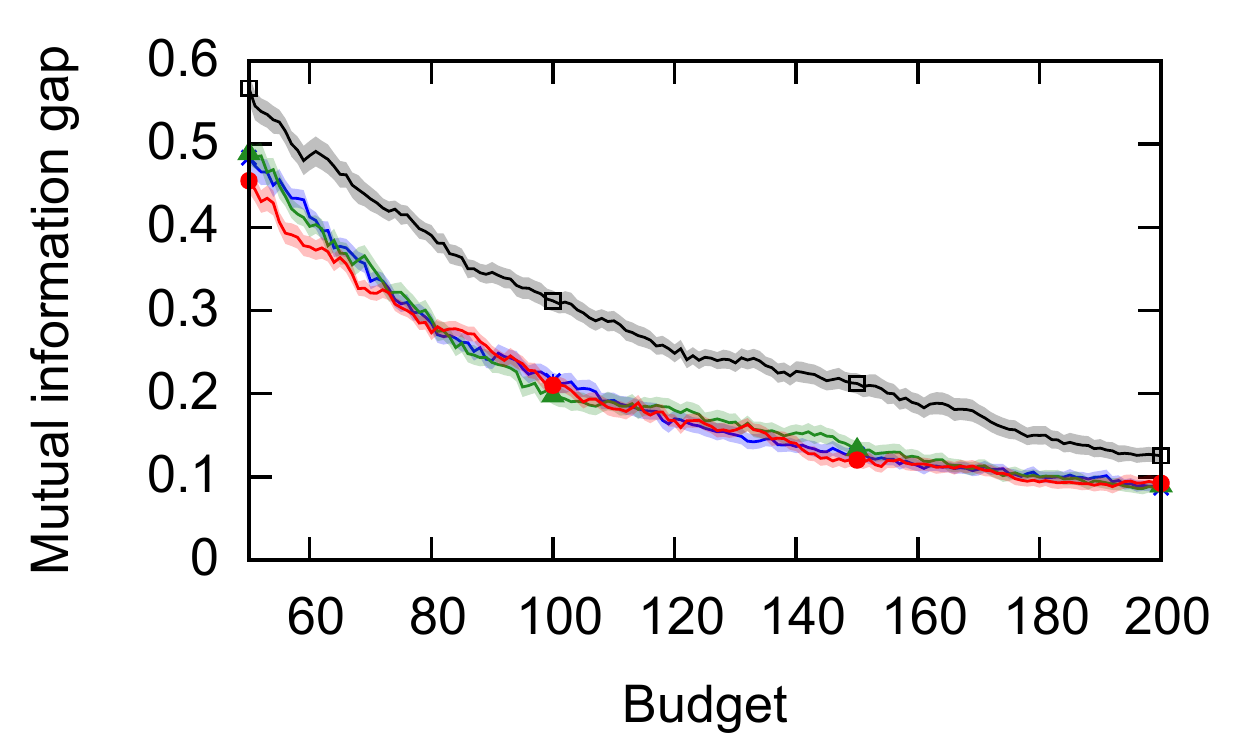}
    }
  \end{center}

\caption{\small Choices of $\psi$: BASEHOCK ($k=20$). Top: full experiment. Bottom: Zoom in.}
\end{figure}

\begin{figure}[h]
  \begin{center}
    \myborder{
    \includegraphics[width = 0.4\textwidth]{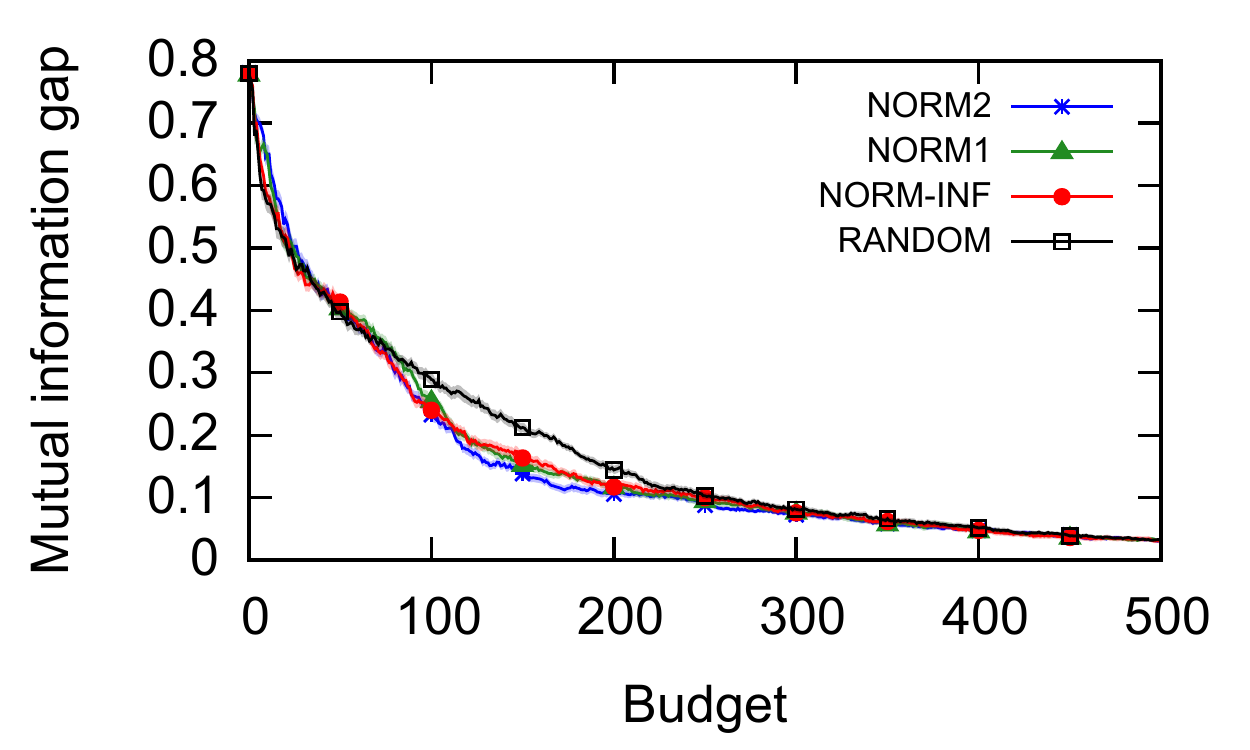} \\
    \includegraphics[width = 0.4\textwidth]{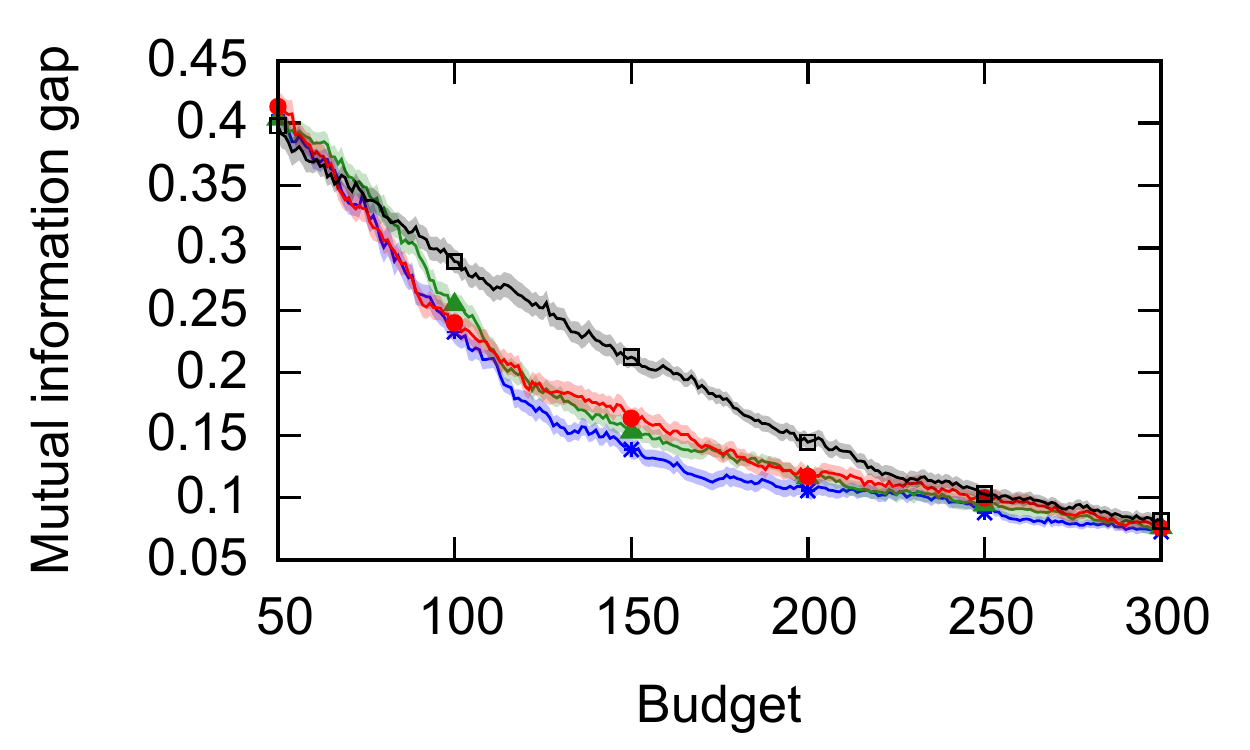}
    }
  \end{center}

\caption{\small Choices of $\psi$: PCMAC ($k=20$). Top: full experiment. Bottom: Zoom in.}
\end{figure}

\begin{figure}[h]
  \begin{center}
    \myborder{
    \includegraphics[width = 0.4\textwidth]{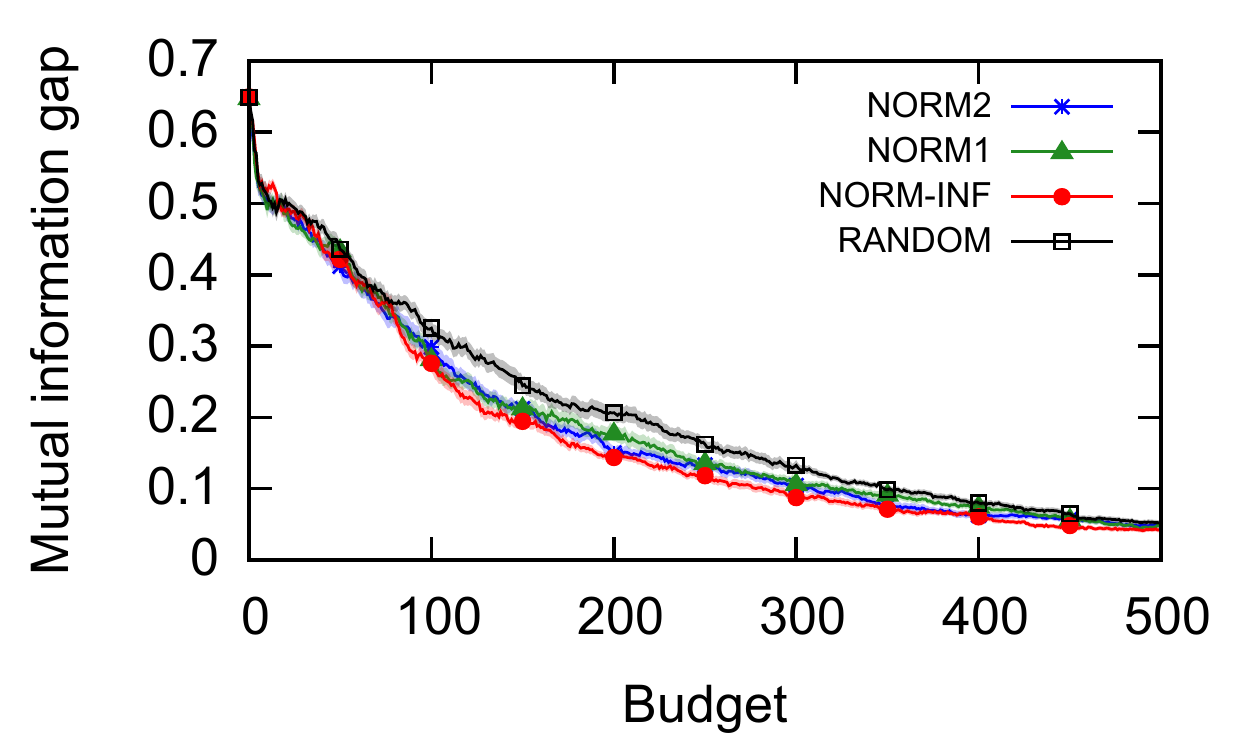} \\
    \includegraphics[width = 0.4\textwidth]{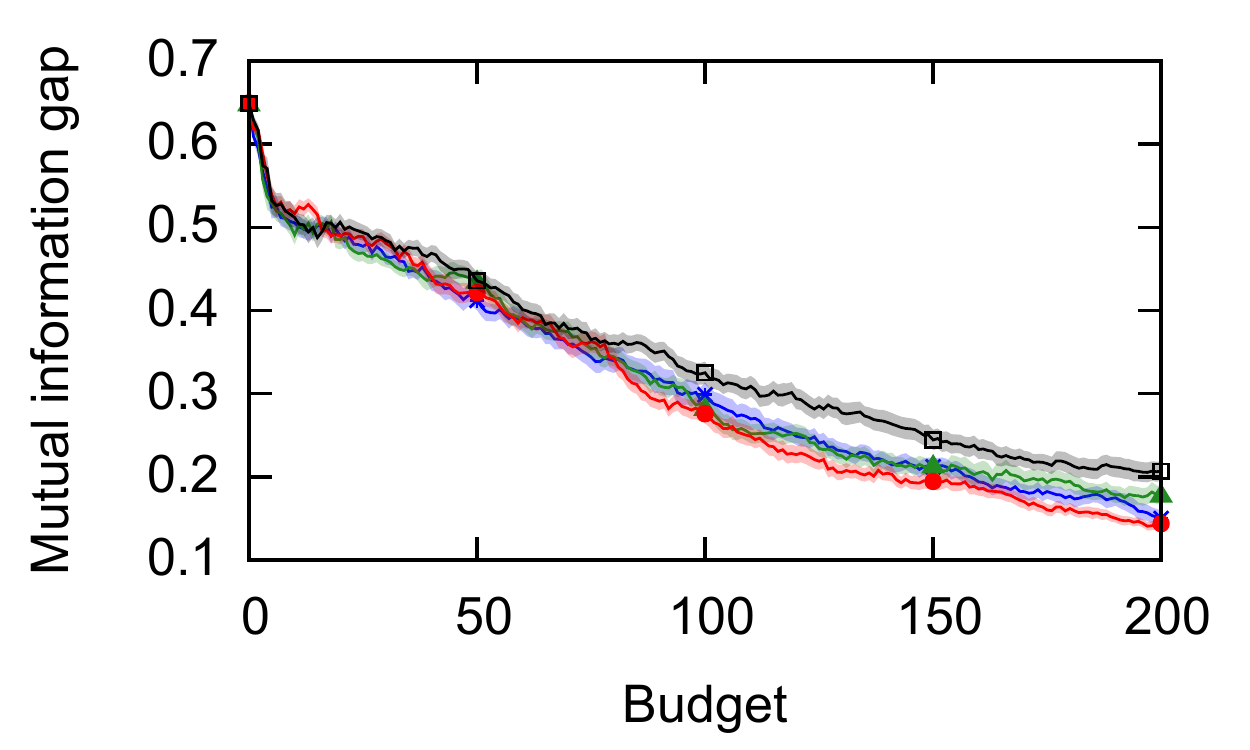}
    }
  \end{center}

\caption{\small Choices of $\psi$: RELATHE ($k=20$). Top: full experiment. Bottom: Zoom in.}
\end{figure}

\begin{figure}[h]
  \begin{center}
    \myborder{
    \includegraphics[width = 0.4\textwidth]{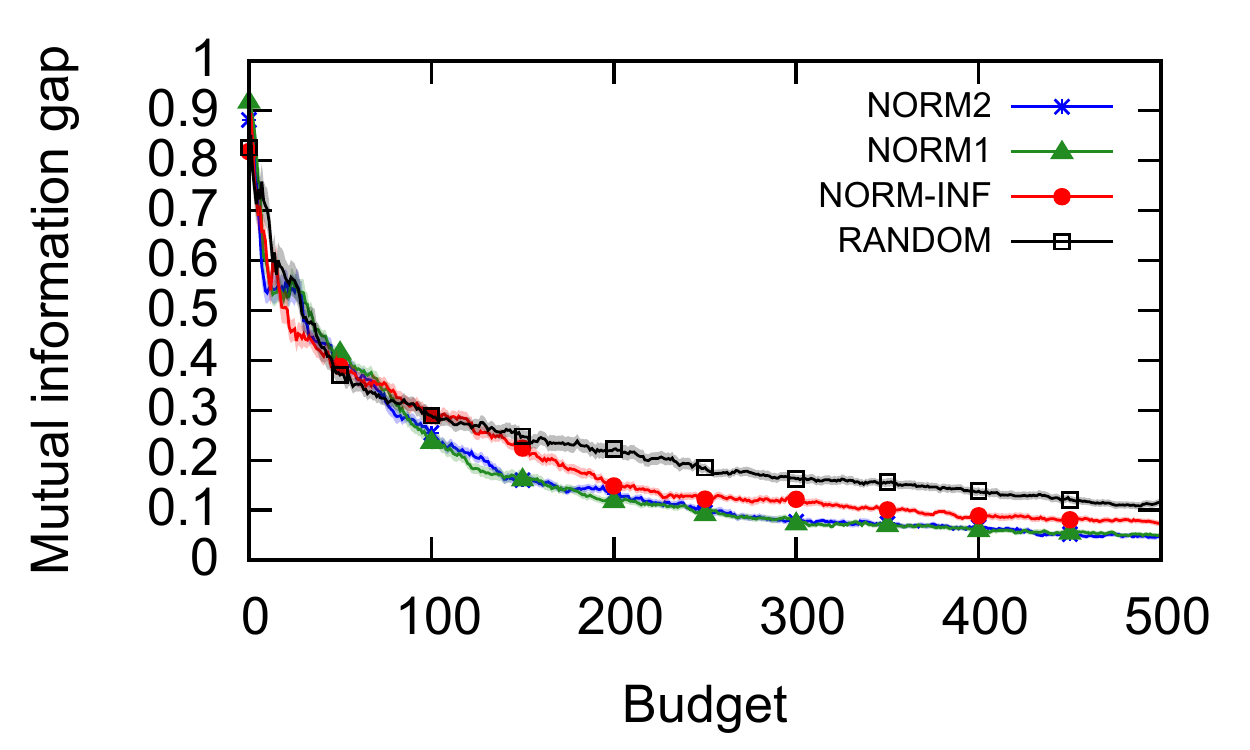} \\
    \includegraphics[width = 0.4\textwidth]{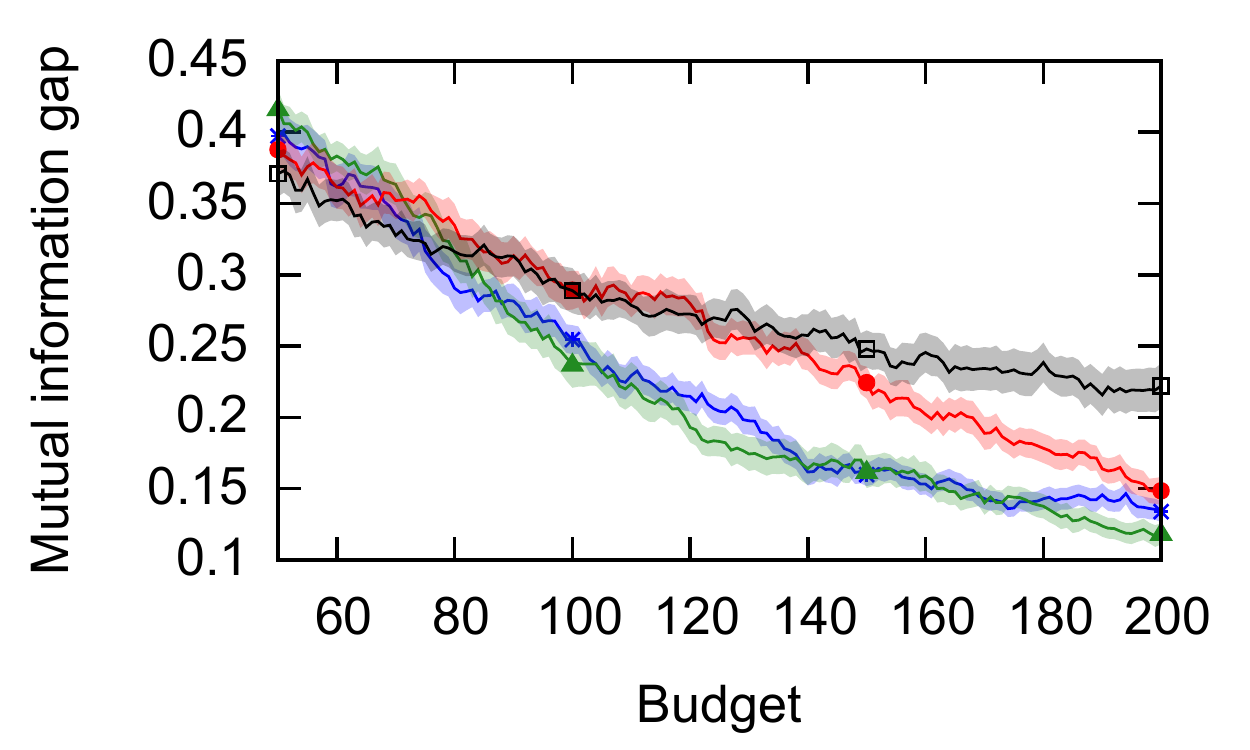}
    }
  \end{center}

\caption{\small Choices of $\psi$: MUSK ($k=20$). Top: full experiment. Bottom: Zoom in.}
\end{figure}

\begin{figure}[h]
  \begin{center}
    \myborder{
    \includegraphics[width = 0.4\textwidth]{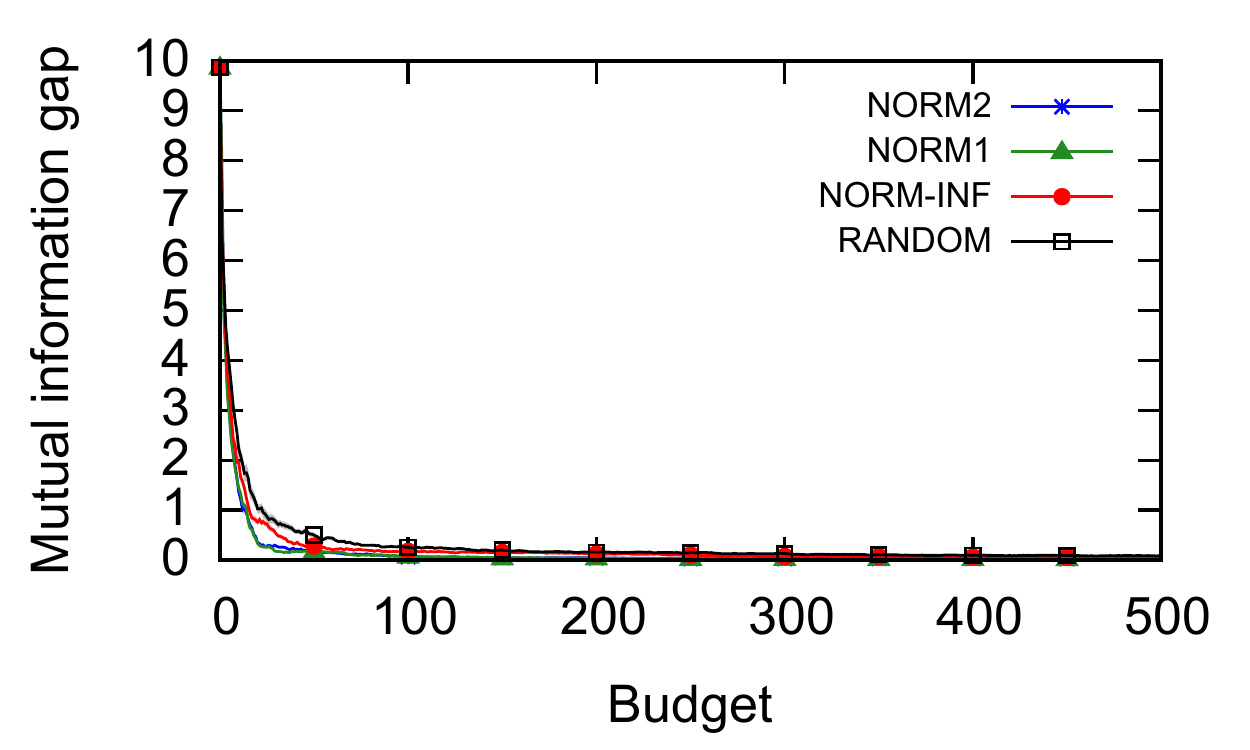} \\
    \includegraphics[width = 0.4\textwidth]{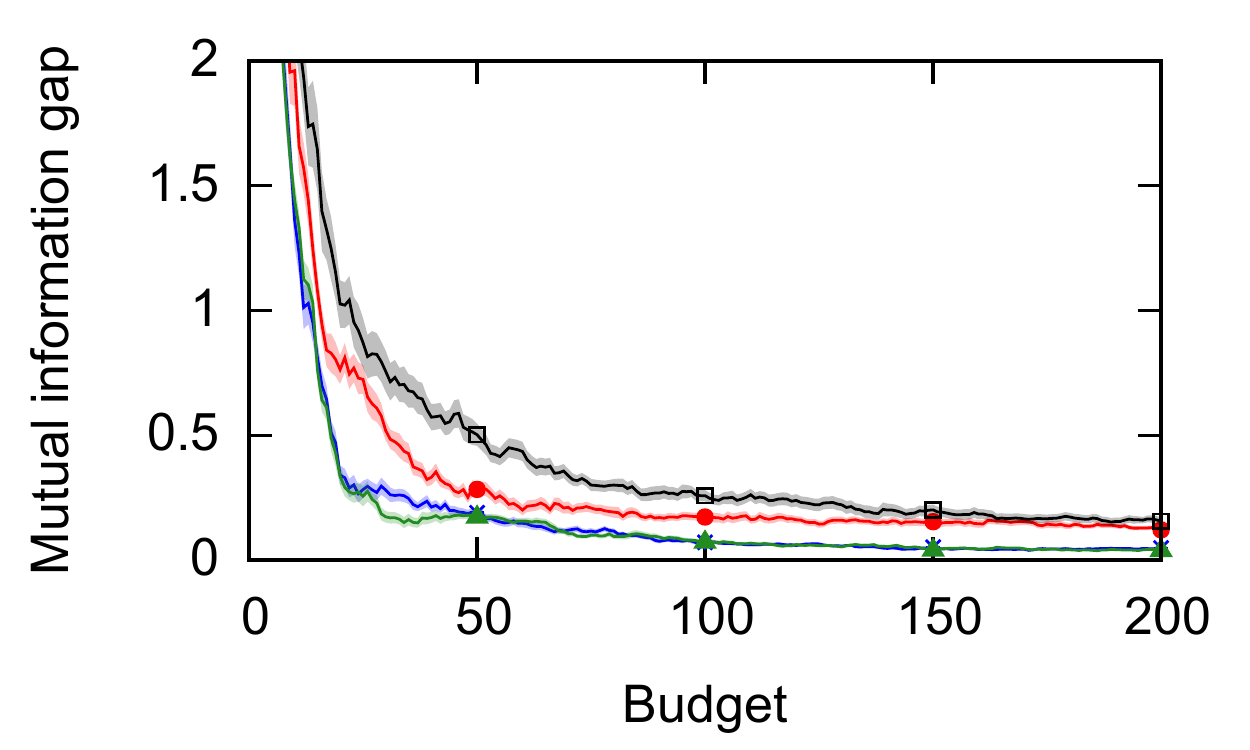}
    }
  \end{center}

\caption{\small Choices of $\psi$: MNIST: 0 vs 1 ($k=20$). Top: full experiment. Bottom: Zoom in.}
\end{figure}

\begin{figure}[h]
  \begin{center}
    \myborder{
    \includegraphics[width = 0.4\textwidth]{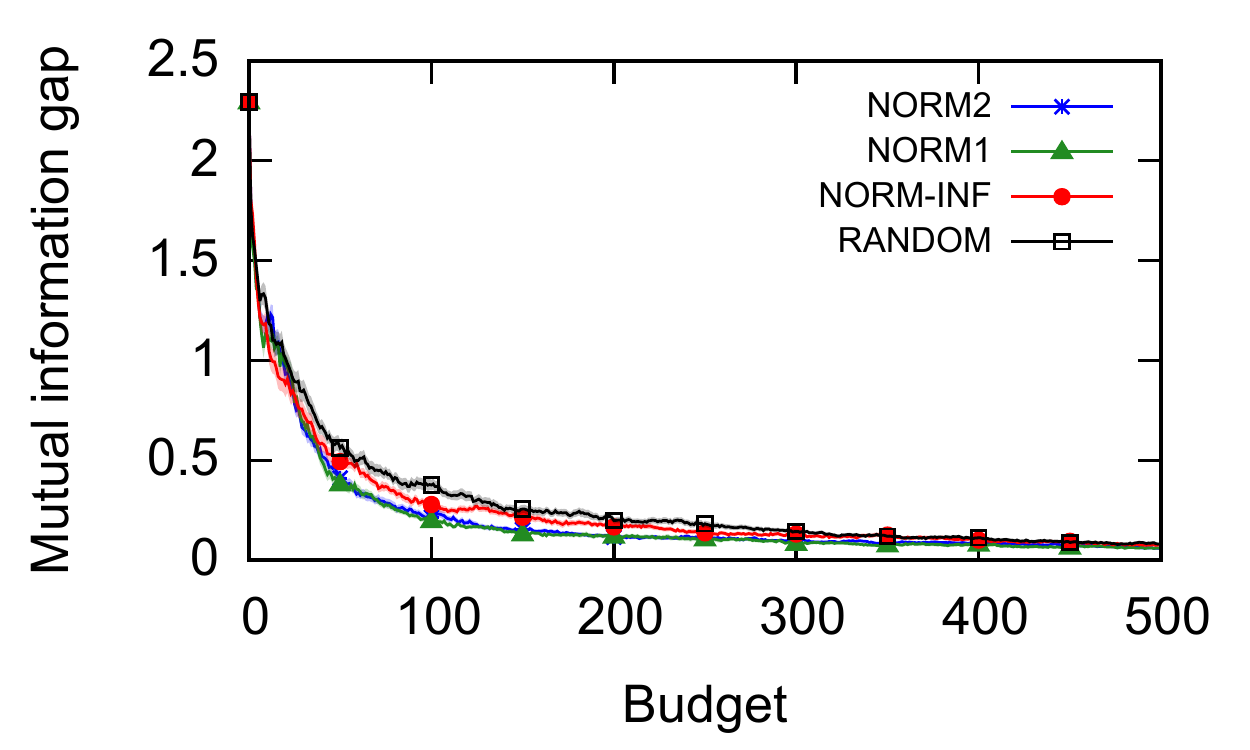} \\
    \includegraphics[width = 0.4\textwidth]{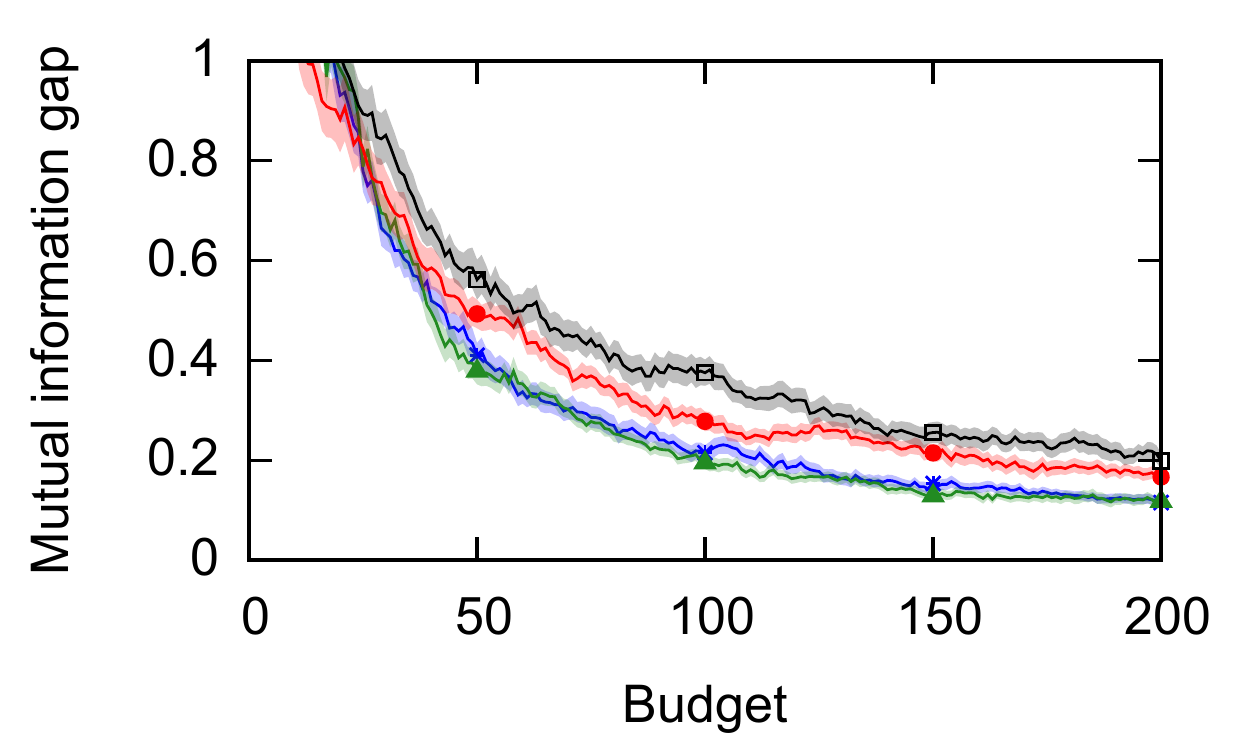}
    }
  \end{center}

\caption{\small Choices of $\psi$: MNIST: 3 vs 5 ($k=20$). Top: full experiment. Bottom: Zoom in.}
\end{figure}

\begin{figure}[h]
  \begin{center}
    \myborder{
    \includegraphics[width = 0.4\textwidth]{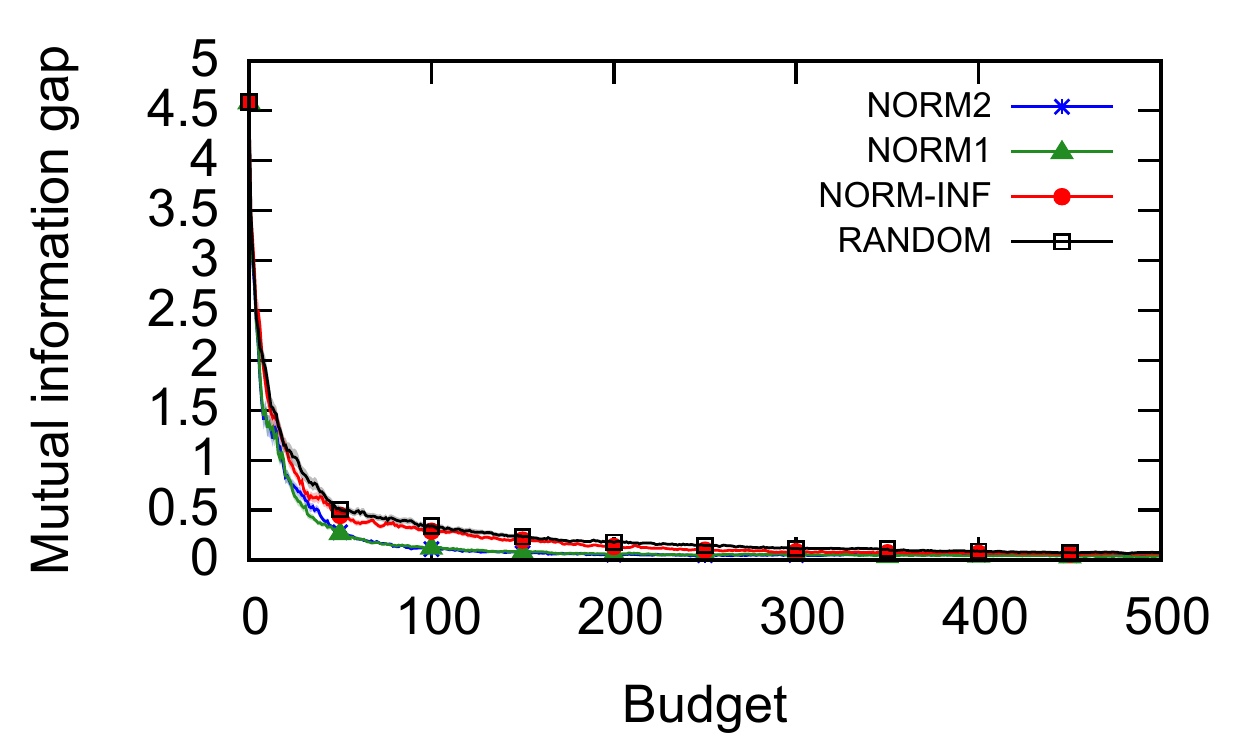} \\
    \includegraphics[width = 0.4\textwidth]{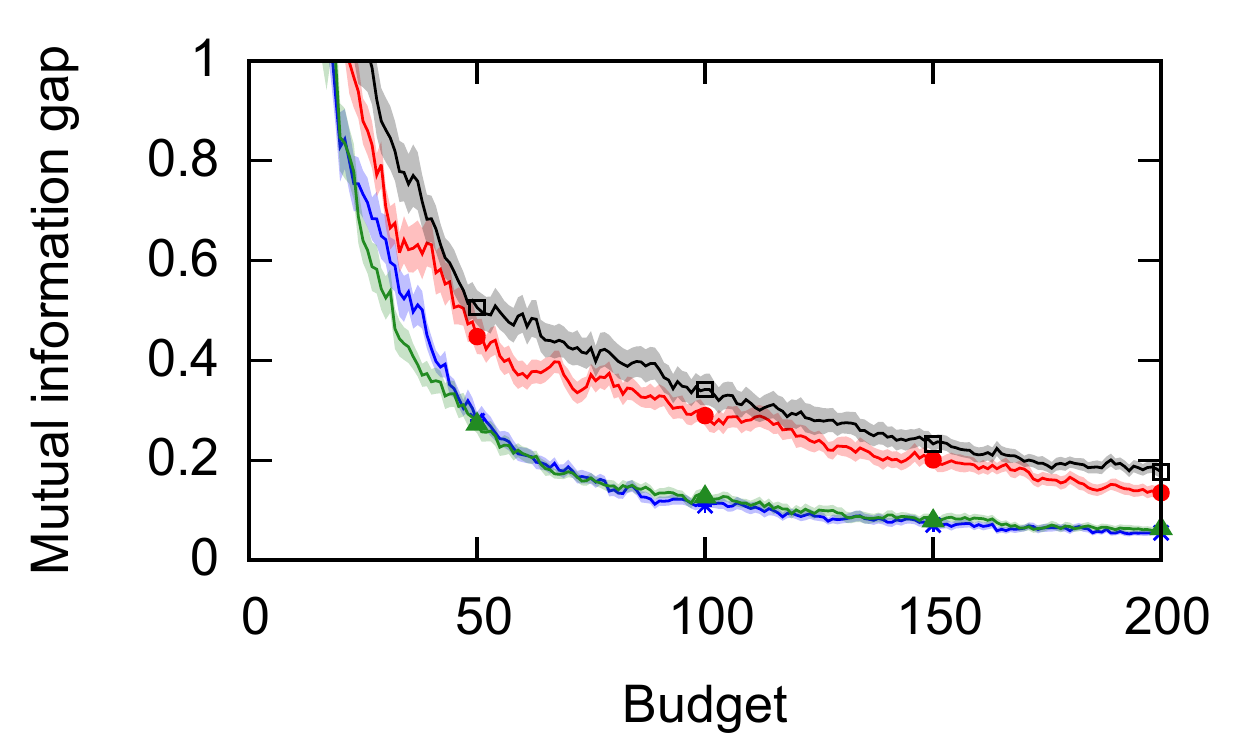}
    }
  \end{center}

\caption{\small Choices of $\psi$: MNIST: 4 vs 6 ($k=20$). Top: full experiment. Bottom: Zoom in.}
\end{figure}

\clearpage

\subsection{Comparing aggregation functions: $k = 10$}
\begin{figure}[h]
  \begin{center}
    \myborder{
    \includegraphics[width = 0.4\textwidth]{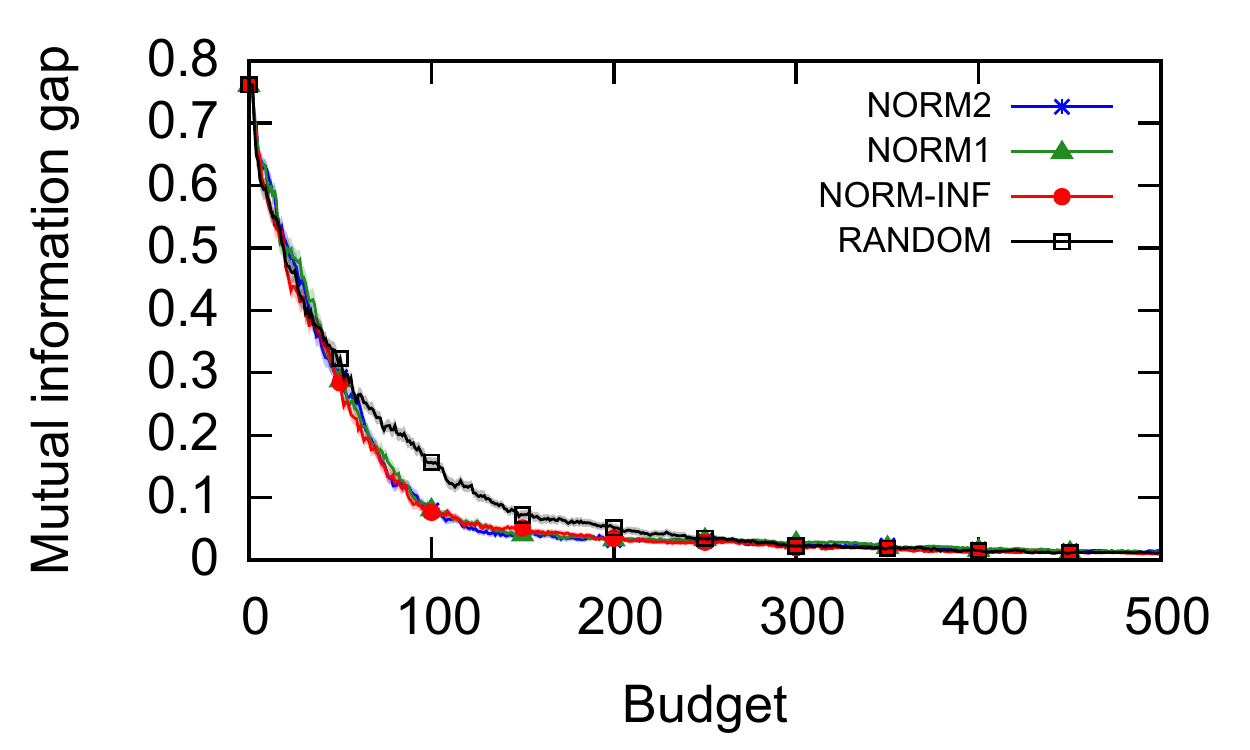} \\
    \includegraphics[width = 0.4\textwidth]{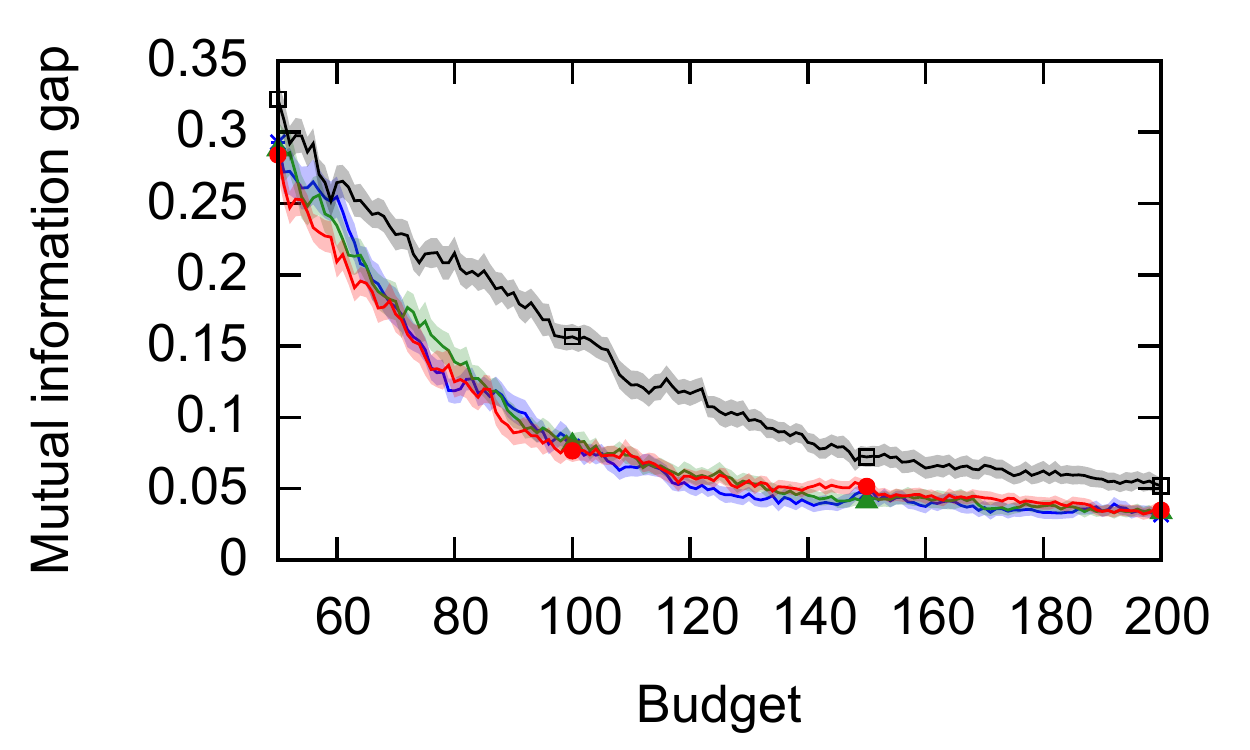}
    }
  \end{center}

\caption{\small Choices of $\psi$: BASEHOCK ($k=10$). Top: full experiment. Bottom: Zoom in.}
\end{figure}

\begin{figure}[h]
  \begin{center}
    \myborder{
    \includegraphics[width = 0.4\textwidth]{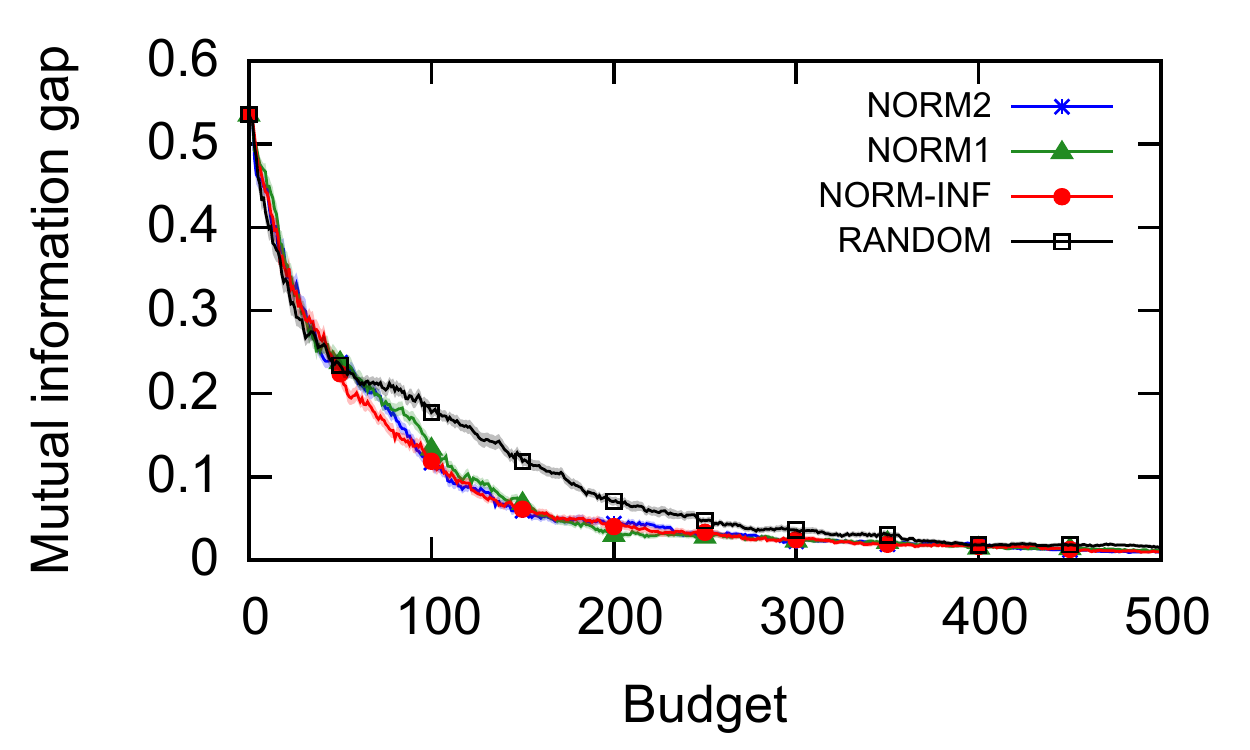} \\
    \includegraphics[width = 0.4\textwidth]{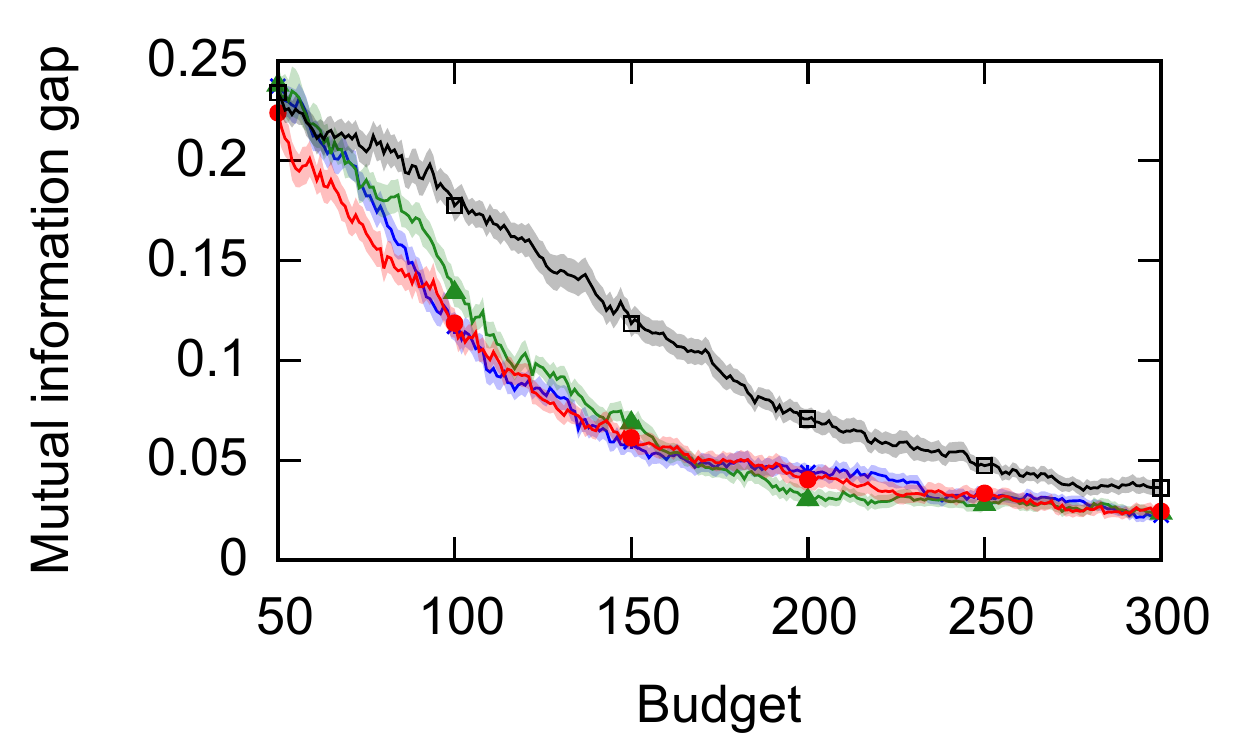}
    }
  \end{center}

\caption{\small Choices of $\psi$: PCMAC ($k=10$). Top: full experiment. Bottom: Zoom in.}
\end{figure}

\begin{figure}[h]
  \begin{center}
    \myborder{
    \includegraphics[width = 0.4\textwidth]{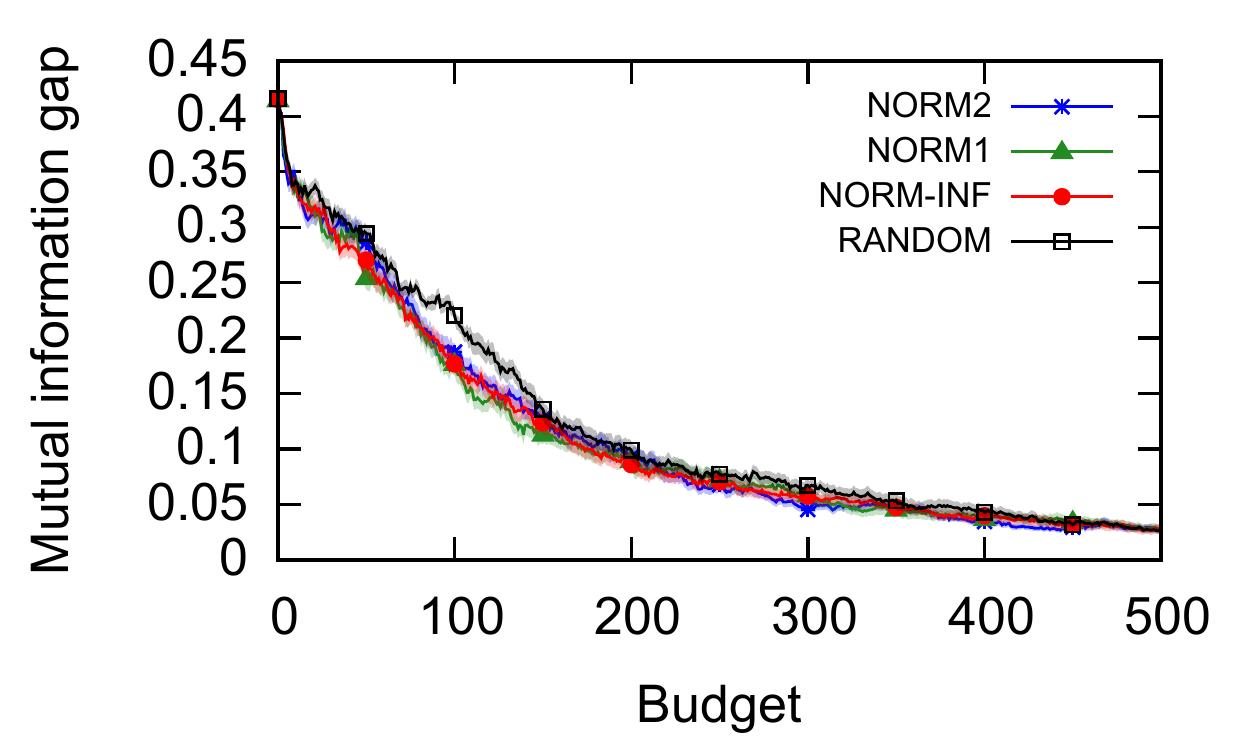} \\
    \includegraphics[width = 0.4\textwidth]{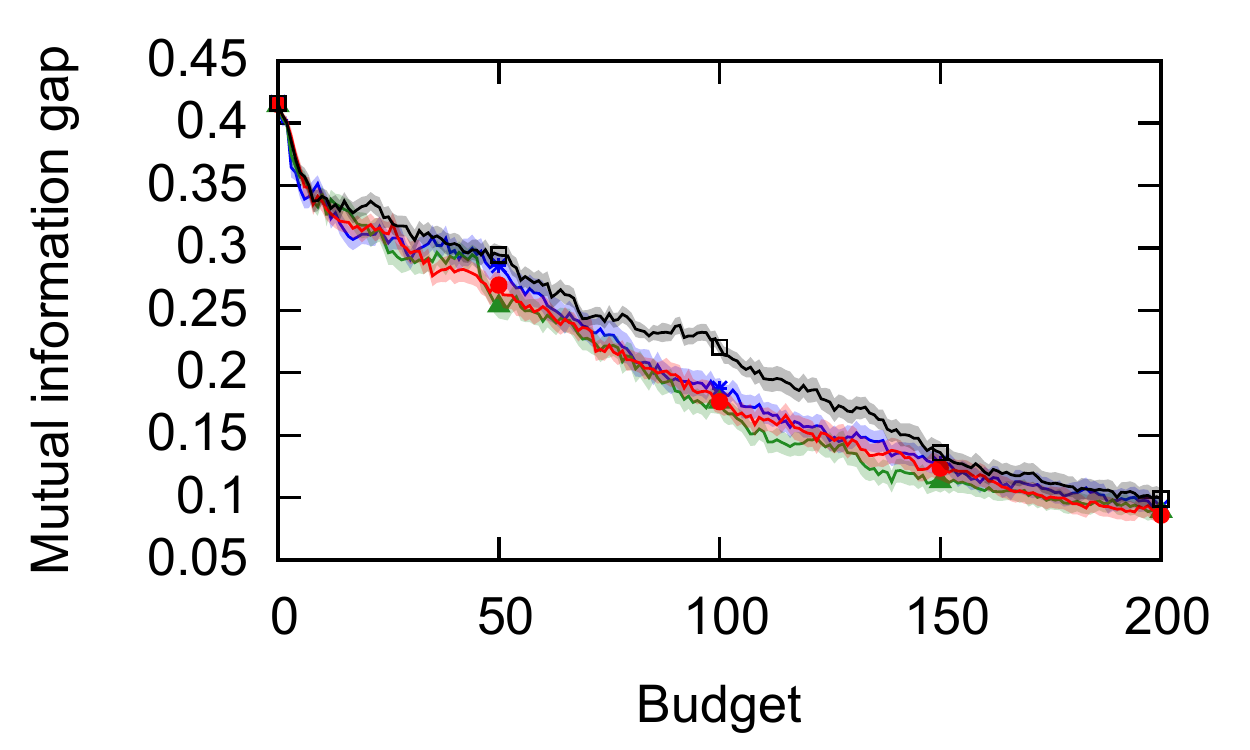}
    }
  \end{center}

\caption{\small Choices of $\psi$: RELATHE ($k=10$). Top: full experiment. Bottom: Zoom in.}
\end{figure}

\begin{figure}[h]
  \begin{center}
    \myborder{
    \includegraphics[width = 0.4\textwidth]{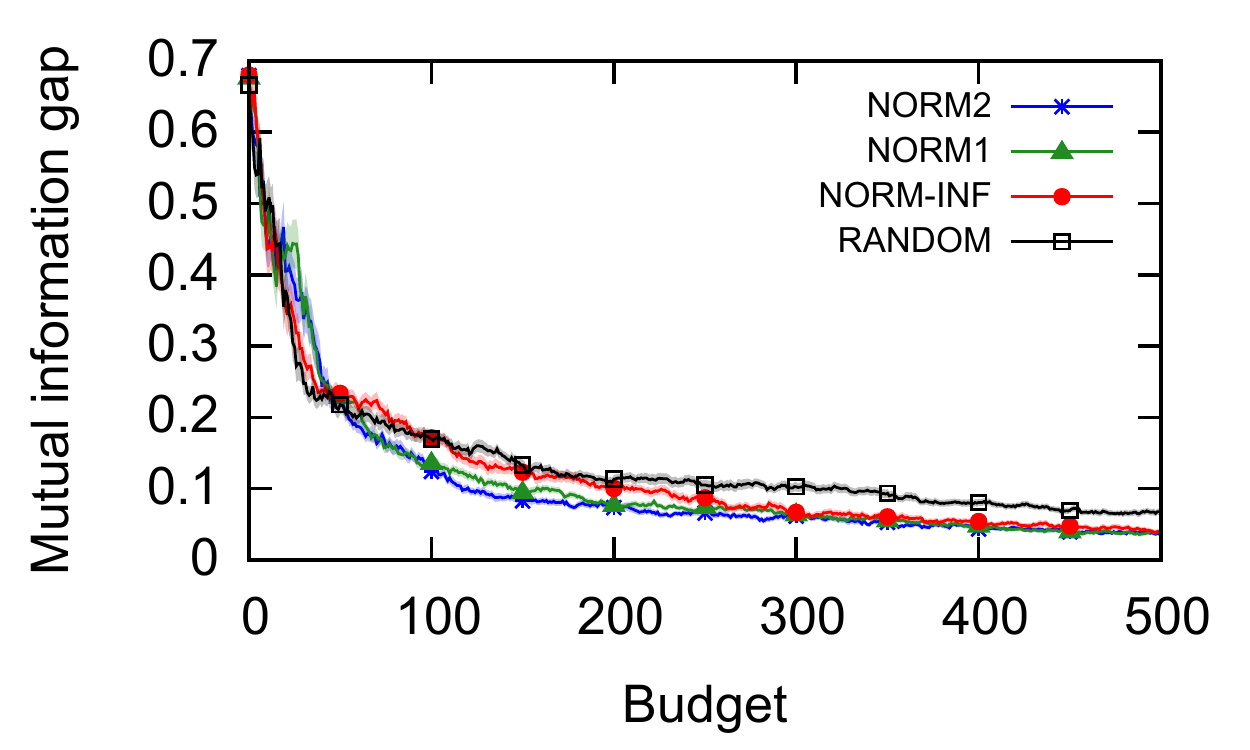} \\
    \includegraphics[width = 0.4\textwidth]{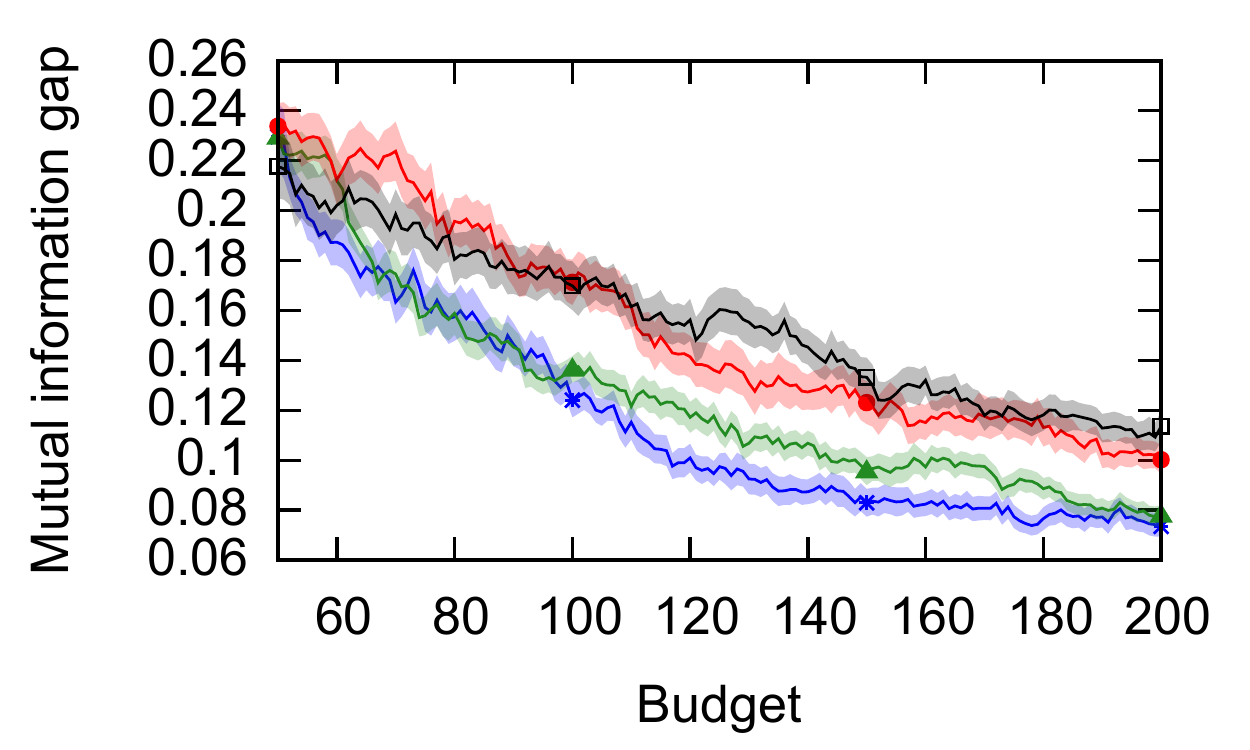}
    }
  \end{center}

\caption{\small Choices of $\psi$: MUSK ($k=10$). Top: full experiment. Bottom: Zoom in.}
\end{figure}

\begin{figure}[h]
  \begin{center}
    \myborder{
    \includegraphics[width = 0.4\textwidth]{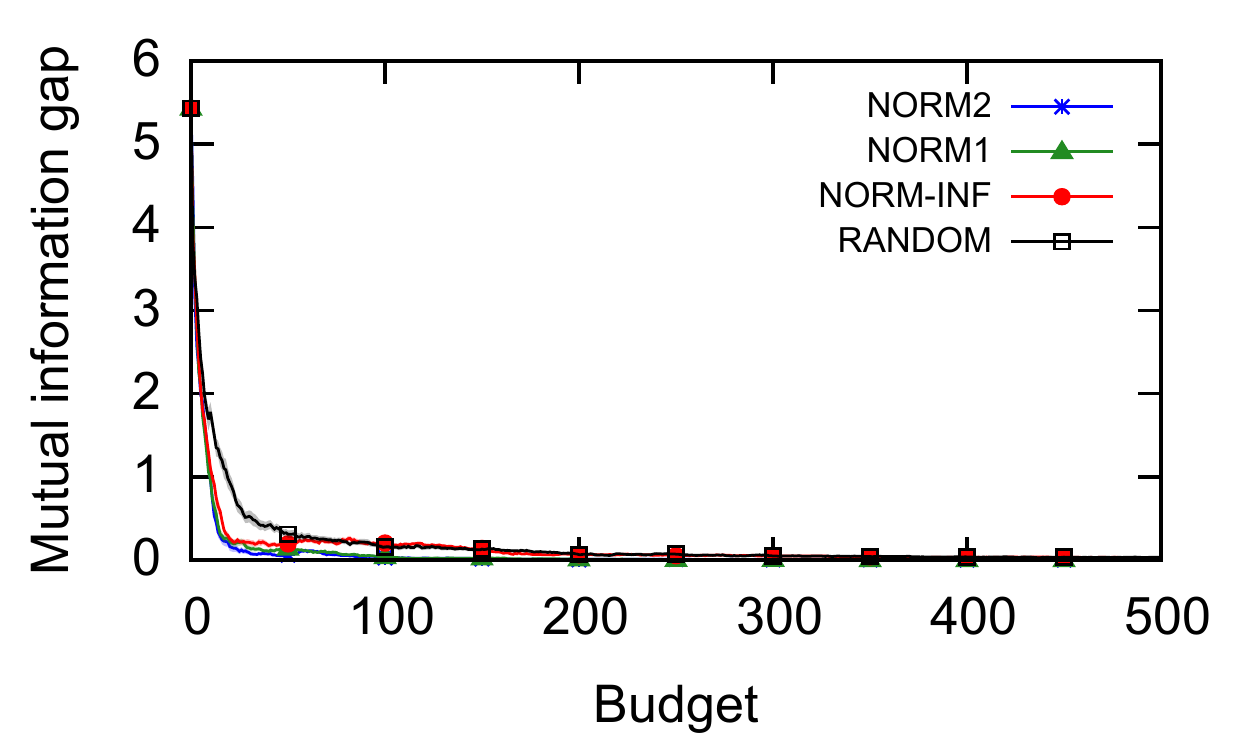} \\
    \includegraphics[width = 0.4\textwidth]{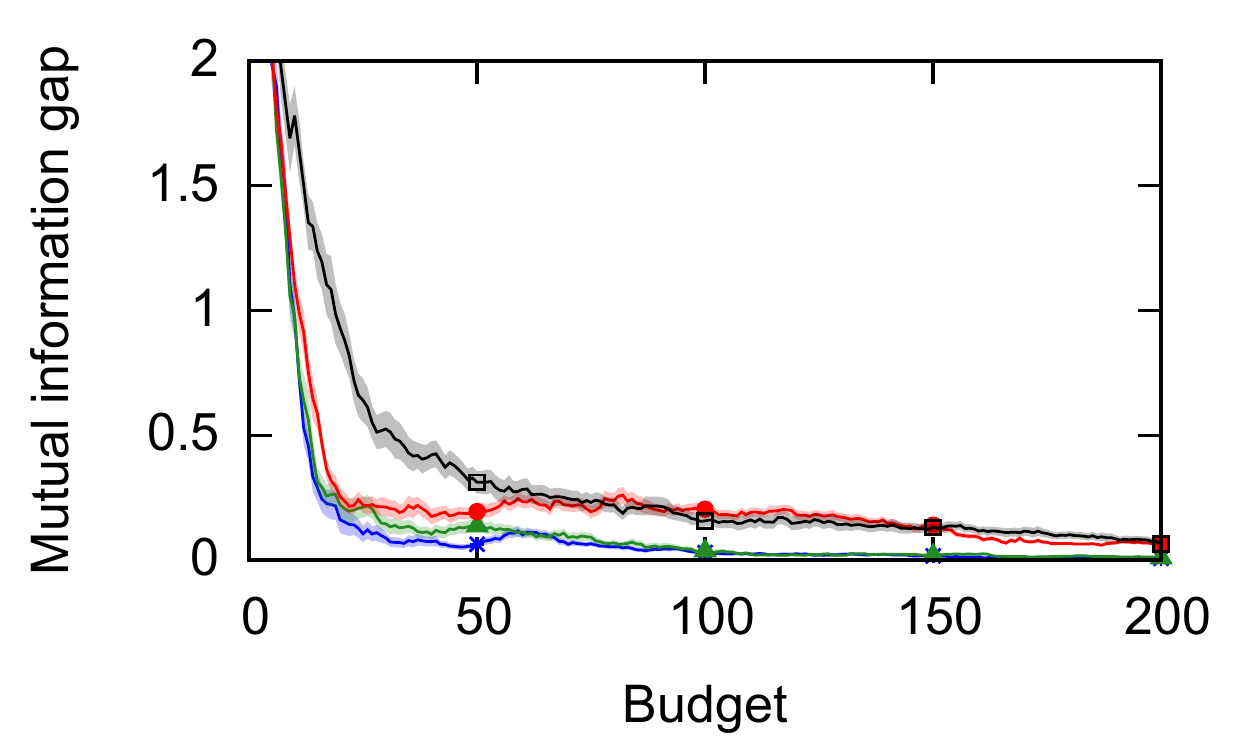}
    }
  \end{center}

\caption{\small Choices of $\psi$: MNIST: 0 vs 1 ($k=10$). Top: full experiment. Bottom: Zoom in.}
\end{figure}

\begin{figure}[h]
  \begin{center}
    \myborder{
    \includegraphics[width = 0.4\textwidth]{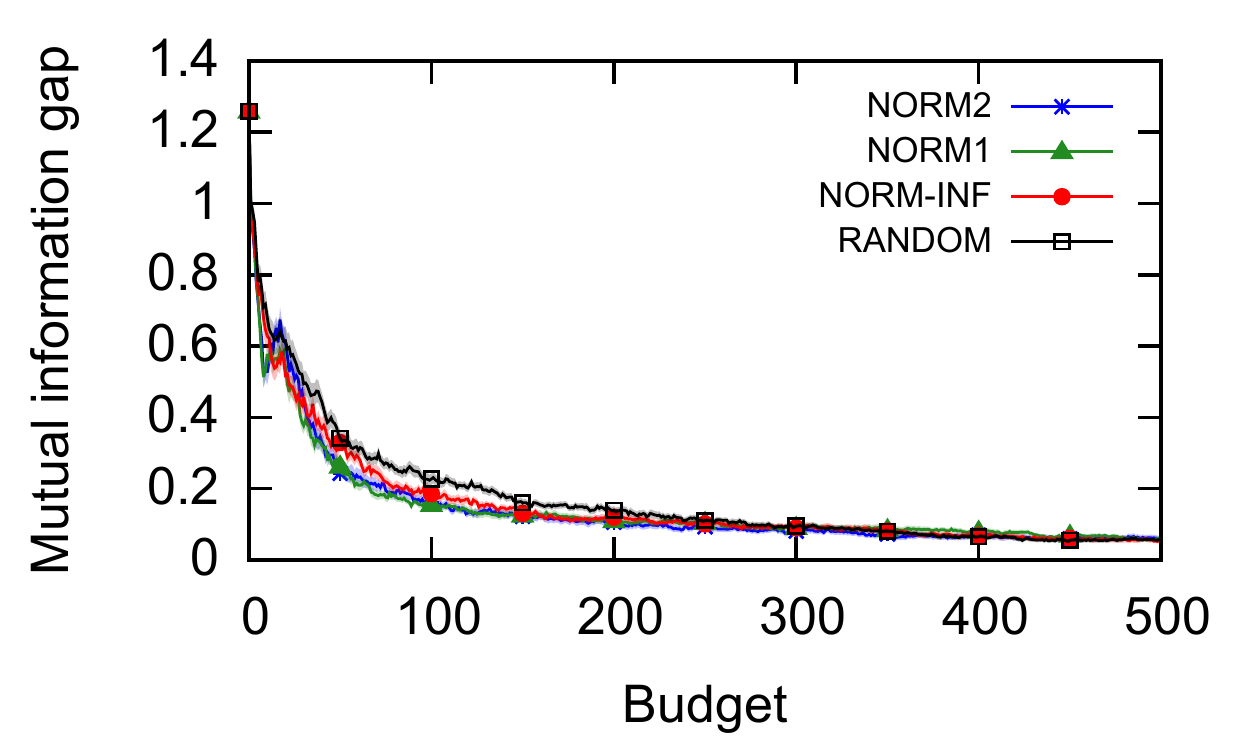} \\
    \includegraphics[width = 0.4\textwidth]{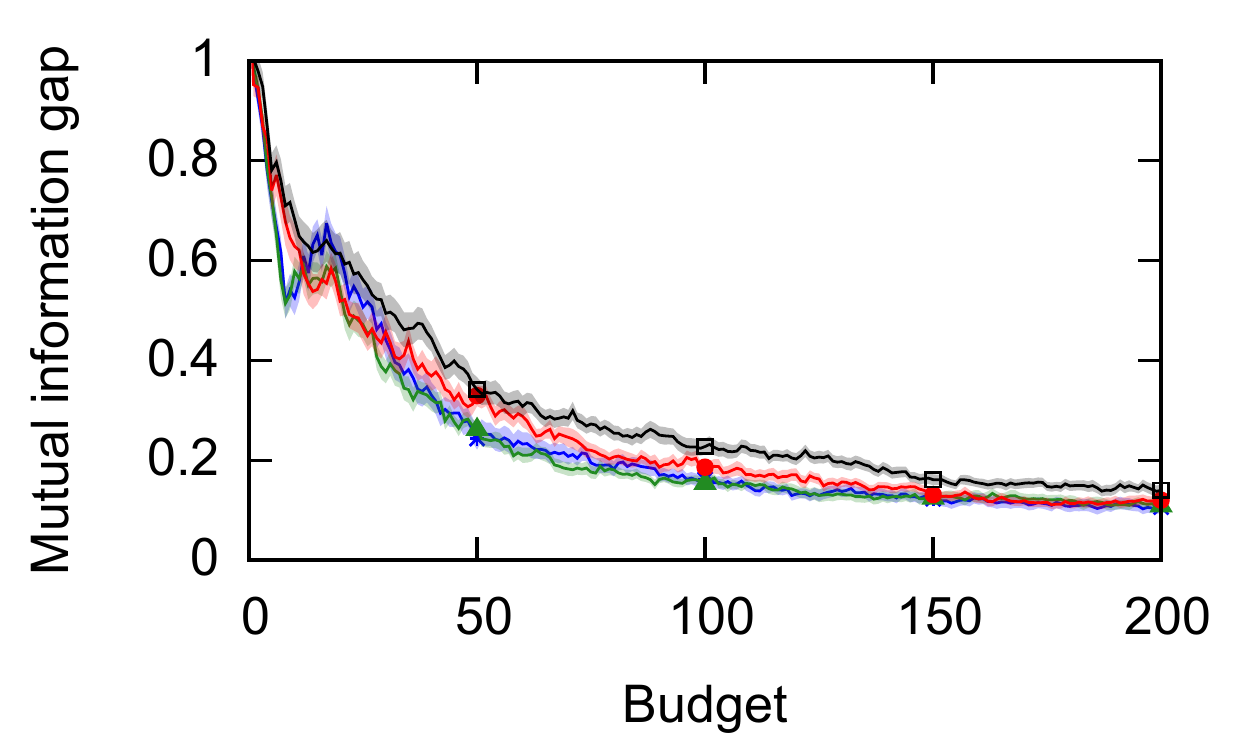}
    }
  \end{center}

\caption{\small Choices of $\psi$: MNIST: 3 vs 5 ($k=10$). Top: full experiment. Bottom: Zoom in.}
\end{figure}

\begin{figure}[h]
  \begin{center}
    \myborder{
    \includegraphics[width = 0.4\textwidth]{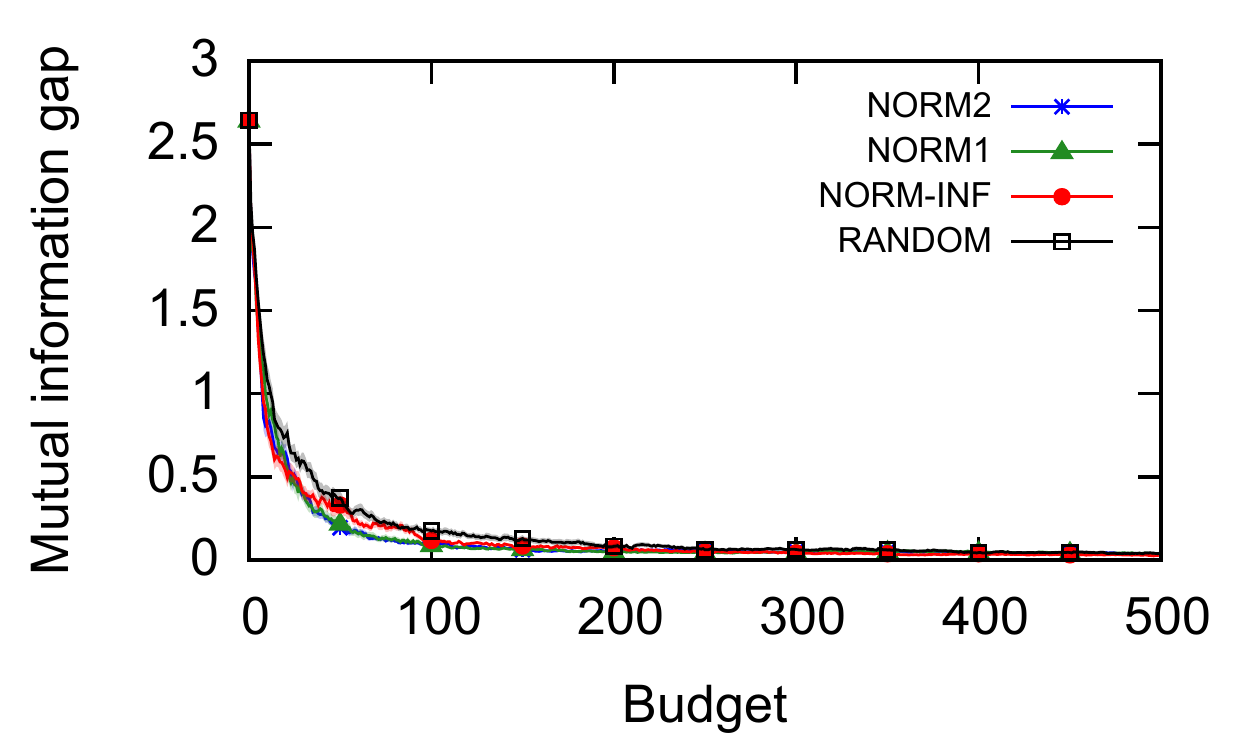} \\
    \includegraphics[width = 0.4\textwidth]{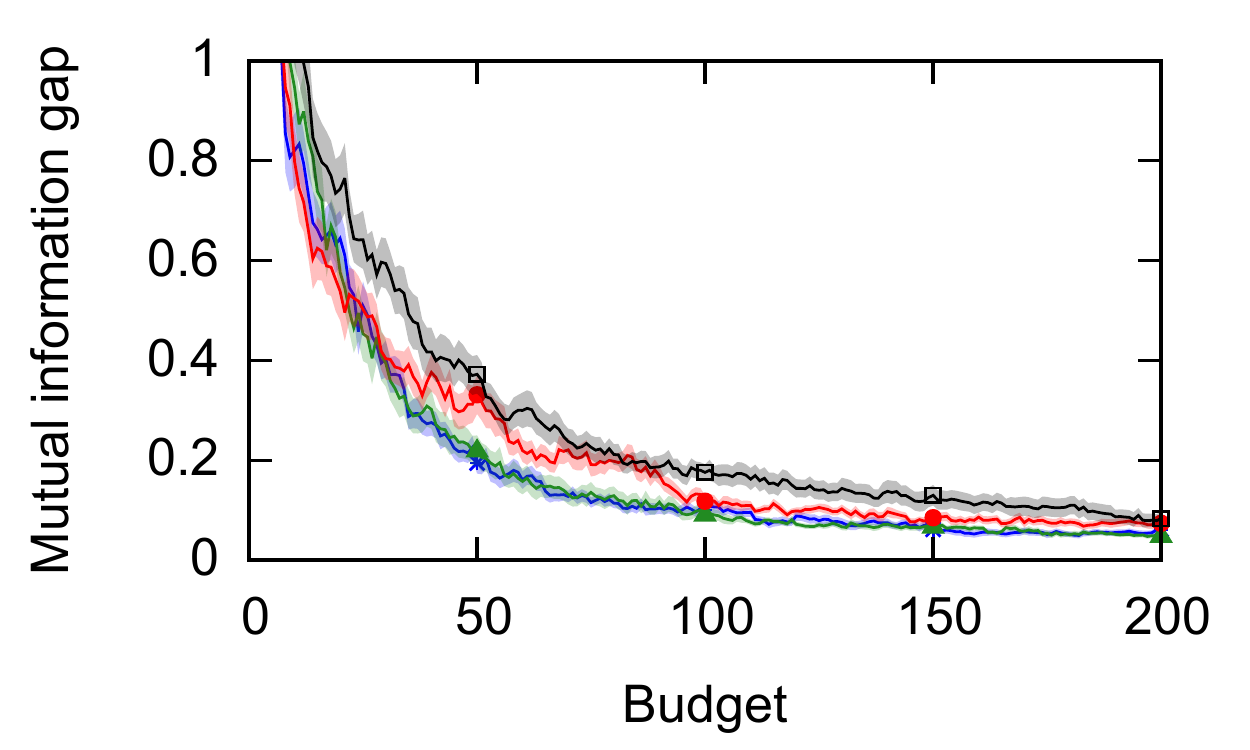}
    }
  \end{center}

\caption{\small Choices of $\psi$: MNIST: 4 vs 6 ($k=10$). Top: full experiment. Bottom: Zoom in.}
\end{figure}

\clearpage
\subsection{Comparing aggregation functions: $k = 5$}

\begin{figure}[h]
  \begin{center}
    \myborder{
    \includegraphics[width = 0.4\textwidth]{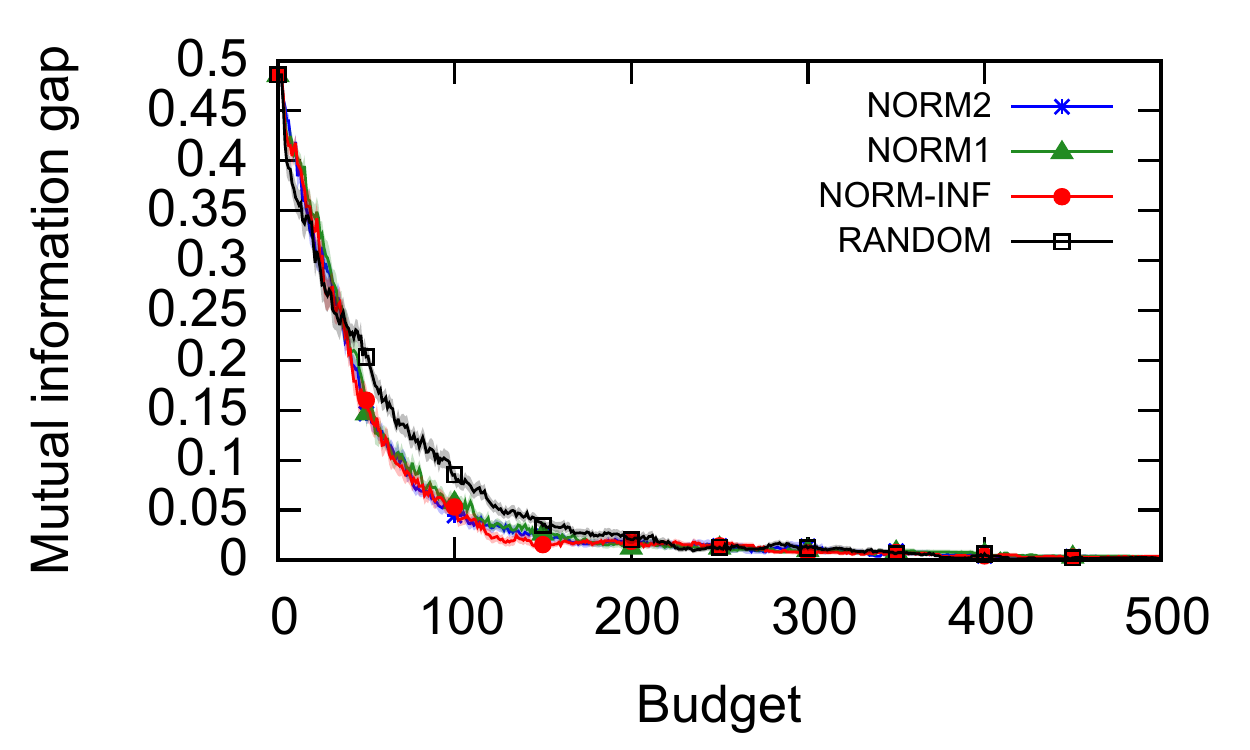} \\
    \includegraphics[width = 0.4\textwidth]{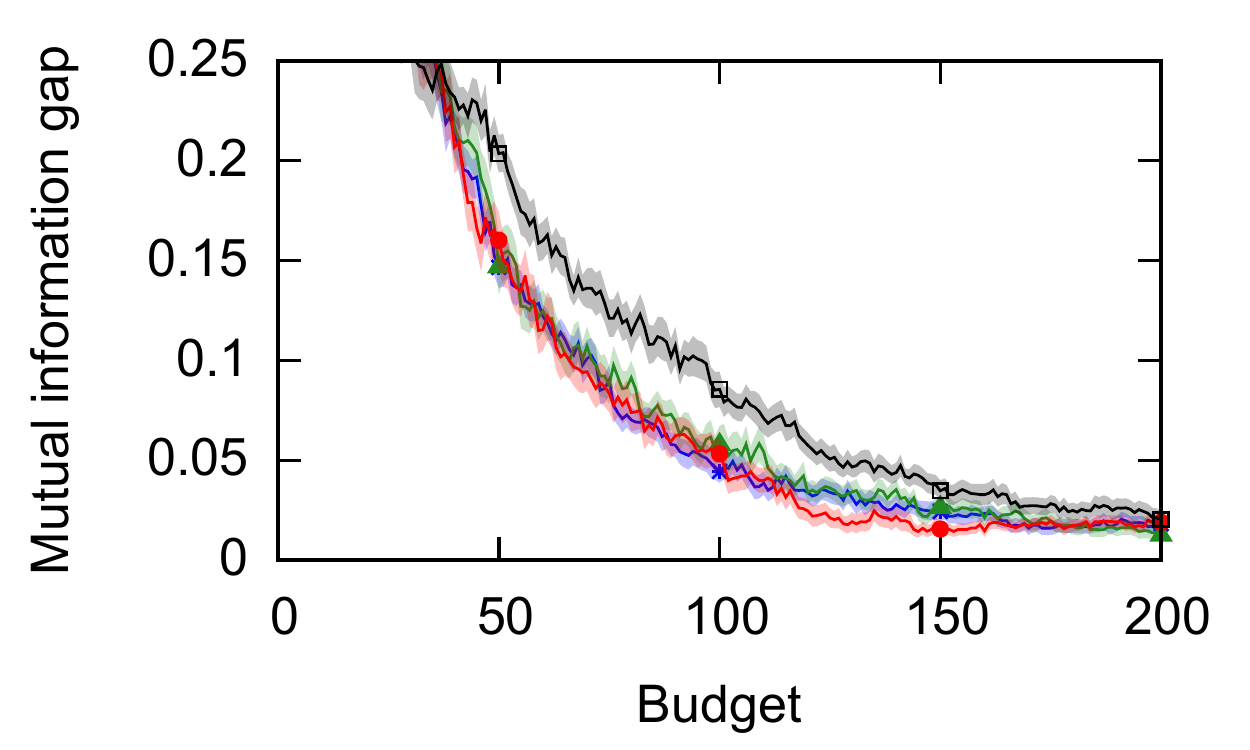}
    }
  \end{center}

\caption{\small Choices of $\psi$: BASEHOCK ($k=5$). Top: full experiment. Bottom: Zoom in.}
\end{figure}

\begin{figure}[h]
  \begin{center}
    \myborder{
    \includegraphics[width = 0.4\textwidth]{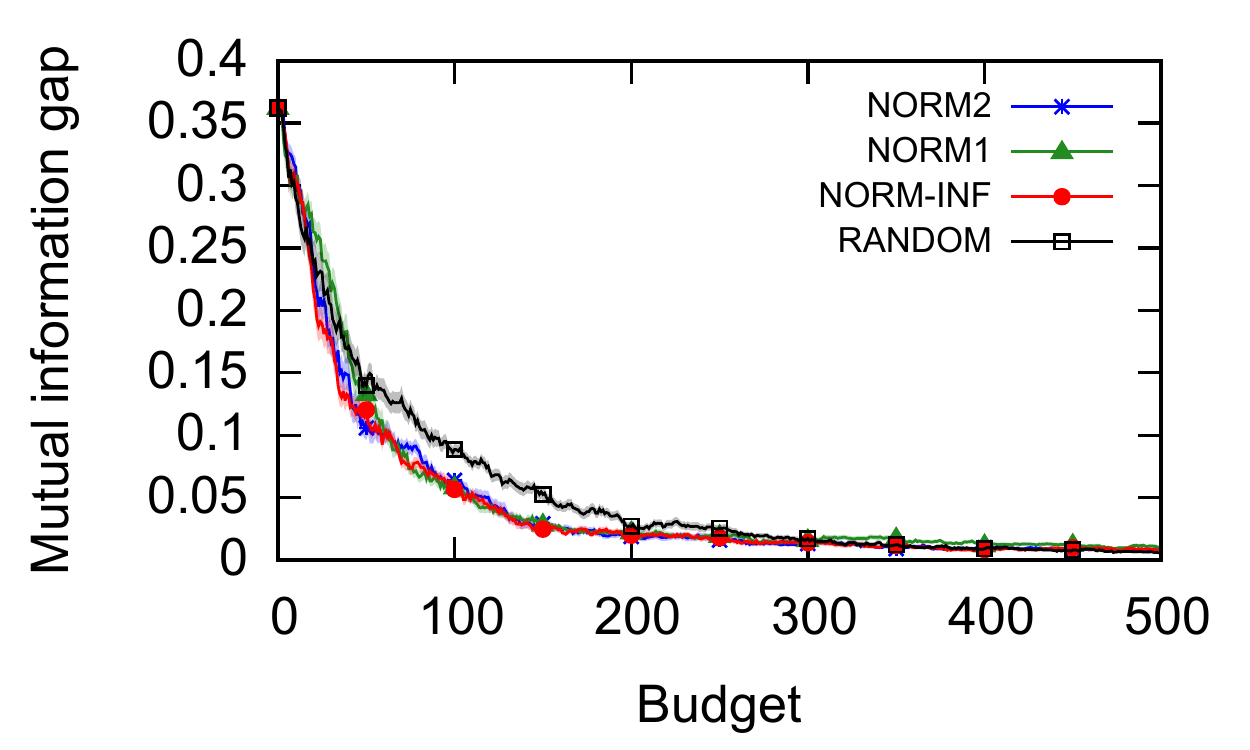} \\
    \includegraphics[width = 0.4\textwidth]{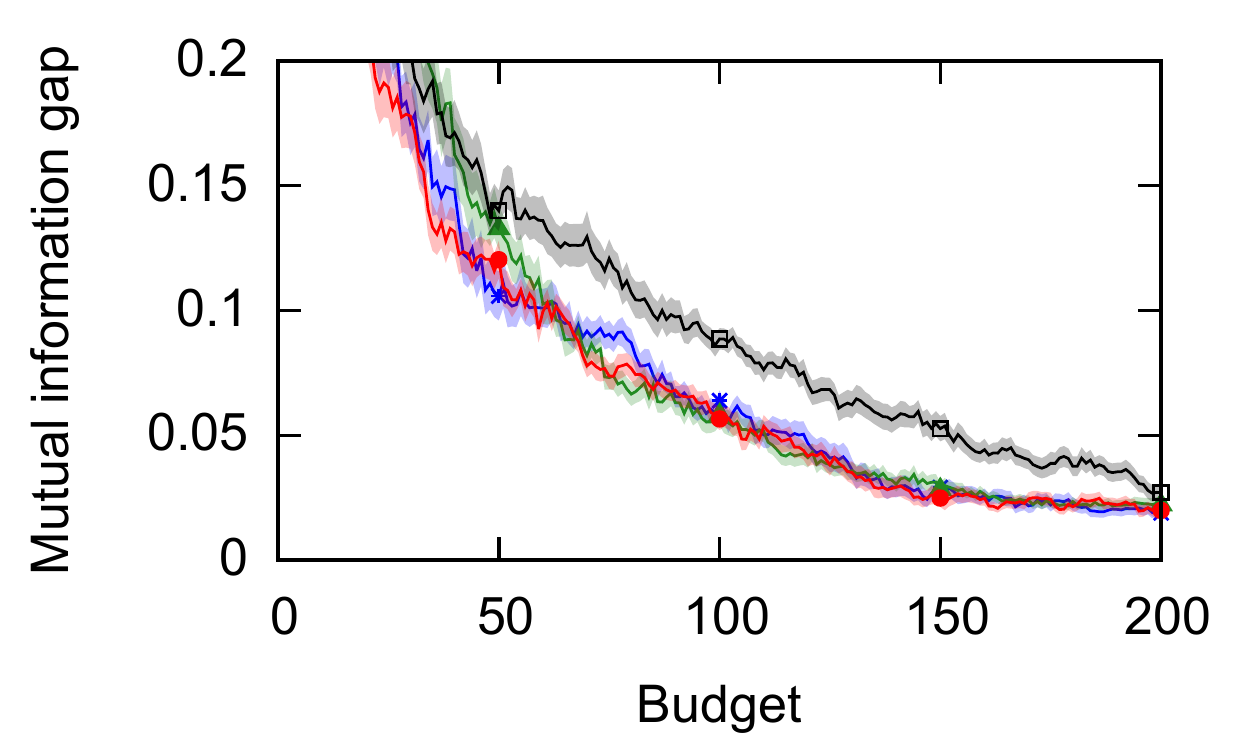}
    }
  \end{center}

\caption{\small Choices of $\psi$: PCMAC ($k=5$). Top: full experiment. Bottom: Zoom in.}
\end{figure}

\begin{figure}[h]
  \begin{center}
    \myborder{
    \includegraphics[width = 0.4\textwidth]{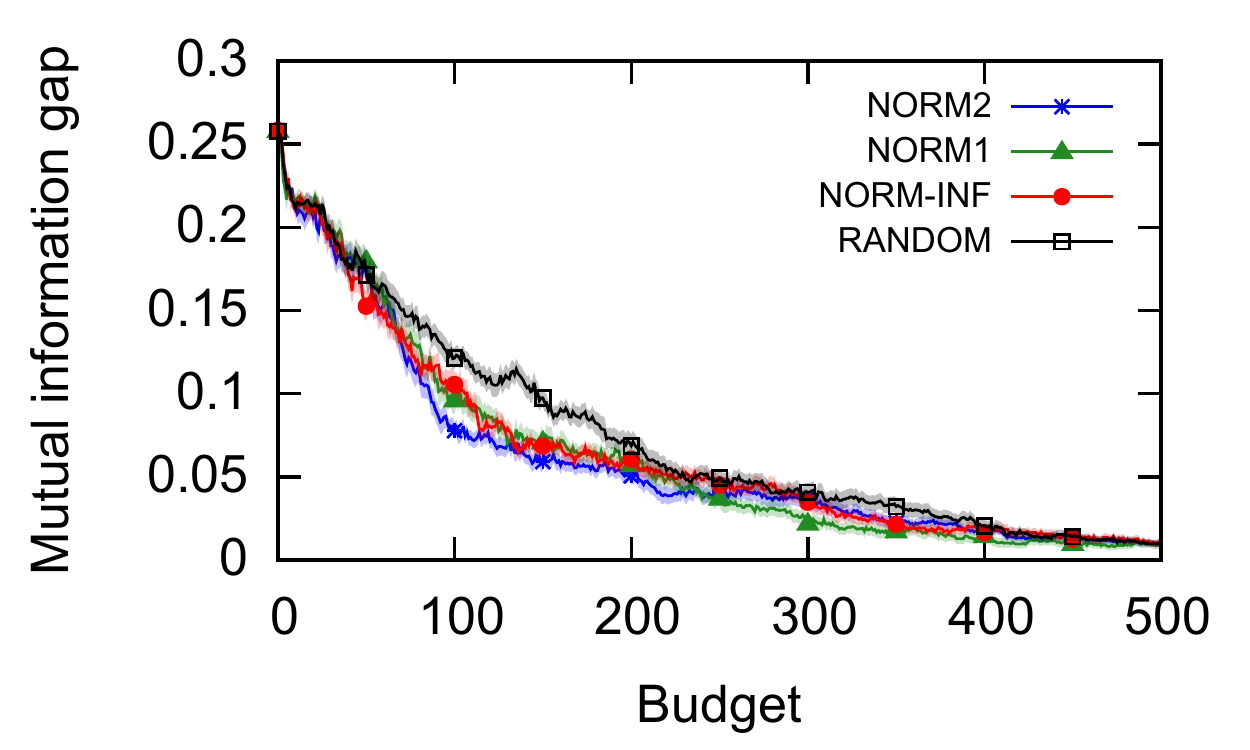} \\
    \includegraphics[width = 0.4\textwidth]{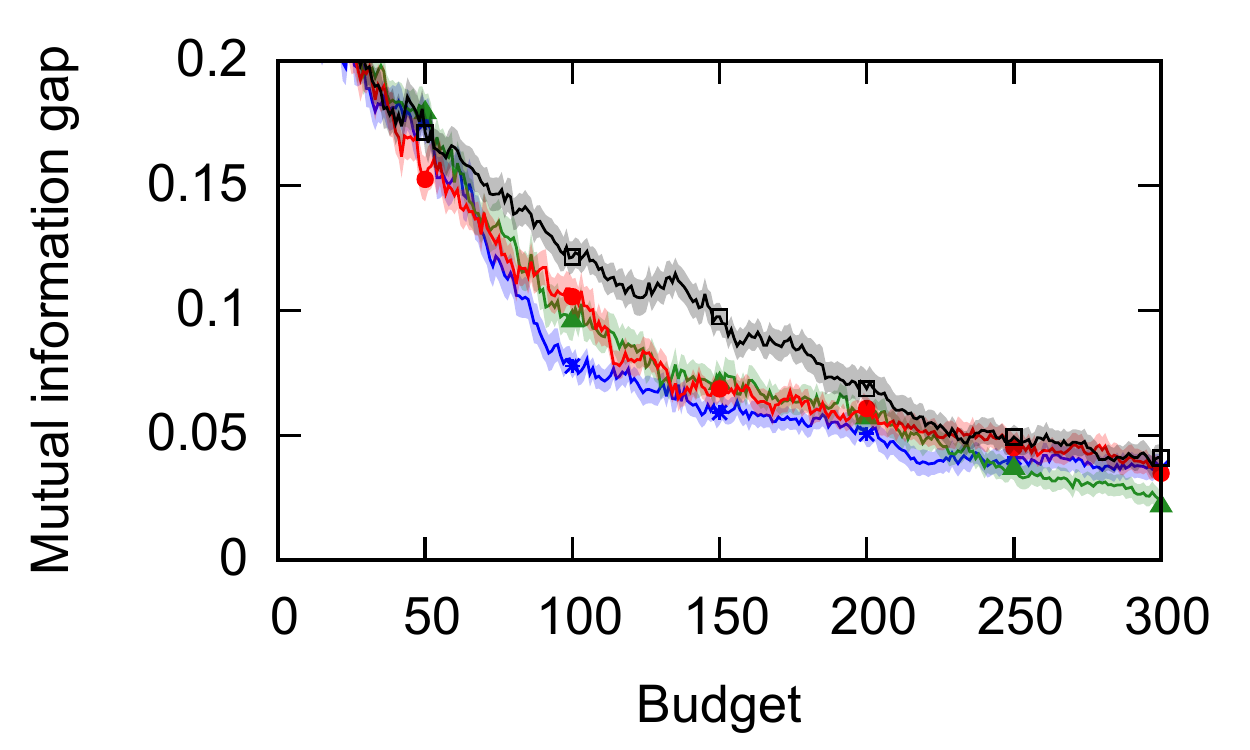}
    }
  \end{center}

\caption{\small Choices of $\psi$: RELATHE ($k=5$). Top: full experiment. Bottom: Zoom in.}
\end{figure}

\begin{figure}[h]
  \begin{center}
    \myborder{
    \includegraphics[width = 0.4\textwidth]{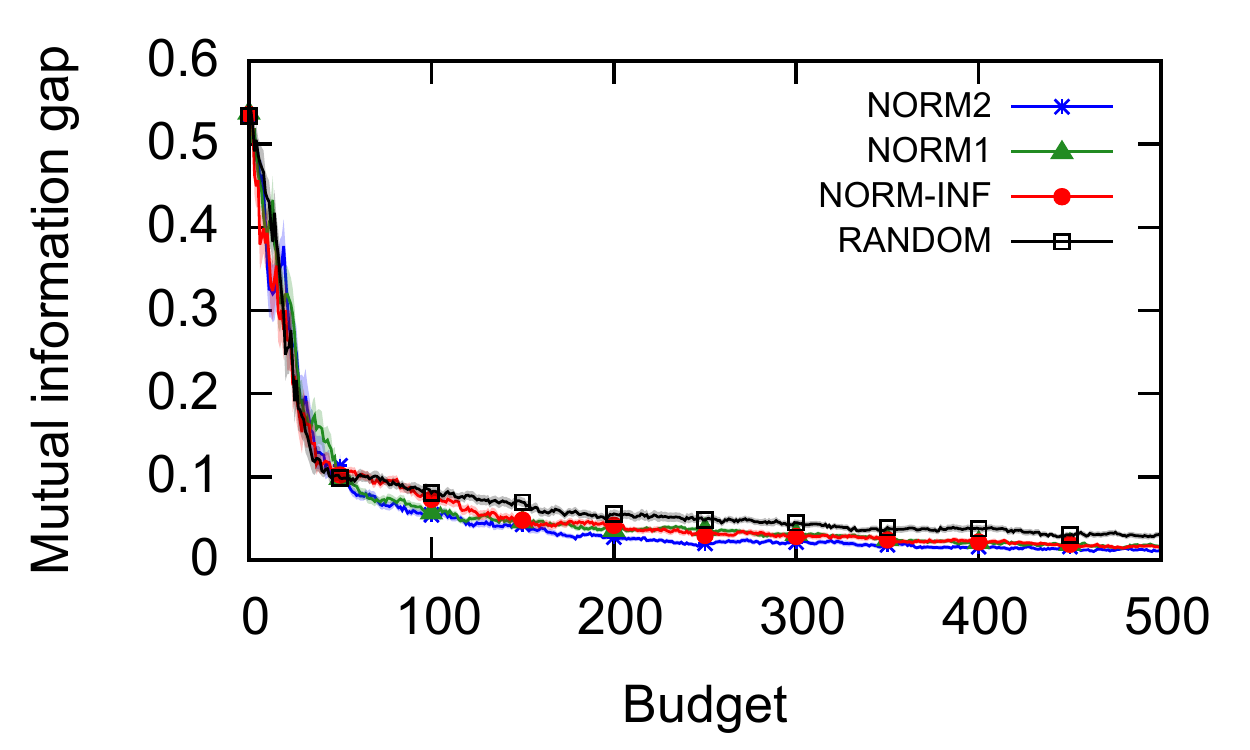} \\
    \includegraphics[width = 0.4\textwidth]{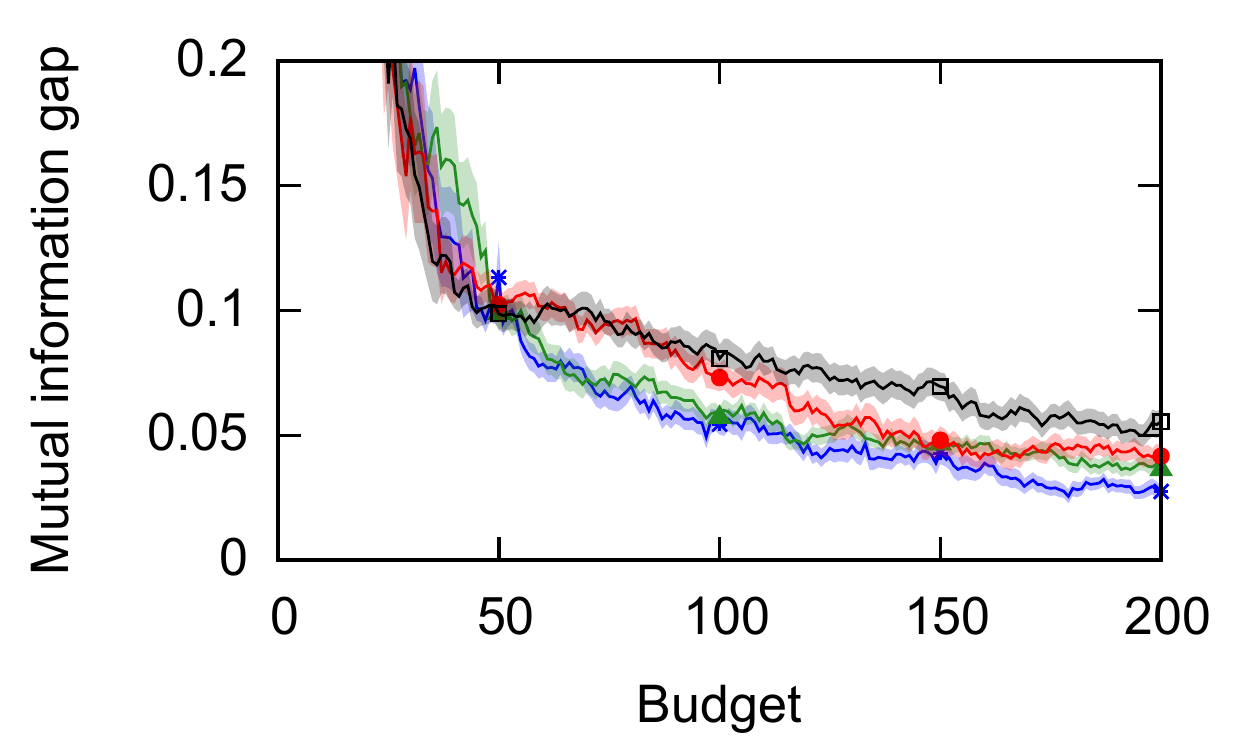}
    }
  \end{center}

\caption{\small Choices of $\psi$: MUSK ($k=5$). Top: full experiment. Bottom: Zoom in.}
\end{figure}

\begin{figure}[h]
  \begin{center}
    \myborder{
    \includegraphics[width = 0.4\textwidth]{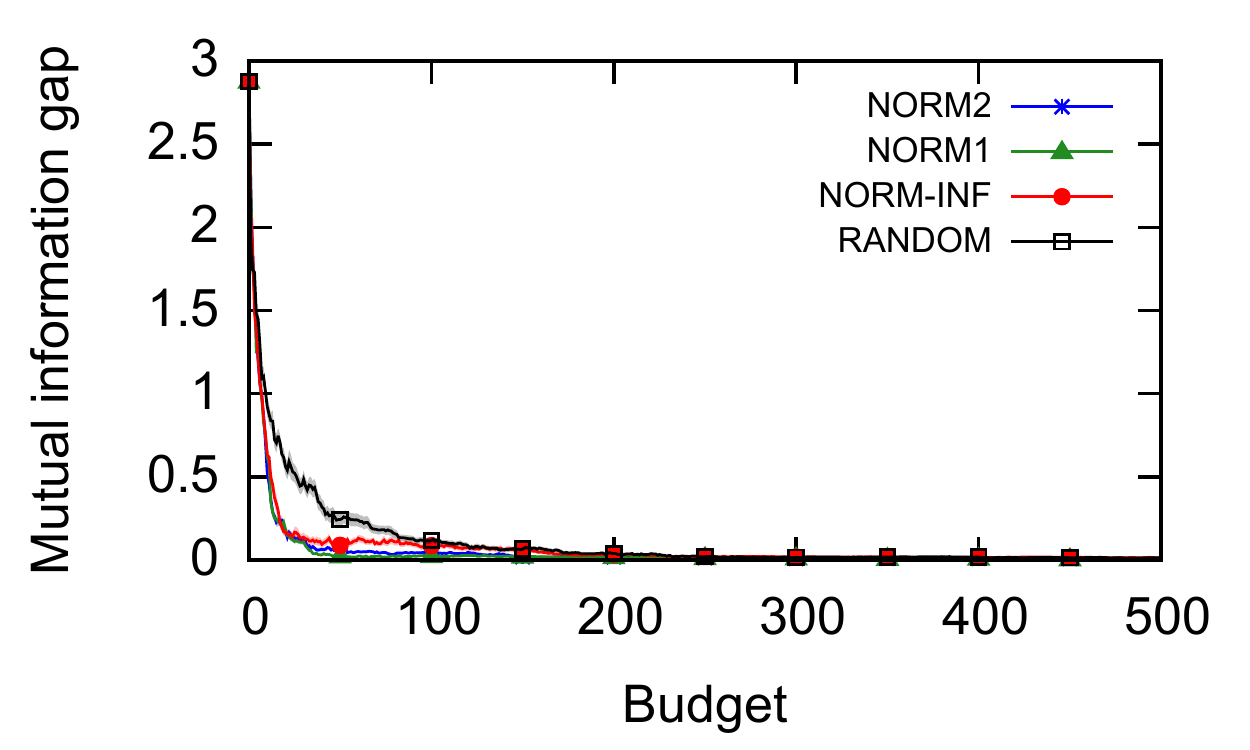} \\
    \includegraphics[width = 0.4\textwidth]{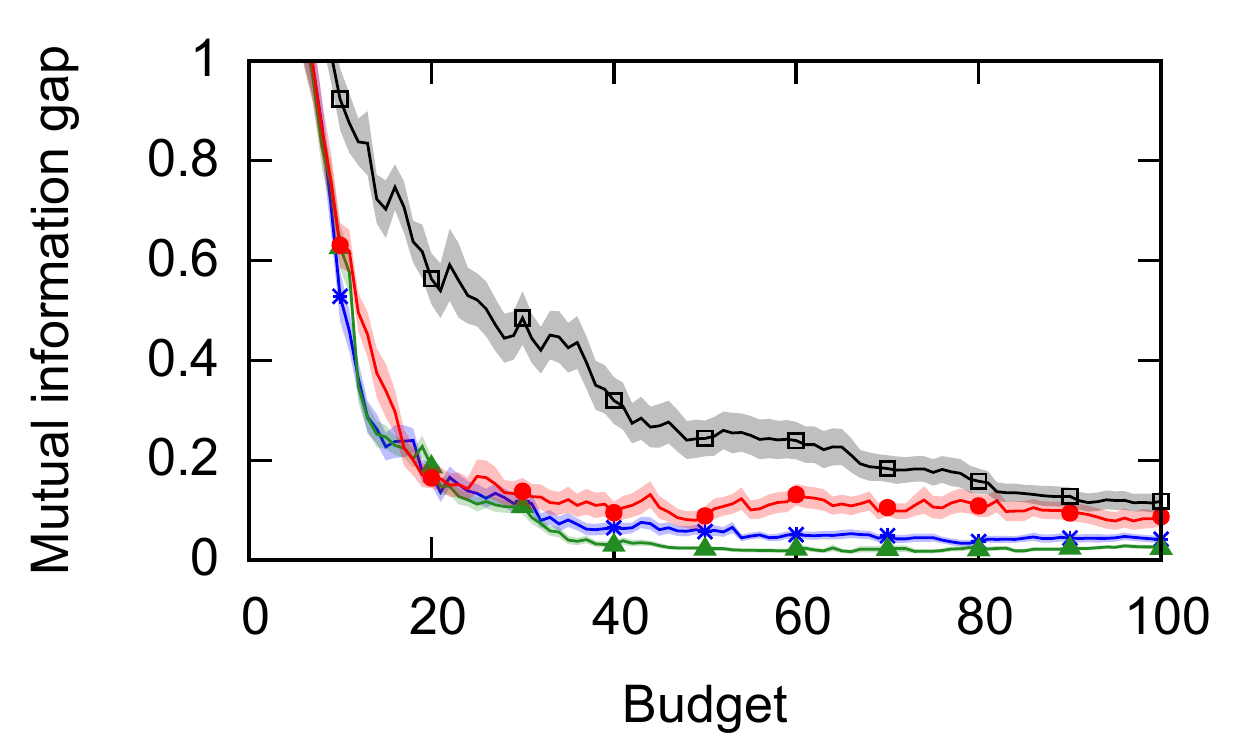}
    }
  \end{center}

\caption{\small Choices of $\psi$: MNIST: 0 vs 1 ($k=5$). Top: full experiment. Bottom: Zoom in.}
\end{figure}

\begin{figure}[h]
  \begin{center}
    \myborder{
    \includegraphics[width = 0.4\textwidth]{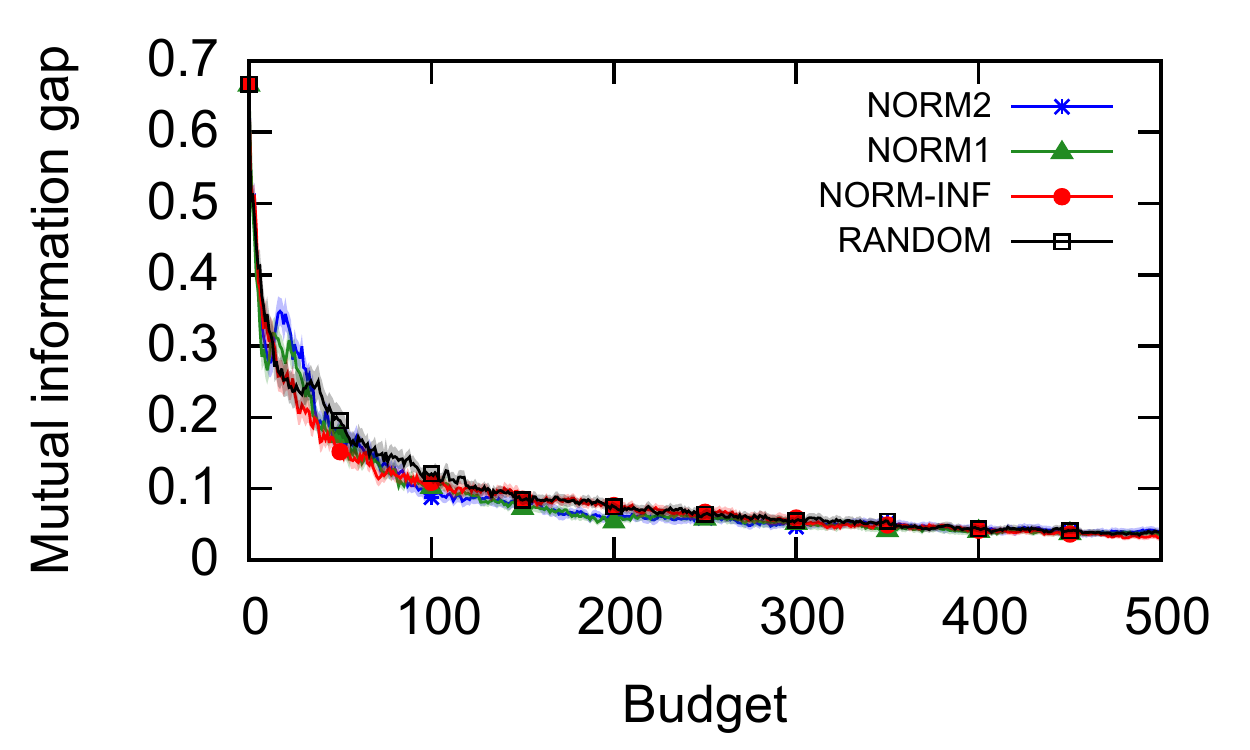} \\
    \includegraphics[width = 0.4\textwidth]{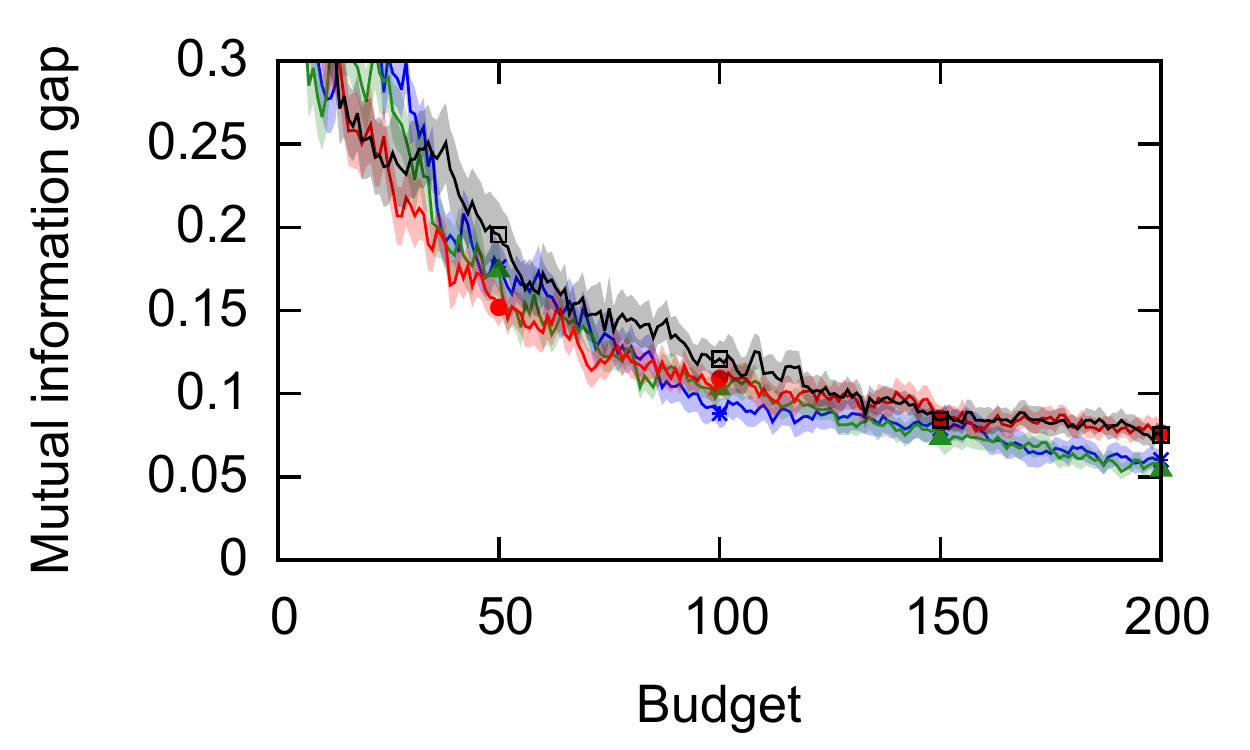}
    }
  \end{center}

\caption{\small Choices of $\psi$: MNIST: 3 vs 5 ($k=5$). Top: full experiment. Bottom: Zoom in.}
\end{figure}

\begin{figure}[h]
  \begin{center}
    \myborder{
    \includegraphics[width = 0.4\textwidth]{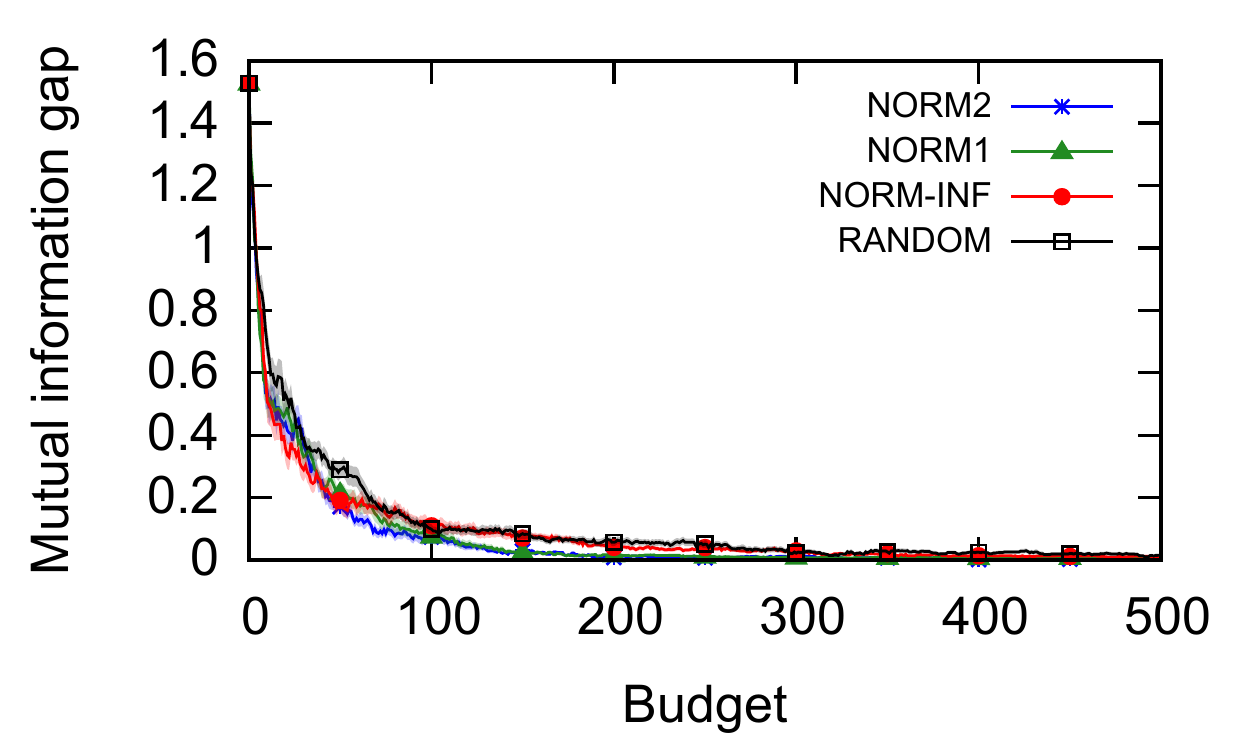} \\
    \includegraphics[width = 0.4\textwidth]{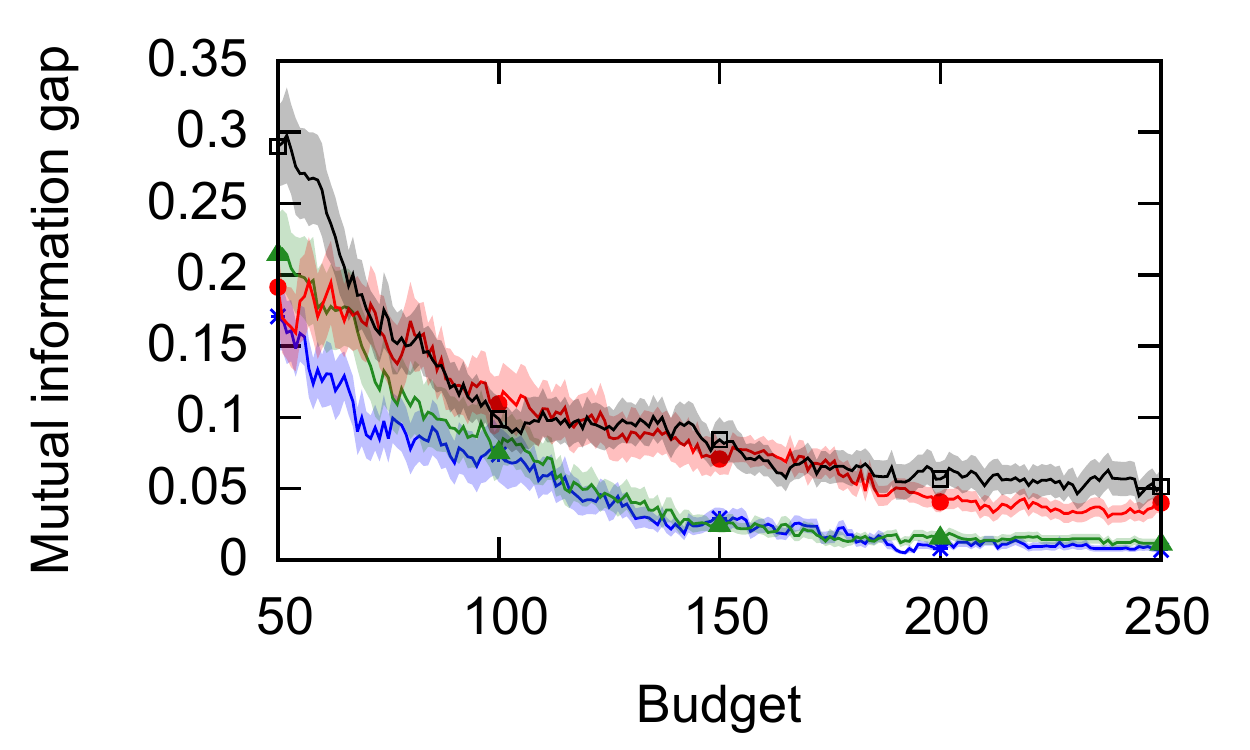}
    }
  \end{center}

\caption{\small Choices of $\psi$: MNIST: 4 vs 6 ($k=5$). Top: full experiment. Bottom: Zoom in.}
\end{figure}

\clearpage
\subsection{Comparing aggregation functions: $k = 1$}

\begin{figure}[h]
  \begin{center}
    \myborder{
    \includegraphics[width = 0.4\textwidth]{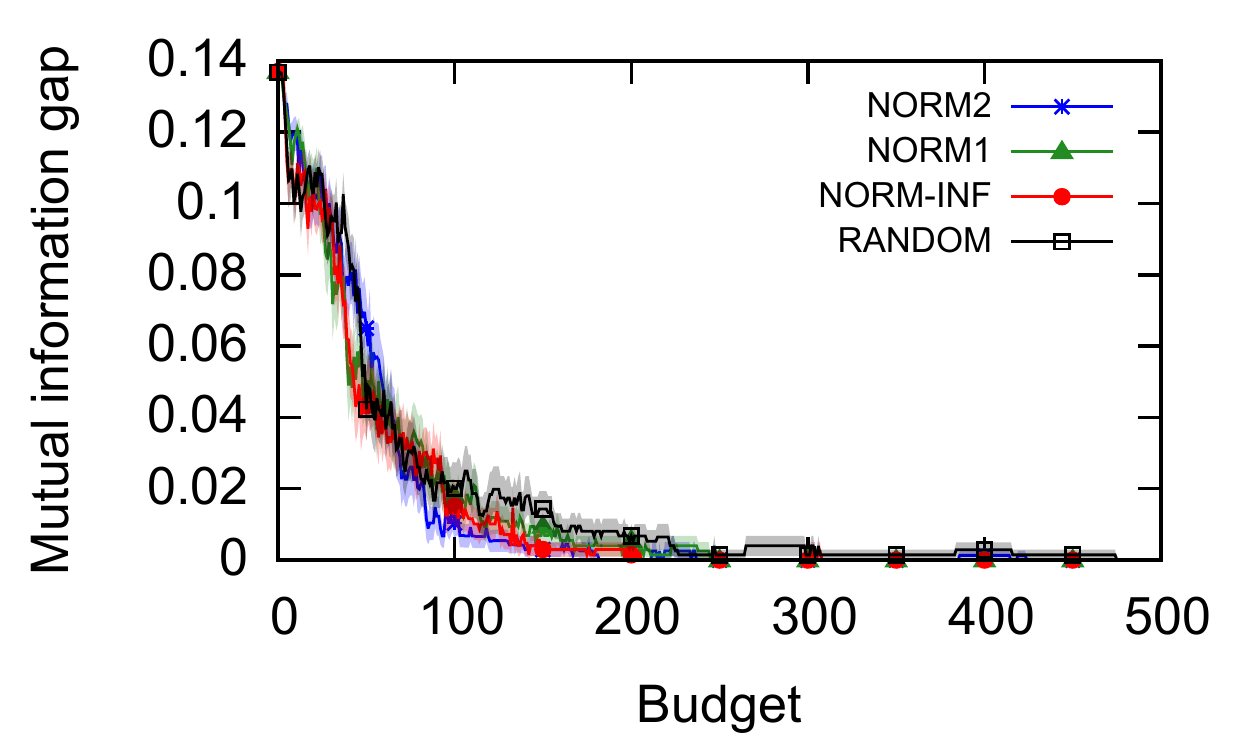} \\
    \includegraphics[width = 0.4\textwidth]{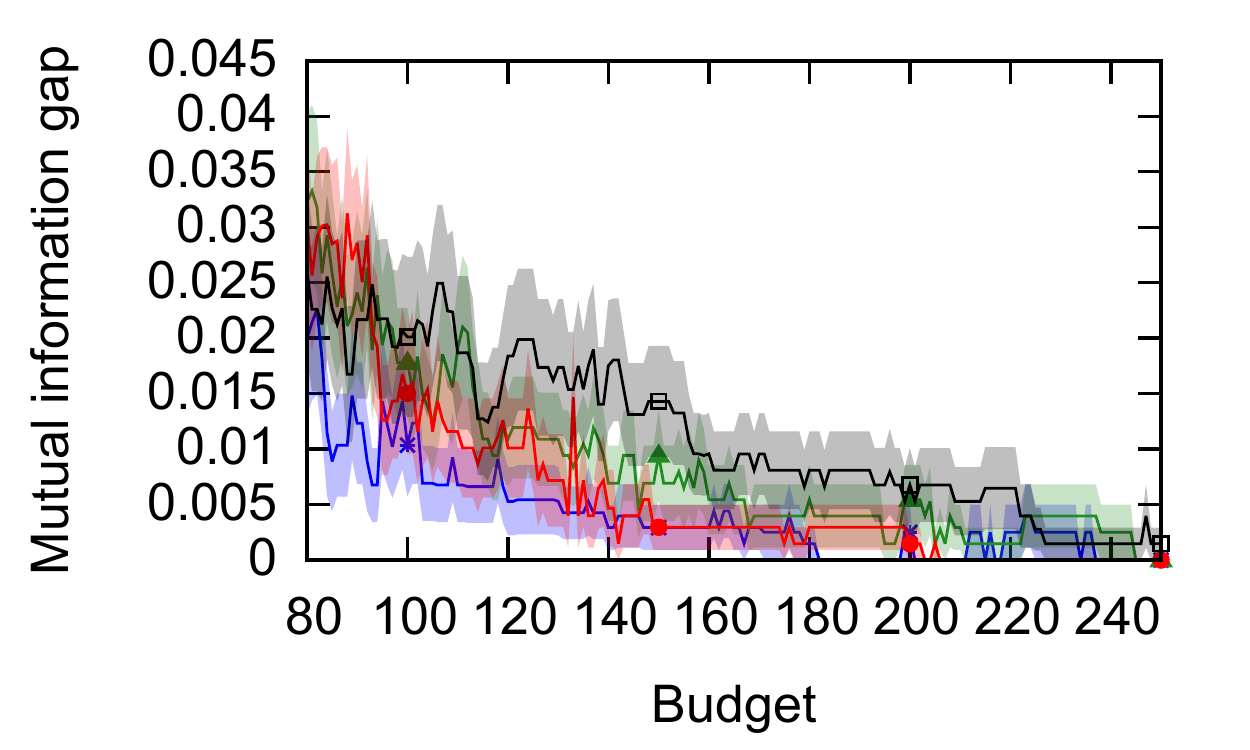}
    }
  \end{center}

\caption{\small Choices of $\psi$: BASEHOCK ($k=1$). Top: full experiment. Bottom: Zoom in.}
\end{figure}

\begin{figure}[h]
  \begin{center}
    \myborder{
    \includegraphics[width = 0.4\textwidth]{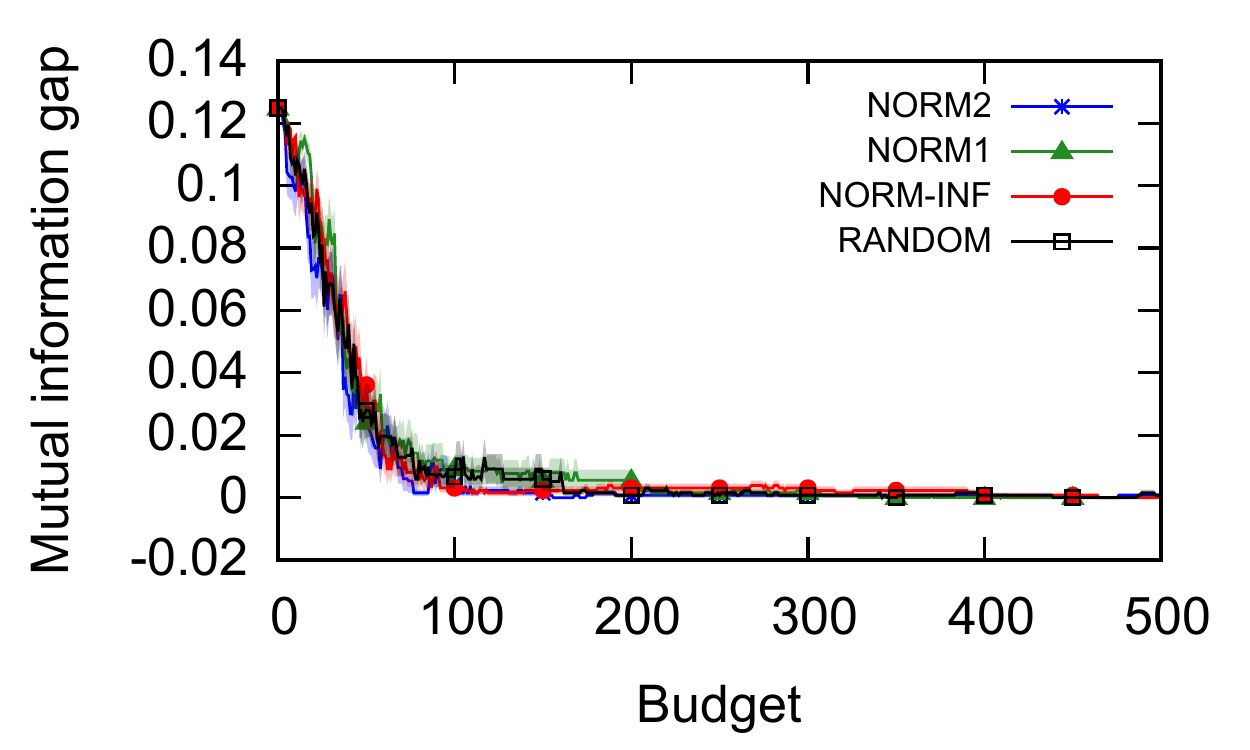} \\
    \includegraphics[width = 0.4\textwidth]{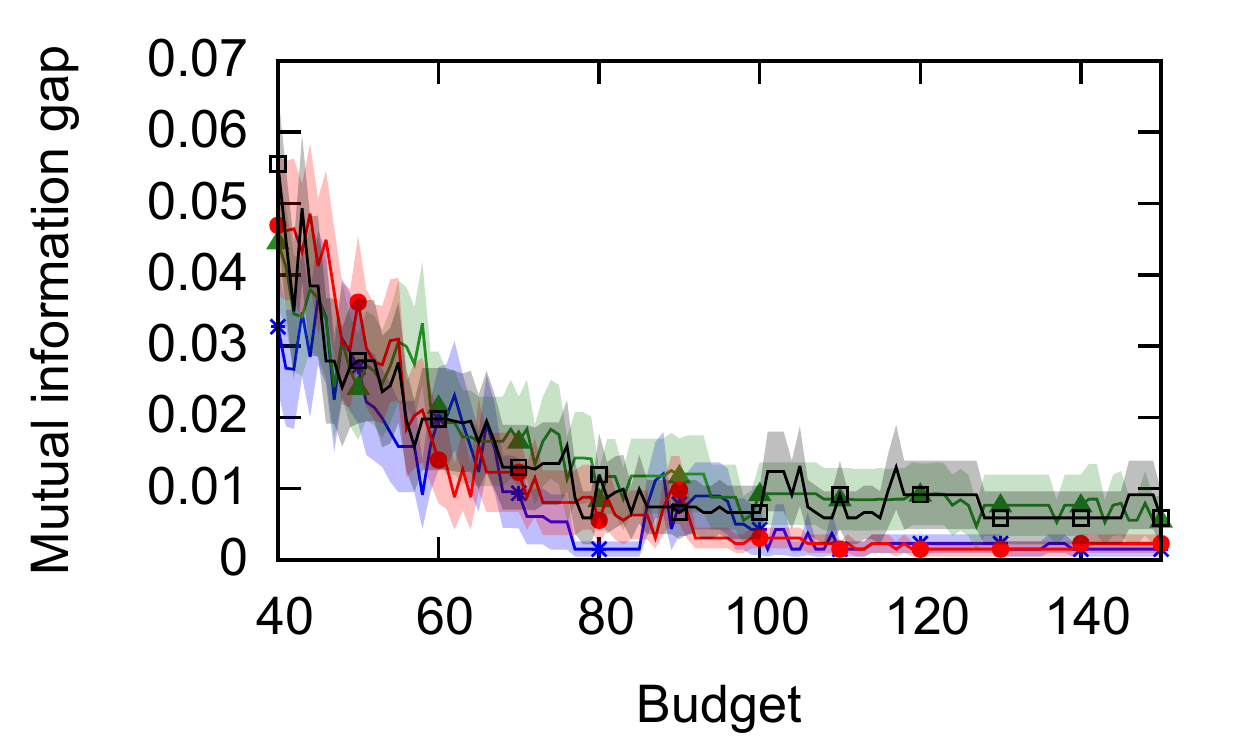}
    }
  \end{center}

\caption{\small Choices of $\psi$: PCMAC ($k=1$). Top: full experiment. Bottom: Zoom in.}
\end{figure}

\begin{figure}[h]
  \begin{center}
    \myborder{
    \includegraphics[width = 0.4\textwidth]{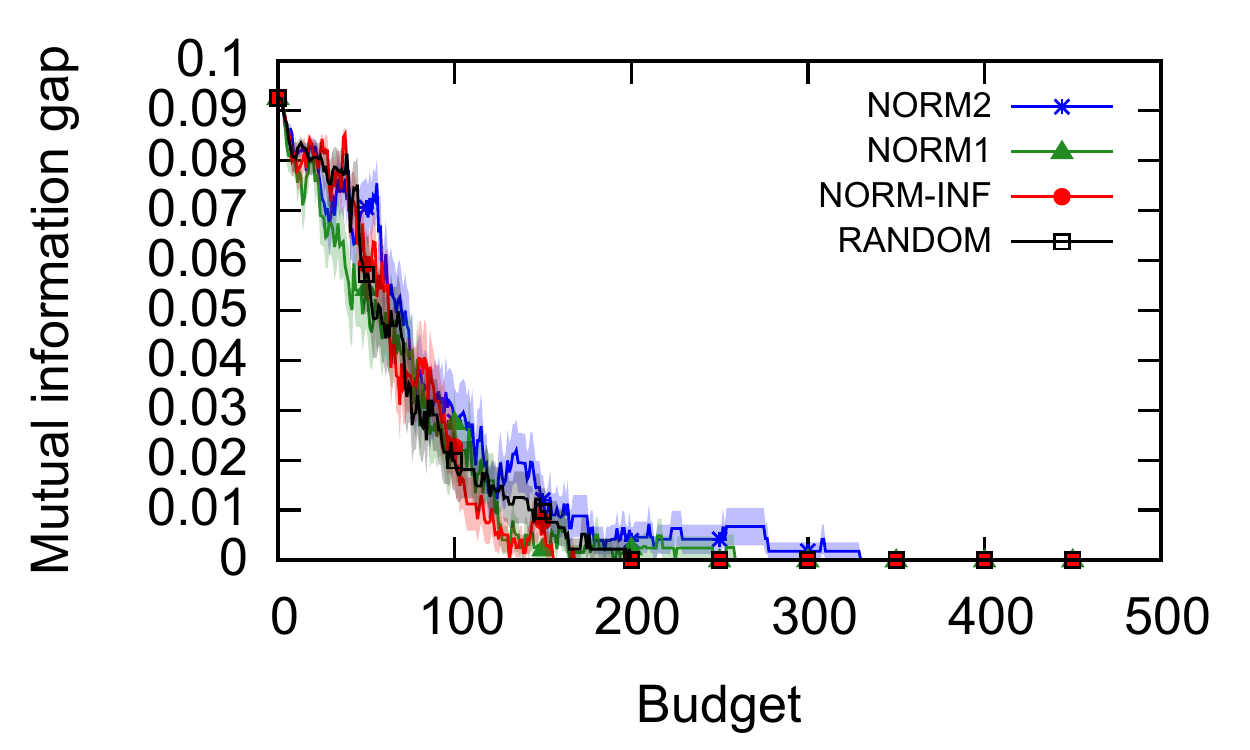} \\
    \includegraphics[width = 0.4\textwidth]{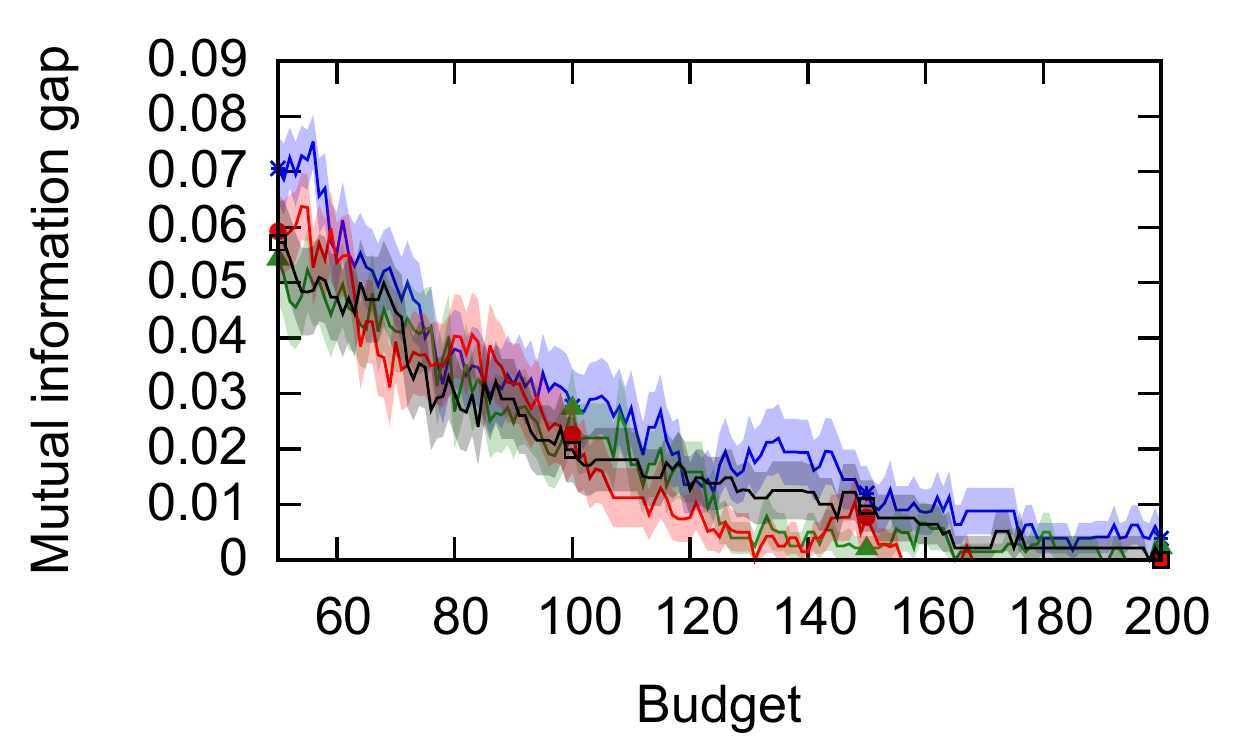}
    }
  \end{center}

\caption{\small Choices of $\psi$: RELATHE ($k=1$). Top: full experiment. Bottom: Zoom in.}
\end{figure}

\begin{figure}[h]
  \begin{center}
    \myborder{
    \includegraphics[width = 0.4\textwidth]{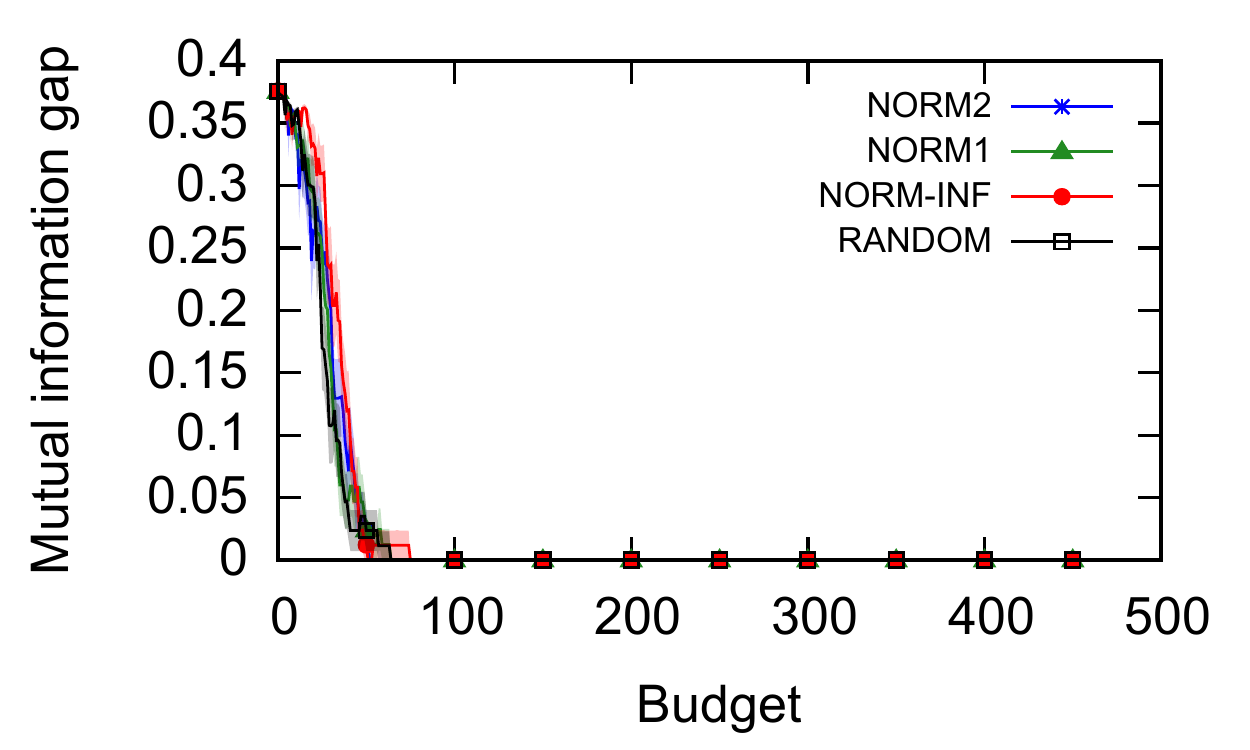} \\
    \includegraphics[width = 0.4\textwidth]{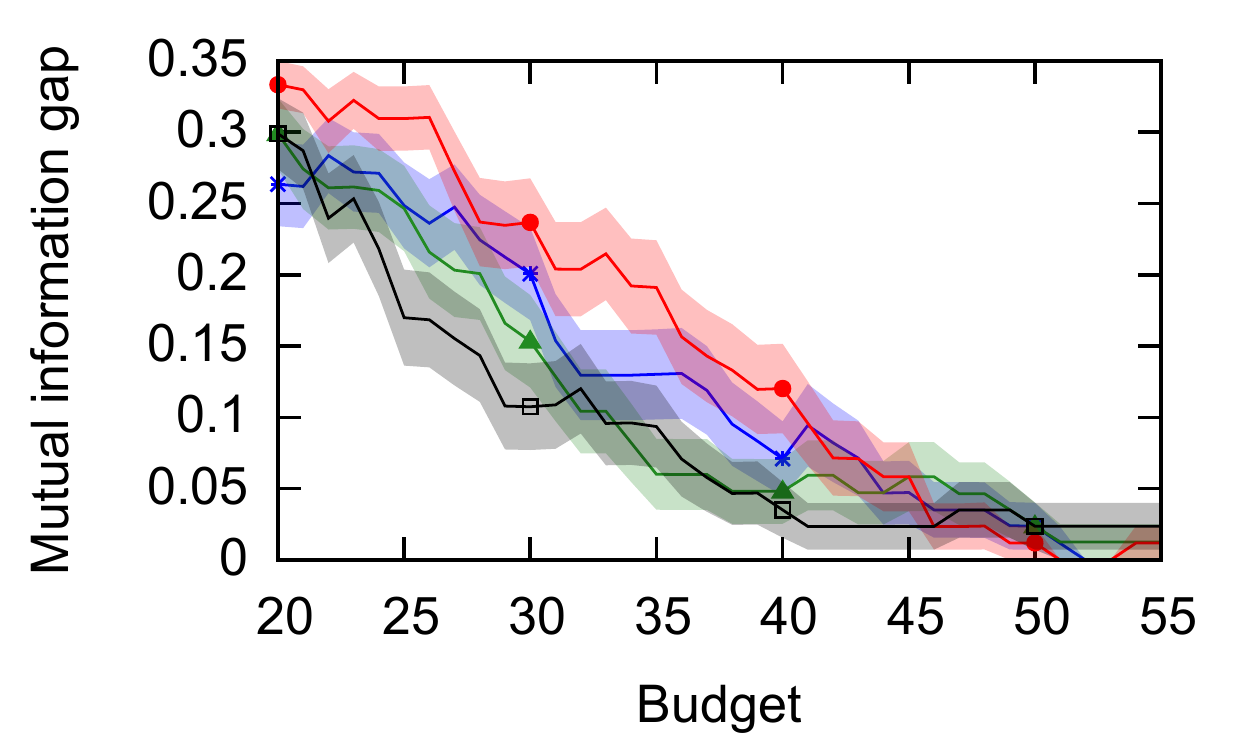}
    }
  \end{center}

\caption{\small Choices of $\psi$: MUSK ($k=1$). Top: full experiment. Bottom: Zoom in.}
\end{figure}

\begin{figure}[h]
  \begin{center}
    \myborder{
    \includegraphics[width = 0.4\textwidth]{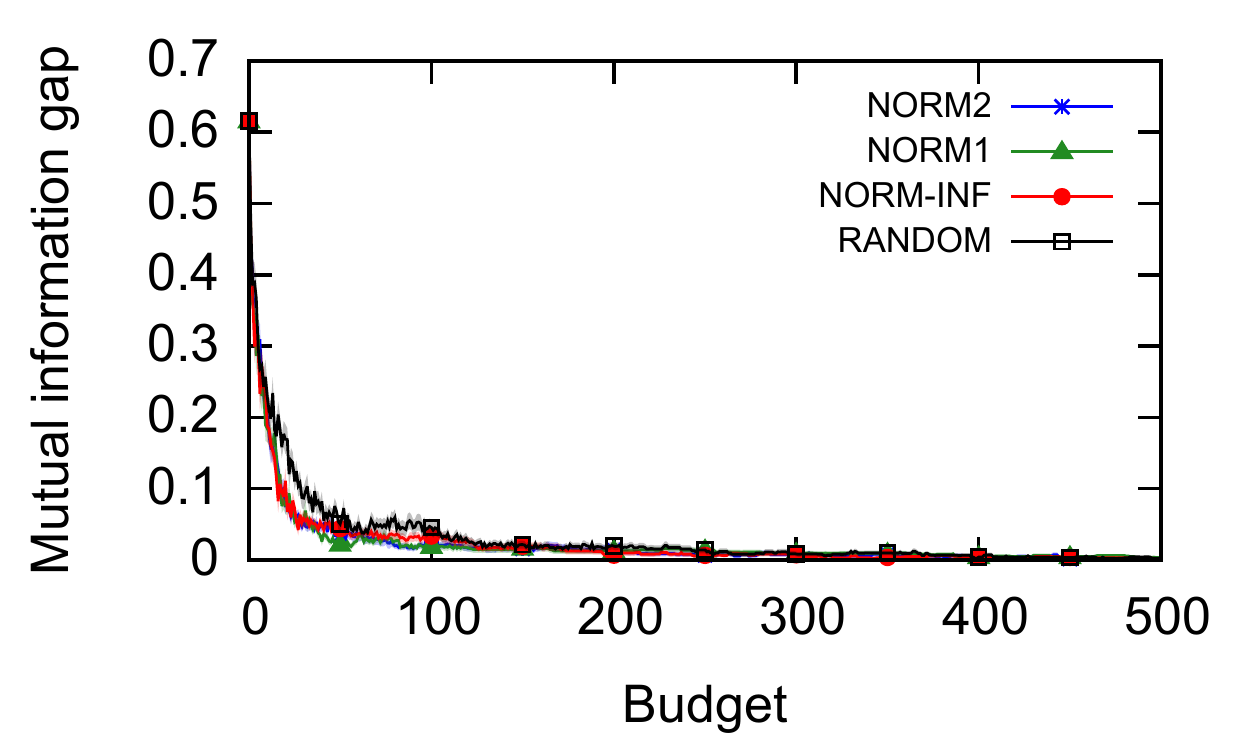} \\
    \includegraphics[width = 0.4\textwidth]{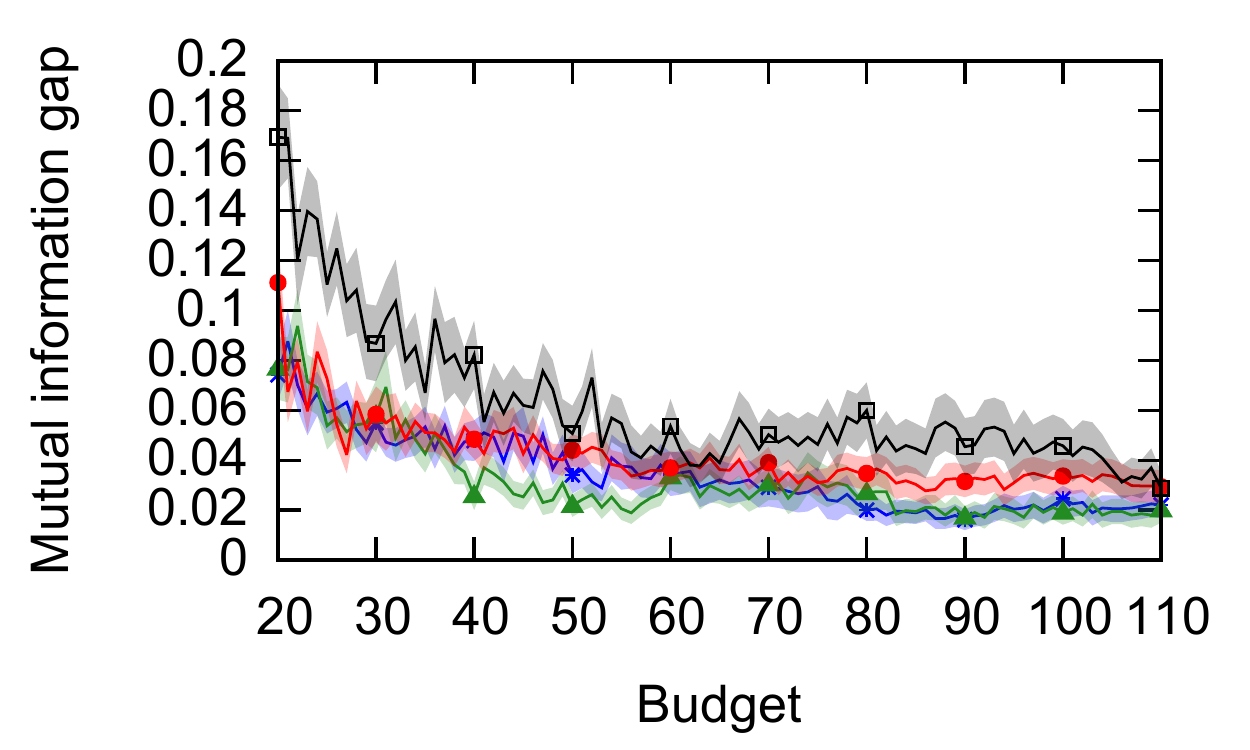}
    }
  \end{center}

\caption{\small Choices of $\psi$: MNIST: 0 vs 1 ($k=1$). Top: full experiment. Bottom: Zoom in.}
\end{figure}

\begin{figure}[h]
  \begin{center}
    \myborder{
    \includegraphics[width = 0.4\textwidth]{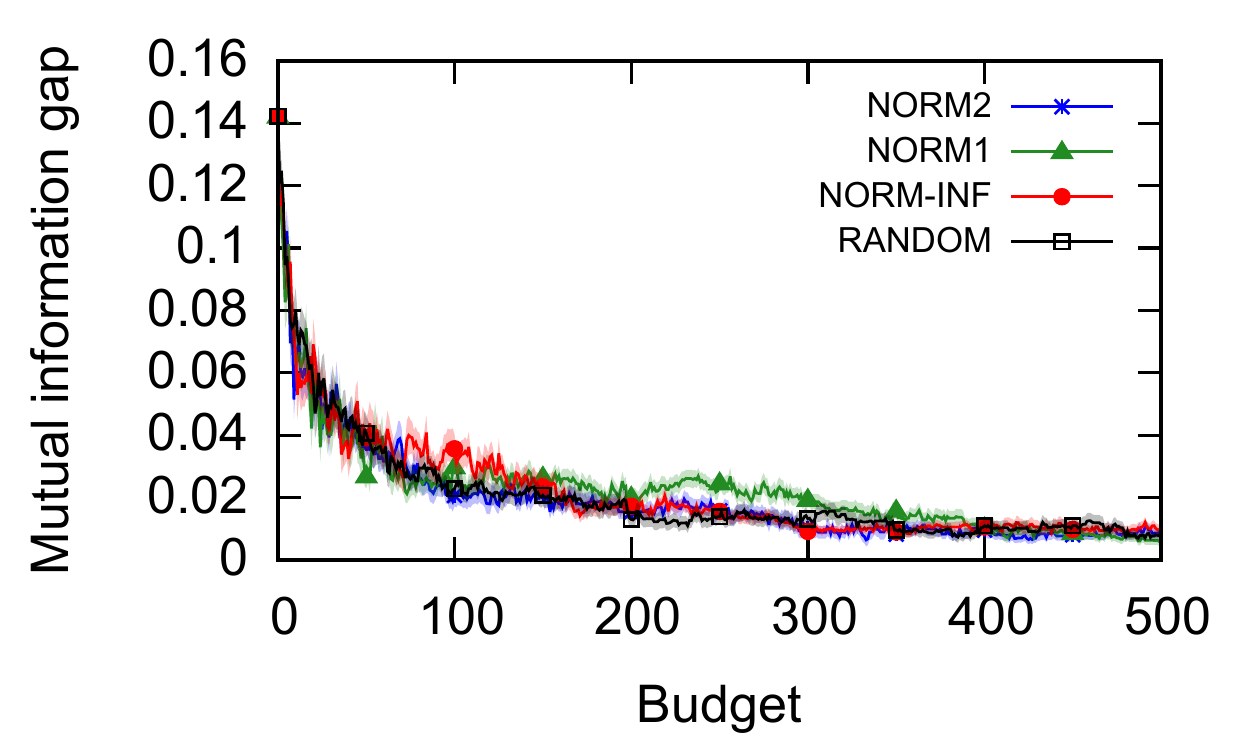} \\
    \includegraphics[width = 0.4\textwidth]{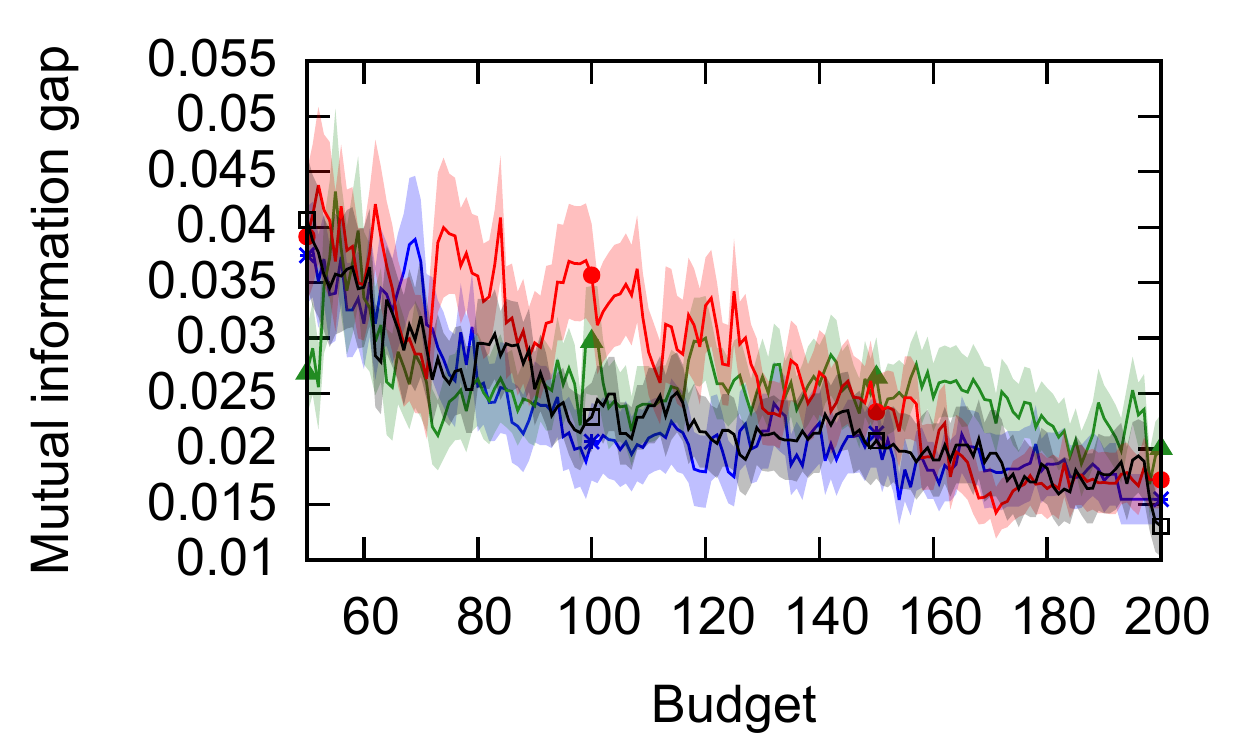}
    }
  \end{center}
\caption{\small Choices of $\psi$: MNIST: 3 vs 5 ($k=1$). Top: full experiment. Bottom: Zoom in.}
\end{figure}

\begin{figure}[h]
  \begin{center}
    \myborder{
    \includegraphics[width = 0.4\textwidth]{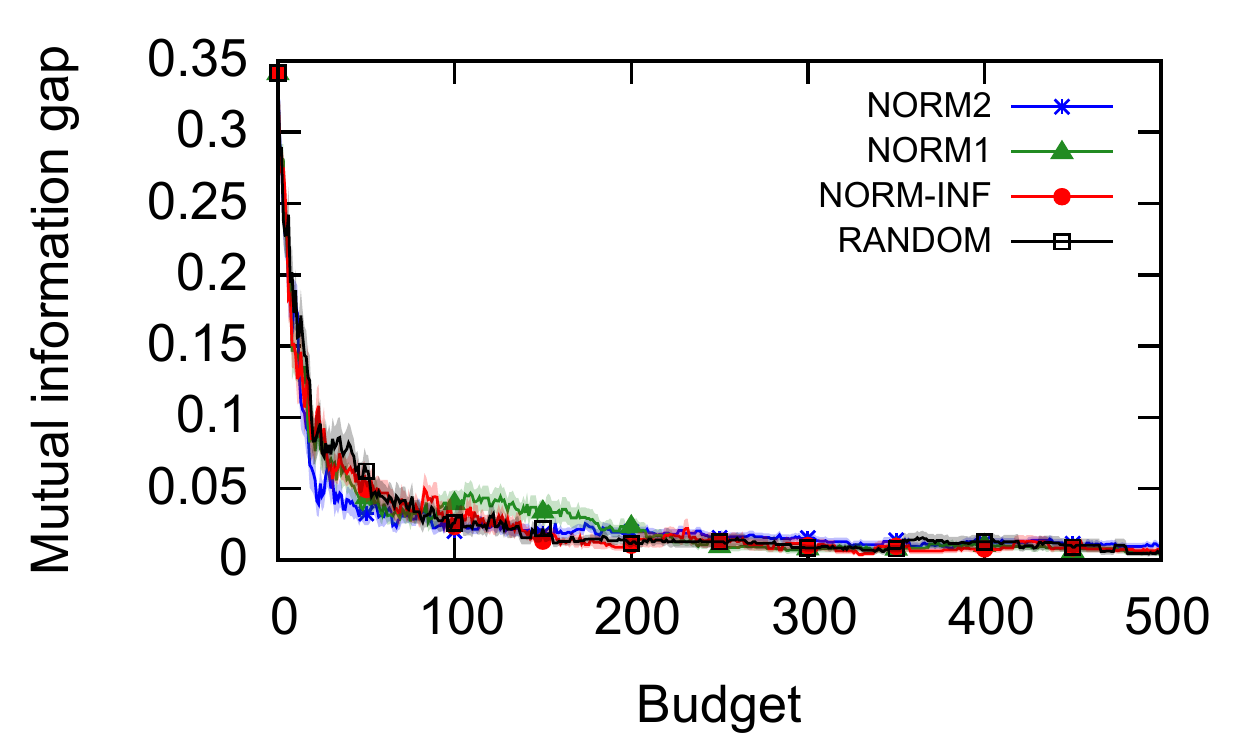} \\
    \includegraphics[width = 0.4\textwidth]{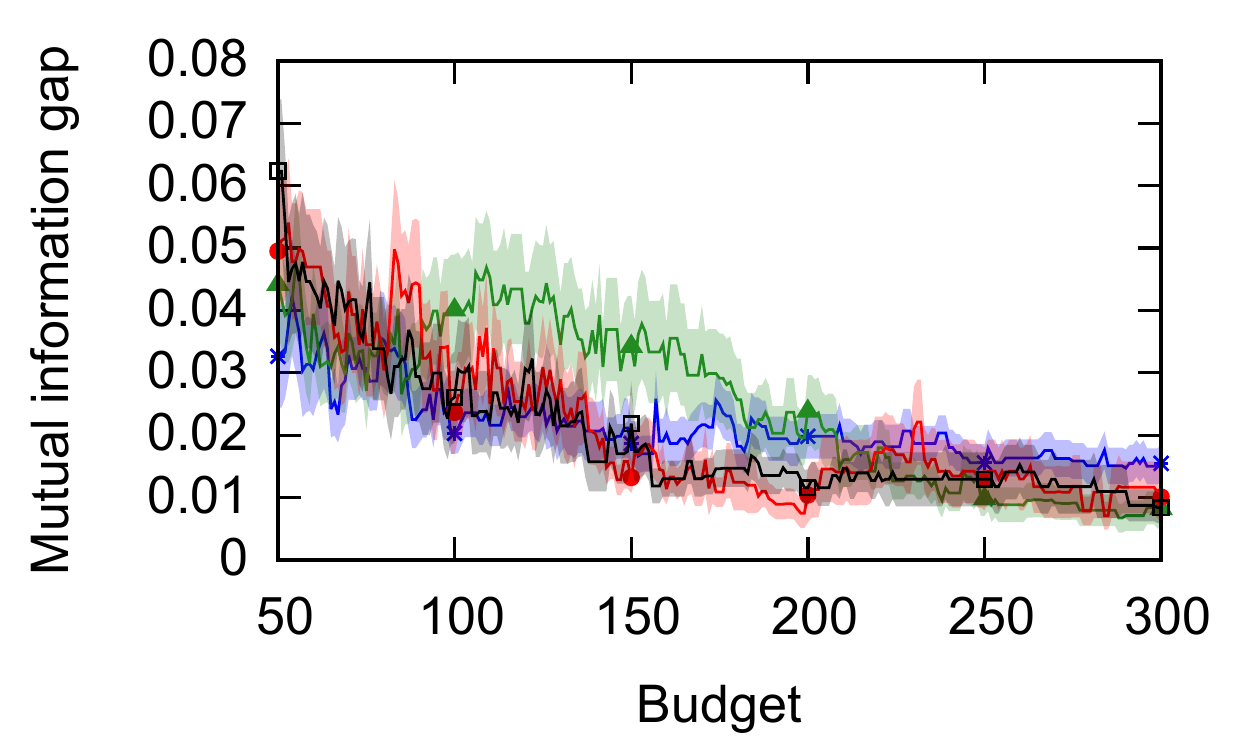}
    }
  \end{center}
\caption{\small Choices of $\psi$: MNIST: 4 vs 6 ($k=1$). Top: full experiment. Bottom: Zoom in.}
\end{figure}

\end{document}

%% file: tables/uniform/uniformtable50.tex
\begin{tabular}{cccccccccc} 
$[q_v]_{v \in V_j}$ & PROP & MAX-H & MAX-B & VAR-H & VAR-B & VAR-CP & I-H & I-B & I-CP \\ 
\toprule
$[0.01, 0.5]$ & $0.0405\pm0.0002$ & $0.0406\pm0.0002$ & $0.0407\pm0.0002$ & $0.0405\pm0.0002$ & $0.0403\pm0.0002$ & $\mathbf{0.0392\pm0.0002}$ & $0.0405\pm0.0002$ & $0.0404\pm0.0002$ & $0.0396\pm0.0002$\\ 
\toprule
$[0.03, 0.14]$ & $\mathbf{0.0692\pm0.0005}$ & $\mathbf{0.0700\pm0.0005}$ & $\mathbf{0.0698\pm0.0005}$ & $\mathbf{0.0692\pm0.0005}$ & $\mathbf{0.0700\pm0.0005}$ & $0.0708\pm0.0005$ & $\mathbf{0.0692\pm0.0005}$ & $\mathbf{0.0696\pm0.0005}$ & $\mathbf{0.0697\pm0.0005}$\\ 
\toprule
$[0.06, 0.14]$ & $\mathbf{0.0738\pm0.0006}$ & $\mathbf{0.0743\pm0.0006}$ & $\mathbf{0.0738\pm0.0006}$ & $\mathbf{0.0738\pm0.0006}$ & $0.0747\pm0.0006$ & $0.0788\pm0.0006$ & $\mathbf{0.0738\pm0.0006}$ & $\mathbf{0.0732\pm0.0006}$ & $0.0746\pm0.0006$\\ 
\toprule
$[0.12, 0.46]$ & $0.0553\pm0.0004$ & $0.0564\pm0.0005$ & $0.0555\pm0.0004$ & $0.0553\pm0.0004$ & $0.0566\pm0.0005$ & $0.0614\pm0.0005$ & $0.0553\pm0.0004$ & $0.0553\pm0.0004$ & $\mathbf{0.0531\pm0.0004}$\\ 
\toprule
$[0.49, 0.26]$ & $0.0397\pm0.0003$ & $0.0406\pm0.0004$ & $0.0397\pm0.0004$ & $0.0397\pm0.0003$ & $0.0401\pm0.0004$ & $0.0412\pm0.0004$ & $0.0397\pm0.0003$ & $0.0389\pm0.0003$ & $\mathbf{0.0381\pm0.0003}$\\ 
\toprule
$[0.05, 0.26, 0.18, 0.15]$ & $\mathbf{0.0704\pm0.0005}$ & $0.0734\pm0.0005$ & $0.0715\pm0.0005$ & $\mathbf{0.0704\pm0.0005}$ & $\mathbf{0.0708\pm0.0005}$ & $0.0790\pm0.0006$ & $\mathbf{0.0704\pm0.0005}$ & $\mathbf{0.0708\pm0.0005}$ & $\mathbf{0.0704\pm0.0005}$\\ 
\toprule
$[0.07, 0.38, 0.29, 0.35]$ & $0.0586\pm0.0005$ & $0.0610\pm0.0005$ & $0.0590\pm0.0005$ & $0.0586\pm0.0005$ & $0.0588\pm0.0005$ & $0.0654\pm0.0005$ & $0.0586\pm0.0005$ & $0.0586\pm0.0005$ & $\mathbf{0.0563\pm0.0005}$\\
\end{tabular} 

%% file: tables/uniform/uniformtable100.tex
\begin{tabular}{cccccccccc} 
$[q_v]_{v \in V_j}$ & PROP & MAX-H & MAX-B & VAR-H & VAR-B & VAR-CP & I-H & I-B & I-CP \\ 
\toprule
$[0.01, 0.5]$ & $0.0304\pm0.0001$ & $0.0318\pm0.0001$ & $0.0308\pm0.0001$ & $0.0304\pm0.0001$ & $0.0322\pm0.0001$ & $0.0332\pm0.0001$ & $0.0304\pm0.0001$ & $0.0288\pm0.0001$ & $\mathbf{0.0278\pm0.0001}$\\ 
\toprule
$[0.03, 0.14]$ & $\mathbf{0.0488\pm0.0004}$ & $\mathbf{0.0491\pm0.0004}$ & $\mathbf{0.0492\pm0.0004}$ & $\mathbf{0.0488\pm0.0004}$ & $\mathbf{0.0494\pm0.0004}$ & $0.0521\pm0.0004$ & $\mathbf{0.0488\pm0.0004}$ & $\mathbf{0.0490\pm0.0004}$ & $\mathbf{0.0491\pm0.0004}$\\ 
\toprule
$[0.06, 0.14]$ & $\mathbf{0.0507\pm0.0004}$ & $\mathbf{0.0511\pm0.0004}$ & $\mathbf{0.0508\pm0.0004}$ & $\mathbf{0.0507\pm0.0004}$ & $0.0519\pm0.0004$ & $0.0540\pm0.0004$ & $\mathbf{0.0507\pm0.0004}$ & $\mathbf{0.0507\pm0.0004}$ & $\mathbf{0.0508\pm0.0004}$\\ 
\toprule
$[0.12, 0.46]$ & $0.0387\pm0.0003$ & $0.0403\pm0.0003$ & $0.0388\pm0.0003$ & $0.0387\pm0.0003$ & $0.0404\pm0.0003$ & $0.0434\pm0.0003$ & $0.0387\pm0.0003$ & $0.0362\pm0.0003$ & $\mathbf{0.0354\pm0.0003}$\\ 
\toprule
$[0.49, 0.26]$ & $0.0272\pm0.0002$ & $0.0275\pm0.0002$ & $0.0274\pm0.0002$ & $0.0272\pm0.0002$ & $0.0278\pm0.0002$ & $0.0279\pm0.0002$ & $0.0272\pm0.0002$ & $0.0259\pm0.0002$ & $\mathbf{0.0253\pm0.0002}$\\ 
\toprule
$[0.05, 0.26, 0.18, 0.15]$ & $\mathbf{0.0480\pm0.0004}$ & $0.0499\pm0.0004$ & $\mathbf{0.0480\pm0.0004}$ & $\mathbf{0.0480\pm0.0004}$ & $0.0495\pm0.0004$ & $0.0560\pm0.0004$ & $\mathbf{0.0480\pm0.0004}$ & $\mathbf{0.0478\pm0.0004}$ & $\mathbf{0.0477\pm0.0004}$\\ 
\toprule
$[0.07, 0.38, 0.29, 0.35]$ & $0.0392\pm0.0003$ & $0.0405\pm0.0003$ & $0.0392\pm0.0003$ & $0.0392\pm0.0003$ & $0.0398\pm0.0003$ & $0.0450\pm0.0003$ & $0.0392\pm0.0003$ & $0.0388\pm0.0003$ & $\mathbf{0.0370\pm0.0003}$\\
\end{tabular} 

%% file: tables/uniform/uniformtable300.tex
\begin{tabular}{cccccccccc} 
$[q_v]_{v \in V_j}$ & PROP & MAX-H & MAX-B & VAR-H & VAR-B & VAR-CP & I-H & I-B & I-CP \\ 
\toprule
$[0.01, 0.5]$ & $0.0159\pm0.0001$ & $0.0173\pm0.0001$ & $0.0162\pm0.0001$ & $0.0159\pm0.0001$ & $0.0186\pm0.0001$ & $0.0216\pm0.0001$ & $0.0159\pm0.0001$ & $\mathbf{0.0133\pm0.0001}$ & $\mathbf{0.0134\pm0.0001}$\\ 
\toprule
$[0.03, 0.14]$ & $\mathbf{0.0280\pm0.0002}$ & $\mathbf{0.0281\pm0.0002}$ & $\mathbf{0.0279\pm0.0002}$ & $\mathbf{0.0280\pm0.0002}$ & $0.0288\pm0.0002$ & $0.0308\pm0.0002$ & $\mathbf{0.0280\pm0.0002}$ & $\mathbf{0.0280\pm0.0002}$ & $\mathbf{0.0280\pm0.0002}$\\ 
\toprule
$[0.06, 0.14]$ & $\mathbf{0.0281\pm0.0002}$ & $\mathbf{0.0284\pm0.0002}$ & $\mathbf{0.0281\pm0.0002}$ & $\mathbf{0.0281\pm0.0002}$ & $\mathbf{0.0284\pm0.0002}$ & $0.0292\pm0.0002$ & $\mathbf{0.0281\pm0.0002}$ & $\mathbf{0.0282\pm0.0002}$ & $\mathbf{0.0282\pm0.0002}$\\ 
\toprule
$[0.12, 0.46]$ & $0.0206\pm0.0002$ & $0.0219\pm0.0002$ & $0.0209\pm0.0002$ & $0.0206\pm0.0002$ & $0.0228\pm0.0002$ & $0.0232\pm0.0002$ & $0.0206\pm0.0002$ & $\mathbf{0.0189\pm0.0002}$ & $\mathbf{0.0188\pm0.0002}$\\ 
\toprule
$[0.49, 0.26]$ & $0.0149\pm0.0001$ & $0.0152\pm0.0001$ & $0.0148\pm0.0001$ & $0.0149\pm0.0001$ & $0.0153\pm0.0001$ & $0.0153\pm0.0001$ & $0.0149\pm0.0001$ & $\mathbf{0.0135\pm0.0001}$ & $\mathbf{0.0135\pm0.0001}$\\ 
\toprule
$[0.05, 0.26, 0.18, 0.15]$ & $\mathbf{0.0267\pm0.0002}$ & $0.0274\pm0.0002$ & $\mathbf{0.0269\pm0.0002}$ & $\mathbf{0.0267\pm0.0002}$ & $0.0284\pm0.0002$ & $0.0304\pm0.0002$ & $\mathbf{0.0267\pm0.0002}$ & $\mathbf{0.0265\pm0.0002}$ & $\mathbf{0.0265\pm0.0002}$\\ 
\toprule
$[0.07, 0.38, 0.29, 0.35]$ & $0.0207\pm0.0002$ & $0.0220\pm0.0002$ & $0.0210\pm0.0002$ & $0.0207\pm0.0002$ & $0.0224\pm0.0002$ & $0.0249\pm0.0002$ & $0.0207\pm0.0002$ & $\mathbf{0.0199\pm0.0002}$ & $\mathbf{0.0197\pm0.0002}$\\
\end{tabular} 

%% file: tables/uniform/uniformtable500.tex
\begin{tabular}{cccccccccc} 
$[q_v]_{v \in V_j}$ & PROP & MAX-H & MAX-B & VAR-H & VAR-B & VAR-CP & I-H & I-B & I-CP \\ 
\toprule
$[0.01, 0.5]$ & $0.0119\pm0.0001$ & $0.0132\pm0.0001$ & $0.0123\pm0.0001$ & $0.0119\pm0.0001$ & $0.0157\pm0.0001$ & $0.0181\pm0.0001$ & $0.0119\pm0.0001$ & $\mathbf{0.0106\pm0.0001}$ & $\mathbf{0.0107\pm0.0001}$\\ 
\toprule
$[0.03, 0.14]$ & $\mathbf{0.0225\pm0.0002}$ & $\mathbf{0.0223\pm0.0002}$ & $\mathbf{0.0226\pm0.0002}$ & $\mathbf{0.0225\pm0.0002}$ & $0.0229\pm0.0002$ & $0.0239\pm0.0002$ & $\mathbf{0.0225\pm0.0002}$ & $\mathbf{0.0226\pm0.0002}$ & $\mathbf{0.0226\pm0.0002}$\\ 
\toprule
$[0.06, 0.14]$ & $\mathbf{0.0219\pm0.0002}$ & $\mathbf{0.0222\pm0.0002}$ & $\mathbf{0.0220\pm0.0002}$ & $\mathbf{0.0219\pm0.0002}$ & $\mathbf{0.0221\pm0.0002}$ & $0.0224\pm0.0002$ & $\mathbf{0.0219\pm0.0002}$ & $\mathbf{0.0219\pm0.0002}$ & $\mathbf{0.0219\pm0.0002}$\\ 
\toprule
$[0.12, 0.46]$ & $0.0165\pm0.0001$ & $0.0174\pm0.0001$ & $0.0167\pm0.0001$ & $0.0165\pm0.0001$ & $0.0178\pm0.0001$ & $0.0179\pm0.0001$ & $0.0165\pm0.0001$ & $\mathbf{0.0142\pm0.0001}$ & $\mathbf{0.0143\pm0.0001}$\\ 
\toprule
$[0.49, 0.26]$ & $0.0117\pm0.0001$ & $0.0119\pm0.0001$ & $0.0117\pm0.0001$ & $0.0117\pm0.0001$ & $0.0119\pm0.0001$ & $0.0119\pm0.0001$ & $0.0117\pm0.0001$ & $\mathbf{0.0105\pm0.0001}$ & $\mathbf{0.0106\pm0.0001}$\\ 
\toprule
$[0.05, 0.26, 0.18, 0.15]$ & $\mathbf{0.0200\pm0.0002}$ & $0.0205\pm0.0002$ & $\mathbf{0.0201\pm0.0002}$ & $\mathbf{0.0200\pm0.0002}$ & $0.0210\pm0.0002$ & $0.0222\pm0.0002$ & $\mathbf{0.0200\pm0.0002}$ & $\mathbf{0.0201\pm0.0002}$ & $\mathbf{0.0201\pm0.0002}$\\ 
\toprule
$[0.07, 0.38, 0.29, 0.35]$ & $0.0166\pm0.0001$ & $0.0171\pm0.0001$ & $0.0168\pm0.0001$ & $0.0166\pm0.0001$ & $0.0179\pm0.0001$ & $0.0188\pm0.0001$ & $0.0166\pm0.0001$ & $\mathbf{0.0155\pm0.0001}$ & $\mathbf{0.0155\pm0.0001}$\\
\end{tabular} 

%% file: tables/min_max_arm/min_max_armstable50.tex
\begin{tabular}{cccccccccc} 
$[q_v]_{v \in V_j}$ & PROP & MAX-H & MAX-B & VAR-H & VAR-B & VAR-CP & I-H & I-B & I-CP \\ 
\toprule
$[0.1, 0.1]$ & $\mathbf{0.0747\pm0.0006}$ & $0.0761\pm0.0006$ & $\mathbf{0.0754\pm0.0006}$ & $\mathbf{0.0747\pm0.0006}$ & $0.0768\pm0.0006$ & $0.0829\pm0.0006$ & $\mathbf{0.0747\pm0.0006}$ & $\mathbf{0.0751\pm0.0006}$ & $\mathbf{0.0751\pm0.0006}$\\ 
\toprule
$[0.1, 0.5]$ & $0.0558\pm0.0004$ & $0.0580\pm0.0004$ & $0.0563\pm0.0004$ & $0.0558\pm0.0004$ & $0.0576\pm0.0005$ & $0.0641\pm0.0005$ & $0.0558\pm0.0004$ & $0.0553\pm0.0004$ & $\mathbf{0.0527\pm0.0004}$\\ 
\toprule
$[0.5, 0.5]$ & $0.0192\pm0.0002$ & $0.0195\pm0.0002$ & $0.0195\pm0.0002$ & $0.0192\pm0.0002$ & $0.0192\pm0.0002$ & $0.0195\pm0.0002$ & $0.0192\pm0.0002$ & $0.0187\pm0.0002$ & $\mathbf{0.0169\pm0.0002}$\\ 
\toprule
$[0.01, 0.5]$ & $0.0408\pm0.0002$ & $0.0407\pm0.0002$ & $0.0410\pm0.0002$ & $0.0408\pm0.0002$ & $0.0403\pm0.0002$ & $\mathbf{0.0398\pm0.0002}$ & $0.0408\pm0.0002$ & $0.0409\pm0.0002$ & $\mathbf{0.0401\pm0.0002}$\\ 
\toprule
$[0.001, 0.5]$ & $0.0156\pm0.0002$ & $0.0149\pm0.0002$ & $0.0156\pm0.0002$ & $0.0156\pm0.0002$ & $0.0148\pm0.0002$ & $\mathbf{0.0131\pm0.0001}$ & $0.0156\pm0.0002$ & $0.0156\pm0.0002$ & $0.0153\pm0.0002$\\ 
\toprule
$[0.01, 0.01]$ & $0.0517\pm0.0002$ & $0.0511\pm0.0002$ & $0.0511\pm0.0002$ & $0.0517\pm0.0002$ & $0.0508\pm0.0002$ & $\mathbf{0.0483\pm0.0002}$ & $0.0517\pm0.0002$ & $0.0514\pm0.0002$ & $0.0512\pm0.0002$\\ 
\toprule
$[0.001, 0.001]$ & $0.0114\pm0.0001$ & $0.0113\pm0.0001$ & $0.0113\pm0.0001$ & $0.0114\pm0.0001$ & $\mathbf{0.0113\pm0.0001}$ & $\mathbf{0.0110\pm0.0001}$ & $0.0114\pm0.0001$ & $0.0114\pm0.0001$ & $0.0113\pm0.0001$\\
\end{tabular} 

%% file: tables/min_max_arm/min_max_armstable50_1.tex
\begin{tabular}{cccccccccc} 
$[q_v]_{v \in V_j}$ & PROP & MAX-H & MAX-B & VAR-H & VAR-B & VAR-CP & I-H & I-B & I-CP \\ 
\toprule
$2 \times 0.01, 6 \times 0.5, $ & $0.0573\pm0.0003$ & $0.0586\pm0.0003$ & $0.0586\pm0.0003$ & $0.0573\pm0.0003$ & $0.0571\pm0.0003$ & $0.0594\pm0.0003$ & $0.0573\pm0.0003$ & $0.0574\pm0.0003$ & $\mathbf{0.0560\pm0.0003}$\\ 
\toprule
$3 \times 0.01, 5 \times 0.5, $ & $0.0547\pm0.0003$ & $0.0560\pm0.0003$ & $0.0560\pm0.0003$ & $0.0547\pm0.0003$ & $0.0548\pm0.0003$ & $0.0559\pm0.0003$ & $0.0547\pm0.0003$ & $\mathbf{0.0546\pm0.0003}$ & $\mathbf{0.0541\pm0.0003}$\\ 
\toprule
$2 \times 0.001, 6 \times 0.5, $ & $\mathbf{0.0507\pm0.0003}$ & $0.0524\pm0.0003$ & $0.0524\pm0.0003$ & $\mathbf{0.0507\pm0.0003}$ & $0.0510\pm0.0003$ & $0.0530\pm0.0003$ & $\mathbf{0.0507\pm0.0003}$ & $0.0510\pm0.0003$ & $\mathbf{0.0503\pm0.0003}$\\ 
\toprule
$4 \times 0.01, 4 \times 0.5, $ & $\mathbf{0.0485\pm0.0003}$ & $0.0496\pm0.0003$ & $0.0494\pm0.0003$ & $\mathbf{0.0485\pm0.0003}$ & $\mathbf{0.0484\pm0.0003}$ & $\mathbf{0.0482\pm0.0003}$ & $\mathbf{0.0485\pm0.0003}$ & $\mathbf{0.0484\pm0.0003}$ & $\mathbf{0.0483\pm0.0003}$\\ 
\toprule
$5 \times 0.01, 3 \times 0.5, $ & $0.0482\pm0.0003$ & $0.0491\pm0.0003$ & $0.0491\pm0.0003$ & $0.0482\pm0.0003$ & $0.0482\pm0.0003$ & $\mathbf{0.0470\pm0.0003}$ & $0.0482\pm0.0003$ & $0.0482\pm0.0003$ & $0.0482\pm0.0003$\\ 
\toprule
$3 \times 0.001, 5 \times 0.5, $ & $\mathbf{0.0442\pm0.0003}$ & $0.0453\pm0.0003$ & $0.0452\pm0.0003$ & $\mathbf{0.0442\pm0.0003}$ & $\mathbf{0.0441\pm0.0003}$ & $\mathbf{0.0439\pm0.0003}$ & $\mathbf{0.0442\pm0.0003}$ & $\mathbf{0.0439\pm0.0003}$ & $\mathbf{0.0438\pm0.0003}$\\ 
\toprule
$6 \times 0.01, 2 \times 0.5, $ & $0.0447\pm0.0003$ & $0.0449\pm0.0003$ & $0.0448\pm0.0003$ & $0.0447\pm0.0003$ & $0.0448\pm0.0003$ & $\mathbf{0.0439\pm0.0003}$ & $0.0447\pm0.0003$ & $0.0448\pm0.0003$ & $0.0445\pm0.0003$\\
\end{tabular} 

%% file: tables/min_max_arm/min_max_armstable100.tex
\begin{tabular}{cccccccccc} 
$[q_v]_{v \in V_j}$ & PROP & MAX-H & MAX-B & VAR-H & VAR-B & VAR-CP & I-H & I-B & I-CP \\ 
\toprule
$[0.1, 0.1]$ & $\mathbf{0.0528\pm0.0004}$ & $\mathbf{0.0527\pm0.0004}$ & $\mathbf{0.0532\pm0.0004}$ & $\mathbf{0.0528\pm0.0004}$ & $\mathbf{0.0533\pm0.0004}$ & $0.0556\pm0.0004$ & $\mathbf{0.0528\pm0.0004}$ & $\mathbf{0.0527\pm0.0004}$ & $\mathbf{0.0533\pm0.0004}$\\ 
\toprule
$[0.1, 0.5]$ & $0.0362\pm0.0003$ & $0.0385\pm0.0003$ & $0.0366\pm0.0003$ & $0.0362\pm0.0003$ & $0.0391\pm0.0003$ & $0.0431\pm0.0004$ & $0.0362\pm0.0003$ & $0.0355\pm0.0003$ & $\mathbf{0.0347\pm0.0003}$\\ 
\toprule
$[0.5, 0.5]$ & $0.0102\pm0.0001$ & $0.0103\pm0.0001$ & $0.0103\pm0.0001$ & $0.0102\pm0.0001$ & $0.0102\pm0.0001$ & $0.0103\pm0.0001$ & $0.0102\pm0.0001$ & $0.0095\pm0.0001$ & $\mathbf{0.0091\pm0.0001}$\\ 
\toprule
$[0.01, 0.5]$ & $0.0291\pm0.0001$ & $0.0306\pm0.0001$ & $0.0293\pm0.0001$ & $0.0291\pm0.0001$ & $0.0311\pm0.0001$ & $0.0327\pm0.0001$ & $0.0291\pm0.0001$ & $0.0278\pm0.0001$ & $\mathbf{0.0267\pm0.0001}$\\ 
\toprule
$[0.001, 0.5]$ & $0.0107\pm0.0001$ & $0.0103\pm0.0001$ & $0.0106\pm0.0001$ & $0.0107\pm0.0001$ & $0.0101\pm0.0001$ & $\mathbf{0.0091\pm0.0001}$ & $0.0107\pm0.0001$ & $0.0108\pm0.0001$ & $0.0108\pm0.0001$\\ 
\toprule
$[0.01, 0.01]$ & $0.0369\pm0.0003$ & $0.0368\pm0.0002$ & $0.0370\pm0.0002$ & $0.0369\pm0.0003$ & $\mathbf{0.0366\pm0.0002}$ & $\mathbf{0.0363\pm0.0002}$ & $0.0369\pm0.0003$ & $0.0371\pm0.0002$ & $0.0370\pm0.0002$\\ 
\toprule
$[0.001, 0.001]$ & $0.0109\pm0.0001$ & $0.0108\pm0.0001$ & $0.0108\pm0.0001$ & $0.0109\pm0.0001$ & $0.0108\pm0.0001$ & $\mathbf{0.0105\pm0.0001}$ & $0.0109\pm0.0001$ & $0.0109\pm0.0001$ & $0.0108\pm0.0001$\\
\end{tabular} 

%% file: tables/min_max_arm/min_max_armstable100_1.tex
\begin{tabular}{cccccccccc} 
$[q_v]_{v \in V_j}$ & PROP & MAX-H & MAX-B & VAR-H & VAR-B & VAR-CP & I-H & I-B & I-CP \\ 
\toprule
$2 \times 0.01, 6 \times 0.5, $ & $0.0332\pm0.0002$ & $0.0341\pm0.0002$ & $0.0343\pm0.0002$ & $0.0332\pm0.0002$ & $0.0334\pm0.0002$ & $0.0338\pm0.0002$ & $0.0332\pm0.0002$ & $0.0333\pm0.0002$ & $\mathbf{0.0308\pm0.0002}$\\ 
\toprule
$3 \times 0.01, 5 \times 0.5, $ & $0.0340\pm0.0002$ & $0.0342\pm0.0002$ & $0.0350\pm0.0002$ & $0.0340\pm0.0002$ & $0.0340\pm0.0002$ & $0.0335\pm0.0002$ & $0.0340\pm0.0002$ & $0.0339\pm0.0002$ & $\mathbf{0.0319\pm0.0002}$\\ 
\toprule
$2 \times 0.001, 6 \times 0.5, $ & $0.0288\pm0.0002$ & $0.0292\pm0.0002$ & $0.0299\pm0.0002$ & $0.0288\pm0.0002$ & $0.0286\pm0.0002$ & $0.0279\pm0.0002$ & $0.0288\pm0.0002$ & $0.0287\pm0.0002$ & $\mathbf{0.0259\pm0.0001}$\\ 
\toprule
$4 \times 0.01, 4 \times 0.5, $ & $0.0325\pm0.0002$ & $0.0330\pm0.0002$ & $0.0329\pm0.0002$ & $0.0325\pm0.0002$ & $0.0326\pm0.0002$ & $0.0324\pm0.0002$ & $0.0325\pm0.0002$ & $0.0326\pm0.0002$ & $\mathbf{0.0311\pm0.0002}$\\ 
\toprule
$5 \times 0.01, 3 \times 0.5, $ & $0.0335\pm0.0002$ & $0.0335\pm0.0002$ & $\mathbf{0.0330\pm0.0002}$ & $0.0335\pm0.0002$ & $0.0334\pm0.0002$ & $0.0337\pm0.0002$ & $0.0335\pm0.0002$ & $0.0335\pm0.0002$ & $\mathbf{0.0328\pm0.0002}$\\ 
\toprule
$3 \times 0.001, 5 \times 0.5, $ & $0.0251\pm0.0002$ & $0.0247\pm0.0002$ & $0.0258\pm0.0002$ & $0.0251\pm0.0002$ & $0.0247\pm0.0002$ & $\mathbf{0.0230\pm0.0001}$ & $0.0251\pm0.0002$ & $0.0251\pm0.0002$ & $\mathbf{0.0232\pm0.0001}$\\ 
\toprule
$6 \times 0.01, 2 \times 0.5, $ & $0.0342\pm0.0002$ & $0.0341\pm0.0002$ & $0.0342\pm0.0002$ & $0.0342\pm0.0002$ & $0.0341\pm0.0002$ & $\mathbf{0.0338\pm0.0002}$ & $0.0342\pm0.0002$ & $0.0340\pm0.0002$ & $\mathbf{0.0335\pm0.0002}$\\
\end{tabular} 

%% file: tables/min_max_arm/min_max_armstable300.tex
\begin{tabular}{cccccccccc} 
$[q_v]_{v \in V_j}$ & PROP & MAX-H & MAX-B & VAR-H & VAR-B & VAR-CP & I-H & I-B & I-CP \\ 
\toprule
$[0.1, 0.1]$ & $\mathbf{0.0312\pm0.0002}$ & $\mathbf{0.0311\pm0.0002}$ & $\mathbf{0.0312\pm0.0002}$ & $\mathbf{0.0312\pm0.0002}$ & $\mathbf{0.0314\pm0.0002}$ & $\mathbf{0.0316\pm0.0002}$ & $\mathbf{0.0312\pm0.0002}$ & $\mathbf{0.0312\pm0.0002}$ & $\mathbf{0.0312\pm0.0002}$\\ 
\toprule
$[0.1, 0.5]$ & $0.0213\pm0.0002$ & $0.0229\pm0.0002$ & $0.0216\pm0.0002$ & $0.0213\pm0.0002$ & $0.0235\pm0.0002$ & $0.0239\pm0.0002$ & $0.0213\pm0.0002$ & $\mathbf{0.0190\pm0.0001}$ & $\mathbf{0.0188\pm0.0001}$\\ 
\toprule
$[0.5, 0.5]$ & $0.0035\pm0.0000$ & $0.0035\pm0.0000$ & $0.0035\pm0.0000$ & $0.0035\pm0.0000$ & $0.0035\pm0.0000$ & $0.0035\pm0.0000$ & $0.0035\pm0.0000$ & $\mathbf{0.0031\pm0.0000}$ & $\mathbf{0.0031\pm0.0000}$\\ 
\toprule
$[0.01, 0.5]$ & $0.0156\pm0.0001$ & $0.0171\pm0.0001$ & $0.0160\pm0.0001$ & $0.0156\pm0.0001$ & $0.0179\pm0.0001$ & $0.0208\pm0.0001$ & $0.0156\pm0.0001$ & $\mathbf{0.0137\pm0.0001}$ & $\mathbf{0.0138\pm0.0001}$\\ 
\toprule
$[0.001, 0.5]$ & $0.0071\pm0.0000$ & $0.0071\pm0.0001$ & $0.0071\pm0.0000$ & $0.0071\pm0.0000$ & $0.0068\pm0.0001$ & $\mathbf{0.0066\pm0.0001}$ & $0.0071\pm0.0000$ & $0.0071\pm0.0000$ & $0.0070\pm0.0000$\\ 
\toprule
$[0.01, 0.01]$ & $\mathbf{0.0220\pm0.0002}$ & $\mathbf{0.0219\pm0.0002}$ & $\mathbf{0.0220\pm0.0002}$ & $\mathbf{0.0220\pm0.0002}$ & $\mathbf{0.0219\pm0.0002}$ & $0.0228\pm0.0002$ & $\mathbf{0.0220\pm0.0002}$ & $\mathbf{0.0220\pm0.0002}$ & $\mathbf{0.0220\pm0.0002}$\\ 
\toprule
$[0.001, 0.001]$ & $0.0098\pm0.0001$ & $0.0097\pm0.0001$ & $0.0097\pm0.0001$ & $0.0098\pm0.0001$ & $0.0096\pm0.0000$ & $\mathbf{0.0092\pm0.0000}$ & $0.0098\pm0.0001$ & $0.0097\pm0.0001$ & $0.0096\pm0.0000$\\
\end{tabular} 

%% file: tables/min_max_arm/min_max_armstable300_1.tex
\begin{tabular}{cccccccccc} 
$[q_v]_{v \in V_j}$ & PROP & MAX-H & MAX-B & VAR-H & VAR-B & VAR-CP & I-H & I-B & I-CP \\ 
\toprule
$2 \times 0.01, 6 \times 0.5, $ & $0.0153\pm0.0001$ & $0.0156\pm0.0001$ & $0.0154\pm0.0001$ & $0.0153\pm0.0001$ & $0.0156\pm0.0001$ & $0.0166\pm0.0001$ & $0.0153\pm0.0001$ & $0.0145\pm0.0001$ & $\mathbf{0.0140\pm0.0001}$\\ 
\toprule
$3 \times 0.01, 5 \times 0.5, $ & $0.0168\pm0.0001$ & $0.0175\pm0.0001$ & $0.0169\pm0.0001$ & $0.0168\pm0.0001$ & $0.0174\pm0.0001$ & $0.0189\pm0.0001$ & $0.0168\pm0.0001$ & $0.0160\pm0.0001$ & $\mathbf{0.0155\pm0.0001}$\\ 
\toprule
$2 \times 0.001, 6 \times 0.5, $ & $0.0111\pm0.0001$ & $0.0110\pm0.0001$ & $0.0110\pm0.0001$ & $0.0111\pm0.0001$ & $0.0107\pm0.0001$ & $0.0103\pm0.0000$ & $0.0111\pm0.0001$ & $0.0103\pm0.0001$ & $\mathbf{0.0098\pm0.0000}$\\ 
\toprule
$4 \times 0.01, 4 \times 0.5, $ & $0.0175\pm0.0001$ & $0.0179\pm0.0001$ & $0.0177\pm0.0001$ & $0.0175\pm0.0001$ & $0.0181\pm0.0001$ & $0.0196\pm0.0001$ & $0.0175\pm0.0001$ & $\mathbf{0.0169\pm0.0001}$ & $\mathbf{0.0168\pm0.0001}$\\ 
\toprule
$5 \times 0.01, 3 \times 0.5, $ & $0.0188\pm0.0001$ & $0.0192\pm0.0001$ & $0.0190\pm0.0001$ & $0.0188\pm0.0001$ & $0.0194\pm0.0001$ & $0.0208\pm0.0001$ & $0.0188\pm0.0001$ & $\mathbf{0.0182\pm0.0001}$ & $\mathbf{0.0181\pm0.0001}$\\ 
\toprule
$3 \times 0.001, 5 \times 0.5, $ & $0.0107\pm0.0001$ & $0.0106\pm0.0001$ & $0.0107\pm0.0001$ & $0.0107\pm0.0001$ & $0.0103\pm0.0001$ & $\mathbf{0.0098\pm0.0000}$ & $0.0107\pm0.0001$ & $0.0102\pm0.0001$ & $0.0099\pm0.0001$\\ 
\toprule
$6 \times 0.01, 2 \times 0.5, $ & $\mathbf{0.0201\pm0.0001}$ & $0.0203\pm0.0001$ & $\mathbf{0.0202\pm0.0001}$ & $\mathbf{0.0201\pm0.0001}$ & $0.0206\pm0.0001$ & $0.0226\pm0.0001$ & $\mathbf{0.0201\pm0.0001}$ & $\mathbf{0.0200\pm0.0001}$ & $\mathbf{0.0199\pm0.0001}$\\
\end{tabular} 

%% file: tables/min_max_arm/min_max_armstable500.tex
\begin{tabular}{cccccccccc} 
$[q_v]_{v \in V_j}$ & PROP & MAX-H & MAX-B & VAR-H & VAR-B & VAR-CP & I-H & I-B & I-CP \\ 
\toprule
$[0.1, 0.1]$ & $\mathbf{0.0230\pm0.0002}$ & $\mathbf{0.0229\pm0.0002}$ & $\mathbf{0.0230\pm0.0002}$ & $\mathbf{0.0230\pm0.0002}$ & $\mathbf{0.0232\pm0.0002}$ & $\mathbf{0.0232\pm0.0002}$ & $\mathbf{0.0230\pm0.0002}$ & $\mathbf{0.0230\pm0.0002}$ & $\mathbf{0.0231\pm0.0002}$\\ 
\toprule
$[0.1, 0.5]$ & $0.0162\pm0.0001$ & $0.0179\pm0.0001$ & $0.0165\pm0.0001$ & $0.0162\pm0.0001$ & $0.0185\pm0.0001$ & $0.0187\pm0.0001$ & $0.0162\pm0.0001$ & $\mathbf{0.0143\pm0.0001}$ & $\mathbf{0.0143\pm0.0001}$\\ 
\toprule
$[0.5, 0.5]$ & $0.0021\pm0.0000$ & $0.0021\pm0.0000$ & $0.0021\pm0.0000$ & $0.0021\pm0.0000$ & $0.0021\pm0.0000$ & $0.0021\pm0.0000$ & $0.0021\pm0.0000$ & $\mathbf{0.0018\pm0.0000}$ & $\mathbf{0.0018\pm0.0000}$\\ 
\toprule
$[0.01, 0.5]$ & $0.0123\pm0.0001$ & $0.0135\pm0.0001$ & $0.0127\pm0.0001$ & $0.0123\pm0.0001$ & $0.0155\pm0.0001$ & $0.0176\pm0.0001$ & $0.0123\pm0.0001$ & $\mathbf{0.0107\pm0.0001}$ & $0.0109\pm0.0001$\\ 
\toprule
$[0.001, 0.5]$ & $0.0061\pm0.0000$ & $0.0063\pm0.0000$ & $0.0061\pm0.0000$ & $0.0061\pm0.0000$ & $0.0063\pm0.0000$ & $0.0060\pm0.0000$ & $0.0061\pm0.0000$ & $\mathbf{0.0057\pm0.0000}$ & $\mathbf{0.0057\pm0.0000}$\\ 
\toprule
$[0.01, 0.01]$ & $\mathbf{0.0175\pm0.0001}$ & $\mathbf{0.0173\pm0.0001}$ & $\mathbf{0.0174\pm0.0001}$ & $\mathbf{0.0175\pm0.0001}$ & $\mathbf{0.0175\pm0.0001}$ & $0.0181\pm0.0001$ & $\mathbf{0.0175\pm0.0001}$ & $\mathbf{0.0174\pm0.0001}$ & $0.0176\pm0.0001$\\ 
\toprule
$[0.001, 0.001]$ & $0.0080\pm0.0000$ & $0.0080\pm0.0000$ & $0.0080\pm0.0000$ & $0.0080\pm0.0000$ & $0.0079\pm0.0000$ & $\mathbf{0.0076\pm0.0000}$ & $0.0080\pm0.0000$ & $0.0080\pm0.0000$ & $0.0078\pm0.0000$\\
\end{tabular} 

%% file: tables/min_max_arm/min_max_armstable500_1.tex
\begin{tabular}{cccccccccc} 
$[q_v]_{v \in V_j}$ & PROP & MAX-H & MAX-B & VAR-H & VAR-B & VAR-CP & I-H & I-B & I-CP \\ 
\toprule
$2 \times 0.01, 6 \times 0.5, $ & $0.0109\pm0.0001$ & $0.0117\pm0.0001$ & $0.0110\pm0.0001$ & $0.0109\pm0.0001$ & $0.0119\pm0.0001$ & $0.0127\pm0.0001$ & $0.0109\pm0.0001$ & $0.0099\pm0.0001$ & $\mathbf{0.0097\pm0.0001}$\\ 
\toprule
$3 \times 0.01, 5 \times 0.5, $ & $0.0125\pm0.0001$ & $0.0130\pm0.0001$ & $0.0126\pm0.0001$ & $0.0125\pm0.0001$ & $0.0135\pm0.0001$ & $0.0151\pm0.0001$ & $0.0125\pm0.0001$ & $\mathbf{0.0116\pm0.0001}$ & $\mathbf{0.0116\pm0.0001}$\\ 
\toprule
$2 \times 0.001, 6 \times 0.5, $ & $0.0072\pm0.0000$ & $0.0072\pm0.0000$ & $0.0072\pm0.0000$ & $0.0072\pm0.0000$ & $0.0071\pm0.0000$ & $0.0069\pm0.0000$ & $0.0072\pm0.0000$ & $0.0067\pm0.0000$ & $\mathbf{0.0064\pm0.0000}$\\ 
\toprule
$4 \times 0.01, 4 \times 0.5, $ & $0.0137\pm0.0001$ & $0.0141\pm0.0001$ & $0.0138\pm0.0001$ & $0.0137\pm0.0001$ & $0.0145\pm0.0001$ & $0.0162\pm0.0001$ & $0.0137\pm0.0001$ & $\mathbf{0.0129\pm0.0001}$ & $\mathbf{0.0129\pm0.0001}$\\ 
\toprule
$5 \times 0.01, 3 \times 0.5, $ & $0.0146\pm0.0001$ & $0.0151\pm0.0001$ & $0.0148\pm0.0001$ & $0.0146\pm0.0001$ & $0.0158\pm0.0001$ & $0.0172\pm0.0001$ & $0.0146\pm0.0001$ & $\mathbf{0.0139\pm0.0001}$ & $\mathbf{0.0140\pm0.0001}$\\ 
\toprule
$3 \times 0.001, 5 \times 0.5, $ & $0.0073\pm0.0000$ & $0.0072\pm0.0000$ & $0.0073\pm0.0000$ & $0.0073\pm0.0000$ & $0.0071\pm0.0000$ & $0.0069\pm0.0000$ & $0.0073\pm0.0000$ & $0.0070\pm0.0000$ & $\mathbf{0.0067\pm0.0000}$\\ 
\toprule
$6 \times 0.01, 2 \times 0.5, $ & $0.0156\pm0.0001$ & $0.0161\pm0.0001$ & $0.0158\pm0.0001$ & $0.0156\pm0.0001$ & $0.0168\pm0.0001$ & $0.0191\pm0.0001$ & $0.0156\pm0.0001$ & $\mathbf{0.0152\pm0.0001}$ & $\mathbf{0.0153\pm0.0001}$\\
\end{tabular} 

%% file: tables/adult/adult_featuretable50.tex
\begin{tabular}{cccccccccc} 
Feature & PROP & MAX-H & MAX-B & VAR-H & VAR-B & VAR-CP & I-H & I-B & I-CP \\ 
\toprule
$0$ & $\mathbf{0.1731\pm0.0004}$ & $0.2219\pm0.0004$ & $0.2232\pm0.0004$ & $\mathbf{0.1731\pm0.0004}$ & $\mathbf{0.1731\pm0.0004}$ & $0.1744\pm0.0004$ & $\mathbf{0.1735\pm0.0004}$ & $\mathbf{0.1735\pm0.0004}$ & $\mathbf{0.1732\pm0.0004}$\\ 
\toprule
$1$ & $\mathbf{0.0708\pm0.0005}$ & $0.1283\pm0.0013$ & $0.1288\pm0.0013$ & $\mathbf{0.0708\pm0.0005}$ & $\mathbf{0.0705\pm0.0005}$ & $\mathbf{0.0708\pm0.0005}$ & $\mathbf{0.0708\pm0.0005}$ & $\mathbf{0.0709\pm0.0005}$ & $\mathbf{0.0709\pm0.0005}$\\ 
\toprule
$2$ & $\mathbf{0.2433\pm0.0003}$ & $0.2939\pm0.0003$ & $0.2939\pm0.0003$ & $\mathbf{0.2433\pm0.0003}$ & $\mathbf{0.2433\pm0.0003}$ & $\mathbf{0.2438\pm0.0003}$ & $\mathbf{0.2435\pm0.0003}$ & $\mathbf{0.2435\pm0.0003}$ & $\mathbf{0.2436\pm0.0003}$\\ 
\toprule
$3$ & $\mathbf{0.0813\pm0.0005}$ & $0.1341\pm0.0010$ & $0.1340\pm0.0010$ & $\mathbf{0.0813\pm0.0005}$ & $0.0819\pm0.0006$ & $0.0892\pm0.0006$ & $\mathbf{0.0813\pm0.0005}$ & $\mathbf{0.0813\pm0.0005}$ & $\mathbf{0.0803\pm0.0005}$\\ 
\toprule
$4$ & $\mathbf{0.0770\pm0.0005}$ & $0.1297\pm0.0010$ & $0.1299\pm0.0010$ & $\mathbf{0.0770\pm0.0005}$ & $\mathbf{0.0776\pm0.0006}$ & $0.0849\pm0.0006$ & $\mathbf{0.0770\pm0.0005}$ & $\mathbf{0.0770\pm0.0005}$ & $\mathbf{0.0767\pm0.0005}$\\ 
\toprule
$5$ & $0.0590\pm0.0004$ & $0.0862\pm0.0005$ & $0.0852\pm0.0005$ & $0.0590\pm0.0004$ & $0.0593\pm0.0004$ & $0.0638\pm0.0004$ & $0.0590\pm0.0004$ & $0.0587\pm0.0004$ & $\mathbf{0.0572\pm0.0004}$\\ 
\toprule
$6$ & $\mathbf{0.0900\pm0.0005}$ & $0.1153\pm0.0007$ & $0.1156\pm0.0007$ & $\mathbf{0.0900\pm0.0005}$ & $\mathbf{0.0900\pm0.0005}$ & $0.0972\pm0.0006$ & $\mathbf{0.0900\pm0.0005}$ & $\mathbf{0.0900\pm0.0005}$ & $\mathbf{0.0892\pm0.0005}$\\
\end{tabular} 

%% file: tables/adult/adult_featuretable100.tex
\begin{tabular}{cccccccccc} 
Feature & PROP & MAX-H & MAX-B & VAR-H & VAR-B & VAR-CP & I-H & I-B & I-CP \\ 
\toprule
$0$ & $\mathbf{0.1326\pm0.0003}$ & $0.1786\pm0.0004$ & $0.1786\pm0.0004$ & $\mathbf{0.1325\pm0.0003}$ & $\mathbf{0.1325\pm0.0003}$ & $0.1332\pm0.0003$ & $\mathbf{0.1326\pm0.0003}$ & $\mathbf{0.1326\pm0.0003}$ & $\mathbf{0.1322\pm0.0003}$\\ 
\toprule
$1$ & $\mathbf{0.0460\pm0.0004}$ & $0.0885\pm0.0009$ & $0.0857\pm0.0008$ & $\mathbf{0.0460\pm0.0004}$ & $\mathbf{0.0454\pm0.0004}$ & $0.0474\pm0.0004$ & $\mathbf{0.0460\pm0.0004}$ & $0.0463\pm0.0004$ & $\mathbf{0.0460\pm0.0004}$\\ 
\toprule
$2$ & $0.2023\pm0.0003$ & $0.2030\pm0.0003$ & $0.2030\pm0.0003$ & $0.2023\pm0.0003$ & $0.2023\pm0.0003$ & $0.2054\pm0.0003$ & $\mathbf{0.2015\pm0.0003}$ & $\mathbf{0.2015\pm0.0003}$ & $\mathbf{0.2015\pm0.0003}$\\ 
\toprule
$3$ & $0.0558\pm0.0004$ & $0.0945\pm0.0008$ & $0.0945\pm0.0008$ & $0.0558\pm0.0004$ & $0.0560\pm0.0004$ & $0.0612\pm0.0004$ & $0.0558\pm0.0004$ & $0.0559\pm0.0004$ & $\mathbf{0.0543\pm0.0004}$\\ 
\toprule
$4$ & $0.0550\pm0.0004$ & $0.0893\pm0.0007$ & $0.0893\pm0.0007$ & $0.0550\pm0.0004$ & $0.0557\pm0.0004$ & $0.0600\pm0.0004$ & $0.0550\pm0.0004$ & $0.0550\pm0.0004$ & $\mathbf{0.0537\pm0.0004}$\\ 
\toprule
$5$ & $0.0417\pm0.0003$ & $0.0576\pm0.0004$ & $0.0564\pm0.0004$ & $0.0417\pm0.0003$ & $0.0426\pm0.0003$ & $0.0492\pm0.0003$ & $0.0417\pm0.0003$ & $0.0407\pm0.0003$ & $\mathbf{0.0398\pm0.0003}$\\ 
\toprule
$6$ & $0.0598\pm0.0004$ & $0.0727\pm0.0005$ & $0.0724\pm0.0005$ & $0.0598\pm0.0004$ & $0.0598\pm0.0004$ & $0.0682\pm0.0004$ & $0.0598\pm0.0004$ & $0.0598\pm0.0004$ & $\mathbf{0.0577\pm0.0004}$\\
\end{tabular} 

%% file: tables/adult/adult_featuretable300.tex
\begin{tabular}{cccccccccc} 
Feature & PROP & MAX-H & MAX-B & VAR-H & VAR-B & VAR-CP & I-H & I-B & I-CP \\ 
\toprule
$0$ & $0.0680\pm0.0002$ & $0.0827\pm0.0003$ & $0.0827\pm0.0003$ & $0.0681\pm0.0002$ & $0.0681\pm0.0002$ & $0.0750\pm0.0002$ & $0.0681\pm0.0002$ & $0.0681\pm0.0002$ & $\mathbf{0.0666\pm0.0002}$\\ 
\toprule
$1$ & $0.0251\pm0.0002$ & $0.0502\pm0.0004$ & $0.0496\pm0.0004$ & $0.0251\pm0.0002$ & $\mathbf{0.0249\pm0.0002}$ & $0.0264\pm0.0002$ & $0.0251\pm0.0002$ & $\mathbf{0.0248\pm0.0002}$ & $\mathbf{0.0246\pm0.0002}$\\ 
\toprule
$2$ & $0.1195\pm0.0003$ & $0.1207\pm0.0003$ & $0.1207\pm0.0003$ & $0.1195\pm0.0003$ & $0.1195\pm0.0003$ & $0.1321\pm0.0003$ & $0.1195\pm0.0003$ & $0.1195\pm0.0003$ & $\mathbf{0.1173\pm0.0003}$\\ 
\toprule
$3$ & $0.0289\pm0.0002$ & $0.0502\pm0.0004$ & $0.0491\pm0.0004$ & $0.0289\pm0.0002$ & $0.0296\pm0.0002$ & $0.0319\pm0.0002$ & $0.0289\pm0.0002$ & $0.0285\pm0.0002$ & $\mathbf{0.0277\pm0.0002}$\\ 
\toprule
$4$ & $0.0307\pm0.0002$ & $0.0482\pm0.0004$ & $0.0468\pm0.0004$ & $0.0307\pm0.0002$ & $0.0310\pm0.0002$ & $0.0336\pm0.0002$ & $0.0307\pm0.0002$ & $0.0301\pm0.0002$ & $\mathbf{0.0292\pm0.0002}$\\ 
\toprule
$5$ & $0.0236\pm0.0002$ & $0.0320\pm0.0002$ & $0.0315\pm0.0002$ & $0.0236\pm0.0002$ & $0.0260\pm0.0002$ & $0.0306\pm0.0002$ & $0.0236\pm0.0002$ & $\mathbf{0.0216\pm0.0002}$ & $\mathbf{0.0214\pm0.0002}$\\ 
\toprule
$6$ & $0.0281\pm0.0002$ & $0.0347\pm0.0003$ & $0.0328\pm0.0003$ & $0.0281\pm0.0002$ & $0.0298\pm0.0002$ & $0.0359\pm0.0003$ & $0.0281\pm0.0002$ & $0.0275\pm0.0002$ & $\mathbf{0.0267\pm0.0002}$\\
\end{tabular} 

%% file: tables/adult/adult_featuretable500.tex
\begin{tabular}{cccccccccc} 
Feature & PROP & MAX-H & MAX-B & VAR-H & VAR-B & VAR-CP & I-H & I-B & I-CP \\ 
\toprule
$0$ & $0.0469\pm0.0002$ & $0.0597\pm0.0002$ & $0.0591\pm0.0002$ & $0.0469\pm0.0002$ & $0.0471\pm0.0002$ & $0.0538\pm0.0002$ & $0.0469\pm0.0002$ & $0.0469\pm0.0002$ & $\mathbf{0.0447\pm0.0002}$\\ 
\toprule
$1$ & $\mathbf{0.0181\pm0.0001}$ & $0.0389\pm0.0003$ & $0.0383\pm0.0003$ & $\mathbf{0.0181\pm0.0001}$ & $0.0187\pm0.0001$ & $0.0196\pm0.0002$ & $\mathbf{0.0181\pm0.0001}$ & $0.0183\pm0.0001$ & $\mathbf{0.0180\pm0.0001}$\\ 
\toprule
$2$ & $0.0828\pm0.0002$ & $0.0840\pm0.0002$ & $0.0840\pm0.0002$ & $0.0828\pm0.0002$ & $0.0828\pm0.0002$ & $0.0977\pm0.0002$ & $0.0828\pm0.0002$ & $0.0828\pm0.0002$ & $\mathbf{0.0811\pm0.0002}$\\ 
\toprule
$3$ & $0.0217\pm0.0002$ & $0.0376\pm0.0003$ & $0.0369\pm0.0003$ & $0.0217\pm0.0002$ & $0.0221\pm0.0002$ & $0.0241\pm0.0002$ & $0.0217\pm0.0002$ & $0.0212\pm0.0002$ & $\mathbf{0.0209\pm0.0002}$\\ 
\toprule
$4$ & $0.0224\pm0.0002$ & $0.0367\pm0.0003$ & $0.0362\pm0.0003$ & $0.0224\pm0.0002$ & $0.0231\pm0.0002$ & $0.0253\pm0.0002$ & $0.0224\pm0.0002$ & $0.0215\pm0.0002$ & $\mathbf{0.0212\pm0.0002}$\\ 
\toprule
$5$ & $0.0180\pm0.0001$ & $0.0246\pm0.0002$ & $0.0234\pm0.0002$ & $0.0180\pm0.0001$ & $0.0204\pm0.0002$ & $0.0234\pm0.0002$ & $0.0180\pm0.0001$ & $\mathbf{0.0164\pm0.0001}$ & $\mathbf{0.0164\pm0.0001}$\\ 
\toprule
$6$ & $0.0204\pm0.0002$ & $0.0249\pm0.0002$ & $0.0238\pm0.0002$ & $0.0205\pm0.0002$ & $0.0225\pm0.0002$ & $0.0273\pm0.0002$ & $0.0204\pm0.0002$ & $\mathbf{0.0198\pm0.0002}$ & $\mathbf{0.0195\pm0.0001}$\\
\end{tabular} 

%% file: activefeatureselection-arxiv.bbl
\begin{thebibliography}{27}
\providecommand{\natexlab}[1]{#1}
\providecommand{\url}[1]{\texttt{#1}}
\providecommand{\urlprefix}{URL }
\expandafter\ifx\csname urlstyle\endcsname\relax
  \providecommand{\doi}[1]{doi:\discretionary{}{}{}#1}\else
  \providecommand{\doi}{doi:\discretionary{}{}{}\begingroup
  \urlstyle{rm}\Url}\fi

\bibitem[{Amaldi and Kann(1998)}]{amaldi1998approximability}
Amaldi, E.; and Kann, V. 1998.
\newblock On the approximability of minimizing nonzero variables or unsatisfied
  relations in linear systems.
\newblock \emph{Theoretical Computer Science} 209(1-2): 237--260.

\bibitem[{Antos, Grover, and Szepesv{\'a}ri(2008)}]{antos2008active}
Antos, A.; Grover, V.; and Szepesv{\'a}ri, C. 2008.
\newblock Active learning in multi-armed bandits.
\newblock In \emph{International Conference on Algorithmic Learning Theory},
  287--302. Springer.

\bibitem[{Benaroya, Han, and Nagurka(2005)}]{benaroya2005probability}
Benaroya, H.; Han, S.~M.; and Nagurka, M. 2005.
\newblock \emph{Probability models in engineering and science}.
\newblock CRC press.

\bibitem[{Bernstein(1924)}]{Bernstein24}
Bernstein, S.~N. 1924.
\newblock On a modification of Chebyshev's inequality and of the error formula
  of Laplace.
\newblock \emph{Annals Science Institute Sav. Ukraine} Sect. Math. 1.

\bibitem[{Carpentier et~al.(2011)Carpentier, Lazaric, Ghavamzadeh, Munos, and
  Auer}]{carpentier2011upper}
Carpentier, A.; Lazaric, A.; Ghavamzadeh, M.; Munos, R.; and Auer, P. 2011.
\newblock Upper-confidence-bound algorithms for active learning in multi-armed
  bandits.
\newblock In \emph{International Conference on Algorithmic Learning Theory},
  189--203. Springer.

\bibitem[{Chen et~al.(2014)Chen, Lin, King, Lyu, and
  Chen}]{chen2014combinatorial}
Chen, S.; Lin, T.; King, I.; Lyu, M.~R.; and Chen, W. 2014.
\newblock Combinatorial pure exploration of multi-armed bandits.
\newblock In \emph{Advances in Neural Information Processing Systems},
  379--387.

\bibitem[{Clopper and Pearson(1934)}]{clopper1934use}
Clopper, C.~J.; and Pearson, E.~S. 1934.
\newblock The use of confidence or fiducial limits illustrated in the case of
  the binomial.
\newblock \emph{Biometrika} 26(4): 404--413.

\bibitem[{Cohn, Atlas, and Ladner(1994)}]{CohnAtLa94}
Cohn, D.; Atlas, L.; and Ladner, R. 1994.
\newblock Improving generalization with active learning.
\newblock \emph{Machine Learning} 15: 201--221.

\bibitem[{Dua and Graff(2019)}]{UCI2019}
Dua, D.; and Graff, C. 2019.
\newblock {UCI} Machine Learning Repository.
\newblock \urlprefix\url{http://archive.ics.uci.edu/ml}.

\bibitem[{Garivier and Kaufmann(2016)}]{garivier2016optimal}
Garivier, A.; and Kaufmann, E. 2016.
\newblock Optimal best arm identification with fixed confidence.
\newblock In \emph{Conference on Learning Theory}, 998--1027.

\bibitem[{Guyon et~al.(2008)Guyon, Gunn, Nikravesh, and
  Zadeh}]{guyon2008feature}
Guyon, I.; Gunn, S.; Nikravesh, M.; and Zadeh, L.~A. 2008.
\newblock \emph{Feature extraction: foundations and applications}, volume 207.
\newblock Springer.

\bibitem[{Hastie, Tibshirani, and Friedman(2001)}]{HastieTiFr01}
Hastie, T.; Tibshirani, R.; and Friedman, J. 2001.
\newblock \emph{The Elements of Statistical Learning}.
\newblock Springer.

\bibitem[{Hoeffding(1963)}]{Hoeffding63}
Hoeffding, W. 1963.
\newblock Probability inequalities for sums of bounded random variables.
\newblock \emph{Journal of the American Statistical Association} 58(301):
  13--30.

\bibitem[{Kalyanakrishnan and Stone(2010)}]{kalyanakrishnan2010efficient}
Kalyanakrishnan, S.; and Stone, P. 2010.
\newblock Efficient Selection of Multiple Bandit Arms: Theory and Practice.
\newblock In \emph{ICML}, volume~10, 511--518.

\bibitem[{Kira and Rendell(1992)}]{kira1992feature}
Kira, K.; and Rendell, L.~A. 1992.
\newblock The feature selection problem: Traditional methods and a new
  algorithm.
\newblock In \emph{Aaai}, volume~2, 129--134.

\bibitem[{LeCun and Cortes(2010)}]{lecun-mnisthandwrittendigit-2010}
LeCun, Y.; and Cortes, C. 2010.
\newblock {MNIST} handwritten digit database.
\newblock http://yann.lecun.com/exdb/mnist/.
\newblock \urlprefix\url{http://yann.lecun.com/exdb/mnist/}.

\bibitem[{Li et~al.(2018)Li, Cheng, Wang, Morstatter, Trevino, Tang, and
  Liu}]{li2018feature}
Li, J.; Cheng, K.; Wang, S.; Morstatter, F.; Trevino, R.~P.; Tang, J.; and Liu,
  H. 2018.
\newblock Feature selection: A data perspective.
\newblock \emph{ACM Computing Surveys (CSUR)} 50(6): 94.
\newblock \urlprefix\url{{http://featureselection.asu.edu}}.

\bibitem[{Liu, Motoda, and Yu(2004)}]{liu2004selective}
Liu, H.; Motoda, H.; and Yu, L. 2004.
\newblock A selective sampling approach to active feature selection.
\newblock \emph{Artificial Intelligence} 159(1-2): 49--74.

\bibitem[{Liu et~al.(2003)Liu, Yu, Dash, and Motoda}]{liu2003active}
Liu, H.; Yu, L.; Dash, M.; and Motoda, H. 2003.
\newblock Active feature selection using classes.
\newblock In \emph{Pacific-Asia Conference on Knowledge Discovery and Data
  Mining}, 474--485. Springer.

\bibitem[{McCallum and Nigam(1998)}]{McCallumNi98}
McCallum, A.~K.; and Nigam, K. 1998.
\newblock Employing {EM} and pool-based active learning for text
  classification.
\newblock In \emph{Proc. International Conference on Machine Learning (ICML)},
  359--367. Citeseer.

\bibitem[{Nguyen and Smeulders(2004)}]{nguyen2004active}
Nguyen, H.~T.; and Smeulders, A. 2004.
\newblock Active learning using pre-clustering.
\newblock In \emph{Proceedings of the twenty-first international conference on
  Machine learning}, 79.

\bibitem[{Paninski(2003)}]{Paninski03}
Paninski, L. 2003.
\newblock Estimation of entropy and mutual information.
\newblock \emph{Neural computation} 15: 1191--1253.

\bibitem[{Ros and Guillaume(2017)}]{ros2017dides}
Ros, F.; and Guillaume, S. 2017.
\newblock DIDES: a fast and effective sampling for clustering algorithm.
\newblock \emph{Knowledge and information systems} 50(2): 543--568.

\bibitem[{Ros and Guillaume(2020)}]{ros2020sampling}
Ros, F.; and Guillaume, S. 2020.
\newblock \emph{Sampling Techniques for Supervised Or Unsupervised Tasks}.
\newblock Springer.

\bibitem[{Sabato and Shalev-Shwartz(2008)}]{SabatoSh08}
Sabato, S.; and Shalev-Shwartz, S. 2008.
\newblock Ranking Categorical Features Using Generalization Properties.
\newblock \emph{Journal of Machine Learning Research} 9: 1083--1114.

\bibitem[{Yang et~al.(2017)Yang, Lee, Chung, Wu, Chen, and Lin}]{YY2017}
Yang, Y.-Y.; Lee, S.-C.; Chung, Y.-A.; Wu, T.-E.; Chen, S.-A.; and Lin, H.-T.
  2017.
\newblock libact: Pool-based Active Learning in Python.
\newblock Technical report, National Taiwan University.
\newblock \urlprefix\url{https://github.com/ntucllab/libact}.
\newblock Available as arXiv preprint \url{https://arxiv.org/abs/1710.00379}.

\bibitem[{Zhang(2018)}]{zhang2018efficient}
Zhang, C. 2018.
\newblock Efficient active learning of sparse halfspaces.
\newblock \emph{arXiv preprint arXiv:1805.02350} .

\end{thebibliography}
